%% file: main.tex
\documentclass[conference]{IEEEtran}
\usepackage{times}

\usepackage[numbers]{natbib}
\usepackage{multicol}
\usepackage[bookmarks=true]{hyperref}

\usepackage{amsmath}
\usepackage{times} 
\usepackage{helvet} 
\usepackage{courier} 
\usepackage{graphicx} 
\usepackage[table]{xcolor}
\usepackage{natbib} 
\usepackage{amsmath}
\usepackage{amsfonts}
\usepackage{subcaption}
\usepackage{pifont}

\newcommand{\cm}{\ding{51}} 
\newcommand{\xm}{\ding{55}} 
\usepackage{caption}
\usepackage{algorithm}
\usepackage{algorithmic}
\usepackage{newfloat}
\usepackage{listings}
\usepackage{booktabs}       
\usepackage{amssymb}
\usepackage{mathtools}
\usepackage{amsthm}
\newtheorem{thm}{Theorem}
\newtheorem{corollary}{Corollary}
\newtheorem{lemma}[thm]{Lemma}
\newtheorem{prop}{Proposition}
\newtheorem{assumption}{Assumption}
\usepackage{mathabx}
\usepackage{array}
\usepackage{multirow}
\usepackage{subcaption}
\usepackage{soul}

\begin{document}

\title{Model-Based Diffusion Optimal Control for Multi-Robot Motion Planning}

\author{Author Names Omitted for Anonymous Review. Paper-ID [add your ID here]}



%
\author{\authorblockN{Zhilin He\textsuperscript{1},
Yorai Shaoul\textsuperscript{1},
Jiaoyang Li\textsuperscript{1}}
\authorblockA{\textsuperscript{1}Carnegie Mellon University, Pittsburgh, PA, USA}
\authorblockA{\texttt{\{hectorh, yshaoul, jiaoyanl\}}@andrew.cmu.edu}
}

\maketitle

\begin{abstract}
Multi-Robot Motion Planning in continuous environments, where robots must generate dynamically feasible, collision-free trajectories, is challenging due to the combinatorial growth of the joint trajectory space and the difficulty of enforcing dynamic feasibility and hard safety constraints. Recent approaches recast trajectory planning as probabilistic inference, sampling from a posterior over trajectories using diffusion models whose score functions are learned from demonstration data. While showing promising performance, these approaches are limited: they often rely on sizable demonstration datasets and struggle to rigorously enforce dynamics and hard safety constraints during sampling. To this end, we introduce Model-Based Diffusion Optimal Control (MDOC), a model-based diffusion planner that efficiently produces dynamically feasible trajectories without relying on data. Crucially, we show that MDOC's safety mechanism---combining known dynamics models with Control Barrier Function-constrained projections---naturally scales to multi-robot planning settings through Conflict-Based Search. Across simulation experiments, this integrated method consistently outperforms representative baseline planners in sample efficiency, geometric smoothness, and success rate, while reducing computation time and producing collision-free trajectories. Code: \href{https://github.com/hhhhzl/mdoc}{\texttt{https://github.com/hhhhzl/mdoc}}
\end{abstract}

\IEEEpeerreviewmaketitle

\section{Introduction}
\quad Consider a warehouse environment with numerous robots, where each robot attempts to reach a specified region while being subject to kinematic and dynamic constraints, such as acceleration limits and turning radius, and must avoid collisions with obstacles and other robots. The challenge lies in efficiently coordinating movements to avoid collisions while navigating long, continuous trajectories under the combinatorial growth of the joint trajectory space in different environments. 

Given the complexity of the full Multi-Robot Motion Planning (MRMP) problem, researchers have often resorted to solving simpler problems that approximate MRMP. One reformulation of the MRMP problem is known as Multi-Agent Path-Finding (MAPF)~\cite{stern2019multi}, where robots move in a discretized space (often a regular grid) and discrete time~\cite{moldagalieva2024db, okumura2023lacam}. 
In particular, the Conflict-Based Search (CBS)~\cite{sharon2015conflict} family of algorithms~\cite{barer2014suboptimal, boyarski2015icbs, li2021eecbs} has been influential due to its strong scalability and safety guarantees. However, this MAPF-based abstraction relies on discretization assumptions, such as constant velocities, rectilinear and grid-based motions, which limit their applicability to continuous MRMP settings with dynamic feasibility and robot interactions.

More recent work has explored motion planning through probabilistic inference, sampling trajectory posteriors with diffusion planners~\cite{carvalho2023motion}. These planners leverage the strong generative capacity of diffusion models to learn a score function from expert demonstrations and represent complex, multimodal trajectory distributions in continuous spaces~\cite{janner2022planning, zhang2024robotdiffuse}. By stochastically denoising trajectory priors, diffusion planners can produce smoother trajectories and avoid the restrictive assumptions imposed by grid-based simplifications. Building on these single-robot advances, recent work extends diffusion planners to MRMP~\cite{dastider2024apex, liang2025simultaneous} and integrates them as low-level continuous-space planners within CBS for multi-robot settings~\cite{shaoul2025multi}.

However, existing diffusion planners remain fundamentally model-free, requiring large expert datasets to learn the score function and struggling to rigorously enforce dynamics during denoising the expert trajectories. As a result, they often rely on extensive learning from demonstrations and neglect existing model information, yielding trajectories that are not genuinely dynamics-aware. Additionally, current diffusion planners handle inter-robot collisions through soft constraints, where the hard safety constraints cannot be guaranteed, or through post-hoc safety projections~\cite{romer2024diffusion, xiao2023safediffuser, zhang2025constrained, christopher2024constrained}, which massively increase computational planning cost on the full trajectory, making planners fail in large-scale multi-robot congested settings.

We propose \textbf{M}odel-Based \textbf{D}iffusion \textbf{O}ptimal \textbf{C}ontrol (MDOC), a training-free diffusion planner that enforces collision avoidance and scales to MRMP without any demonstration data. Building on Model-Based Diffusion (MBD)~\cite{pan2024model}, MDOC analytically estimates diffusion scores via Monte Carlo score ascent under a known dynamics model, interpreting denoising as stochastic optimal control so avoiding demonstration learning. Crucially, MDOC extends MBD to enforce rigorous collision avoidance through Control Barrier Function (CBF)-constrained projections directly inside the model-based diffusion rollouts. Capitalizing on these properties, we further propose MDOC-CBS, a scalable MRMP algorithm that resolves inter-robot collisions through the same safety mechanism while retaining the efficiency of high-level CBS.
The contributions of this work are threefold:

\begin{itemize}
    \item We develop MDOC, a model-based diffusion planner that enforces safety during sampling via CBF-constrained projections inside model-based rollouts, yielding dynamically feasible and collision-free trajectories.
    \item We propose MDOC-CBS, a CBS-based coordination that scales to 20 robots (40 in larger maps) while resolving inter-robot collisions via the same CBF-constrained projection by MDOC as the low-level planner.
    \item We evaluate our methods across diverse environments, start-goal configurations, constraint settings, and horizons and show that it improves success rate and trajectory quality while reducing computation and maintaining collision-free execution compared to representative baselines.
\end{itemize}

\section{Related Work}
We review prior work on single-robot motion planning and multi-robot motion planning.

\subsection{Single-Robot Motion Planning}
Arguably, the most popular paradigm for single-robot motion planning is sampling. Sampling-based motion planning has been extensively studied through methods such as probabilistic roadmaps (PRM)~\cite{kavraki1996probabilistic, kavraki1998analysis, la2011motion} and rapidly exploring random trees (RRT)~\cite{lavalle1998rapidly, kuffner2000rrt, kleinbort2018probabilistic}.
Although they provide probabilistic completeness~\cite{karaman2011sampling, solovey2016finding} and finite sampling guarantees~\cite{dayan2023near}, those methods struggle as problem geometry becomes more constrained and dynamics become more involved.

Trajectory optimization, mostly from control theory, formulates nonlinear robot motion planning problems directly over trajectories with constraints and dynamics. Sequential Linear Quadratic methods~\cite{sideris2005efficient, farshidian2017real} have been successfully used for legged robots~\cite{katayama2023model, grandia2023perceptive}, while other works directly embed kinematic and dynamic limits for trajectory optimization~\cite{chen2015decoupled, luis2020online}. These methods can generate dynamically feasible trajectories, but solving large nonlinear programs becomes computationally demanding in cluttered environments.

Diffusion-based methods mitigate expensive optimization by learning a trajectory distribution and sampling candidates via Denoising Diffusion Probabilistic Models (DDPMs)~\cite{ho2020denoising, janner2022planning, chi2025diffusion}. Following these developments, Motion Planning Diffusion (MPD)~\cite{carvalho2023motion} proposes learning and sampling diffusion models as prior and posterior trajectories for robot motion planning. Despite their efficiency, MPD still requires learning from large collections of expert trajectories. More recent methods combine diffusion models with control on Model Predictive Control~\cite{zhou2024diffusion}, Model-Based Diffusion (MBD)~\cite{pan2024model} for trajectory optimization, and diffusion-based constraints~\cite{romer2024diffusion}. However, safety mechanisms in those methods are either underexplored or enforced via inefficient projections.

\subsection{Multi-Robot Motion Planning}
Early works address the PSPACE-hardness of MRMP~\cite{hopcroft1986reducing} by extending single-robot sampling-based planners into the composite configuration space, where each sample lies in the high-dimensional joint configuration space of all robots.
Although those methods provide asymptotic optimality~\cite{shome2020drrt}, they struggle in multi-robot settings because their search space grows exponentially with the number of robots.

To improve scalability, recent works utilized the Conflict-Based Search (CBS) framework~\cite{sharon2015conflict} and decoupled the multi-robot planning problem into low-level single-robot planning and high-level conflict resolution.
For example, KCBS~\cite{kottinger2022conflict} and db-CBS~\cite{moldagalieva2024db} design sampling-based low-level planners with kinodynamic constraints, aiming for dynamic feasibility but still suffering from sampling inefficiency of kinodynamic, sampling-based tree expansion.

Multi-Robot Motion Planning with Diffusion Models (MMD)~\cite{shaoul2025multi} combines MPD with CBS to provide collision-free paths in scalable multi-robot environments. Model-based approaches aim for trajectory optimization within three-dimensional holonomic cases~\cite{zhang2025d4orm} and annealing legs~\cite{xue2024full} without learning, but these methods still optimize all robots with a joint cost function, which limits scalability and makes hard safety constraints difficult to enforce.

\textbf{Summary.} We are the first to propose CBF-constrained projection with model-based diffusion as a single-robot planner, yielding dynamically feasible trajectories and ensuring safety during diffusion sampling, and to scale it to multi-robot settings through CBS.


\section{Preliminaries}
\label{sec:preliminaries}
Let us turn to formally defining the problem at hand and cover the relevant background. In what follows, we use a lower right index $i$ to denote the robot identity, an upper right index $h$ for the time step within the planning horizon, and an upper left index $k$ for the diffusion step (e.g., ${}^{k}\tau_i^h$).

\subsection{Problem Statement}
\label{subsec:problem_setup}

In Multi-Robot Motion Planning (MRMP), we aim to convey a team of $N$ robots from their start states to goal states while respecting each robot's dynamics and avoiding collisions with obstacles and each other. Let $s_i$ encode the state of robot $i$, such as its configuration and velocity, and $u_i$ encode its control input. A trajectory can be represented as a sequence of state and control input pairs over a finite time horizon $H$ as
\begin{math}
\tau_i = \big[(s_i^1, u_i^1), (s_i^2, u_i^2) \dots, (s_i^H, u_i^H)\big].
\end{math}
We denote by $\boldsymbol{s}^h := \{s^h_i\}_{i=1}^N$ and $\boldsymbol{u}^h := \{u^h_i\}_{i=1}^N$ the joint state and control of all robots at time $h$. Let $\boldsymbol{\tau}^{1:H} := \{\tau^{1:H}_i\}_{i=1}^N$ denote a multi-robot joint trajectory over time horizon 1 to $H$. We define the problem objective as
\begin{equation}
\mathcal J(\boldsymbol{\tau}^{1:H}) := \sum_{i=1}^N \Big(\gamma_i \sum_{h=1}^H c(\tau_i^{h})\Big) \;+\; \sum_{i=1}^N \beta_i\, c_f(s_i^{H}),
\label{eq:objective}
\end{equation}
where $\mathcal{J}$ indicates the objective function, $c$ and $c_f$ denote the intermediate and terminal cost functions (e.g., distance traveled between consecutive states and terminal distance to the goal),  and $\gamma_i $ and $\beta_i$ are their associated positive weighting coefficients. Then, the MRMP problem is
\begin{equation}
\begin{aligned}
\boldsymbol{\tau}^* &\in \arg\min_{\boldsymbol{\tau}^{1:H}}\ \mathcal J(\boldsymbol{\tau}^{1:H}) \\
\text{s.t.}\quad
& s_i^{h+1} = f(s_i^h,u_i^h),\quad \forall i,\ h=1,\dots,H-1,\\
& g^{h}(\boldsymbol{s}^{h}, \boldsymbol{u}^{h}) \le 0,\quad h=1,\dots,H,
\end{aligned}
\end{equation}
with $s_i^{h+1} = f(s_i^h,u_i^h)$ representing the discrete-time dynamics, and $g^{h}(\boldsymbol{s}^{h}, \boldsymbol{u}^{h}) \le 0$ generically collecting all obstacle-avoidance and inter-robot collision-avoidance constraints at time $h$ to ensure safety. 

\subsection{Model-Based Diffusion for Trajectory Sampling}
\label{subsec:mbd}

The trajectory for robot $i$, $\tau^{1:H}_i$, can be treated as a random variable and sampled by iterative denoising from Gaussian noise. Since this subsection focuses on single-robot motion planning, we omit the robot index $i$ for brevity from now on.

We follow the standard DDPMs and denote by ${}^{k}\tau^{1:H}$ the noisy trajectory at diffusion step $k$:
\begin{equation}
{}^{k+1}\tau^{1:H} \;=\; \sqrt{\bar\alpha_k}\,{}^{1}\tau^{1:H} \;+\; \sqrt{1-\bar\alpha_k}\,\epsilon,
\quad \epsilon\sim\mathcal N(0,\mathbf I),
\label{eq:ddpm_reparam}
\end{equation}
where $\alpha_k := 1-\beta_k$ controls the amount of signal preserved at diffusion step $k$ under the noise-schedule coefficient $\beta_k$, $\bar\alpha_k:=\prod_{n=1}^{k}\alpha_n$, and $K$ is the total number of diffusion steps. The noise schedule is chosen such that the final noisy trajectory ${}^{K}\tau^{1:H}$ is close to standard Gaussian noise and can be initialized from $\mathcal N(0,\mathbf I)$. Let ${}^1p({}^1\tau^{1:H})$ denote the target distribution over clean trajectories. We define the noisy marginals using the forward noising kernel $q$:
\begin{equation}
{}^kp({}^{k}\tau^{1:H}) := \int q({}^{k}\tau^{1:H}\mid {}^{1}\tau^{1:H})\,{}^1p({}^{1}\tau^{1:H})\,d{}^{1}\tau^{1:H}.
\label{eq:pk_def}
\end{equation}

Based on this noisy marginal, a standard result in score-based diffusion is that the optimal reverse-time drift depends on the score function, 
\begin{math}
\mathbf S_k({}^{k}\tau^{1:H})
\;:=\;
\nabla_{{}^{k}\tau^{1:H}} \log {}^kp({}^{k}\tau^{1:H}),
\label{eq:score_def}
\end{math}
where $\nabla_{{}^{k}\tau^{1:H}}$ denotes the gradient with respect to ${}^{k}\tau^{1:H}$. The score guides denoising toward higher-density regions of the noisy marginal ${}^kp$ at diffusion step $k$ while removing noise. Consequently, the reverse update can be written in the score-guided form
\begin{equation}
{}^{k-1}\tau^{1:H}
=
\frac{1}{\sqrt{\alpha_k}}
\left(
{}^{k}\tau^{1:H} + (1-\bar\alpha_k)\,\hat{\mathbf S}_k
\right),
\label{eq:reverse_update_prelim}
\end{equation}
where $\hat{\mathbf S}_k\approx \mathbf S_k$ is a score estimate at step $k$.
In standard DDPMs, $\hat{\mathbf S}_k$ is obtained from a learned noise predictor $\epsilon_\theta({}^{k}\tau^{1:H},k)$, yielding the approximation
\begin{equation}
\mathbf S_k({}^{k}\tau^{1:H})
\approx
-\frac{1}{\sqrt{1-\bar\alpha_k}}\ \epsilon_\theta({}^{k}\tau^{1:H},k).
\label{eq:score_eps}
\end{equation}

Learning $\epsilon_\theta$ (or a score network) typically requires a large amount of expert trajectories. When a dynamic model is available, MBD~\cite{pan2024model} avoids learning $\epsilon_\theta$ by sampling control sequences ${}^{k}\bar{u}^{1:H-1}$ and using the known dynamics to roll them out into state trajectories ${}^{k}\tau^{1:H}$, then analytically constructing $\hat{\mathbf S}_k$ with denoised rollouts. Concretely, at reverse step $k$, we first sample denoised candidates from the Gaussian proposal induced by the forward kernel

\begin{equation}
\tilde{\tau}^{1:H} \sim \Omega_k
:= \mathcal{N}\!\left(
\frac{{}^{k}\tau^{1:H}}{\sqrt{\bar{\alpha}_k}},
\left(\frac{1}{\bar{\alpha}_k} - 1\right)\mathbf{I}
\right).
\label{eq:proposal_prelim}
\end{equation}
Let $\{\tilde{\tau}_m^{1:H}\}_{m=1}^{M}$ be $M$ samples from $\Omega_k$.
We then evaluate each candidate under a target density ${}^{1}p(\cdot)$ and form the importance-weighted average
\begin{equation}
{}^{k}\bar{\tau}^{1:H}
=
\frac{\sum_{m=1}^{M} {}^{1}p(\tilde{\tau}_m^{1:H})\,\tilde{\tau}_m^{1:H}}
{\sum_{m=1}^{M} {}^{1}p(\tilde{\tau}_m^{1:H})},
\label{eq:mc_avg_prelim}
\end{equation}
which yields the Monte Carlo Score-Ascent (MCSA) estimator
\begin{equation}
\hat{\mathbf{S}}_k
\approx
-\frac{{}^{k}\tau^{1:H}}{1-\bar{\alpha}_k}
+\frac{\sqrt{\bar{\alpha}_k}}{1-\bar{\alpha}_k}\,
{}^{k}\bar{\tau}^{1:H}.
\label{eq:mc_score_prelim}
\end{equation}
Together with Eqs~\eqref{eq:reverse_update_prelim}--\eqref{eq:mc_score_prelim}, this yields a training-free diffusion sampler that iteratively (\textbf{1}) proposes candidates, (\textbf{2}) reweights them by ${}^1p({}^{k}\tau^{1:H})$, and (\textbf{3}) updates ${}^{k}\tau^{1:H}$ toward high-density trajectories. The remaining question is how to instantiate ${}^{1}p({}^{k}\tau^{1:H})$ so that high-probability samples correspond to low-cost, dynamically feasible, and collision-free trajectories. This can be answered through a planning-as-inference target distribution.

\subsection{Planning as Inference}
\label{subsec:pai}
In the following, we omit both robot index $i$ and diffusion step $k$ for brevity. That is, we write the trajectory for robot $i$ at diffusion step $k$, ${}^{k}\tau^{1:H}_i$, as $\tau^{1:H}$.

A standard relaxation of deterministic optimal control is to model soft optimality through a Boltzmann distribution, extensively studied in~\cite{toussaint2009robot, levine2018reinforcement, peters2010relative, watson2020stochastic, watson2023inferring}:
\begin{equation}
p_J(\tau^{1:H})
\propto
\exp\!\left(-\frac{\mathcal{J}(\tau^{1:H})}{\lambda}\right),
\label{eq:boltzmann_cost_prelim}
\end{equation}
where $\lambda>0$ is a temperature parameter.
As $\lambda\rightarrow 0$, $p_J$ concentrates around minimizers of $\mathcal{J}$.

Dynamic feasibility and safety can be encoded by a feasibility prior $p_{\mathrm{feas}}(\tau^{1:H})$, so that
\begin{math}
{}^{1}p(\tau^{1:H})
\;\propto\;
p_{\mathrm{feas}}(\tau^{1:H})\,
p_J(\tau^{1:H}),
\label{eq:target_dist}
\end{math}
whose support is restricted to dynamically feasible and collision-free trajectories. Conceptually, a hard-feasibility prior can be written as an indicator product
\begin{subequations}
\begin{align}
p_{\mathrm{feas}}(\tau^{1:H})
&\propto p_d(\tau^{1:H})\,p_g(\tau^{1:H}), \\
p_d(\tau^{1:H})
&:= \prod_{h=1}^{H-1}\mathbf{1}\!\left(s^{h+1}=f(s^{h},u^{h})\right),\\
p_g(\tau^{1:H})
&:= \prod_{h=1}^{H}\mathbf{1}\!\left(g^{h}(s^{h},u^{h})\le 0\right).
\end{align}
\end{subequations}
However, directly using hard indicators is brittle in constrained environments: most candidates receive zero weight (``dead samples''), providing no useful score guidance for MCSA~\cite{mishra2025eb}.
This motivates handling feasibility explicitly rather than through hard rejection. In practice, it is desirable to work with candidates that lie in the feasible set $\mathcal D\cap\mathcal C$, where $\mathcal D$ denotes dynamic feasibility by rollouts and $\mathcal C := \{\tau^{1:H}\mid g^h(s^h,u^h)\le 0,\ \forall h=1,\ldots,H\}$ denotes the safety set induced by the inequality constraints. When trajectories are feasible, the MCSA importance weights in Eq.~\eqref{eq:mc_avg_prelim} can be computed using only the optimality term,
\begin{math}
w_m \propto p_J(\tilde\tau_m^{1:H}),
\end{math}
without assigning zero weights to infeasible samples.

\section{Methodology}
\label{sec:method}
We present our Model-Based Diffusion Optimal Control (MDOC) planner and its multi-robot extension, MDOC-CBS. Given a known dynamics model, diffusion rollouts maintain dynamic feasibility, yielding trajectories in $\mathcal D$. Our contribution is to enforce safety during sampling, then scales to MRMP, by introducing a feasibility operator $\mathcal F$ that maps each candidate into $\mathcal D\cap\mathcal C$. Specifically, $\mathcal F$ applies CBF-constrained projections on controls within the model-based diffusion rollouts.

\input{algos/algorithm1}

\subsection{Model-Based Diffusion Optimal Control (MDOC)}
\label{subsec:cbf}

MDOC is a single-robot motion planner, and thus we omit the robot index $i$ for brevity in the following. 
MDOC follows the diffusion paradigm. At each diffusion step $k$, it invokes a sampler that proposes a nominal control sequence ${}^{k}\bar{u}^{1:H-1}$ (with ${}^{k}\bar{u}^{h}$ at time $h$), which may violate safety constraints. To maintain safety, MDOC projects ${}^{k}\bar{u}^{1:H-1}$ to a safe sequence ${}^{k}u^{1:H-1}$ that satisfies $\mathcal{C}$ via a feasibility operator $\mathcal F$, which enforces a discrete-time CBF condition during the model-based rollout. 

\input{algos/algorithm2}

Concretely, we require a barrier function $b({}^ks^{h})$ (e.g., $b({}^ks^{h})=\|{}^ks^{h}-s_{\mathrm{obs}}\|^2-r^2$, so $b({}^ks^{h})\ge 0$ enforces a minimum separation $r$ from the obstacle) that satisfies the discrete-time CBF condition
$b({}^ks^{h+1}) \ge (1-\gamma \Delta h)\,b({}^ks^{h})$, where $\Delta h$ is the time step size, meaning that the safety margin $b(\cdot)$ should not decay faster than rate $\gamma$. We locally linearize the dynamics and the barrier function around $({}^{k}s^{h}, {}^{k}\bar{u}^{h})$:
\begin{equation}
{}^ks^{h+1} \approx {}^{k}\hat{s}^{h+1} + {}^{k}B^{h}({}^{k}u^{h} - {}^{k}\bar{u}^{h}),\
{}^{k}\hat{s}^{h+1} := f({}^{k}s^{h}, {}^{k}\bar{u}^{h}),
\end{equation}
where ${}^{k}B^{h} := \left.\frac{\partial f(s,u)}{\partial u}\right|_{({}^{k}s^{h}, {}^{k}\bar{u}^{h})}$ is the Jacobian that locally maps control deviations to changes in the next state. Applying the same first-order approximation to $b(\cdot)$ yields the control--affine inequality~\cite{agrawal2017discrete,ames2014control, ames2019control} 
\begin{equation}
\begin{aligned}
& \nabla b({}^{k}\hat{s}^{h+1})^{\!\top} {}^{k}B^{h} {}^ku^{h}
\;\ge\; \\
& (1 - \gamma \Delta h)\, b({}^{k}s^{h})
 - b({}^{k}\hat{s}^{h+1})
 + \nabla b({}^{k}\hat{s}^{h+1})^{\!\top} {}^{k}B^{h} {}^{k}\bar{u}^{h},
\label{eq:disc-lin}
\end{aligned}
\end{equation}
which is equivalent to ${}^{k}a^{h\top}{}^ku^{h} \ge {}^{k}d^{h}$, where
\begin{equation}
\begin{aligned}
& {}^{k}a^{h\top}:=\nabla b({}^{k}\hat s^{h+1})^\top {}^{k}B^h,
\\
& {}^{k}d^h := (1-\gamma\Delta h)b({}^{k}s^h)-b({}^{k}\hat s^{h+1})+{}^{k}a^{h\top}{}^{k}\bar u^h.
\label{eq:cbf_rhs_def}
\end{aligned}
\end{equation}
These inequalities define a safe-control set in the control space
\(
{}^{k}\mathcal K^h := \{u \mid {}^{k}A^h u \ge {}^{k}D^h\},
\)
where ${}^{k}D^h$ is the vector collecting all scalar offsets ${}^{k}d^h$.

Stacking the half-spaces from all obstacles yields the polyhedral safe-control set
\begin{math}
{}^{k}\mathcal K^{h}
=
\left\{u\ \middle|\ {}^{k}A^{h}u \ge {}^{k}D^{h}\right\},
\label{eq:cbf_polytope}
\end{math}
where each row ${}^{k}a_{j}^{h\top}$ of ${}^{k}A^{h}$ and entry ${}^{k}d_{j}^{h}$
of ${}^{k}D^{h}$ correspond to one linearized CBF constraint. For speed inside diffusion sampling, we enforce each violated half-space via a closed-form projection.
Given a single constraint ${}^ka^{\!h\top}{}^{k}u^{h}\ge {}^{k}d^h$, the projection for nominal control ${}^{k}\bar u^{h}$ is
\begin{equation}
{}^{k}u^{h} \leftarrow
\begin{cases}
{}^{k}\bar{u}^{h}, & {}^ka^{h\top} {}^{k}\bar{u}^{h} \ge {}^{k}d^h, \\
{}^{k}\bar{u}^{h}
+ \dfrac{{}^{k}d^h - {}^ka^{h\top} {}^{k}\bar{u}^{h}}{\|{}^{k}a^{h}\|^{2}}\, {}^{k}a^{h},
& \text{otherwise}.
\end{cases}
\label{eq:cbf_proj_method}
\end{equation}
The correction term moves ${}^{k}\bar u^{h}$ along the constraint normal ${}^{k}a^{h}$ by the minimal amount so that ${}^ka^{h\top }{}^{k}u^{h}={}^{k}d^h$, which is the Euclidean projection onto the corresponding half-space and solves $\min_{{}^{k}u^{h}}\ \tfrac12\|{}^{k}u^{h}-{}^{k}\bar u^{h}\|^2$ subject to ${}^ka^{h\top} {}^{k}u^{h}\ge {}^{k}d^h$.

Therefore, given ${}^{k}\bar u^{1:H-1}$, we compute a feasible rollout by projecting per step ${}^{k}\bar u^{h}$ to ${}^{k}u^{h}$ through the half-space projection implied by Eq.~\eqref{eq:cbf_proj_method}.
In MDOC, $\mathcal F$ is thus implemented by iterating over $H$:
(\textbf{1}) projecting each proposed control ${}^{k}\bar u^{h}$ onto ${}^{k}\mathcal K^{h}$ to enforce safety and
(\textbf{2}) rolling out the known dynamics one step to obtain ${}^{k}s^{h+1}$, so dynamic feasibility is not broken.


We now combine (\textbf{1}) the MCSA sampler from Section~\ref{subsec:mbd} and (\textbf{2}) the feasibility operator $\mathcal F$.
At each diffusion step $k$, MDOC samples $M$ potentially unsafe candidates $\{\tilde\tau_m^{1:H}\}_{m=1}^M$ from the Gaussian proposal with Eq.~\eqref{eq:proposal_prelim} parameterized by controls ${}^{k}\bar{u}^{1:H-1}$, then recovers dynamically feasible and CBF-constrained trajectories $\{\tilde\tau_m^{1:H}\}_{m=1}^M$ by rolling out ${}^{k}u_m^{1:H-1}$ through the known dynamics and applying the projection with Eq.~\eqref{eq:cbf_proj_method}, namely $\tau_m^{1:H} = \mathcal F(\tilde\tau_m^{1:H})$,  computes Monte Carlo mean with 
\begin{equation}
{}^{k}\bar{\tau}^{1:H}=
\frac{\sum_{m=1}^{M} {}^1p(\tau_m^{1:H})\,\tau_m^{1:H}}
{\sum_{m=1}^{M} {}^1p(\tau_m^{1:H})},
\label{eq:mc_avg_prelim2}
\end{equation}
and forms the score estimate from Eq.~\eqref{eq:mc_score_prelim} to update the noisy trajectory through Eq.~\eqref{eq:reverse_update_prelim}, as summarized in Algorithm~\ref{alg:mdoc}. As a result, all trajectories used to compute the Monte Carlo estimate in Eq.~\eqref{eq:mc_score_prelim} satisfy the dynamics and constraints.

\subsection{MDOC-CBS}
\label{sec:method_mdoc_cbs}
We now show how the mechanisms in MDOC can directly generalize to support planning for multi-robot settings via CBS~\cite{sharon2015conflict}. Briefly, the CBS algorithm decomposes the MRMP problem into two levels: a low-level where each robot plans for itself, and a high-level, where the planned trajectories are surveyed, conflicts (i.e., inter-robot collisions) are identified, and constraints are imposed on colliding robots such that they will avoid the collision configurations upon replanning. 

Specifically, MDOC-CBS, as shown in Algorithm~\ref{alg:mdoc_cbs}, first creates a root node $V_{\text{root}}$ and stores it in a priority queue termed the constraint tree (CT). The root node includes (\textbf{1}) an constraint set  $V_{\text{root}}.\mathcal{C}_i$ for each robot $i$, initialized with obstacle-avoidance constraints, and (\textbf{2}) a joint trajectory $V_{\text{root}}.\boldsymbol{\tau}^{1:H}$, constructed by planning for each robot with MDOC over the horizon $H$.

CBS iteratively selects the least-cost node $V$ from the CT (in our case, the one with the fewest conflicts) and evaluates it. If $V.\boldsymbol{\tau}^{1:H}$ is conflict-free, it returns it as a solution. Otherwise, if a conflict is found between robots $i$ and $j$ at time $h$, CBS branches node $V$ into two new CT nodes, $V_i$ and $V_j$, for robots $i$ and $j$: each child node copies the constraint sets and trajectories from $V$ and adds a new constraint forbidding robot $i$ (resp.\ $j$) from entering a small workspace sphere centered at robot $j$'s state $s_j^h$ at time $h$, with a safe margin $r$. CBS then replans the trajectories of robots $i$ and $j$ using the low-level MDOC planner under the updated constraint sets $V_i.\mathcal{C}_i$ and $V_j.\mathcal{C}_j$, enforced by the single-robot feasibility operator $\mathcal{F}$ via CBF-constrained projections within the model-based diffusion rollouts, as illustrated in Algorithm~\ref{alg:mdoc}. When replanning robot $i$, each CBS constraint generated from a conflict with robot $j$ is represented as a time-indexed forbidden sphere centered at the conflicting state of robot $j$. We convert the pairwise CBF $b_{ij}(s_i,s_j)$ into a single-robot barrier by fixing the second argument to this sphere center, that $b_{i\mid j}^h(s_i^h) = b_{ij}(s_i^h,s_j^h)$
, where $s_j^h$ is fixed by the CBS constraint. If $b_{ij}(s_i^h,s_j^h)=\|s_i^h-s_j^h\|^2-r^2$, then $b_{i\mid j}^h(s_i^h)=\|s_i^h-s_j^h\|^2-r^2\ge 0$, enforced by $\mathcal F$ with $b_{i\mid j}^{h+1}\ge(1-\gamma\Delta h)b_{i\mid j}^{h}$ during diffusion rollouts. The two new CT nodes, with updated trajectories $V_i.\boldsymbol{\tau}^{1:H}$ and $V_j.\boldsymbol{\tau}^{1:H}$, are added to the CT priority queue. CBS then pops the node with the fewest conflicts for further expansion. 

\section{Experimental Analysis}
We aim to explore the performance of our method through various environments to address the following three questions:

\begin{itemize}
    \item What advantages does MDOC offer over state-of-the-art (SOTA) single-robot planners?
    \item How does MDOC-CBS compare to SOTA MRMP planners in terms of success rate, trajectory smoothness, and the ability to ensure constraints?
    \item Is MDOC-CBS robust and scalable with respect to map size, planning horizon, and number of robots?
\end{itemize}

\textbf{Experimental Setup.} We evaluate our method MDOC and its MRMP version MDOC-CBS on various map types (illustrated in  Fig.~\ref{fig:2x2}): (\textbf{1}) we use the Narrow map of size $2\times2$  to evaluate MDOC in single-robot settings; 
(\textbf{2}) we use the Empty map, Conveyor map~\cite{shaoul2025multi}, Drop-Region map~\cite{shaoul2025multi}, and Random Map (with an obstacle density of 13.7\%), all of size 2$\times$2, to evaluate MDOC-CBS in multi-robot settings; 
(\textbf{3}) we use Empty Large maps of size 4$\times$4 and 6$\times$6 for multi-robot scalability evaluation. The robot radius is fixed at 0.05 for all maps.

We consider three start-goal pairs setups: (\textbf{1}) Circle Setup requires robots to symmetrically swap positions between opposite points on the perimeter; (\textbf{2}) Weave Setup requires robots to exchange positions along uniformly spaced boundary points; (\textbf{3}) Random Setup randomly generates start-goal pairs in maps. Setup demonstrations are shown 


\input{figures/tex/envs}

\textbf{Baselines.} We compare our method, \textbf{MDOC}, with SOTA trajectory optimization and planning methods: (\textbf{1}) CEM~\cite{botev2013cross}, a sampling-based trajectory optimizer using importance-weighted elite selection; (\textbf{2}) MPPI~\cite{williams2018information}, a stochastic model-predictive control method based on path integral sampling; and (\textbf{3}) RRT*~\cite{karaman2011anytime}, a sampling-based asymptotically optimal motion planner. We also compare \textbf{MDOC-CBS} with SOTA CBS-based MRMP methods: (\textbf{4}) CBS~\cite{sharon2015conflict} and its variant Enhanced CBS (ECBS)~\cite{barer2014suboptimal}, multi-agent path finding planners with grid-based A* in the discretized configuration space as its low-level planner; (\textbf{5}) KCBS~\cite{kottinger2022conflict}, a continuous-space variant of CBS with kinematic RRT* as its lower-level planner;
(\textbf{6}) MMD-CBS~\cite{shaoul2025multi}, a continuous-space variant of CBS with model-free diffusion, trained on RRT* trajectories, as its low-level planner. 
Note that we mainly focus on CBS-based MRMP planners, since MMD-CBS outperforms ``composite'' methods, which directly optimize joint constraints and objectives without CBS~\cite{shaoul2025multi}.

\input{figures/tex/single}
\input{figures/tex/traj}
\input{figures/tex/emptyconveyor}

\subsection{Experimental Results}
We conduct experiments on an NVIDIA RTX A4500 GPU with 20 GB of memory and 12 CPU cores. We run 10 trials per method and map, each with a different random seed, under identical map conditions. We use a single-integrator model in the main experiments for clarity.

\textbf{Why Model-Based Diffusion in Motion Planning?} As illustrated in the right panel of Fig.~\ref{fig:single_agent}, we showcase the sample distribution at 90\%, 50\%, and 10\% of the diffusion denoising process. We observe that the early-stage (90\%) distribution is wide, covering multiple homotopy classes and exploring diverse passages. As iterations progress (50\%), the distribution is denoised towards feasible and low-cost paths by score guidance augmented with dynamics and constraint projections, traversing the bottleneck in both maps and safely avoiding obstacles in More-Constrained Narrow (bottom row), and finally converges to a high-quality feasible trajectory at the late stage (10\%). CEM and MPPI refine the control sequence through importance weighting or elite resampling, but mainly rely on repeated stochastic resampling, which often leads to getting stuck in wrong homotopy classes or failing in highly multimodal or constrained environments. MDOC establishes a well-conditioned gradient structure over the trajectory distribution and explicitly incorporates dynamics and constraints into the denoising process, so that every reverse step is model-informed guidance rather than a blind perturbation. Together, these properties turn motion planning into a model-based diffusion process, where the dynamics model and constraints shape the score instead of relying solely on sampled rollouts. This allows each candidate trajectory to become progressively more feasible and cost-efficient within a single multi-step refinement procedure, achieving higher effectiveness under the same sampling budget compared to other optimizers that depend purely on random perturbation resampling. ``Pass\&Free''-Yield (PF-Yield) from Table~\ref{tab:pf} further verifies that MDOC achieves a 100.0\% effective sample rate, while CEM and MPPI remain significantly lower on both narrow maps.

\input{figures/tex/random}
\input{figures/tex/emptylarge}

As a tree-search-based geometric planner, RRT* explores reachable paths by randomly sampling states and incrementally expanding a search tree, but it does not directly optimize trajectory cost or enforce dynamic feasibility. To quantify its effective sampling efficiency in narrow passages, we compute PF-Yield for RRT* as the fraction of expanded nodes whose associated paths both traverse the bottleneck and remain collision-free (Pass \& Free) over all expanded nodes. Comparing PF-Yield across RRT* and MDOC therefore places a geometric planner and a trajectory optimizer on the same footing, by asking: out of all samples or rollouts under a fixed budget, how many actually produce useful, bottleneck-traversing, collision-free candidates? Under this metric, RRT* attains PF-Yield values of 66.1\% and 41.7\% on the two maps, substantially lower than MDOC’s 100\%. This indicates that, under a limited sample budget, while RRT* possesses global reachability and probabilistic completeness, it still wastes many samples in unproductive homotopy classes and struggles to efficiently converge to a feasible and smooth solution in bottlenecked environments. In contrast, MDOC, through diffusion denoising in trajectory space, achieves continuous convergence from global exploration to local refinement guided by dynamics and constraints, thereby completing path selection and trajectory optimization within a single model-based diffusion process. Therefore, we consider MDOC as a model-based diffusion planner that inherits the global exploration behavior of RRT* while also enjoying the optimization capabilities of MPPI and CEM, using the underlying dynamics model and hard constraints to steer the diffusion updates, thereby achieving higher feasibility and convergence efficiency.

\textbf{Performance for MRMP on 2$\times$2 Maps.} We first evaluate the scalability of MDOC-CBS on the Empty map and Conveyor map with the Circle and Weave setups up to 20 robots (Fig.~\ref {fig:empty_conveyor}). The experiment uses a fixed horizon $H=64$. Overall, the results show that MDOC-CBS (blue lines) outperforms baseline methods under the same map and start-goal conditions in most cases. In contrast, MMD-CBS fails beyond 10 robots in the Empty Circle setup and 12 in the Conveyor Circle setup, while KCBS’s success rate steadily degrades after 5 robots. At the same time, MDOC-CBS achieves 5$\times$–10$\times$ lower planning time than MMD-CBS at comparable CT expansion levels, since it computes the trajectory score analytically through MCSA with model-based rollouts rather than repeatedly invoking a large diffusion network.

Moreover, our MDOC-CBS consistently produces shorter and smoother trajectories than KCBS and MMD-CBS (Table~\ref{tab:path}). This improvement stems from the model-based diffusion objective with MCSA, which drives rich exploration directly in the continuous state space while regularizing dynamic coherence. KCBS, which relies on discrete RRT* primitives, tends to generate irregular, jittery motions, and MMD-CBS inherits these artifacts because its denoising objective is trained to imitate RRT*-generated demonstrations. In contrast, MDOC-CBS actively exploits open free space and promotes dynamically coordinated motion across agents (Fig.~\ref{fig:path}), yielding trajectories that are globally more symmetric and better synchronized. The difference is most pronounced in the Conveyor map, as shown in the bottom row of Fig.~\ref{fig:path}. MMD-CBS inherits the demonstration’s ``queue-through-the-corridor'' behavior: all 6 robots enter the narrow passage, causing oscillation and congestion at the bottleneck, which inflates path length and degrades geometric smoothness. 
In contrast, MDOC-CBS does not force all robots into the same homotopy. Through model-based score ascent and continuous-space exploration, only 2 robots choose the corridor, while the remaining robots naturally distribute themselves around the outer free space. This leads to significantly shorter average path lengths, fewer high-level CT expansions, and substantially smoother continuous trajectories while ensuring a safety margin. These gains arise from combining diffusion-based exploration with a model-based low-level objective within CBS, thereby improving scalability in dense settings.

We further evaluate MDOC-CBS on a Drop-Region map with four tightly packed central obstacles and a Random map with uniformly sampled start-goal pairs (Fig.~\ref{fig:drop}). In both settings, MDOC-CBS and MMD-CBS achieve similar success rates and clearly outperform KCBS, while MDOC-CBS consistently attains lower planning time at comparable CT expansion levels, indicating robustness across different map structures. However, MDOC-CBS exhibits higher variability: its success rate does not always reach 100\% in random maps, and most failures occur when a robot collides with an obstacle and no valid rollout is returned, reflecting higher Monte Carlo score ascent variance in compressed free space with tight random constraints. This suggests a nontrivial trade-off between open-space exploration and aggressive safety tightening within the CBF-constrained projection during model-based diffusion rollouts. Even so, MDOC-CBS solves instances with up to 15 robots, and we expect stability to improve with better variance control and constraint projections in future work.

\textbf{Longer-Horizon MRMP on Larger Maps.} We present a robustness study of MDOC-CBS on larger maps with longer horizons ($H=128$ for larger $4\times4$ and $H=196$ for larger $6\times6$). As shown in Fig.~\ref{fig:long_horizon}, MDOC-CBS remains robust as the problem size grows, achieving the highest success rate up to 40 robots. In contrast, MMD-CBS degrades sharply beyond 30 robots: increasing $H$ directly lengthens the denoising chain and inflates the inference cost of its model-free single-robot diffusion model, so many runs hit the time limit before CBS can complete the high-level search. MDOC-CBS, by comparison, scales much more gracefully with horizon length. Because MDOC performs model-based diffusion rollouts in parallel under known dynamics, the per-iteration cost grows sub-linearly with $H$ compared to model-free denoising, and even when CBS introduces additional constraints on larger maps, the computation can be amortized across robots and time steps. This makes MDOC-CBS substantially more robust for long-horizon planning and explains its superior performance on the $6\times6$ map.

\section{Conclusion} 
\label{sec:conclusion}
In this work, we propose MDOC, a CBF-constrained model-based diffusion planner that satisfies dynamical feasibility and safety constraints simultaneously. We also extend the planner to the multi-robot settings to solve MRMP. 
Results showcase the superior performance of our method in various environmental setups. 
Future directions include optimizing Monte Carlo variance and constraint projection efficiency, adapting the planner to online tasks with receding-horizon methods, and exploring richer dynamics models and multi-robot variants such as Prioritized Planning.

\section*{Acknowledgments}
This work was partially supported by the National Science Foundation under Grants \#$2328671$ and \#$2441629$. 

\bibliographystyle{plainnat}
\bibliography{references}

\clearpage
\input{appendix}

\end{document}

%% file: algos/algorithm1.tex
\begin{algorithm}[tb!]
\caption{Model-Based Diffusion Optimal Control}
\label{alg:mdoc}
\small
\begin{algorithmic}[1]
\STATE \textbf{Input:} Single robot start state $s^1$, goal state $s^H$, dynamics $f(\cdot)$, safety set $\mathcal{C}$, and finite time horizon $H$.
\STATE \textbf{Output:} Clean denoised trajectory ${}^{1}{\tau}^{1:H}$.
\STATE Sample an initial noisy trajectory
${}^{K}{\tau}^{1:H} \sim \mathcal{N}({0}, I)$.
\FOR {$k = K$ to $1$}
    \FOR {$m = 1$ to $M$}
    \STATE 
    Sample a candidate trajectory:
    \\$  \tilde{\tau}_m^{1:H} \sim
      \mathcal{N}\!\left(
        \frac{{}^{k}{\tau}^{1:H}}{\sqrt{\bar{\alpha}_k}},
        \left(\frac{1}{\bar{\alpha}_k} - 1\right) \mathbf{I}
      \right).$
    \STATE Recover a dynamically feasible and safe trajectory: \\
      $\quad\tau_m^{1:H} \leftarrow
      \textsc{RolloutWithCBF}(\tilde{\tau}_m^{1:H}, s^1, s^H, f(\cdot), \mathcal{C}).$
    \ENDFOR 
    \STATE Calculate Monte Carlo mean $\bar{\tau}^{1:H}$ via Eq.~\eqref{eq:mc_avg_prelim2}.
    \STATE Estimate score $\hat{\mathbf{S}}_k$ via Eq.~\eqref{eq:mc_score_prelim}.
    \STATE Compute Monte Carlo score ascent via \eqref{eq:reverse_update_prelim}: \\
        \(
        \begin{aligned}
        &\quad{}^{k-1}{\tau}^{1:H} \leftarrow \frac{1}{\sqrt{\alpha_k}} \left( 
        {}^{k}{\tau}^{1:H} + (1 - \bar{\alpha}_k) 
        \hat{\mathbf{S}}_k \right).
        \end{aligned}
        \) \label{alg:reverse_to}
\ENDFOR  
\STATE \textbf{return ${}^{1}{\tau}^{1:H}$}. 
\end{algorithmic}
\end{algorithm}

%% file: algos/algorithm2.tex

\begin{algorithm}[tb!]
\caption{MDOC-CBS}
\label{alg:mdoc_cbs}
\small
\begin{algorithmic}[1]
\STATE \textbf{Input:} $N$ robots with their start states $\boldsymbol{s}^1$, goal states $\boldsymbol{s}^H$, dynamics $f(\cdot)$, safety set $\mathcal C_{\mathrm{obs}}$ that encode obstacle-avoidance constraints, and finite time horizon $H$. 
\STATE \textbf{Output:} Trajectories $\mathbf{\tau}^{1:H}$.
\STATE Create CT root node $V_{\text{root}}$.
\STATE CT $\gets \{V_{\text{root}}\}$.
\FOR{$i = 1$ to $N$}
    \STATE $V_{\text{root}}.\mathcal{C}_{i} \gets \mathcal C_{\mathrm{obs}}$.
    \STATE $V_{\text{root}}.\tau_i^{1:H} \gets \textbf{MDOC}(s^{1}_i, s^{H}_i, f(\cdot), V_{\text{root}}.\mathcal{C}_i, H)$.
    \IF{$V_{\text{root}}.\tau_i^{1:H}$ is not feasible}
        \STATE \textbf{return} ``no solution''.
    \ENDIF
\ENDFOR
\WHILE{CT $\neq \emptyset$}
    \STATE $V \gets \arg\min\limits_{V' \in \text{CT}} \textsc{numConflicts}(V'.\boldsymbol{\tau}^{1:H})$.
    \STATE Remove $V$ from CT.
    \IF{$V.\boldsymbol{\tau}^{1:H}$ is conflict-free}
        \STATE \textbf{return} $V.\boldsymbol{\tau}^{1:H}$.
    \ENDIF
    \STATE $s_i^h, s_j^h, h, i, j \gets \textsc{getOneConflictPair}(V.\boldsymbol{\tau}^{1:H})$.
    \FOR{$z \in \{i, j\}$}
        \STATE $V_z \gets V.\text{copy}()$.
        \STATE $V_z.\mathcal{C}_z \gets V_z.\mathcal{C}_z \cup \{\langle z, \mathrm{Sphere}(s_{i+j-z}^h), h\rangle\}$.
        \STATE $V_z.\mathbf{\tau}_z^{1:H} \gets \textbf{MDOC}(s^{1}_z, s^{H}_z, f(\cdot),  V_z.\mathcal{C}_z, H)$.
        \IF{$V_z.\mathbf{\tau}_z^{1:H}$ is feasible}
            \STATE Add $V_z$ to CT.
        \ENDIF
    \ENDFOR
\ENDWHILE
\STATE \textbf{return} ``no solution''.
\end{algorithmic}
\end{algorithm}

%% file: figures/tex/envs.tex
\begin{figure}[tb!]
  \centering
  \begin{subfigure}[b]{0.15\textwidth}
    \centering
    \includegraphics[width=\linewidth]{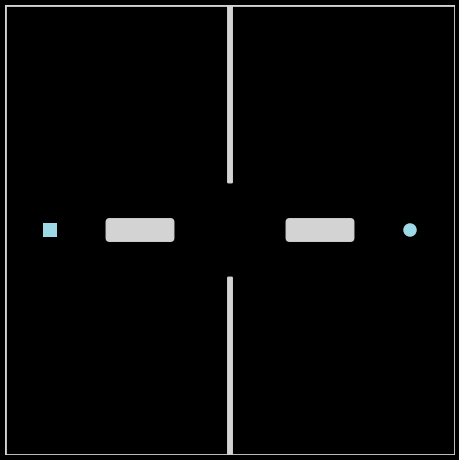}
  \end{subfigure}\hfill
  \begin{subfigure}[b]{0.15\textwidth}
    \centering
    \includegraphics[width=\linewidth]{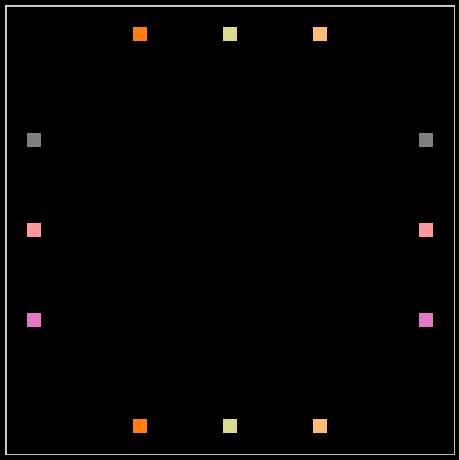}
  \end{subfigure}\hfill
  \begin{subfigure}[b]{0.15\textwidth}
    \centering
    \includegraphics[width=\linewidth]{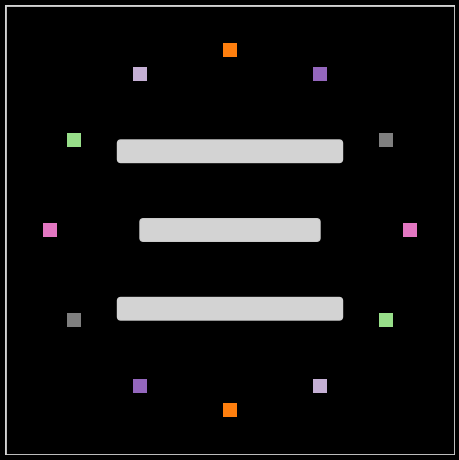}
  \end{subfigure} 
  \par\vspace{10pt}
  \begin{subfigure}[b]{0.15\textwidth}
    \centering
    \includegraphics[width=\linewidth]{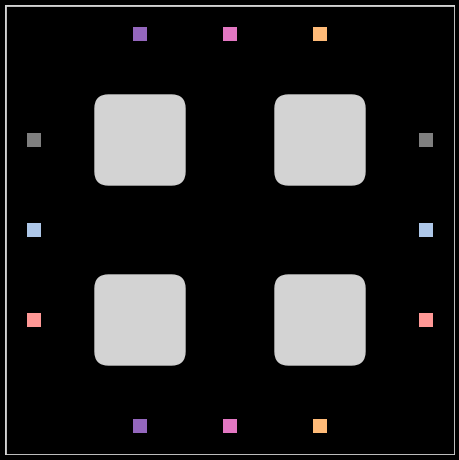}
  \end{subfigure}\hfill
  \begin{subfigure}[b]{0.15\textwidth}
    \centering
    \includegraphics[width=\linewidth]{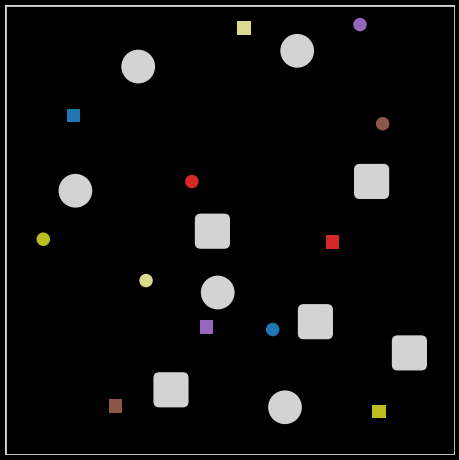}
  \end{subfigure}\hfill
  \begin{subfigure}[b]{0.15\textwidth}
    \centering
    \includegraphics[width=\linewidth]{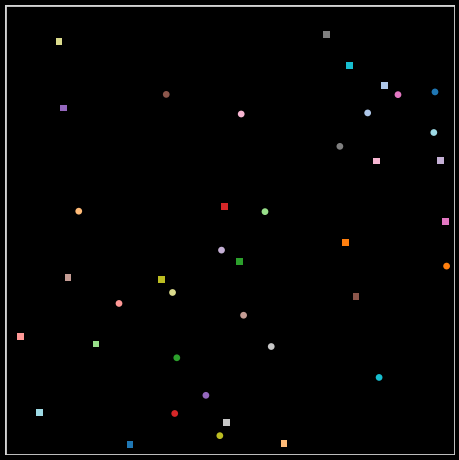}
  \end{subfigure}\hfill
  \caption{Illustrations of Narrow map, Empty map, Conveyor map (Top Row), Drop-Region map, Random map, and Empty Large map (Below) used in MRMP experiments where gray objects represent obstacles. Colorful squares and spheres indicate the start and goal of robots. For the Empty and Conveyor maps, we consider a dual swap (Circle Setup and Weave Setup) between two robots positioned on squares of the same color.}
  \label{fig:2x2}
\end{figure}



%% file: figures/tex/single.tex
\begin{figure}[tb!]
  \centering
  \caption{Results of MDOC and baseline methods on Narrow map (top row) and More-Constrained Narrow map (bottom row). The left panels plot, for each planner, the collision-free trajectory with the highest success probability. Right panels depict MDOC rollouts at 90\%, 50\%, and 10\% of the diffusion denoising process, showing how the distribution converges toward successful passages.}
  \label{fig:single_agent}
  \begin{subfigure}[b]{0.19\textwidth}
    \centering
    \includegraphics[width=\linewidth]{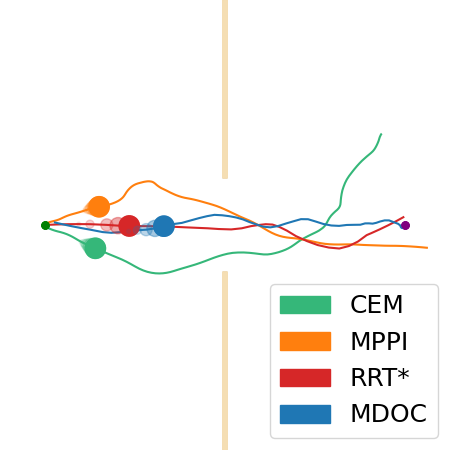}
    \label{fig:2}
  \end{subfigure}\hfill
  \begin{subfigure}[b]{0.25\textwidth}
    \centering
    \includegraphics[width=\linewidth]{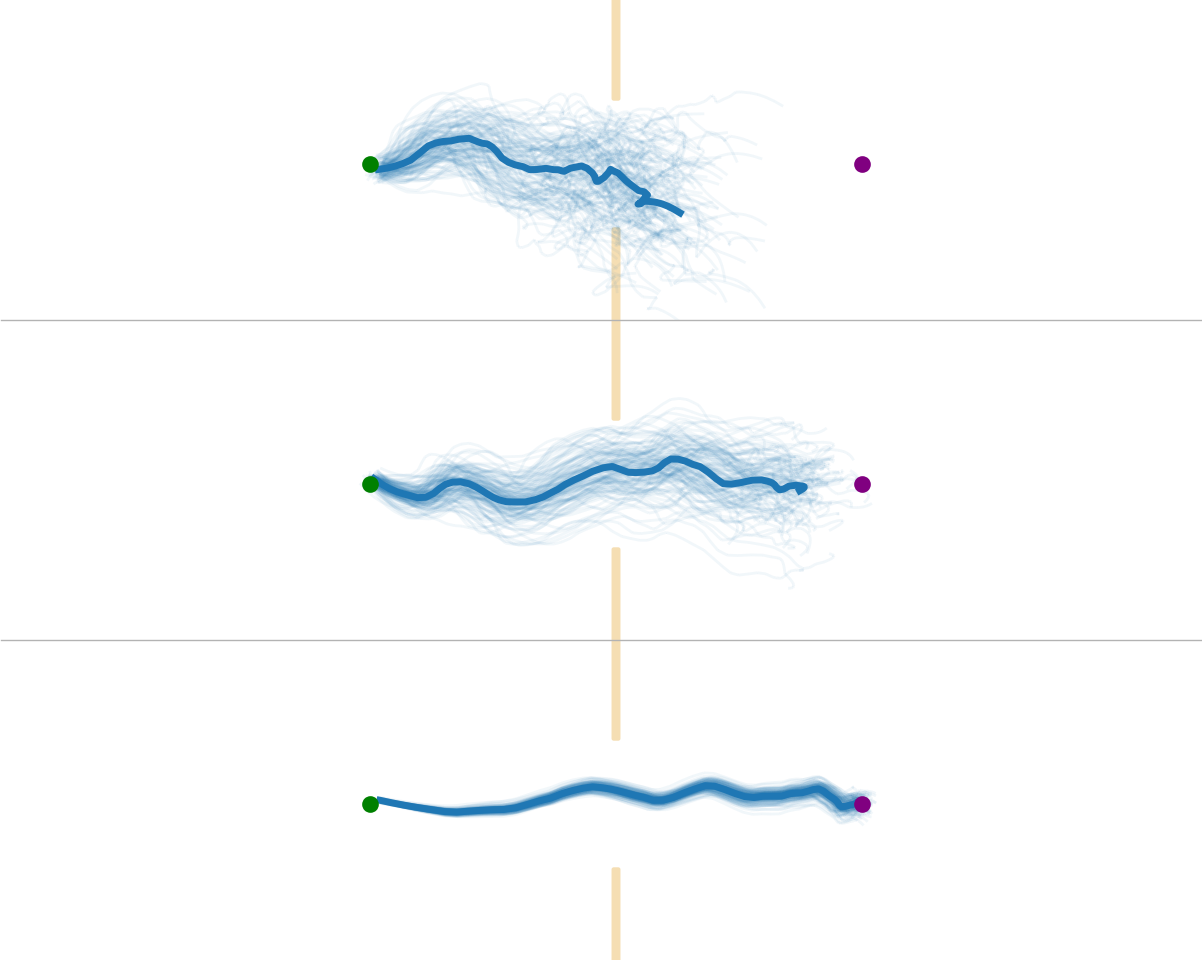}
    \label{fig:3}
  \end{subfigure}\hfill
  \begin{subfigure}[b]{0.19\textwidth}
    \centering
    \includegraphics[width=\linewidth]{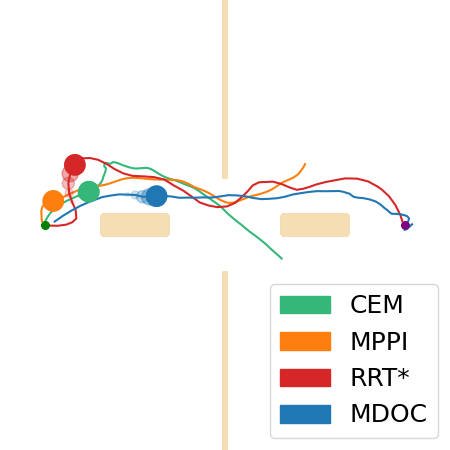}
    \label{fig:4}
  \end{subfigure}\hfill
  \begin{subfigure}[b]{0.25\textwidth}
    \centering
    \includegraphics[width=\linewidth]{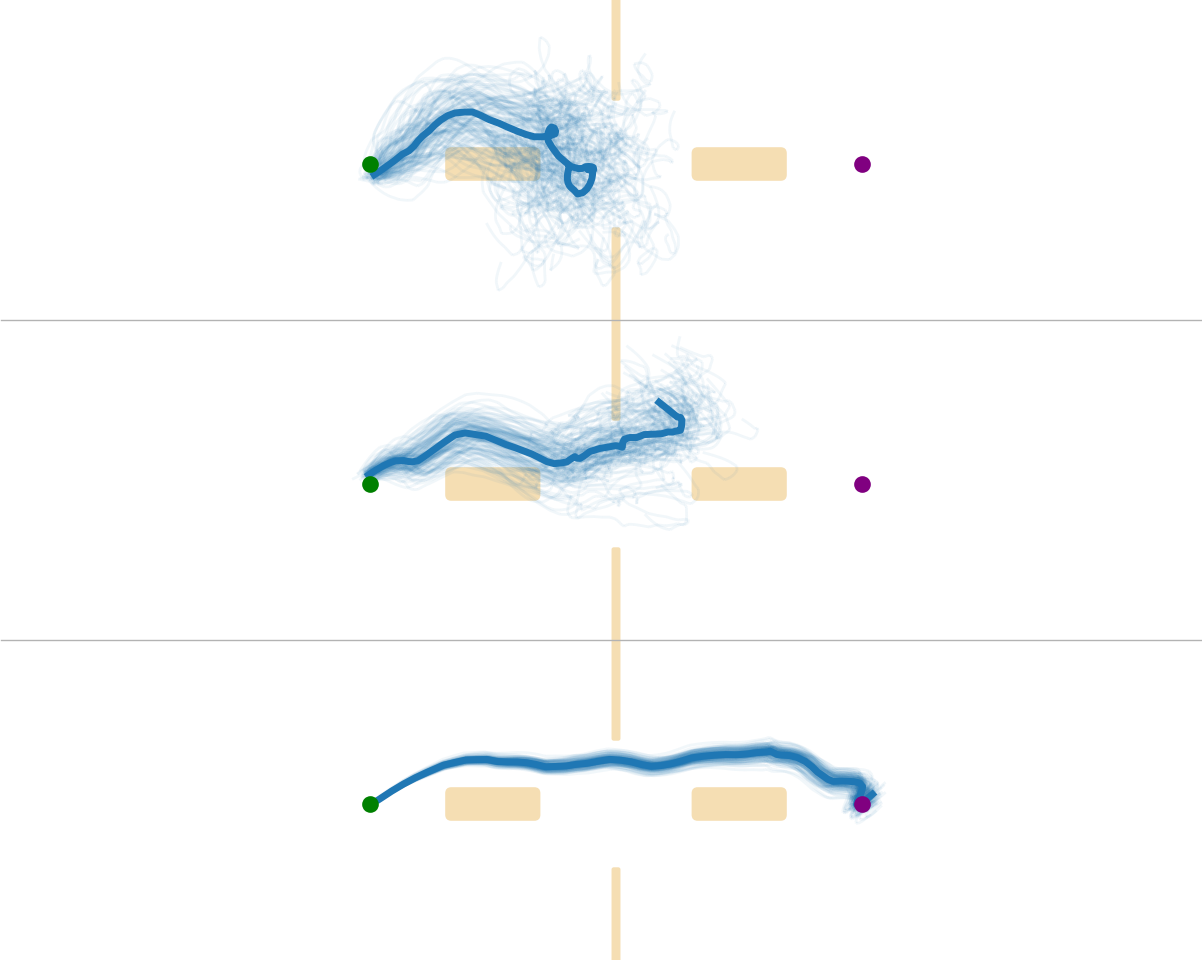}
    \label{fig:5}
  \end{subfigure}
  \captionsetup{type=table}
  \begin{tabular}{c|c|c|c|c}
  \toprule
    & CEM & MPPI & RRT* & MDOC \\
  \midrule
  Top  & 2.1 $\pm$ 0.2 & 3.2 $\pm$ 0.3 & 66.1 $\pm$ 6.0 & \textbf{100.0} \\
  Bottom  & 0.2 $\pm$ 0.1 & 0.4 $\pm$ 0.1  & 41.7 $\pm$ 3.4  & \textbf{100.0} \\
  \bottomrule
  \end{tabular}
  \caption{``Pass\&Free''-Yield (\%) ($\uparrow$) of MDOC and baseline methods, which is defined as the percentage of candidate trajectories that both traverse the bottleneck (Pass) and remain collision-free (Free) under a fixed proposal sample size $M=128$ per planning step.}
  \label{tab:pf}
\end{figure}

%% file: figures/tex/traj.tex
\begin{figure}[tb!]
  \centering
  \caption{Results on Empty map (Top Row) and Conveyor map (Below) with 6 robots on Circle setup. Columns display 3 algorithms: KCBS (Left), MMD-CBS (Mid), and our MDOC-CBS (Right). We show trajectories without post-smoothness.}
  \label{fig:path}
  \begin{subfigure}[b]{0.15\textwidth}
    \centering
    \includegraphics[width=\linewidth]{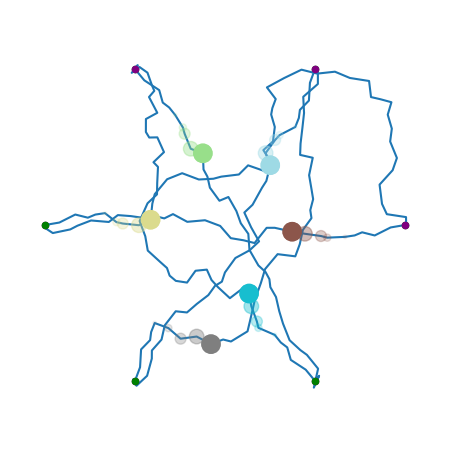}
  \end{subfigure}\hfill
  \begin{subfigure}[b]{0.15\textwidth}
    \centering
    \includegraphics[width=\linewidth]{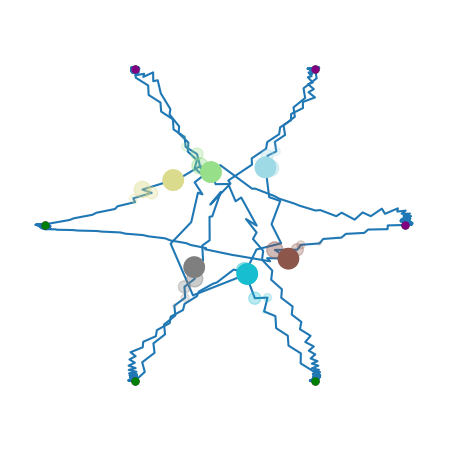}
  \end{subfigure}\hfill
  \begin{subfigure}[b]{0.15\textwidth}
    \centering
    \includegraphics[width=\linewidth]{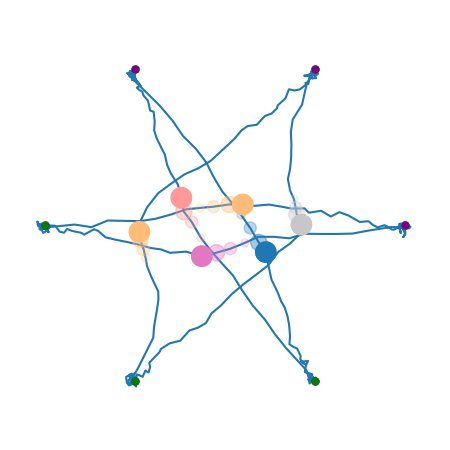}
  \end{subfigure} \hfill
  \begin{subfigure}[b]{0.15\textwidth}
    \centering
    \includegraphics[width=\linewidth]{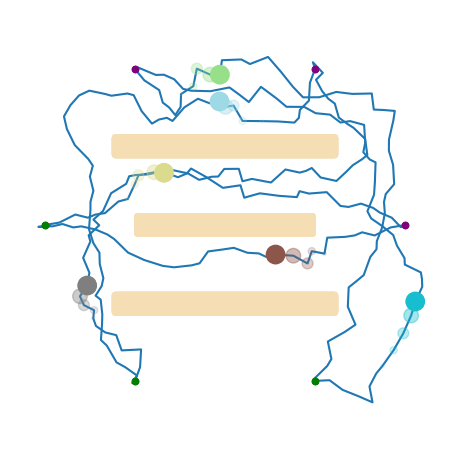}
  \end{subfigure}\hfill
  \begin{subfigure}[b]{0.15\textwidth}
    \centering
    \includegraphics[width=\linewidth]{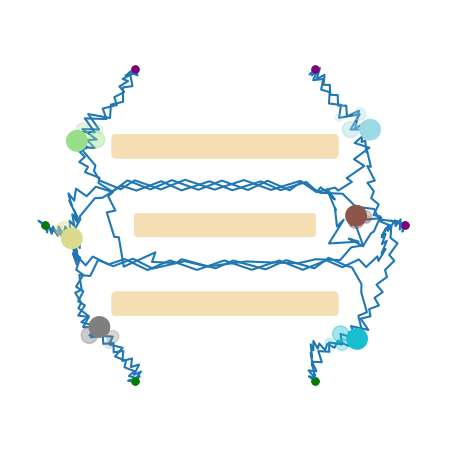}
  \end{subfigure}\hfill
  \begin{subfigure}[b]{0.15\textwidth}
    \centering
    \includegraphics[width=\linewidth]{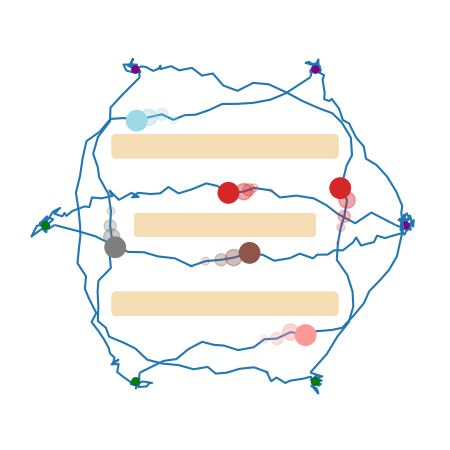}
  \end{subfigure}\hfill
  \captionsetup{type=table}
  \begin{tabular}{c|c|c|r}
    \toprule
    Map & Method & {\textbf{P}$\downarrow$} & {\textbf{G }(1e-4)$\downarrow$} \\
    \midrule
    \multirow{3}{*}{Empty} 
      & KCBS      & 2.2 $\pm$ 0.07 & 20.5 $\pm$ 2.75 \\
      & MMD-CBS   & 2.9 $\pm$ 0.06 & 60.6 $\pm$ 5.33  \\
      & MDOC-CBS  & \textbf{1.8 $\pm$ 0.03} & \textbf{6.7 $\pm$ 0.05}  \\
    \midrule
    \multirow{3}{*}{Conveyor} 
      & KCBS      & \textbf{2.7 $\pm$ 0.10} & 20.7 $\pm$ 2.12 \\
      & MMD-CBS   & 3.9 $\pm$ 0.11 & 85.0 $\pm$ 5.45 \\
      & MDOC-CBS  & 2.8  $\pm$ 0.04 & \textbf{12.9 $\pm$ 1.40} \\
    \bottomrule
    \end{tabular}
  \caption{Comparison of methods in terms of average path length \textbf{P} and geometric (Laplacian) smoothness \textbf{G} over success trails of all agent numbers, evaluated on the Empty map and Conveyor map in the Circle setup.}
  \label{tab:path}
\end{figure}

%% file: figures/tex/emptyconveyor.tex
\begin{figure*}[t]
  \centering

  \newcommand{\LeftW}{0.20\textwidth}
  \newcommand{\RightW}{0.79\textwidth}
  \newcommand{\CellW}{0.33\linewidth}
  \newcommand{\CellP}{0.48\textwidth}
  \newcommand{\Legend}{1\textwidth}

  \begin{minipage}[t]{\LeftW}
    \vspace{0pt}%

    \begin{minipage}[t]{\CellP}
      \vspace{0pt}\includegraphics[width=\linewidth]{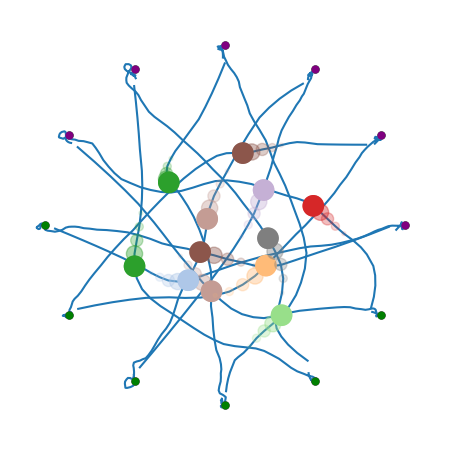}
    \end{minipage}\hfill
    \begin{minipage}[t]{\CellP}
      \vspace{0pt}\includegraphics[width=\linewidth]{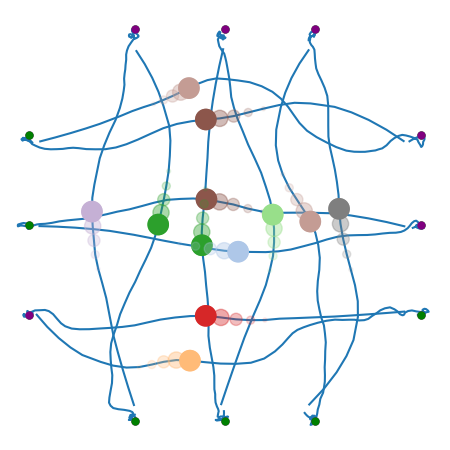}
    \end{minipage}\hfill\vspace{10pt}%

    \begin{minipage}[t]{\CellP}
      \vspace{0pt}\includegraphics[width=\linewidth]{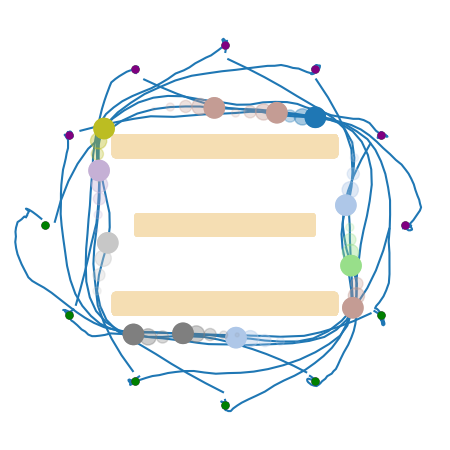}
      \centering
      \textbf{Circle Setup}
    \end{minipage}\hfill
    \begin{minipage}[t]{\CellP}
      \vspace{0pt}\includegraphics[width=\linewidth]{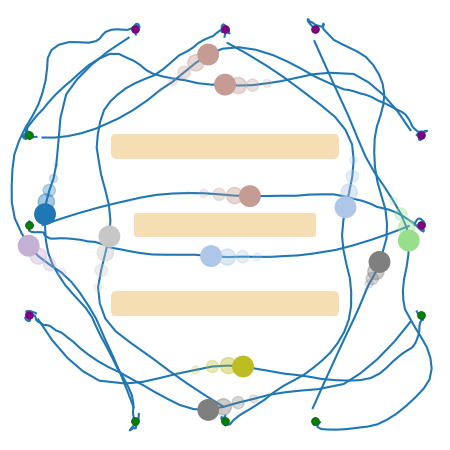}
      \centering
      \textbf{Weave Setup}
    \end{minipage}\hfill

    \begin{minipage}[t]{\Legend}
      \vspace{10pt}\includegraphics[width=\linewidth]{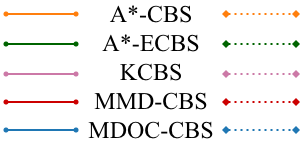} 
    \end{minipage}\hfill
    
  \end{minipage}%
  \hfill
  \begin{minipage}[t]{\RightW}
    \vspace{0pt}%

    \begin{minipage}[t]{\CellW}
      \vspace{0pt}\includegraphics[width=\linewidth]{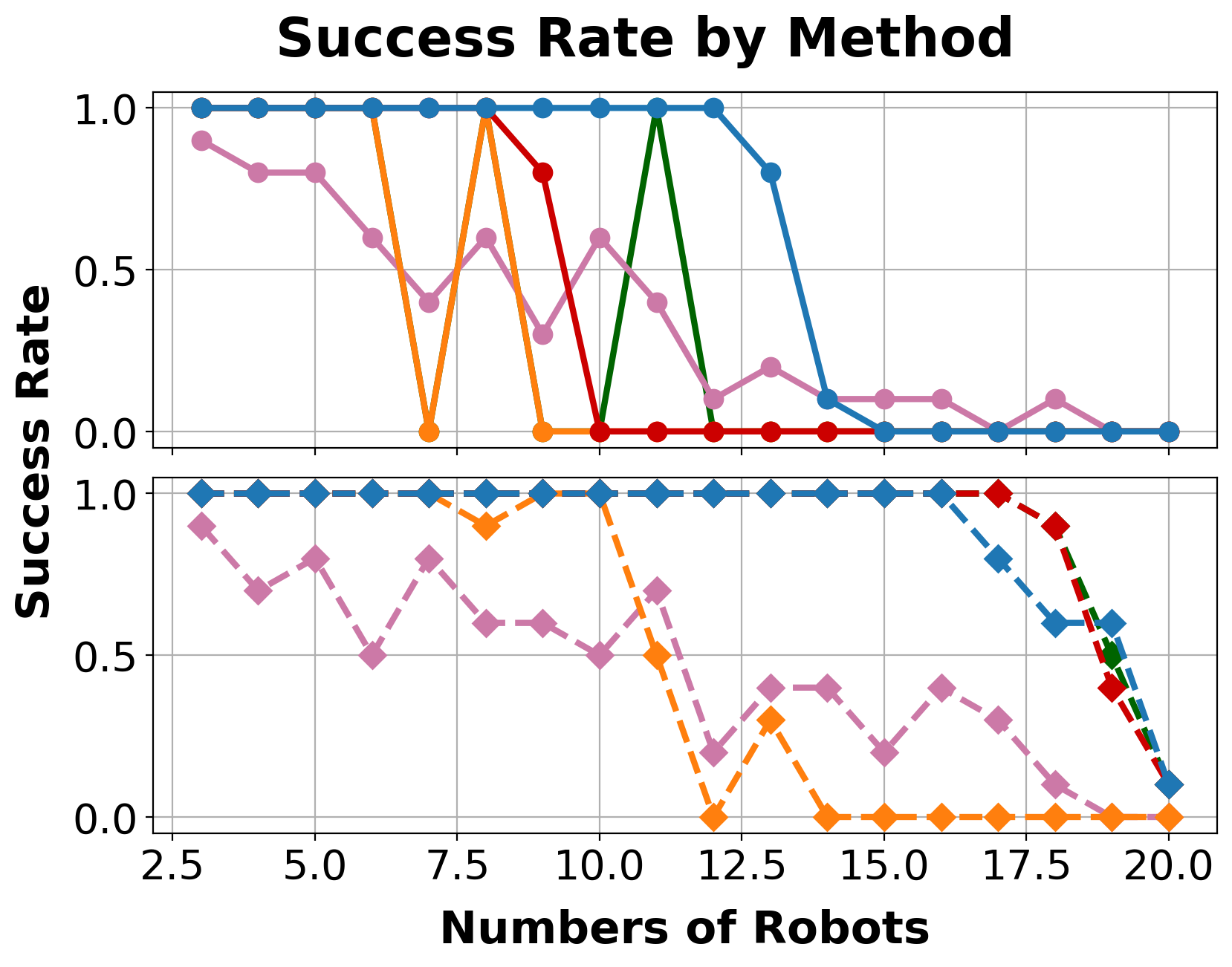}
    \end{minipage}\hfill
    \begin{minipage}[t]{\CellW}
      \vspace{0pt}\includegraphics[width=\linewidth]{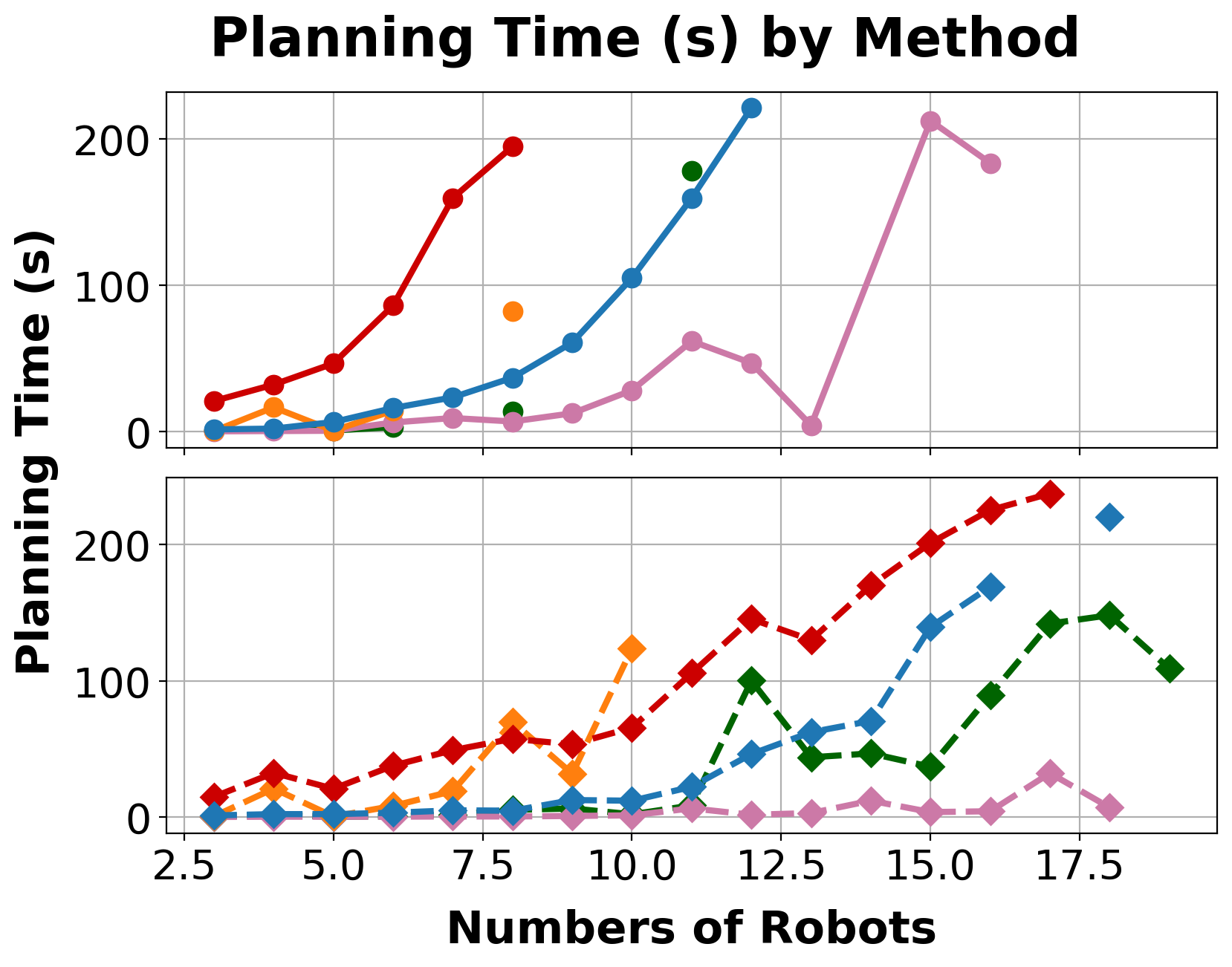}
    \end{minipage}\hfill
    \begin{minipage}[t]{\CellW}
      \vspace{0pt}\includegraphics[width=\linewidth]{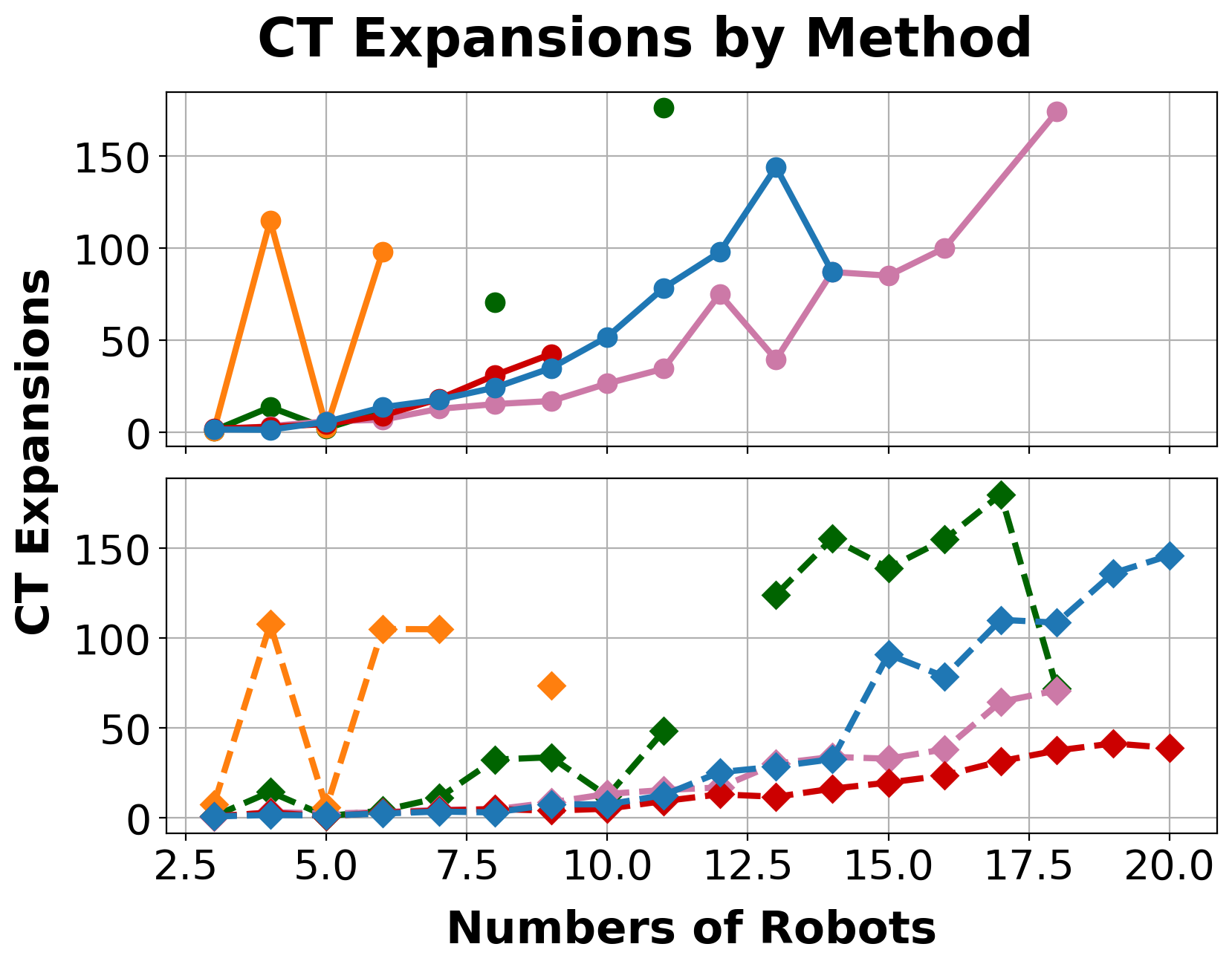}
    \end{minipage}

    \vspace{0.6em}

    \begin{minipage}[t]{\CellW}
      \vspace{0pt}\includegraphics[width=\linewidth]{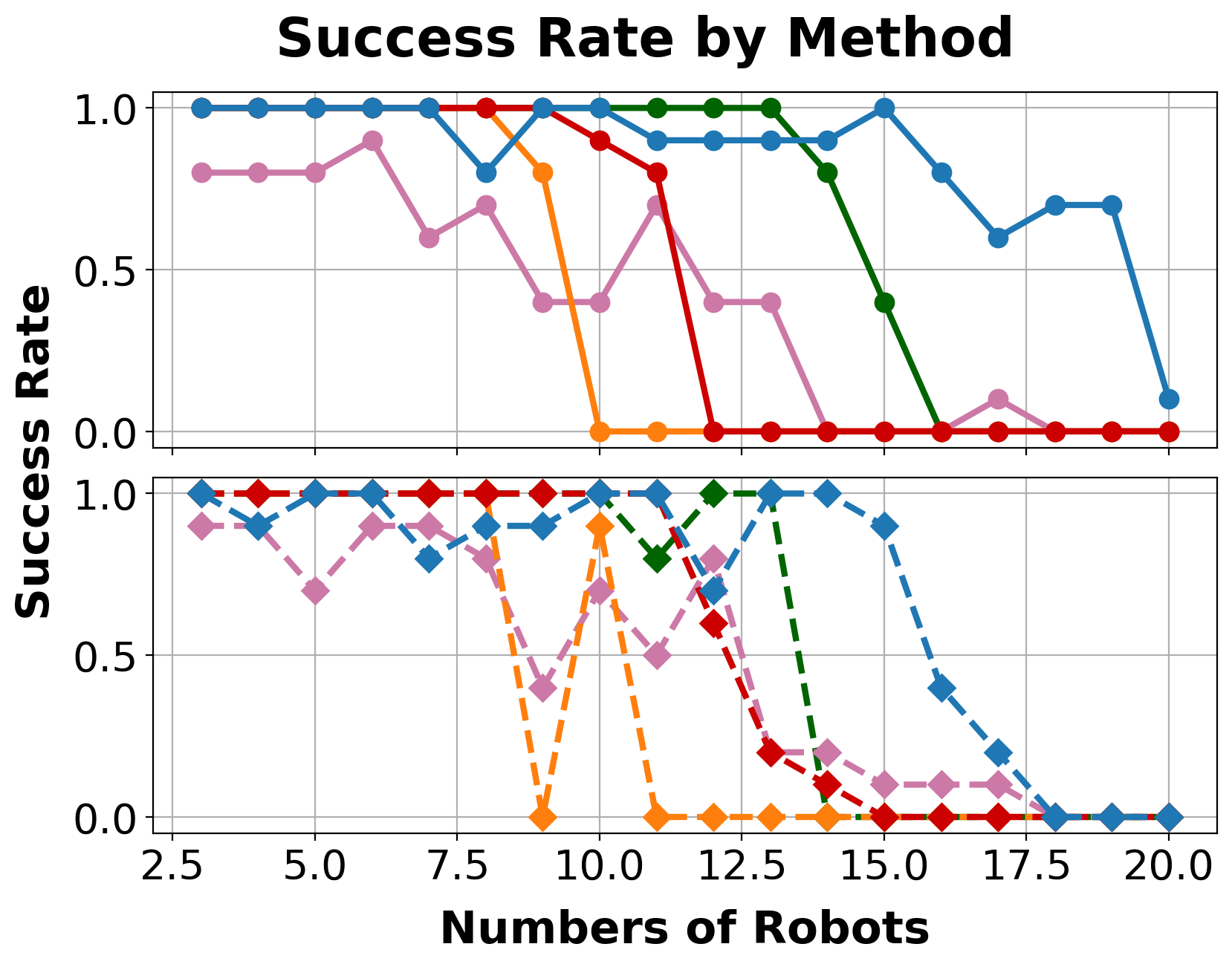}
    \end{minipage}\hfill
    \begin{minipage}[t]{\CellW}
      \vspace{0pt}\includegraphics[width=\linewidth]{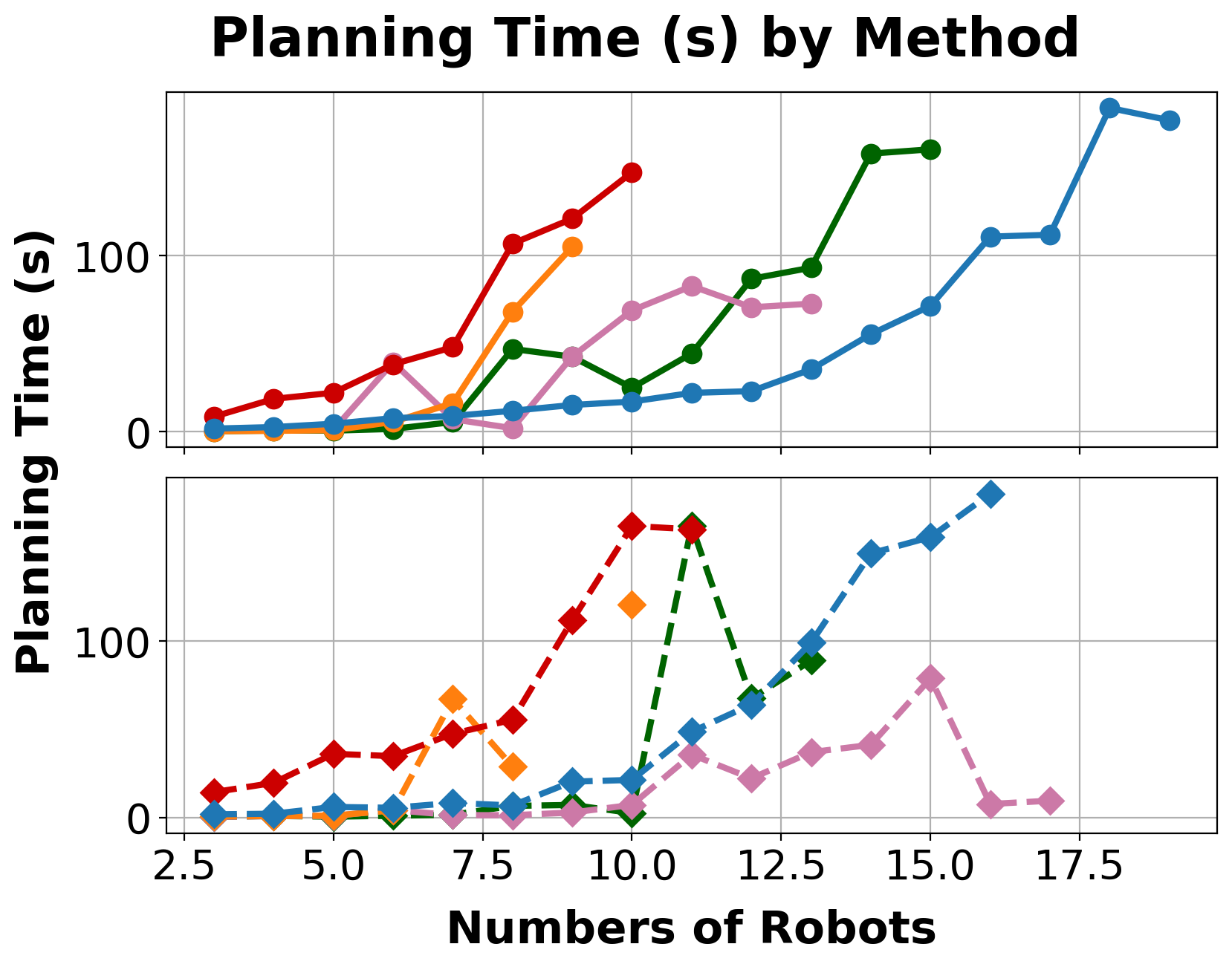}
    \end{minipage}\hfill
    \begin{minipage}[t]{\CellW}
      \vspace{0pt}\includegraphics[width=\linewidth]{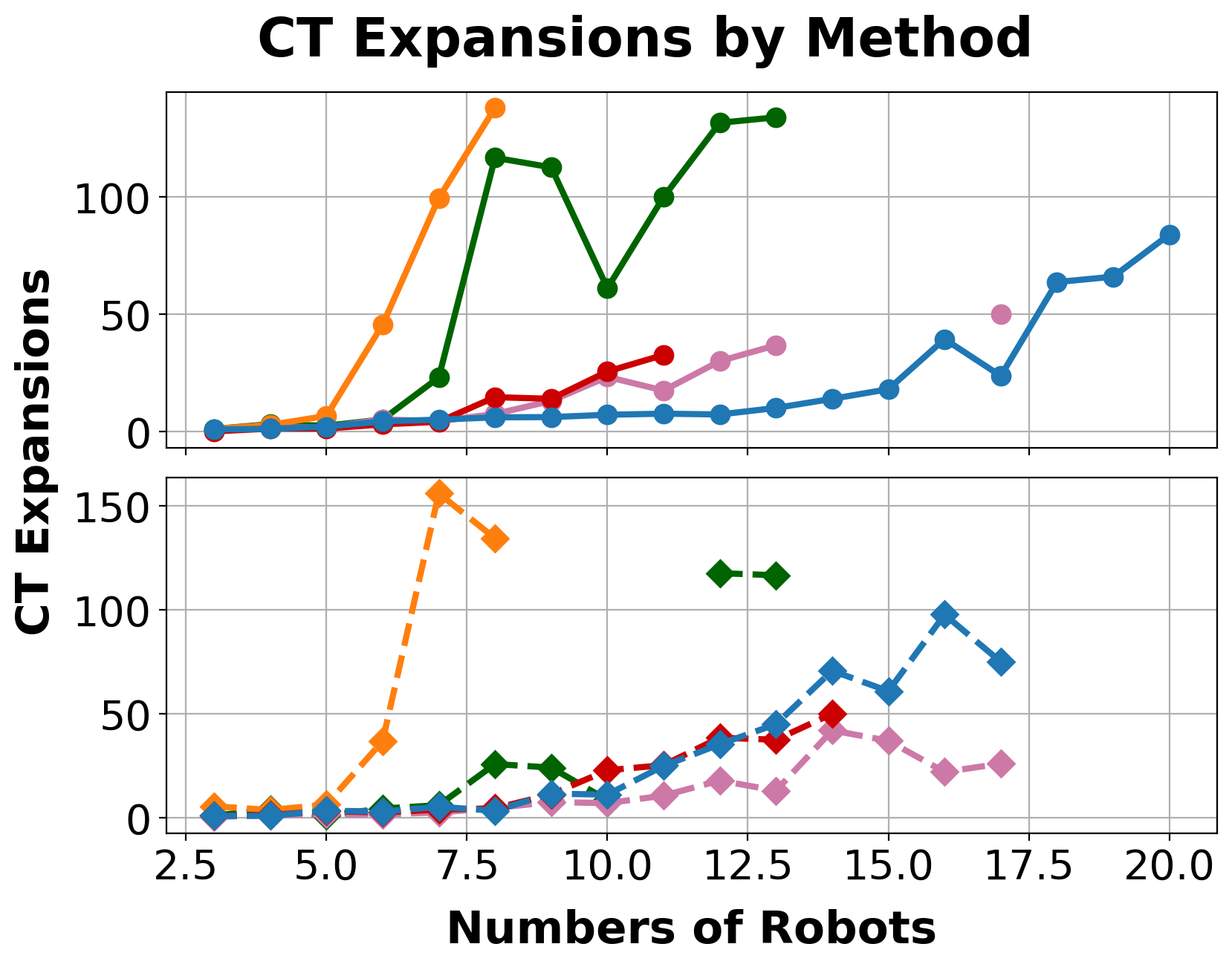}
    \end{minipage}
  \end{minipage}

  \caption{Performance of MRMP methods on Empty map (top row) and Conveyor map (bottom row) with a 500-second time limit. For each subfigure, the top plot shows the Circle Setup (circular markers with solid lines), and the bottom plot shows the Weave Setup (diamond markers with dashed lines). The planning time and CT expansions are averaged on successful trials, and outliers are included in the averages but omitted beyond the 95th percentile for better visualization, especially in CT expansions.}
  \label{fig:empty_conveyor}
\end{figure*}

%% file: figures/tex/random.tex
\begin{figure*}[t]
  \centering

  \newcommand{\LeftW}{0.16\textwidth}
  \newcommand{\RightW}{0.80\textwidth}
  \newcommand{\CellW}{0.32\linewidth}

  \begin{minipage}[t]{\LeftW}
    \vspace{0pt}%
    \centering
    \includegraphics[width=0.7\linewidth]{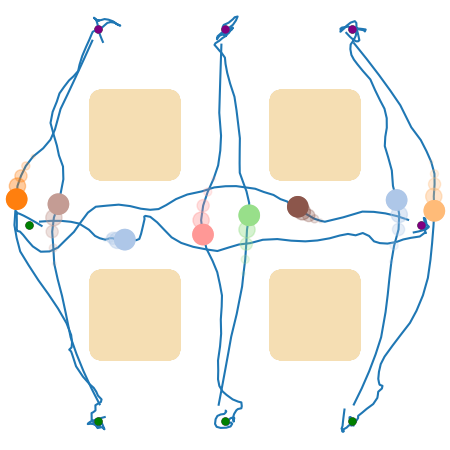}\\
    \textbf{Drop-Region Map}\par\vspace{5pt}%
    \includegraphics[width=0.7\linewidth]{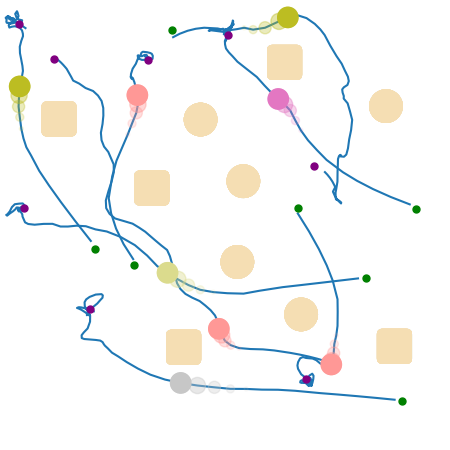}\\
    \textbf{Random Map}\par\vspace{5pt}
    \includegraphics[width=0.9\linewidth]{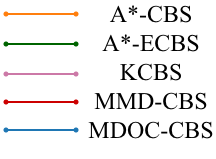} \\
  \end{minipage}%
  \hfill
  \begin{minipage}[t]{\RightW}
    \vspace{0pt}%

    \begin{minipage}[t]{\CellW}
      \vspace{0pt}\includegraphics[width=\linewidth]{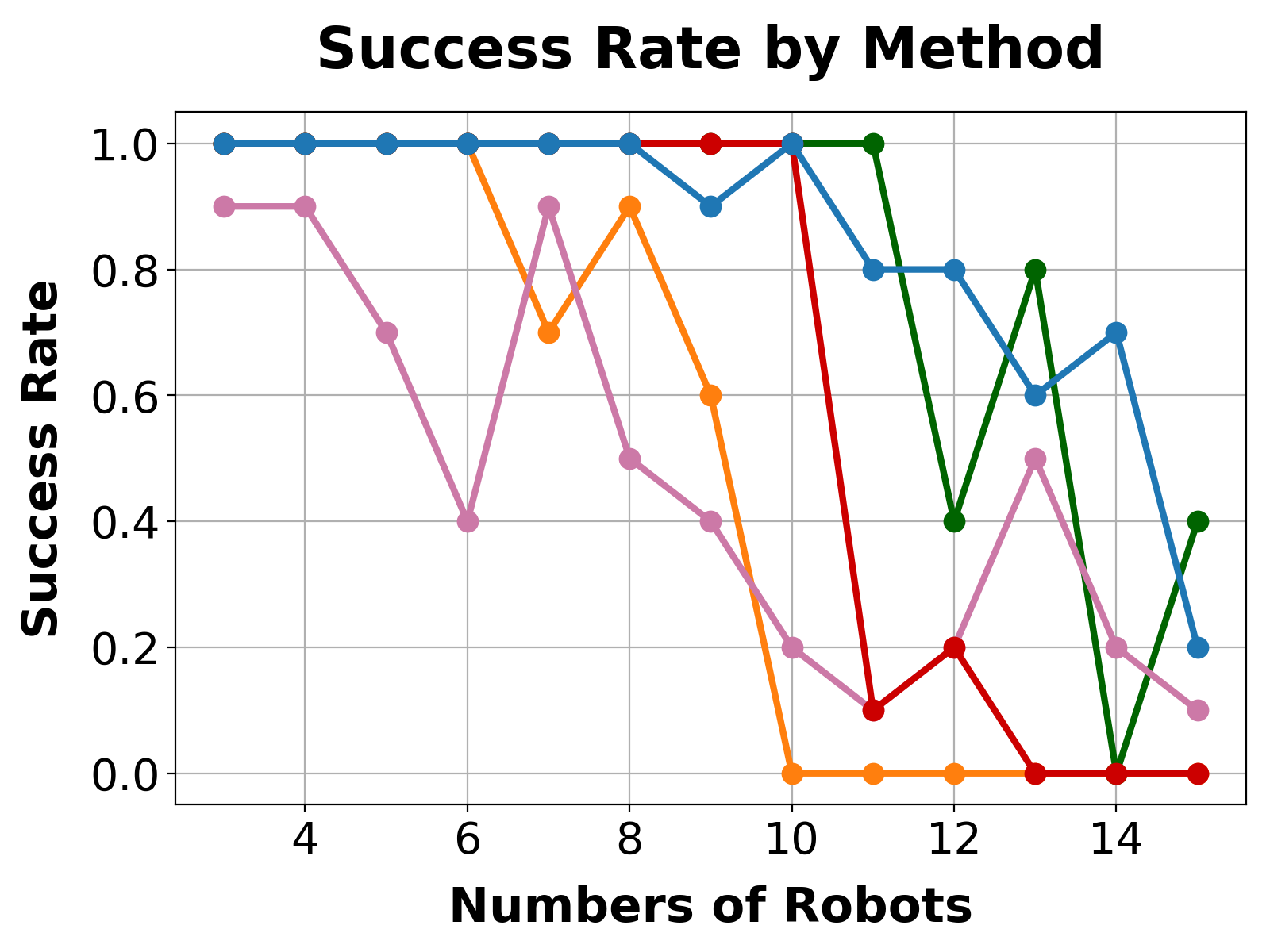}
    \end{minipage}\hfill
    \begin{minipage}[t]{\CellW}
      \vspace{0pt}\includegraphics[width=\linewidth]{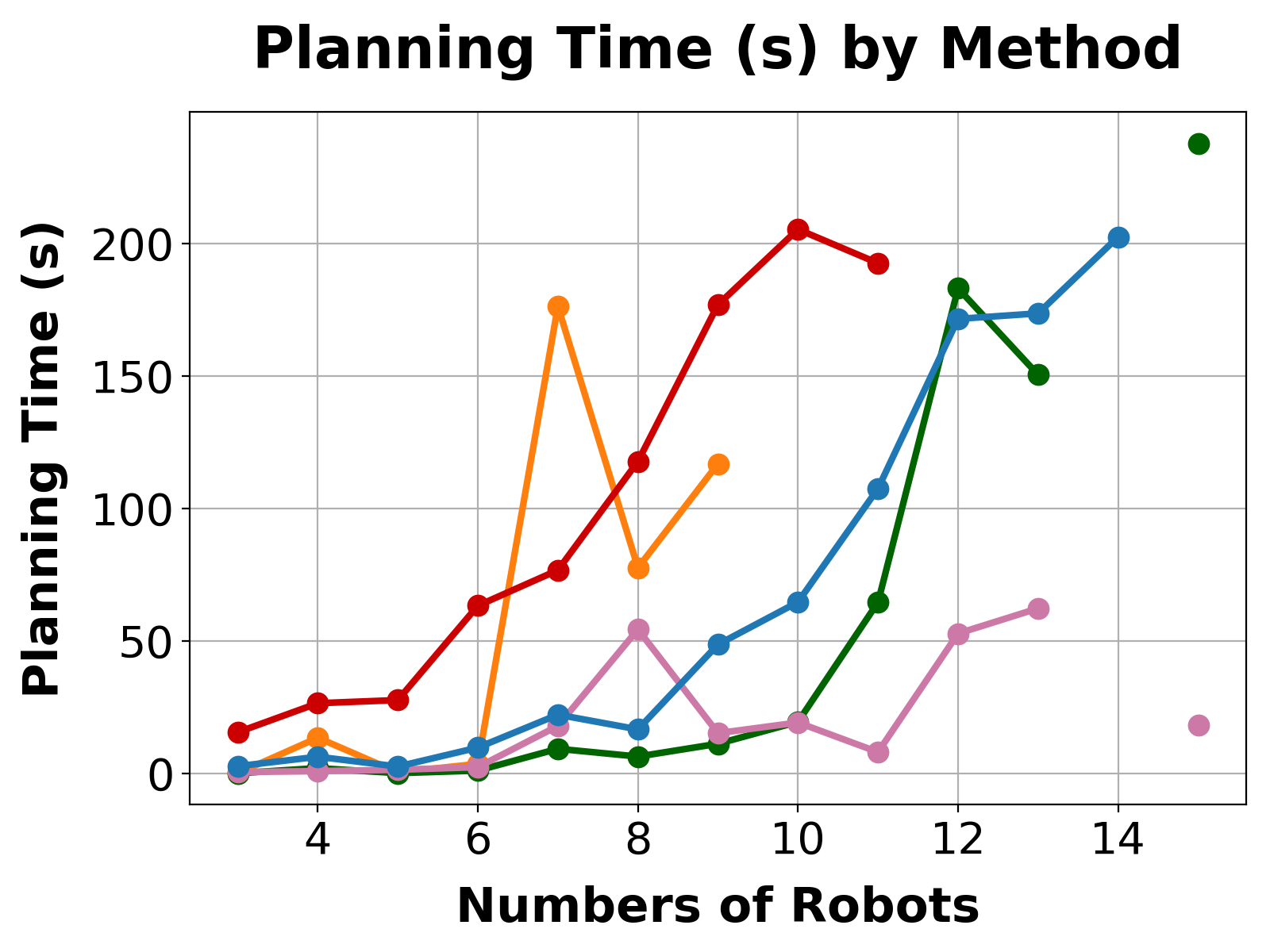}
    \end{minipage}\hfill
    \begin{minipage}[t]{\CellW}
      \vspace{0pt}\includegraphics[width=\linewidth]{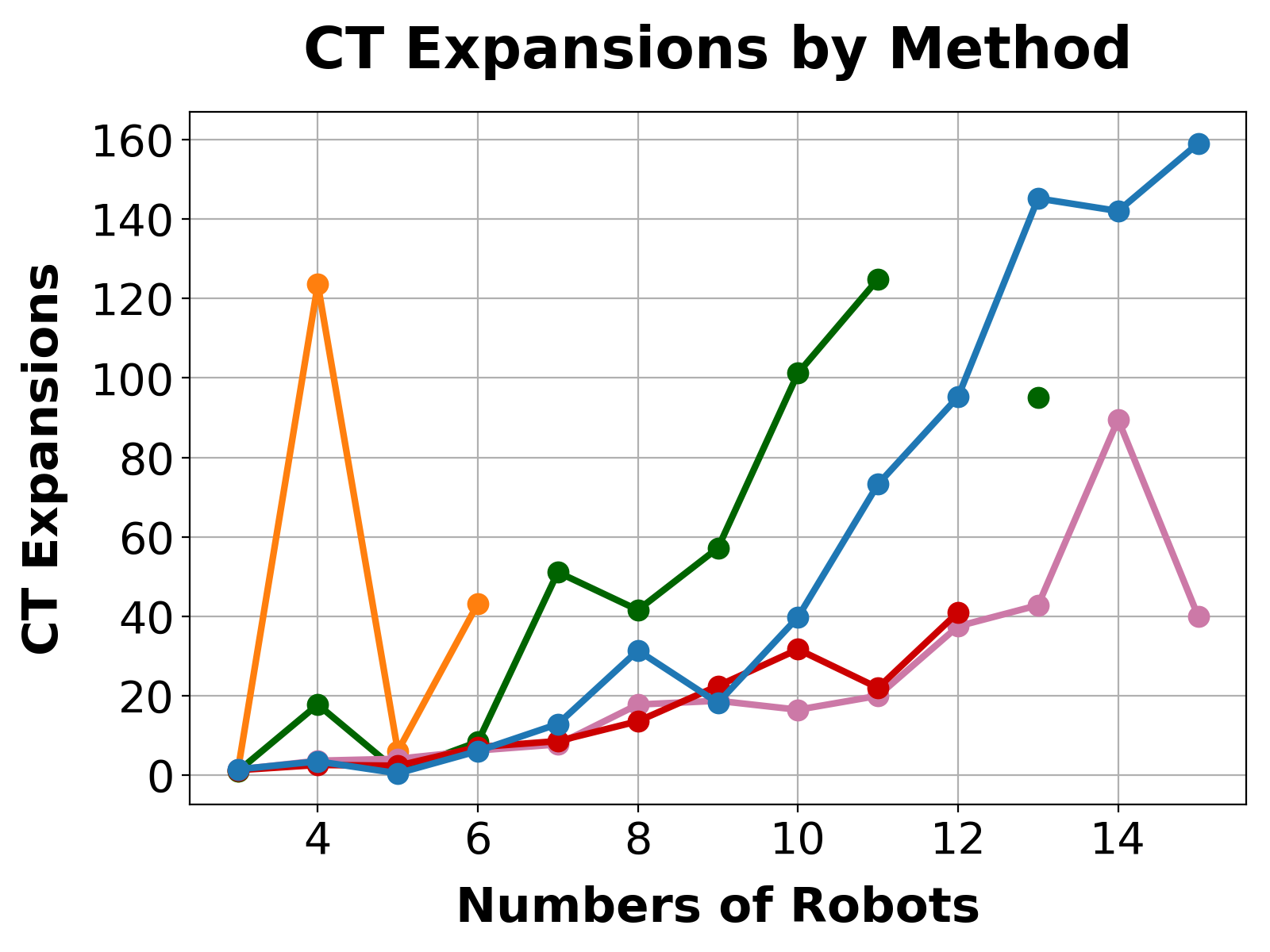}
    \end{minipage}

    \vspace{0.6em}

    \begin{minipage}[t]{\CellW}
      \vspace{0pt}\includegraphics[width=\linewidth]{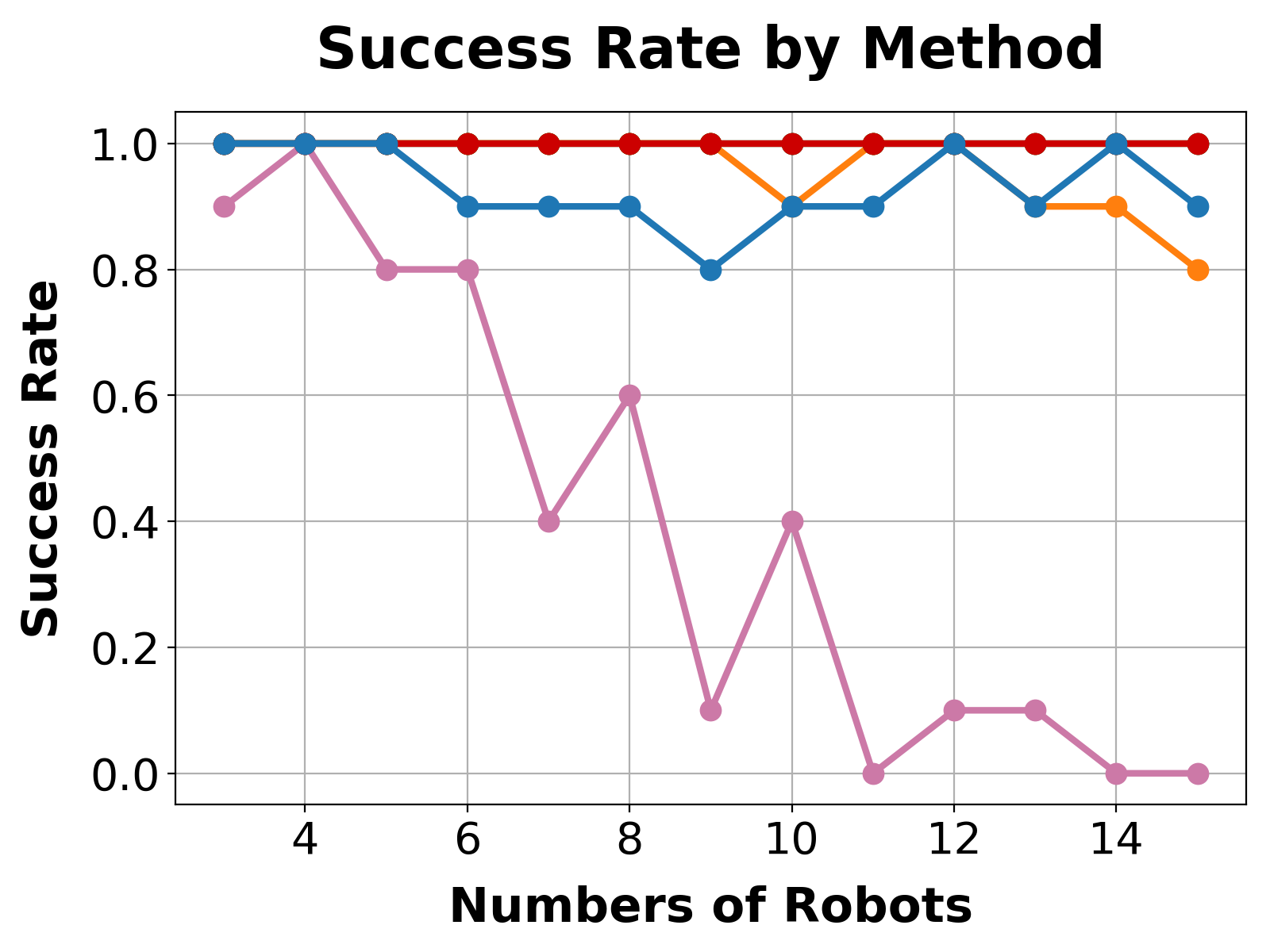}
    \end{minipage}\hfill
    \begin{minipage}[t]{\CellW}
      \vspace{0pt}\includegraphics[width=\linewidth]{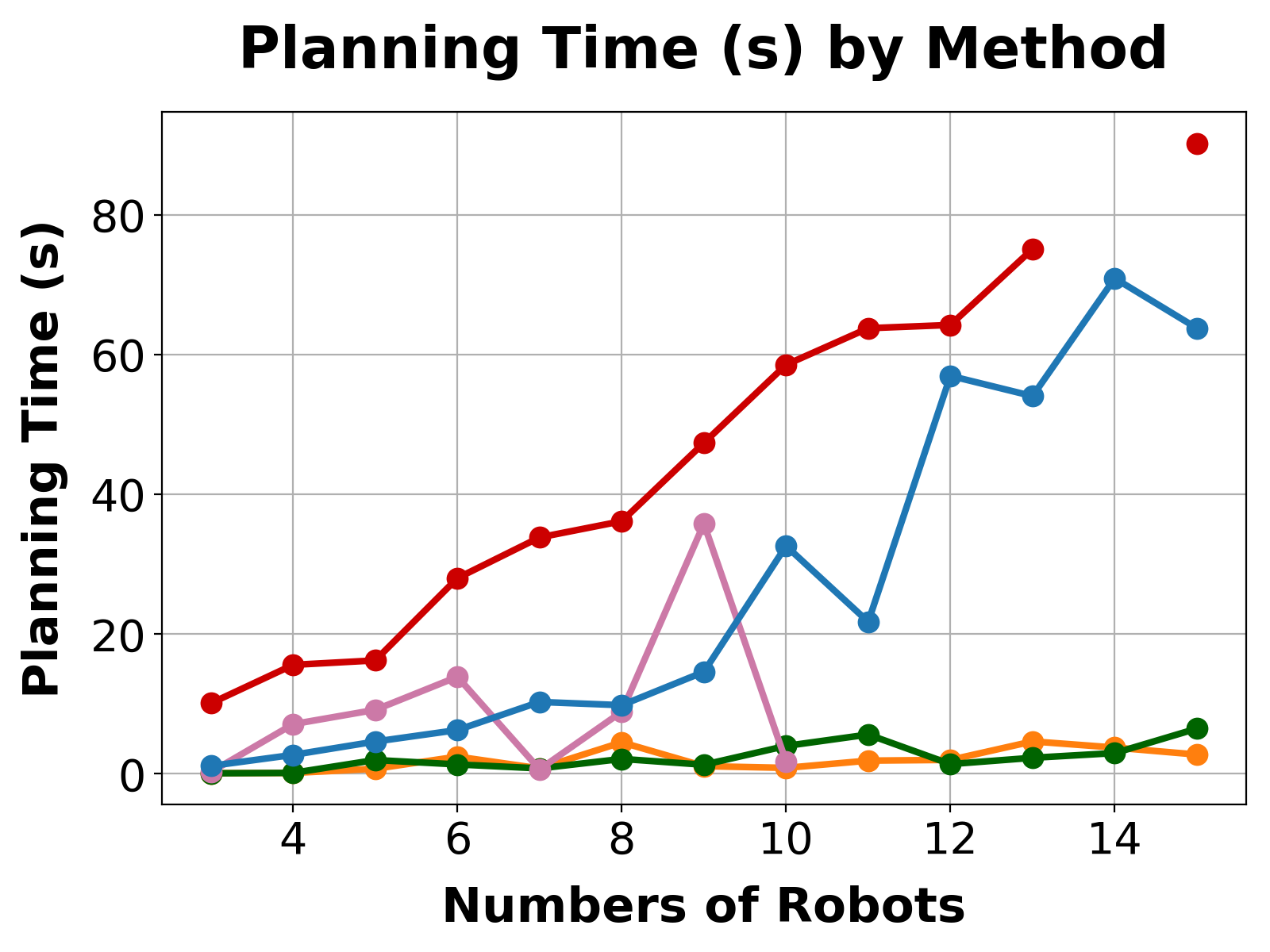}
    \end{minipage}\hfill
    \begin{minipage}[t]{\CellW}
      \vspace{0pt}\includegraphics[width=\linewidth]{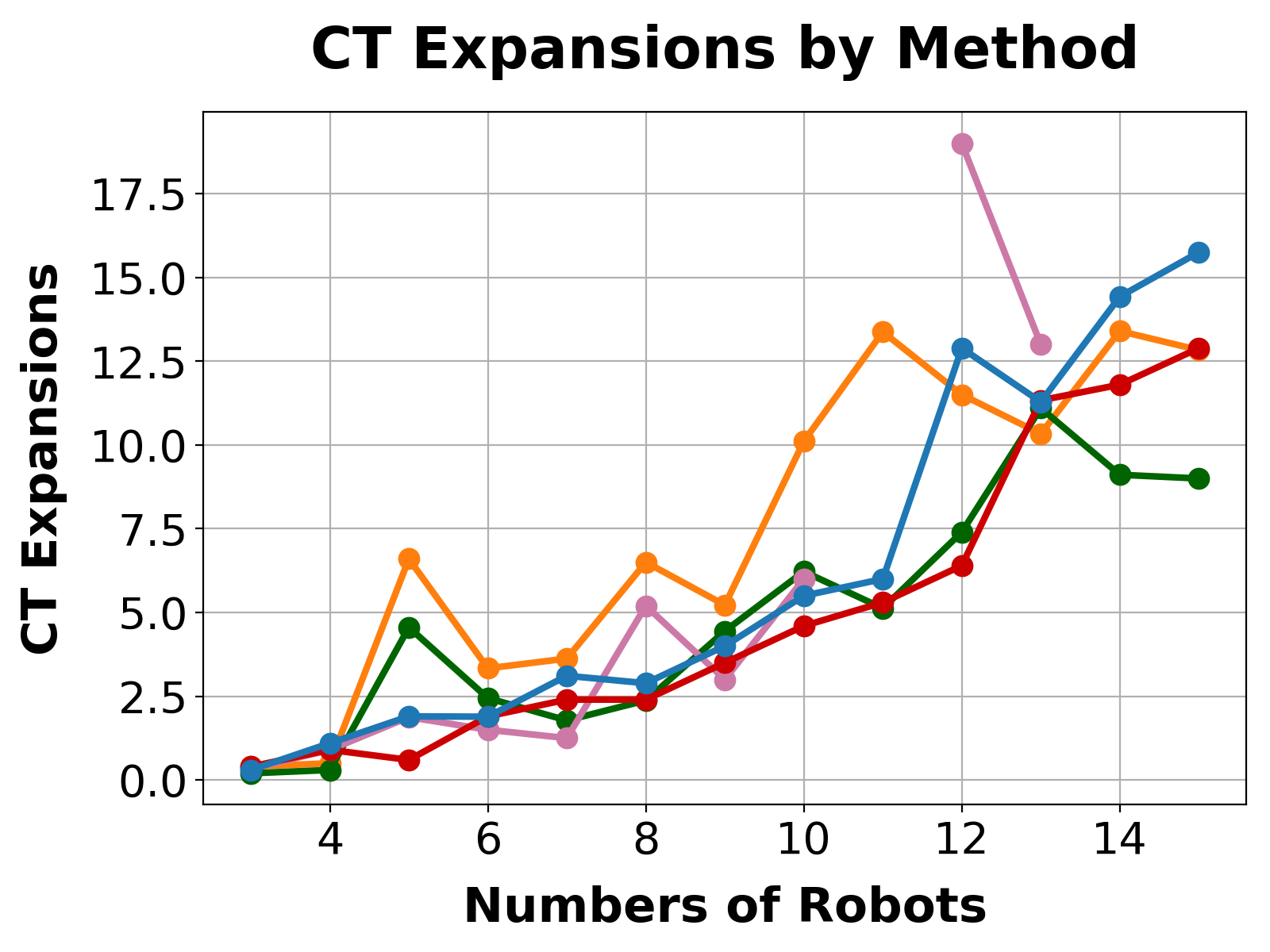}
    \end{minipage}
  \end{minipage}

  \caption{Performance of MRMP methods on Drop-Region map with the Weave Setup (top row) and Random map with the Random Setup(bottom row) with a 500-second time limit. The planning time and CT expansions are averaged on successful trials, and outliers are included in the averages but omitted beyond the 95th percentile for better visualization, especially in CT expansions.
  }
  \label{fig:drop}
\end{figure*}

%% file: figures/tex/emptylarge.tex
\begin{figure*}[t]
  \centering

  \newcommand{\LeftW}{0.16\textwidth}
  \newcommand{\RightW}{0.80\textwidth}
  \newcommand{\CellW}{0.32\linewidth}

  \begin{minipage}[t]{\LeftW}
    \vspace{0pt}%
    \centering
    \includegraphics[width=0.7\linewidth]{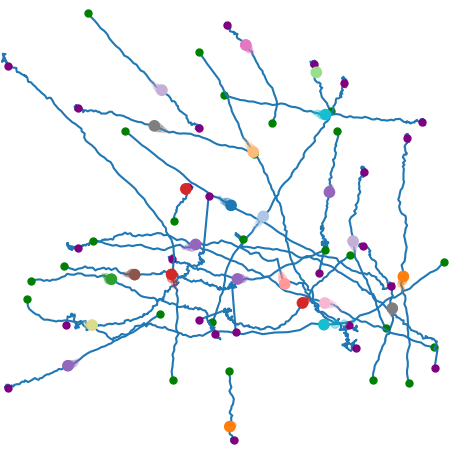}\\
    \textbf{Empty Map 4$\times$4}\par\vspace{5pt}%
    \includegraphics[width=0.7\linewidth]{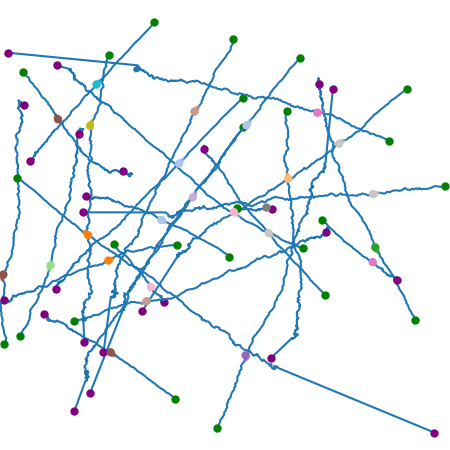}\\
    \textbf{Empty Map 6$\times$6}\par\vspace{5pt}%
    \includegraphics[width=0.9\linewidth]{figures/exp2/exp2_legend_s.png} \\
  \end{minipage}%
  \hfill
  \begin{minipage}[t]{\RightW}
    \vspace{0pt}%

    \begin{minipage}[t]{\CellW}
      \vspace{0pt}\includegraphics[width=\linewidth]{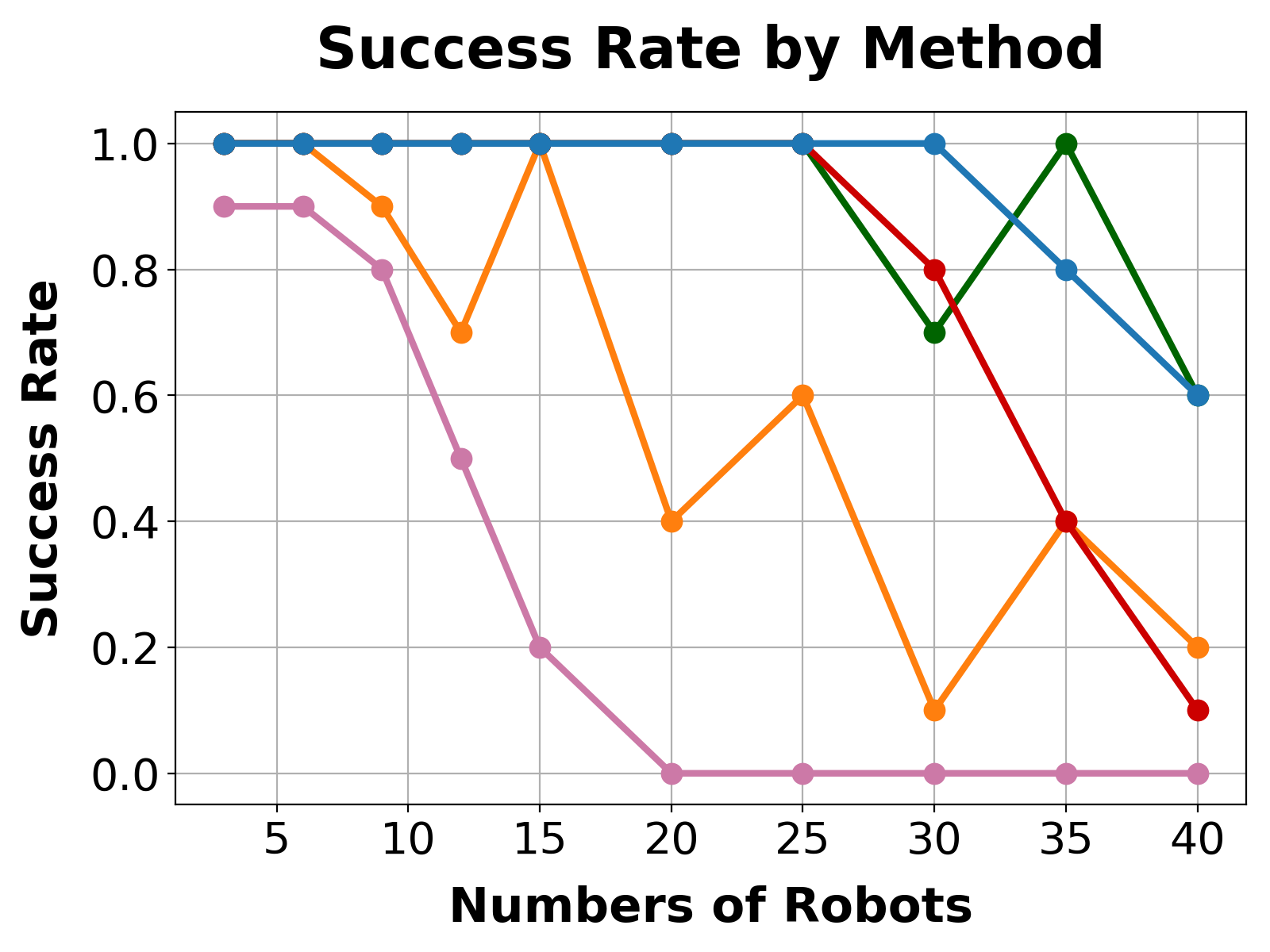}
    \end{minipage}\hfill
    \begin{minipage}[t]{\CellW}
      \vspace{0pt}\includegraphics[width=\linewidth]{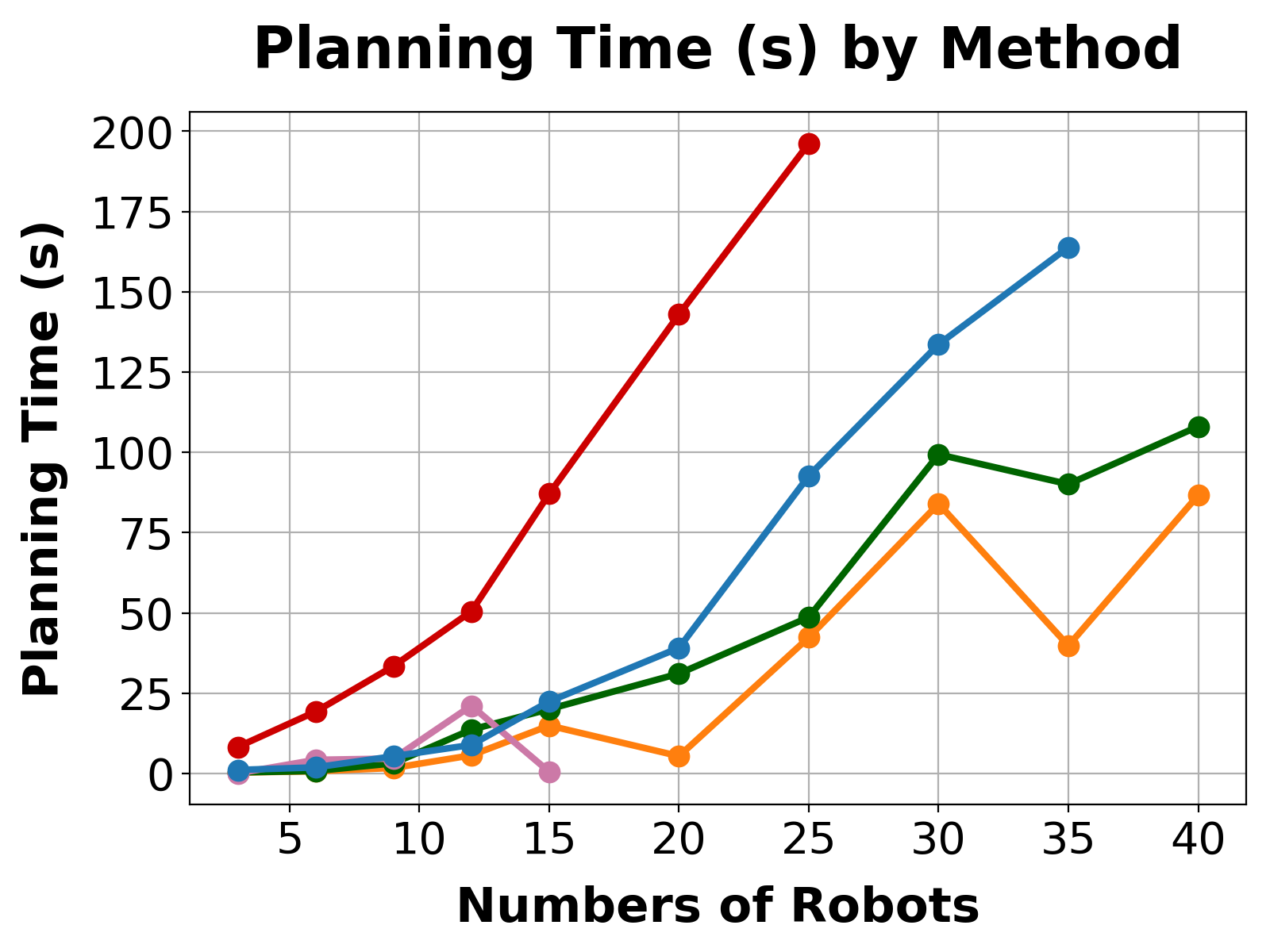}
    \end{minipage}\hfill
    \begin{minipage}[t]{\CellW}
      \vspace{0pt}\includegraphics[width=\linewidth]{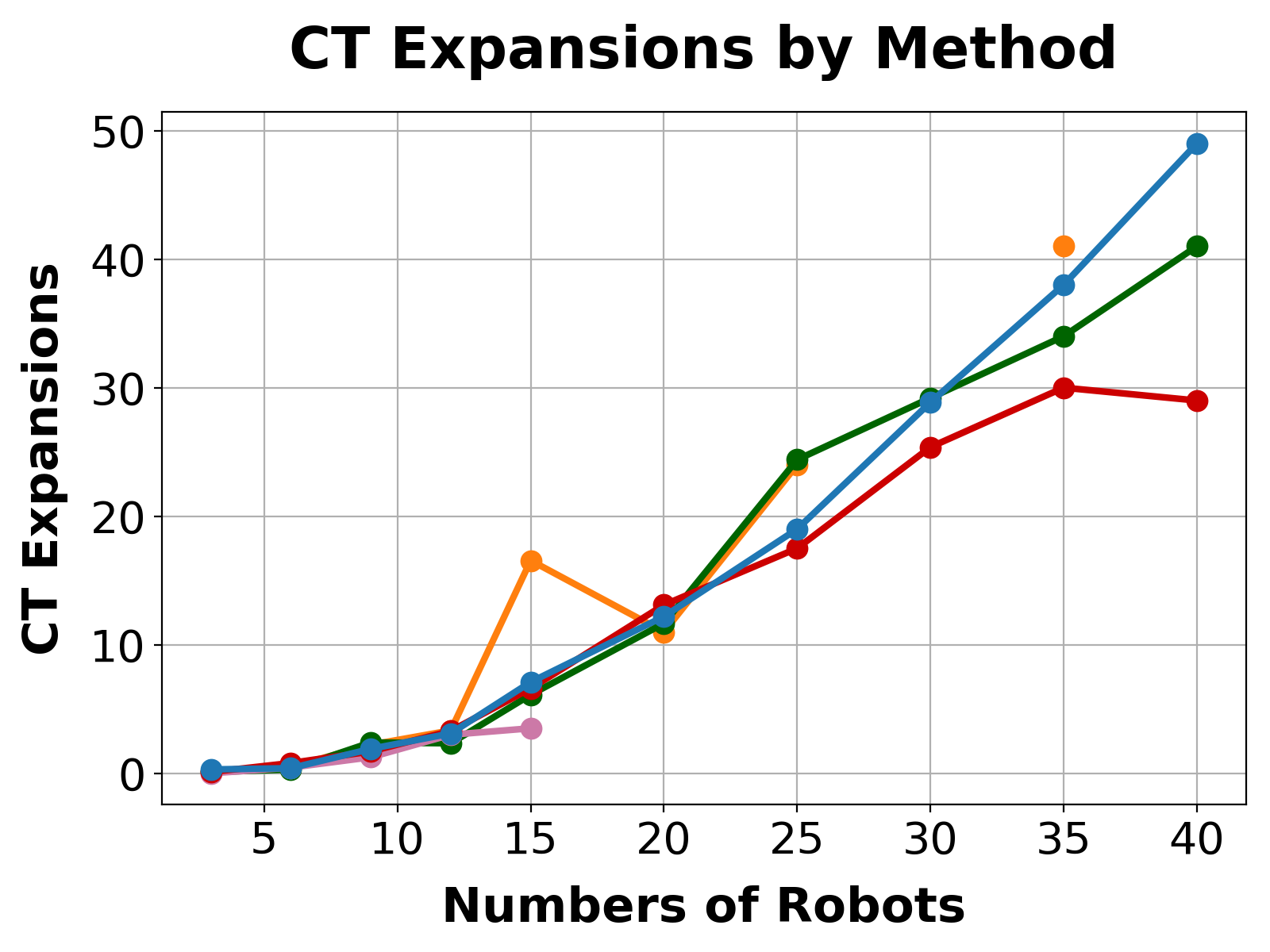}
    \end{minipage}

    \vspace{0.6em}

    \begin{minipage}[t]{\CellW}
      \vspace{0pt}\includegraphics[width=\linewidth]{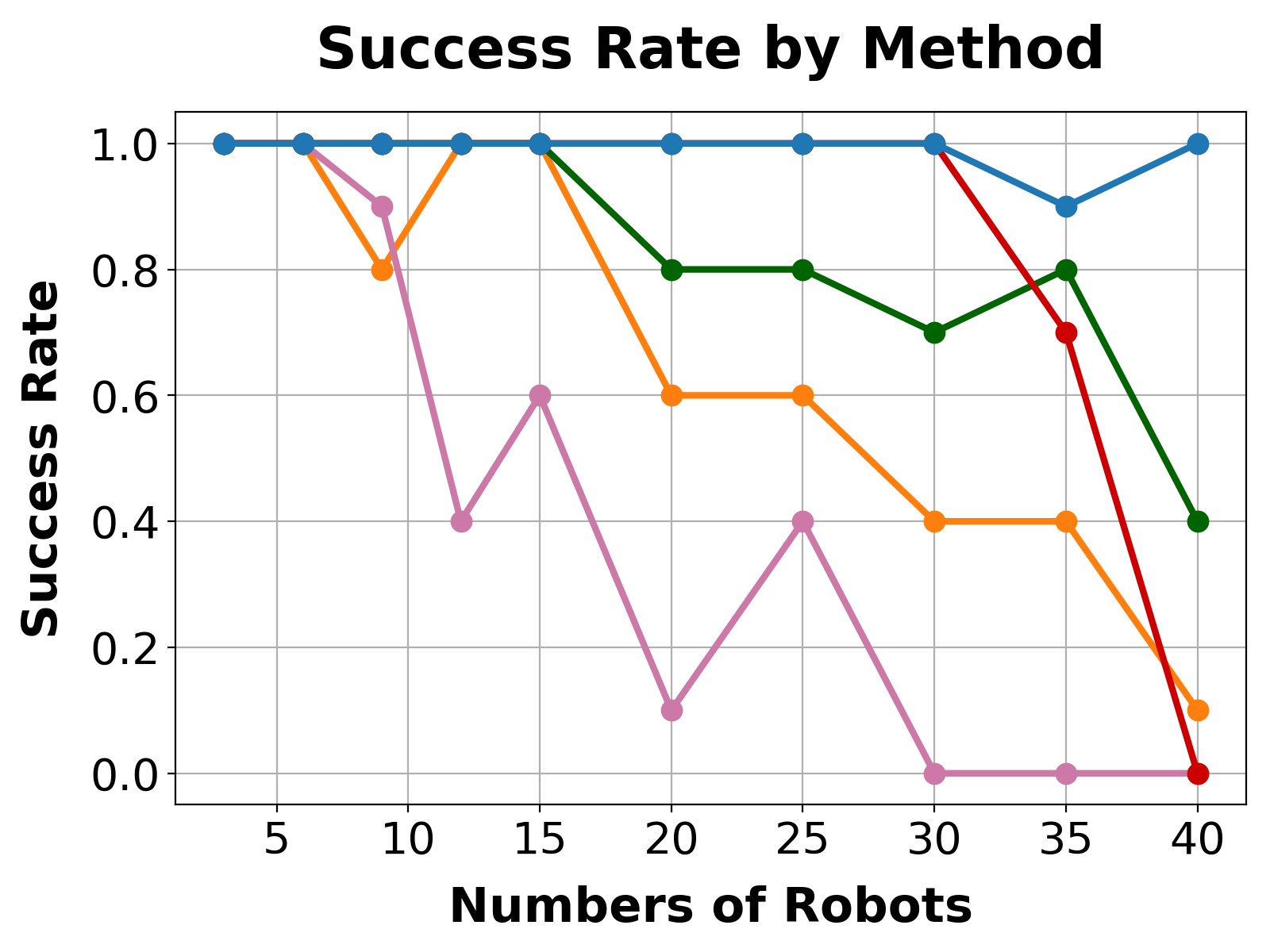}
    \end{minipage}\hfill
    \begin{minipage}[t]{\CellW}
      \vspace{0pt}\includegraphics[width=\linewidth]{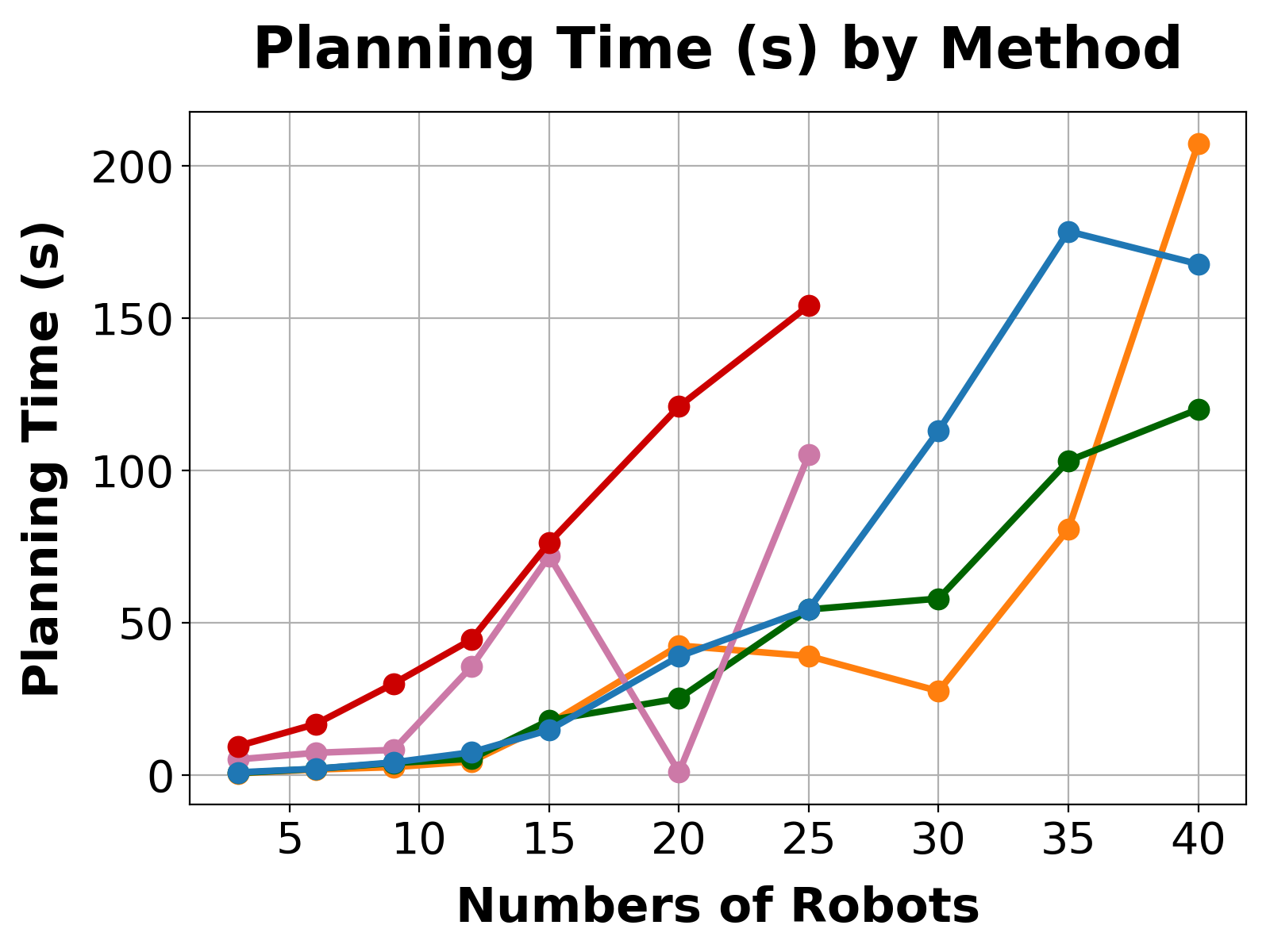}
    \end{minipage}\hfill
    \begin{minipage}[t]{\CellW}
      \vspace{0pt}\includegraphics[width=\linewidth]{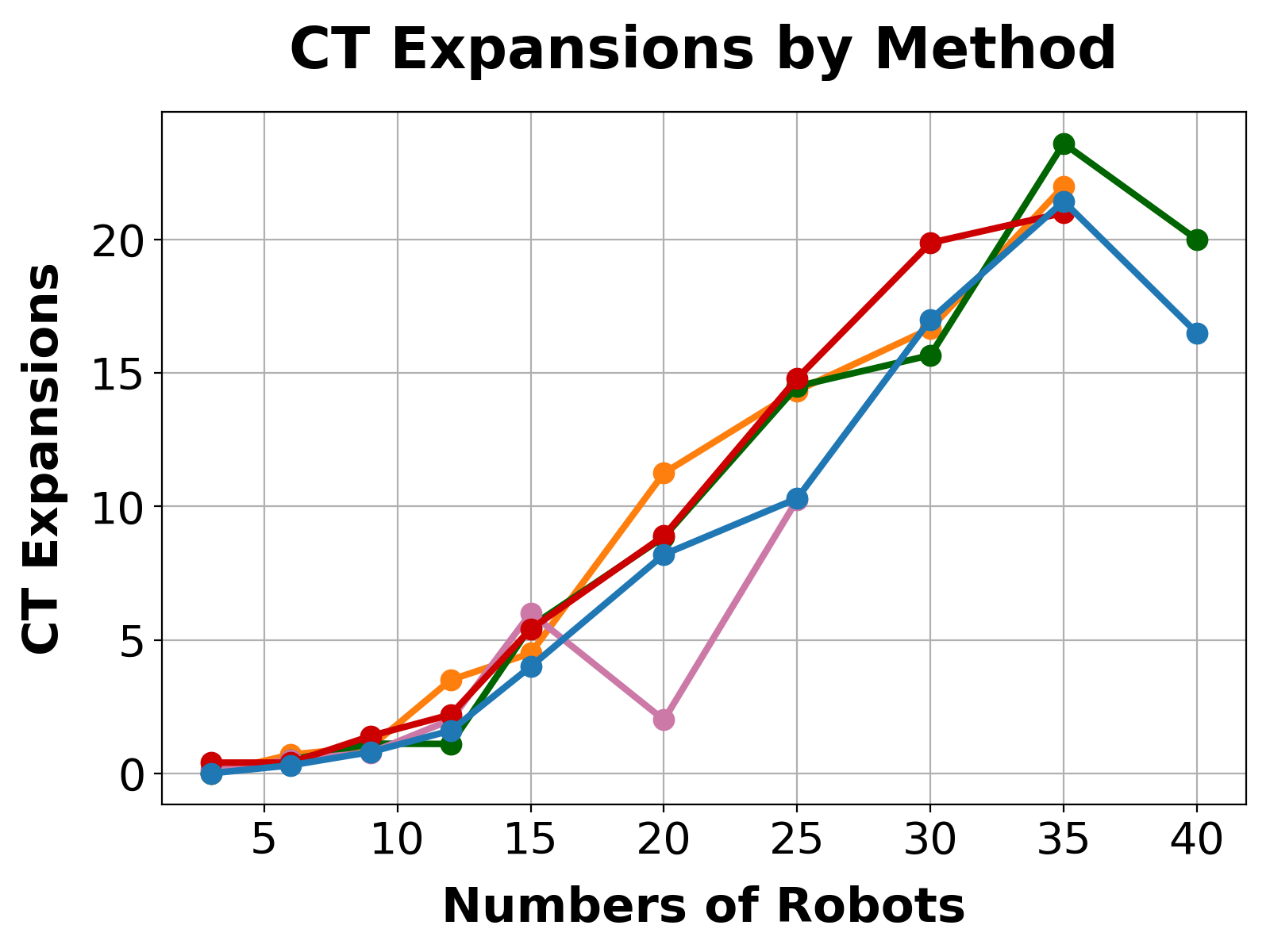}
    \end{minipage}
  \end{minipage}

  \caption{Performance of MRMP methods on the larger 4$\times$4 (top row) and 6$\times$6 (bottom row) Empty maps with a 500-second time limit, both with the Random Setup. The planning time and CT expansions are averaged on successful trials, and outliers are included in the averages but omitted beyond the 95th percentile for better visualization, especially in CT expansions.
  }
  \label{fig:long_horizon}
\end{figure*}

%% file: appendix.tex
\appendices

\section{Theoretical Analysis}
\subsection{Consistency of Constrained Monte Carlo Score Estimation}
\label{app:snis}

For brevity, we write $\tau=\tau^{1:H}$. In MDOC, we consider a single robot $i$ and, following model-based diffusion~\cite{pan2024model}, use a Monte Carlo score ascent approximation of the score function at diffusion step $k$. Consider a fixed reverse diffusion step $k$ and the Gaussian proposal $\Omega_k$ in Eq.~\eqref{eq:proposal_prelim}. We draw i.i.d.\ samples $\{\tilde\tau_m\}_{m=1}^M \sim \Omega_k$ and map each sample to the feasible set via the feasibility operator $\mathcal{F}$ introduced in Section~\ref{subsec:pai}--\ref{subsec:cbf}:
\begin{equation}
\tau_m \;:=\; \mathcal F(\tilde\tau_m), \qquad m=1,\dots,M,
\label{eq:pushforward_samples}
\end{equation}
so that $\tau_m \in \mathcal D\cap\mathcal C$ by construction. Consequently, all candidates used for Monte Carlo estimation are dynamically feasible and constrained, and the importance weights only need the soft-optimality term in Eq.~\eqref{eq:boltzmann_cost_prelim}:
\begin{equation}
w_m \;\propto\; p_J(\tau_m)
\;\propto\;
\exp\!\left(-\frac{\mathcal J(\tau_m)}{\lambda}\right).
\label{eq:weights_pJ_app}
\end{equation}
The weighted mean trajectory used in Eq.~\eqref{eq:mc_avg_prelim2} is then
\begin{equation}
{}^{k}\bar\tau
\;=\;
\frac{\sum_{m=1}^{M} w_m\,\tau_m}{\sum_{m=1}^{M} w_m}.
\label{eq:snis_estimator_app}
\end{equation}

Since $\mathcal F$ is deterministic, the random variable $\tau=\mathcal F(\tilde\tau)$ with $\tilde\tau\sim\Omega_k$ induces a probability measure on $\mathcal D\cap\mathcal C$, namely the pushforward measure
\begin{equation}
\Omega_k^{\mathcal F}(A)
\;\triangleq\;
\Omega_k\!\left(\mathcal F^{-1}(A)\right),
\quad \forall\,A\subseteq \mathcal D\cap\mathcal C.
\label{eq:app_pushforward_def}
\end{equation}
Equivalently, $\tau_m$ in Eq.~\eqref{eq:pushforward_samples} are i.i.d.\ samples from $\Omega_k^{\mathcal F}$.

We weight feasible trajectories using the soft-optimality term in Eq.~\eqref{eq:boltzmann_cost_prelim}.
Define the normalizer
\begin{equation}
Z_k \triangleq \int_{\mathcal D\cap\mathcal C} p_J(\tau)\, d\Omega_k^{\mathcal F}(\tau),
\label{eq:app_Zk}
\end{equation}
and define the step-$k$ planning distribution $\pi_k$ on $\mathcal D\cap\mathcal C$ by
\begin{equation}
\pi_k(d\tau)
\;\triangleq\;
\frac{p_J(\tau)}{Z_k}\,\Omega_k^{\mathcal F}(d\tau).
\label{eq:app_pi_def}
\end{equation}
That is, $\pi_k$ is the $p_J$-tilted version of the feasible-sample proposal $\Omega_k^{\mathcal F}$.
Intuitively, $\Omega_k^{\mathcal F}$ captures the feasible trajectories produced by the model-based diffusion rollout with CBF-constrained projection, while $p_J$ prefers a lower-cost trajectory distribution.

\begin{prop}[Consistency of the self-normalized feasible-trajectory estimator]
\label{prop:snis_app}
Assume that (i) $0<Z_k<\infty$ in Eq.~\eqref{eq:app_Zk}, and (ii)
\[
\mathbb E_{\Omega_k^{\mathcal F}}\!\left[\|p_J(\tau)\,\tau\|\right] < \infty.
\]
Then, for i.i.d.\ samples $\{\tilde\tau_m\}_{m=1}^M\sim\Omega_k$ and $\tau_m=\mathcal F(\tilde\tau_m)$, the estimator in Eq.~\eqref{eq:snis_estimator_app} satisfies
\begin{equation}
{}^{k}\bar\tau
\;\xrightarrow[M\to\infty]{\mathrm{a.s.}}\;
\mathbb E_{\pi_k}[\tau],
\label{eq:snis_convergence_app}
\end{equation}
i.e., ${}^{k}\bar\tau$ is a consistent self-normalized importance sampling estimator of the mean trajectory under $\pi_k$.
\end{prop}

\begin{proof}
Define $X_m := p_J(\tau_m)\,\tau_m$ and $Y_m := p_J(\tau_m)$.
By Eq.~\eqref{eq:pushforward_samples} and Eq.~\eqref{eq:app_pushforward_def}, $\{\tau_m\}$ are i.i.d.\ under $\Omega_k^{\mathcal F}$, hence $\{X_m\}$ and $\{Y_m\}$ are i.i.d.
Assumption (ii) implies $\mathbb E[\|X_1\|]<\infty$, and assumption (i) gives
\begin{equation}
\mathbb E[Y_1]
=
\mathbb E_{\Omega_k^{\mathcal F}}[p_J(\tau)]
=
Z_k \in (0,\infty).
\end{equation}
By the Strong Law of Large Numbers (SLLN), we have
\begin{equation}
\frac{1}{M}\sum_{m=1}^{M}X_m \xrightarrow{\mathrm{a.s.}} \mathbb E[X_1],
\end{equation}
and 
\begin{equation}
\frac{1}{M}\sum_{m=1}^{M}Y_m \xrightarrow{\mathrm{a.s.}} \mathbb E[Y_1]=Z_k>0.
\end{equation}
Therefore, the ratio converges almost surely:
\begin{equation}
{}^{k}\bar\tau
=
\frac{\frac{1}{M}\sum_{m=1}^{M}X_m}{\frac{1}{M}\sum_{m=1}^{M}Y_m}
\xrightarrow{\mathrm{a.s.}}
\frac{\mathbb E[X_1]}{\mathbb E[Y_1]}.
\label{ratio}
\end{equation}
Using Eq.~\eqref{eq:app_pi_def} and the definition of expectation under $\pi_k$,
\begin{align}
\mathbb E_{\pi_k}[\tau]
&=
\int_{\mathcal D\cap\mathcal C} \tau\,\pi_k(d\tau)
\label{eq:app_step1}
\\
&=
\int_{\mathcal D\cap\mathcal C} \tau\,
\frac{p_J(\tau)}{Z_k}\,\Omega_k^{\mathcal F}(d\tau)
\label{eq:app_step2}
\\
&=
\frac{\int_{\mathcal D\cap\mathcal C} \tau\,p_J(\tau)\,d\Omega_k^{\mathcal F}(\tau)}
{\int_{\mathcal D\cap\mathcal C} p_J(\tau)\,d\Omega_k^{\mathcal F}(\tau)}
\label{eq:app_step3}
\\
&=
\frac{\mathbb E_{\Omega_k^{\mathcal F}}\!\big[p_J(\tau)\,\tau\big]}
{\mathbb E_{\Omega_k^{\mathcal F}}\!\big[p_J(\tau)\big]}
\label{eq:app_step4}
\\
&=
\frac{\mathbb E[X_1]}{\mathbb E[Y_1]}.
\label{eq:app_step5}
\end{align}
Here, Eq.~\eqref{eq:app_step1} is the definition of $\mathbb E_{\pi_k}[\tau]$,
Eq.~\eqref{eq:app_step2} substitutes $\pi_k(d\tau)=\frac{p_J(\tau)}{Z_k}\Omega_k^{\mathcal F}(d\tau)$ from Eq.~\eqref{eq:app_pi_def},
Eq.~\eqref{eq:app_step3} uses the definition of $Z_k$ in Eq.~\eqref{eq:app_Zk},
Eq.~\eqref{eq:app_step4} follows from the definition of expectation under $\Omega_k^{\mathcal F}$,
and Eq.~\eqref{eq:app_step5} uses $X_1=p_J(\tau_1)\tau_1$ and $Y_1=p_J(\tau_1)$ with $\tau_1\sim\Omega_k^{\mathcal F}$.
Combining Eq.~\eqref{ratio} with
$\;\mathbb E_{\pi_k}[\tau]=\frac{\mathbb E[X_1]}{\mathbb E[Y_1]}$ yields
${}^{k}\bar\tau \xrightarrow{\mathrm{a.s.}} \mathbb E_{\pi_k}[\tau]$, which is Eq.~\eqref{eq:snis_convergence_app}.
\end{proof}

\begin{corollary}[Consistency of the affine MCSA score estimator]
\label{cor:score_consistency_app}
Assume the conditions of Proposition~\ref{prop:snis_app} hold.
Let the MCSA score estimator at diffusion step $k$ be the affine function of ${}^{k}\bar\tau$ given by Eq.~\eqref{eq:mc_score_prelim}, namely
\begin{equation}
\hat{\mathbf S}_k
=
-\frac{{}^{k}\tau}{1-\bar\alpha_k}
+
\frac{\sqrt{\bar\alpha_k}}{1-\bar\alpha_k}\,{}^{k}\bar\tau.
\label{eq:score_affine_app}
\end{equation}
Then
\begin{equation}
\hat{\mathbf S}_k
\xrightarrow[M\to\infty]{\mathrm{a.s.}}
-\frac{{}^{k}\tau}{1-\bar\alpha_k}
+
\frac{\sqrt{\bar\alpha_k}}{1-\bar\alpha_k}\,\mathbb E_{\pi_k}[\tau].
\label{eq:score_consistency_app}
\end{equation}
That is, the step-$k$ affine MCSA score estimator is almost surely consistent with respect to the planning distribution $\pi_k$.
\end{corollary}

\begin{proof}
By Proposition~\ref{prop:snis_app},
\[
{}^{k}\bar\tau \xrightarrow[M\to\infty]{\mathrm{a.s.}} \mathbb E_{\pi_k}[\tau].
\]
Since the MCSA score estimator in Eq.~\eqref{eq:score_affine_app} is an affine function of ${}^{k}\bar\tau$, the conclusion follows immediately by continuity of affine maps:
\[
\hat{\mathbf S}_k
=
-\frac{{}^{k}\tau}{1-\bar\alpha_k}
+
\frac{\sqrt{\bar\alpha_k}}{1-\bar\alpha_k}\,{}^{k}\bar\tau
\xrightarrow[M\to\infty]{\mathrm{a.s.}}
-\frac{{}^{k}\tau}{1-\bar\alpha_k}
+
\frac{\sqrt{\bar\alpha_k}}{1-\bar\alpha_k}\,\mathbb E_{\pi_k}[\tau].
\]
This is exactly Eq.~\eqref{eq:score_consistency_app}.
\end{proof}

Proposition~\ref{prop:snis_app} implies that increasing the rollout batch size $M$ makes the Monte Carlo mean ${}^{k}\bar\tau$ in Eq.~\eqref{eq:mc_avg_prelim2} converge to the planning mean under $\pi_k$. Corollary~\ref{cor:score_consistency_app} further shows that, since the MCSA score estimator in Eq.~\eqref{eq:mc_score_prelim} is an affine function of ${}^{k}\bar\tau$, it inherits the same almost-sure consistency with respect to $\pi_k$. This establishes the consistency of the step-$k$ constrained Monte Carlo score construction, rather than global optimality of the full reverse chain.

\subsection{Safety of CBF-Constrained MBD Rollouts}
\label{app:cbf_safety}

We formalize the safety guarantee provided by the per-step CBF-constrained projection on control input ${}^k\bar{u}^{h}$ inside the model-based diffusion rollouts.

Recall the safe set
\(
\mathcal C := \{\, s \mid b({}^{k}s^h)\ge 0 \,\}
\)
induced by a continuously differentiable barrier function $b$.
MDOC enforces the discrete-time CBF condition
\begin{equation}
b({}^{k}s^{h+1}) \;\ge\; (1-\gamma\Delta h)\,b({}^{k}s^{h}),
\quad \gamma>0,
\label{eq:disc_cbf_app}
\end{equation}
at diffusion step $k$ and uses the local linearization in Eq.~\eqref{eq:disc-lin} to obtain a
control-affine half-space constraint at each time step $h$.
At diffusion step $k$, the resulting CBF-feasible control set is the
polyhedron (Eq.~\eqref{eq:cbf_polytope})
\begin{equation}
{}^{k}\mathcal K^{h}
=
\left\{u\ \middle|\ {}^{k}A^{h}u \ge {}^{k}d^{h}\right\}.
\label{eq:Kkh_app}
\end{equation}
Each row $(a_j^\top,\,b_j)$ of $\bigl({}^{k}A^{h},{}^{k}d^{h}\bigr)$ defines
a half-space $a_j^\top u \ge b_j$ corresponding to one linearized CBF
constraint (e.g., one obstacle or one pairwise robot separation constraint).

Given a nominal control ${}^{k}\bar u^{h}$ proposed by diffusion, MDOC
computes a safe control ${}^{k}u^{h}$ by projecting ${}^{k}\bar u^{h}$ onto
the CBF half-spaces. For a single half-space
\(H(a,d):=\{u\mid a^\top u\ge d\}\), the projection of the nominal input is
\begin{equation}
{}^{k}u^{h}
=
\Pi_{H(a,d)}\!\left({}^{k}\bar u^{h}\right)
=
\begin{cases}
{}^{k}\bar u^{h}, & a^\top {}^{k}\bar u^{h} \ge d,\\[4pt]
{}^{k}\bar u^{h} + \dfrac{d-a^\top {}^{k}\bar u^{h}}{\|a\|^2}\,a, & a^\top {}^{k}\bar u^{h} < d,
\end{cases}
\label{eq:halfspace_proj_app}
\end{equation}
which is the unique minimizer of
\(\min_{{}^ku^h}\ \tfrac12\|{}^ku^h-{}^{k}\bar u^{h}\|^2\) subject to ${}^ku^h\in H(a,d)$.

\begin{lemma}[Projection onto the linearized feasible set]
\label{lem:projection_app}
Assume ${}^{k}\mathcal K^{h}$ is nonempty, closed, and convex.
Then the Euclidean projection
\begin{equation}
{}^{k}u^{h}
=
\Pi_{{}^{k}\mathcal K^{h}}\!\left({}^{k}\bar u^{h}\right)
\label{eq:proj_Kkh_app}
\end{equation}
exists uniquely and satisfies
\begin{equation}
\bigl\|{}^{k}u^{h}-{}^{k}\bar u^{h}\bigr\|
=
\min_{u\in{}^{k}\mathcal K^{h}}
\bigl\|u-{}^{k}\bar u^{h}\bigr\|.
\label{eq:proj_min_app}
\end{equation}
In particular, if ${}^{k}\bar u^{h}\in{}^{k}\mathcal K^{h}$, then
\begin{equation}
{}^{k}u^{h} = {}^{k}\bar u^{h}.
\label{eq:proj_identity_app}
\end{equation}
\end{lemma}

\begin{proof}
Since ${}^{k}\mathcal K^{h}$ is assumed to be nonempty, closed, and convex,
the Euclidean projection of ${}^{k}\bar u^{h}$ onto ${}^{k}\mathcal K^{h}$
exists and is unique. By definition of Euclidean projection, the point
${}^{k}u^{h}=\Pi_{{}^{k}\mathcal K^{h}}({}^{k}\bar u^{h})$ is the unique minimizer of
\[
\min_{u\in{}^{k}\mathcal K^{h}} \tfrac12 \|u-{}^{k}\bar u^{h}\|^2,
\]
which gives Eq.~\eqref{eq:proj_min_app}. If ${}^{k}\bar u^{h}\in{}^{k}\mathcal K^{h}$,
then the minimizer is attained at ${}^{k}\bar u^{h}$ itself, hence
${}^{k}u^{h} = {}^{k}\bar u^{h}$.
\end{proof}

\begin{assumption}[Rollout safety assumptions]
\label{ass:rollout_safety}
Fix a diffusion step $k$. Assume:
\begin{enumerate}
\item[(i)] Safe Initialization: ${}^{k}s^{1}\in\mathcal C$.
\item[(ii)] Nonempty Feasible Set: ${}^{k}\mathcal K^{h}\neq\emptyset$ for all rollout steps $h=1,\dots,H-1$.
\item[(iii)] Constraint Satisfaction: At each time step $h$, MDOC applies a projected control ${}^{k}u^{h}$ satisfying ${}^{k}u^{h}\in {}^{k}\mathcal K^{h}$.
\item[(iv)] Local Linearization Validity: The linearized CBF constraint obtained from Eq.~\eqref{eq:disc-lin} is locally valid around the nominal rollout point, with a bounded approximation error that can be absorbed by a tightening margin.
\item[(v)] $\gamma\Delta h \in (0,1]$.
\end{enumerate}
\end{assumption}

To account for the local linearization error in Assumption~\ref{ass:rollout_safety}(iv),
we introduce a tightened linearized feasible set
\begin{equation}
{}^{k}\widetilde{\mathcal K}^{h}
=
\left\{u\ \middle|\ {}^{k}A^{h}u \ge {}^{k}d^{h} + {}^{k}\eta^{h}\right\},
\label{eq:tightened_Kkh_app}
\end{equation}
where ${}^{k}\eta^{h}\ge 0$ denotes a tightening margin accounting for the local approximation error.

\begin{lemma}[Tightened linearized CBF constraints imply true one-step barrier inequality]
\label{lem:tightened_cbf_app}
Under Assumption~\ref{ass:rollout_safety}(iv), suppose ${}^{k}u^{h}\in{}^{k}\widetilde{\mathcal K}^{h}$.
Then the true next state ${}^{k}s^{h+1}=f({}^{k}s^{h},{}^{k}u^{h})$ satisfies
\begin{equation}
b({}^{k}s^{h+1})
\;\ge\;
(1-\gamma\Delta h)\,b({}^{k}s^{h}).
\label{eq:cbf_true_step_app}
\end{equation}
\end{lemma}

\begin{proof}
By Eq.~\eqref{eq:disc-lin}, the discrete-time CBF condition is locally linearized at step $(k,h)$ into the half-space constraint
\[
{}^{k}A^{h}u \ge {}^{k}d^{h},
\]
which is a first-order approximation of the true nonlinear condition in Eq.~\eqref{eq:disc_cbf_app}.
Under Assumption~\ref{ass:rollout_safety}(iv), the approximation error is locally bounded and can be absorbed by the tightening margin ${}^{k}\eta^{h}$ in Eq.~\eqref{eq:tightened_Kkh_app}.
Therefore, if ${}^{k}u^{h}\in{}^{k}\widetilde{\mathcal K}^{h}$, i.e.,
\[
{}^{k}A^{h}{}^{k}u^{h} \ge {}^{k}d^{h}+{}^{k}\eta^{h},
\]
then the true nonlinear discrete-time CBF condition holds for the successor state
${}^{k}s^{h+1}=f({}^{k}s^{h},{}^{k}u^{h})$, namely
\[
b({}^{k}s^{h+1})
\;\ge\;
(1-\gamma\Delta h)\,b({}^{k}s^{h}).
\]
This is exactly Eq.~\eqref{eq:cbf_true_step_app}.
\end{proof}

\begin{prop}[Discrete-time safety over the planning horizon]
\label{prop:cbf_safety}
Consider the discrete-time dynamics $s^{h+1}=f(s^{h},u^{h})$ and the safe set
$\mathcal C=\{s\mid b(s)\ge 0\}$.
Fix a diffusion step $k$ and assume Assumption~\ref{ass:rollout_safety} holds.
If at each time step $h$, MDOC applies a projected control satisfying
\[
{}^{k}u^{h}\in {}^{k}\widetilde{\mathcal K}^{h},
\]
then the rollout generated by applying $\{{}^{k}u^{h}\}_{h=1}^{H-1}$ satisfies
\[
{}^{k}s^{h}\in\mathcal C,\qquad \forall h=1,\dots,H,
\]
i.e., all states remain in the safe set at every discrete step over the planning horizon.
\end{prop}

\begin{proof}
Fix a diffusion step $k$. By assumption, for each time index
$h=1,\dots,H-1$, the control applied by MDOC satisfies
${}^{k}u^{h}\in{}^{k}\widetilde{\mathcal K}^{h}$. Therefore, by Lemma~\ref{lem:tightened_cbf_app}, the corresponding successor state
${}^{k}s^{h+1}=f({}^{k}s^{h},{}^{k}u^{h})$ satisfies the discrete-time CBF inequality
\begin{equation}
b({}^{k}s^{h+1}) \;\ge\; (1-\gamma\Delta h)\,b({}^{k}s^{h}).
\label{eq:cbf_step_app}
\end{equation}
Under Assumption~\ref{ass:rollout_safety}(v), we have $1-\gamma\Delta h\in[0,1)$, hence whenever
$b({}^{k}s^{h})\ge 0$ it follows from Eq.~\eqref{eq:cbf_step_app} that
$b({}^{k}s^{h+1})\ge 0$, i.e., ${}^{k}s^{h+1}\in\mathcal C$.

We now prove ${}^{k}s^{h}\in\mathcal C$ for all $h=1,\dots,H$ by induction:

\textbf{Base case:} by Assumption~\ref{ass:rollout_safety}(i), ${}^{k}s^{1}\in\mathcal C$, i.e.,
$b({}^{k}s^{1})\ge 0$.

\textbf{Induction step:} assume for some $h\in\{1,\dots,H-1\}$ that
${}^{k}s^{h}\in\mathcal C$. Applying the implication
above with ${}^{k}u^{h}={}^{k}u^{h}$ yields $b({}^{k}s^{h+1})\ge 0$, hence
${}^{k}s^{h+1}\in\mathcal C$.
By the principle of mathematical induction, ${}^{k}s^{h}\in\mathcal C$ for all
$h=1,\dots,H$.
\end{proof}

Proposition~\ref{prop:cbf_safety} establishes discrete-time safety over the planning horizon under controls that satisfy the tightened linearized CBF constraints at every rollout step. In practice, the guarantee may fail if ${}^{k}\mathcal K^{h}$ is empty, or if the local linearization mismatch is too large such that satisfying
${}^{k}A^{h}u\ge {}^{k}d^{h}+{}^{k}\eta^{h}$ no longer implies the true nonlinear condition from Eq.~\eqref{eq:disc_cbf_app}. In implementation, if ${}^{k}\mathcal K^{h}$ is empty, we enlarge the clearance buffer $\epsilon$ and re-run the trial until a feasible update is found.

\subsection{Properties of MDOC-CBS}
\label{app:mdoc_cbs}

We integrate our single-robot
planner, MDOC, as the low-level solver inside a
Conflict-Based Search (CBS) framework to obtain MDOC-CBS for
multi-robot motion planning (MRMP). We formalize the properties of this combination.

\begin{assumption}[Low-level probabilistic completeness]
\label{ass:low_level_pc}
For any robot $i$ and any finite constraint set
$\mathcal{C}_i$, the goal is to find the $\tau_i$ that satisfies dynamics $f(\cdot)$ and $\mathcal{C}_i$.

Assume that whenever this problem admits a feasible solution, the probability that MDOC returns some feasible trajectory
$\tau_i$ converges to $1$ as the computational budget (e.g., number of rollouts or diffusion steps) $B \to \infty$.
\end{assumption}

In addition, Proposition~\ref{prop:cbf_safety} guarantees that the trajectory returned by MDOC is safe with respect to the encoded obstacle and CBF constraints over the planning horizon.
Thus, we only need to verify the interaction conflicts of robots that
are handled by the high-level CBS constraints.

\begin{prop}[Soundness of MDOC-CBS]
\label{prop:mdoc_cbs_soundness}
Suppose that MDOC-CBS terminates at some CT node $V$ and returns
the joint trajectory $V.\tau$.
Then $V.\tau$ is a feasible joint trajectory, i.e.,
\begin{enumerate}
    \item each $V.\tau_i$ satisfies the robot dynamics, static
    obstacle constraints, and CBF safety constraints over the planning horizon, and
    \item $V.\tau$ contains no interaction conflicts for the
    entire planning horizon.
\end{enumerate}
\end{prop}

\begin{proof}
At any CT node $N$, the low-level planner MDOC
is invoked for each robot $i$ with the constraints
$N.\mathcal{C}_i$, together with the robot dynamics and
obstacle constraints. By Proposition~\ref{prop:cbf_safety}, the resulting trajectory $N.\tau_i$ is collision-free along the
entire horizon with respect to the encoded obstacle and CBF constraints.

The constraint sets $\{V.\mathcal{C}_i\}$ are constructed by
CBS as follows:
whenever a conflict between robots $i$ and $j$ is detected at
some ancestor node $V'$, CBS generates two child nodes by adding
a new constraint to either $V'.\mathcal{C}_i$ or
$V'.\mathcal{C}_j$, forbidding that specific conflict.
All constraints along the path from the root to $V$ are
accumulated in $\{V.\mathcal{C}_i\}$.

At the node $V$ where MDOC-CBS terminates, CBS has checked
$V.\tau$ and found no remaining robot--robot conflicts.
This implies that for all robot pairs $(i,j)$ and all times $h$
within the horizon, the trajectories $V.\tau_i$ and $V.\tau_j$
respect the separation constraints; in other words, $V.\tau$ is free of interaction conflicts.

Combining the low-level safety with the absence of interaction conflicts
established by CBS at termination, we conclude that $V.\tau$ is
a feasible joint trajectory.
\end{proof}

\begin{corollary}[Terminal conflict-free CT node implies a collision-free joint plan]
\label{cor:mdoc_cbs_terminal}
If MDOC-CBS terminates at a CT node $V$ such that each low-level trajectory $V.\tau_i$ satisfies its corresponding robot-specific constraints and $V.\tau$ contains no remaining interaction conflicts, then the returned joint trajectory $V.\tau$ is collision-free over the planning horizon.
\end{corollary}

\begin{proof}
This follows directly from Proposition~\ref{prop:mdoc_cbs_soundness}.
If each low-level trajectory satisfies the encoded single-robot constraints and the terminal CT node is conflict-free, then the joint trajectory is feasible and free of robot--robot conflicts over the planning horizon.
\end{proof}

Note that conditional CBFs enforce separation with respect to other robots' fixed CT trajectories during replanning; CBS termination additionally certifies that the final joint trajectories are mutually conflict-free.

\begin{prop}[Probabilistic completeness of MDOC-CBS]
\label{prop:mdoc_cbs_pc}
Assume that there exists at least one feasible joint trajectory
$\tau^{1:H}$ over the discrete horizon $h=1,\dots,H$ that satisfies the robot
dynamics, obstacle CBF constraints, and inter-robot separation
$\|{}^1s_i^{h}-{}^1s_j^{h}\|\ge r$ for all pairs $(i,j)$ and all $h$.
Assume further that the high-level CBS search expands CT nodes fairly, so that any CT node at finite depth will eventually be expanded.
Under Assumption~\ref{ass:low_level_pc}, MDOC-CBS is probabilistically complete:
as the low-level computational budget $B\to\infty$, the probability that
MDOC-CBS finds feasible joint trajectories converge to $1$.
\end{prop}

\begin{proof}
Consider an arbitrary feasible joint trajectory $\tau^{1:H}$ and the conflict
tree (CT) induced by the CBS branching rules used in MDOC-CBS.
Conflicts are checked only on the discrete grid $h=1,\dots,H$, hence $\tau^{1:H}$
contains finitely many conflict events, and the induced CT path has finite depth.

Whenever CBS detects a conflict between robots $i$ and $j$ at time $h$ at a node $V$,
it generates two child nodes by adding one of the following constraints:
(\textbf{i}) forbid robot $i$ from entering the workspace ball centered at ${}^1s_j^h$ with radius $r$ at time $h$, or (\textbf{ii}) forbid robot $j$ from entering the ball centered at ${}^1s_i^h$ with radius $r$ at time $h$.
Equivalently, these constraints require
$\|{}^1s_i^h-{}^1s_j^h\|\ge r$ at that time step for one of the robots.
Since $\tau^{1:H}$ is conflict-free, it satisfies $\|{}^1s_i^{h}-{}^1s_j^{h}\|\ge r$.
Therefore, at least one of the two child constraints is consistent with $\tau^{1:H}$,
so there exists a CT path from the root along which $\tau^{1:H}$ remains feasible.

Let $V$ be the finite-depth node at the end of this path. By construction,
the per-robot constraint sets $\{V.\mathcal C_i\}$ are all consistent with $\tau^{1:H}$.
When CBS expands $V$, MDOC is invoked for each robot $i$ under constraints $V.\mathcal C_i$.
In MDOC-CBS, inter-robot constraints at specified times are enforced in the low-level
via conditional CBFs $b_{i\mid j}^h(s_i^h)\ge 0$ with the other robots' CT trajectories fixed,
together with the discrete CBF condition
$b_{i\mid j}^{h+1}\ge (1-\gamma\Delta h)\,b_{i\mid j}^{h}$ inside the diffusion rollouts.
Because $\tau_i^{1:H}$ is feasible under these constraints, the single-robot problem at $V$
admits at least one feasible solution. By Assumption~\ref{ass:low_level_pc},
for each robot $i$, the probability that MDOC returns a feasible trajectory converges to $1$
as $B\to\infty$. Since the robot set is finite, by a union bound the probability that
MDOC succeeds for all robots at $V$, and also converges to $1$.

If the joint trajectory at $V$ is still not conflict-free, CBS will branch again.
However, the above argument still guarantees that at each detected conflict, at least one child
remains consistent with $\tau^{1:H}$. Hence, with probability approaching $1$ as $B\to\infty$,
CBS will eventually expand a finite-depth node on this path at which the low-level MDOC returns
a jointly conflict-free solution, and MDOC-CBS terminates successfully. Therefore, as $B\to\infty$, the probability that MDOC-CBS finds feasible joint trajectories converges to $1$.
\end{proof}

Proposition~\ref{prop:mdoc_cbs_soundness} shows that any solution
returned by MDOC-CBS is guaranteed to be collision-free and
feasible with respect to all encoded constraints over the planning horizon.
Corollary~\ref{cor:mdoc_cbs_terminal} makes explicit that a terminal conflict-free CT node induces a collision-free joint plan.
Proposition~\ref{prop:mdoc_cbs_pc} further establishes that,
under a natural probabilistic completeness assumption on the
low-level MDOC planner, the combined MDOC-CBS algorithm retains the probabilistic completeness property of CBS. Together with the single-robot safety results in
Proposition~\ref{prop:cbf_safety}, this provides a theoretical
justification for using MDOC as a continuous, model-based
diffusion low-level planner inside CBS for MRMP.

\input{figures/tex/baselines_table}

\section{Environmental Setup}
\subsection{Map Setup}
We consider 5 types of environment: Narrow, Empty, Conveyor, Drop-Region, and Random, with different sizes to create maps:

\textbf{Empty 2$\times$2.}
2-D Empty 2$\times$2 is our simplest map that a square $[-1, 1]^2$ with empty setup.

\textbf{Narrow 2$\times$2.}
To evaluate the performance of traversing a narrow bottleneck, we design a 2-D Narrow 2$\times$2 map. The workspace is a square $[-1, 1]^2$ containing a horizontal corridor (narrow bottleneck) of width 0.42, bounded by two long wall segments of thickness 0.02. To make the problem more complicated, we place two obstacles symmetrically near the center, each with a size of (0.3 × 0.1), 0.4 to the center of the map, creating a local bottleneck narrower than the rest of the passage but still traversable. 

\textbf{Conveyor 2$\times$2.}
To evaluate the performance of narrow passages with directional motion, we use a 2-D Conveyor 2$\times$2 map that places the robot in a workspace of $[-1, 1]^2$ with a set of long, thin rectangular obstacles arranged like conveyor belts. Three horizontal boxes are centered at $y=0$, $y=+0.35$, and $y=-0.35$, with lengths between 0.8 and 1.0 and a uniform thickness of 0.1.

\textbf{Drop-Region 2$\times$2.} 2-D Drop-Region 2$\times$2 map simulates package drop-off chutes with workspace $[-1, 1]^2$ that populated by four large box obstacles centered at the four quadrants $\{\,(\pm 0.4,\, \pm 0.4)\,\}$, each with size 0.4$\times$0.4. This layout yields a cluttered, symmetric environment where navigation requires maneuvering around large corner blocks and reaching specific narrow regions between them.

\textbf{Random 2$\times$2.} 2-D Random Basic 2$\times$2 map places robot in a workspace of $[-1, 1]^2$ with 5 square obstacles (0.15 width), and 5 sphere obstacles (0.15 diameters). We evaluate obstacle density by inflating all obstacles by the robot radius
($r=0.05$). The workspace area is $A_{\text{ws}}=4$. After Minkowski inflation,
each square becomes $0.25\times0.25$, and each circle becomes radius $0.125$,
giving a total occupied area
\begin{equation}
A_{\text{obs}}
= 5(0.25^2) + 5\!\left[\pi (0.125)^2\right]
\approx 0.558.
\end{equation}
Thus, the obstacle density is
\begin{equation}
\rho = A_{\text{obs}} / A_{\text{ws}} \approx 0.137.
\end{equation}


\textbf{Empty Large 4$\times$4.}
2-D Empty Large 4$\times$4 environment is the map that a square $[-2, 2]^2$ with empty setup for long horizon robustness evaluation.


\textbf{Empty Extra Large 6$\times$6.}
2-D Empty Extra Large 6$\times$6 environment is the map that a square $[-3, 3]^2$ with empty setup for long horizon robustness evaluation.


\subsection{Start and Goal Setup}
We consider 3 types of setups: Random, Circle, and Weave to initialize start and goal positions for all robots. 

\textbf{Circle Setup.}
All robots are placed uniformly on a circle of radius $r$, with start positions
\begin{equation}
s_i = r\bigl(\cos(2\pi i / N),\; \sin(2\pi i / N)\bigr),
\end{equation}
and goals located at the antipodal points on the same circle,
\begin{equation}
g_i = r\bigl(\cos(2\pi i / N + \pi),\; \sin(2\pi i / N + \pi)\bigr).
\end{equation}
Thus, each robot performs a symmetric dual-swap with its opposite partner.

\textbf{Weave Setup.} Robots are first sampled on a circle of radius $0.8$ and then projected onto the boundary of the square by snapping the dominant coordinate to $\pm\,\text{dist}$. For each robot $i$ with start $s_i$, the goal is defined by 
reflecting it across the center along its dominant axis:
\begin{equation}
g_i = \bigl( \;\text{sign}(s_i^{(1)})\,(-|s_i^{(1)}|) ,\;
          \text{sign}(s_i^{(2)})\,(-|s_i^{(2)}|) \bigr),
\end{equation}
effectively sending each robot to the opposite side of the boundary and creating
long-range cross-boundary swap motions.

\textbf{Random Setup.}
Start and goal positions are randomly generated in the map for all robots, and positions should (\textbf{1}) not on the obstacles; (\textbf{2}) have a margin to others (we fix the margin as 0.15 in all experiments).

\section{Baseline Details}
This section summarizes all baseline methods evaluated in our experiments. We group the methods into (\textbf{i}) single-robot planners, which generate trajectories for an individual robot under dynamics or geometric constraints, and (\textbf{ii}) MRMP methods that coordinate multiple robots. Each method is described with its paradigm, modeling assumptions, and constraint-handling mechanism, consistent with Table.~\ref{tab:baselines}.

\subsection{Single-Robot Motion Planning} 
\textbf{CEM.} The Cross-Entropy Method (CEM) is a sampling-based stochastic optimal control algorithm. CEM samples a batch $M$ of action sequences $\mathcal{U} = \{u_{m}^{1:H-1}\}_{m=1}^{M}$ is under the known dynamics $x_{t+1} = f(x_t, u_t)$, and yields trajectories $\tau$ with cost
$J(\tau) = \sum_{h=1}^{H-1} \ell(s^h, u^h)$. The top-$K$ elite samples update the sampling distribution via cross-entropy minimization. CEM is model-based and dynamics-aware, but handles collisions only via 
soft penalties in $J(\tau)$ and does not guarantee constraint satisfaction. Model-based diffusion will reduce to the CEM if the diffusion step is set to 1, proven in MBD~\cite{pan2024model}. 

\textbf{MPPI.} Model Predictive Path Integral (MPPI) control perturbs a nominal control sequence
and computes importance weights $w \propto \exp\!\left( -\frac{1}{\lambda} J(\tau) \right)$, which are used to update the nominal sequence. MPPI is fully model-based and relies on forward simulation of the system dynamics. Like CEM, collision constraints are implemented by soft cost shaping in the cost function, and no smoothness nor hard constraint guarantees are provided.

\textbf{RRT*.} RRT* is a geometric sampling-based planner that incrementally grows 
a tree in the configuration space. The planner produces purely geometric, piecewise-linear paths and thus does not model robot dynamics explicitly. RRT* is dynamics-agnostic and does not enforce smoothness or hard constraints beyond geometric collision checking. We include RRT* as one of our single-planar baselines for two reasons: (\textbf{1}) KCBS relies on the kinematic RRT* family for lower planar. (\textbf{2}) MMD-CBS uses RRT*-generated trajectories as demonstrations when training its model-free diffusion model.

\input{figures/tex/diffusion}

\subsection{Multi-Robot Motion Planning}
\textbf{A*-CBS/ECBS.} Conflict-Based Search (CBS) solves Multi-Agent Path Finding (MAPF) on a discrete grid. A low-level A* planner computes a path $\pi_i$ for each robot.
Enhanced CBS (ECBS) applies focal search to obtain bounded-tex/.
Both methods operate on discrete state-time graphs, thus are not dynamics-aware, and assume rectilinear motions with constant velocities. When a collision is detected at vertex $N$ and time $h$, 
both methods introduce a hard constraint $\langle R_i, N, h\rangle$ preventing agent $i$ from occupying $N$ at time $h$. For A*-CBS and A*-ECBS in our experiments, we randomly choose a node from the CT (instead of choosing the first node in CT) for different random seeds.

\textbf{KCBS.} Kinodynamic Conflict-Based Search (KCBS) extends CBS to continuous-time,
continuous-state MRMP with native kinodynamic feasibility.
Each robot follows nonlinear dynamics $\dot{s}_i = f_i(s_i, u_i)$,
and KCBS uses a kinodynamic single-robot planner RRT*
as its low-level solver.
A collision between robots $i$ and $j$ is represented as a time interval $K = \langle i, j, [h_s, h_e] \rangle ,$ inducing a moving-obstacle constraint 
\begin{equation}
\mathcal{C}_i = \big\{ \langle i,\; S_j(T_j(h)),\; h \rangle : h \in [h_s, h_e] \big\},
\end{equation}
which must be satisfied by the low-level kinodynamic planner.
KCBS therefore provides hard constraint satisfaction in continuous space and produces kinodynamically feasible trajectories.

\textbf{MMD-CBS.} MMD-CBS integrates single-robot diffusion models with CBS-style conflict resolution.
For robot $i$, the diffusion model $\theta$ generates a trajectory 
${}^{k}\tau_i = ({}^{k}\tau_i^{1:H})$.

Each denoising update is
\begin{equation}
    {}^{k-1}\tau_i \;\sim\;
    \mathcal{N}\!\Big(
        \mu_\theta({}^{k}\tau_i)
        \,+\,
        \eta\beta_k \nabla_\tau J\!\big(\mu_\theta({}^{k}\tau_i)\big),
        \;\beta_k
    \Big),
\end{equation}
where $J$ includes task terms, smoothness terms, and constraint-guidance terms.

CBS detects robot--robot conflicts and adds soft spatial constraints.  
A sphere constraint at location $p$ with radius $r$ is imposed through the guidance
\begin{equation}
    J_c({}^{k}\tau_i^{h})
    = 
    \max\!\big(
        \epsilon r - d\!\left(R_i({}^{k}\tau_i^{h}),\, p\right),
        0
    \big).
\end{equation}
Thus, MMD-CBS follows the CBS high-level branching structure, 
while each low-level plan is performed by running the constrained diffusion process 
using the indices ${}^{k}\tau_i^{t}$. Note that there are more advanced MMD-based multi-robot variants that exist, but we do
not include them here, as our goal is to isolate the effect of model-based versus model-free diffusion under a fixed high-level CBS coordination scheme.

\section{Differences Between MMD-CBS and MDOC-CBS}
Both works use diffusion as a lower-level path-generative method. However, while MMD-CBS leverages a model-free approach to train a diffusion network, MDOC-CBS relies on a dynamic model (model-based) for trajectory generation. We will discuss the differences between the two methods in this section.

\subsection{Objective} 
In MMD-CBS, the diffusion model learns a data-driven prior over feasible single-robot trajectories. The denoising update for robot $i$ at diffusion step $k$ is
\begin{equation}
    {}^{k-1}\tau_i \sim 
    \mathcal{N}\!\Big(
        \mu_\theta({}^{k}\tau_i)
        + \eta\beta_k \nabla_{\tau} J_{\text{MMD}}
        \!\left(\mu_\theta({}^{k}\tau_i)\right),
        \ \beta_k
    \Big),
\end{equation}
where $J_{\text{MMD}}$ is a composite soft objective whose primary role is to bias the generated trajectory toward data-adherent motions and away from collisions. 
Thus, the objective reflects a trade-off between dataset homotopy classes and feasibility.  

In MDOC-CBS, we instantiate the intermediate and terminal costs in
Eq.\eqref{eq:objective} to encode control effort, goal tracking, temporal shaping,
velocity smoothness, early-arrival penalties, and obstacle avoidance. 

\input{figures/tex/double}
\input{figures/tex/double_traj}

For robot $i$, let the state at step $h$ be $s_i^h = (q_i^h, \dots)$ with
position $q_i^h \in \mathbb{R}^2$ and control $u_i^h$, and let
$q_{i,\mathrm{goal}}$ denote its goal position. Define the distance to the
goal:
\begin{equation}
    d_i^h \triangleq \bigl\| q_i^h - q_{i,\mathrm{goal}} \bigr\|_2,
    \qquad
    d_{i,0} \triangleq \bigl\| q_i^1 - q_{i,\mathrm{goal}} \bigr\|_2 .
\end{equation}
Let the nominal arrival step be
\(
    H^\star \triangleq \max\{1, H-1\}
\)
with discrete time index $h \in \{1,\dots,H\}$.
We choose the stage cost:
\begin{equation}
\begin{aligned}
    c(\tau_i^h)
    &= w_u \,\|u_i^h\|_2^2
     \;+\; w_d \, d_i^h \\[0.2em]
    &\quad
     +\, w_{\mathrm{time}}
        \bigl(d_i^h - \hat d_i^h\bigr)^2 \\
    &\quad
     +\, w_{\mathrm{vel}}
        \bigl(\|u_i^h\|_2 - \bar v_i^h\bigr)^2 \\
    &\quad
     +\, w_{\mathrm{early}} \,
        \mathbf{1}\!\left[d_i^h \le \varepsilon_g,\; h < H^\star \right]
        \frac{H^\star - h}{H} \\[0.2em]
    &\quad
     +\, w_{\mathrm{sdf}} \,
        \phi_{\mathrm{sdf}}(q_i^h) ,
\end{aligned}
\end{equation}
where:
\begin{equation}
    \hat d_i^h
    \;=\;
    d_{i,0}\,\max\!\left(1 - \frac{h-1}{H^\star},\,0\right)
\end{equation}
is a linearly decaying reference distance encouraging an approximately
constant closing rate, and
\begin{equation}
    \rho^h \;=\; \min\!\left(\frac{h-1}{H^\star},\,1\right),
    \quad
    \bar v_i^h
    \;=\;
    \frac{d_{i,0}}{H^\star \Delta t}\, \rho^h (1 - \rho^h)
\end{equation}
is a bell-shaped desired speed profile.
The ``early arrival'' term penalizes staying within a small goal ball of radius
$\varepsilon_g$ too early in the horizon, with a penalty weight that decays as
$h$ increases.

To encode static obstacle avoidance, we use a signed-distance field (SDF)
penalty. Let $\mathrm{sdf}(q)$ denote the signed distance to obstacles in
workspace coordinates (positive outside obstacles, negative inside), $R$ the
robot radius, and $\delta>0$ a safety margin. We set:
\begin{equation}
    \phi_{\mathrm{sdf}}(q)
    \;=\;
    \bigl(\max\{0,\; R + \delta - \mathrm{sdf}(q)\}\bigr)^2,
\end{equation}
so that configurations that penetrate the inflated obstacles incur a
quadratic cost, while collision-free configurations incur zero cost.

The terminal cost in \eqref{eq:objective} is chosen as:
\begin{equation}
    c_f(s_i^H)
    = \bigl\| q_i^H - q_{i,\mathrm{goal}} \bigr\|_2 .
\end{equation}

\subsection{Implementation Difference}
MMD-CBS requires pre-training a neural diffusion model on RRT* generated dataset trajectories for a single planner. Replanning during CBS branches simply re-runs the denoiser with modified soft guidance terms. The method, therefore, requires sufficiently diverse demonstrations to generalize to new layouts. In the experiments, we generate datasets and retrain the diffusion model on each map. Without retraining, performance will downgrade massively. \cite{liang2025simultaneous} reports that MMD-CBS only solves 6-10 robots in random maps without retraining models on random setups. MDOC-CBS, by contrast, implements diffusion entirely in model space: each denoising update integrates the robot dynamics forward, computes the model-based guidance, and applies a CBF projection. This eliminates the need for dataset pretraining and enables generalization to arbitrary map configurations, but it requires explicit knowledge of the dynamics $f(\cdot)$ and the CBF geometry. Implementation-wise, MMD-CBS resembles a learned generative prior and soft constraint guidance, whereas MDOC-CBS resembles a model-based optimal control solver embedded with CFB-constrained projection inside diffusion rollouts.

\subsection{Single Planner Planning Time} Because MMD-CBS relies on a neural diffusion model, its planning time is dominated 
by neural forward passes, and therefore varies depending on model complexity and the 
computing device used. For a fair comparison, MDOC-CBS also leverages GPU 
acceleration: all rollout components (dynamics integration, cost evaluation, and CBF 
projection) are executed inside a single parallel large tensor. As a result, the only 
factor that directly affects MDOC-CBS planning time is the number of diffusion 
steps. This also suggests that MDOC-CBS is more amenable to 
parallelization within CBS, as many agents can be batched directly on the GPU, 
whereas MMD-CBS must additionally handle multi-model inferences during tree 
expansion.

\section{Ablation Study}

We also conduct an ablation study to examine how dynamics, diffusion hyperparameters, and the high-level planner jointly affect MRMP performance: (\textbf{1}) with a double-integrator model, in both Circle and Weave setups, robots spontaneously form a coherent circulating ring that swaps positions around the outer boundary instead of pushing through the congested center, yielding longer paths but substantially lower planning time and fewer CT expansions; (\textbf{2}) increasing the number of diffusion steps broadens the exploration cone and helps escape suboptimal homotopies, but also raises rollout time, variance, and collision rates;  (\textbf{3}) using ECBS as the high-level planner does reduce CT expansions, yet its relaxed branching increases the difficulty of each CBF-projected rollout, effectively neutralizing the gains from exploring fewer nodes.

\input{figures/tex/ecbs}

\subsection{Performance of Double Integrator}
We study how different dynamic models affect the behavior of robots. With single-integrator dynamics, the Monte Carlo Score Ascent of MDOC often favors center-cutting trajectories that quickly traverse the intersection zone. The high-level CBS must then frequently insert and adjust timings near the center, leading to more CT expansions and higher planning time. Double-integrator dynamics, on the other hand, introduce velocity as an additional state and thus penalize rapid changes in velocity/acceleration, making sudden stops and sharp turns near the center costly. As a result, MDOC tends to organize robots into a smooth following loop along the circumference, using coherent rotation for repositioning (Figure~\ref{fig:double_path}). This circular cooperative pattern matches the Circle setup: instead of resolving conflicts by rushing into the center, robots swap positions along the ring, which substantially reduces interaction conflicts. Consequently, in Empty Circle settings, double-integrator dynamics sustain higher robot counts with higher success rates while significantly reducing CT expansions and planning time, even though the physical paths are slightly longer (Figure~\ref{fig:double_empty_conveyor}). Overall, dynamics can induce qualitatively different coordination patterns, and their benefits depend on map topology.

In the Conveyor Circle with obstacles, the same circular flow concentrates trajectories on a smooth ring outside the obstacles, yielding a shorter and more stable joint solution than the single-integrator behavior that effectively ``draws a rectangle'' through narrow passages. It also reduces repeated CBF corrections near passage boundaries. However, this advantage largely exploits the symmetry of the Circle setting: when the topology is closer to a Weave, the coordinated circular flow no longer aligns with obstacle geometry, leading to performance similar to single-integrator dynamics in Empty Weave or even degraded performance in more complex Conveyor Weave (Figure~\ref{fig:double_empty_conveyor}). We further hypothesize that this mismatch would be amplified on random maps or random start--goal pairs where such symmetry is absent.

\subsection{Performance of Different Diffusion Steps}
We evaluated MDOC-CBS with different numbers of diffusion steps on the Conveyor map under Circle and Weave setups. In the Circle setup, which induces fewer CT expansions and sparser constraints, increasing the number of diffusion steps from 200 to 400 allows Monte Carlo Score Ascent to optimize longer within a relatively clean geometry, enabling trajectories to converge toward safe and dynamically feasible solutions gradually. MDOC-CBS with 400 diffusion steps achieves higher success rates than the 200- and 300-step variants once the number of robots exceeds 15. In this regime, the CBF-constrained projection typically applies only minor safety corrections, so longer rollouts improve solution quality with only a slight increase in planning time.

In contrast, the Weave setup requires substantially more expansions and imposes denser constraints. Here, score ascent enforces consistency at every diffusion step, while the CBF-constrained projection performs hard corrections to maintain safety in narrow regions. This interaction accumulates projection bias and increasingly drives trajectories toward corners of feasible regions or obstacle boundaries, leading to local jitter, instability, or forced edge-hugging as the number of diffusion steps increases. Therefore, the 200-step chain, being shorter, is less prone to accumulating projection-induced artifacts and exhibits the most stable performance under high-constraint conditions. Overall, more diffusion steps help in low-constraint regimes but can hurt when safety projections are frequent and strong.

\begin{table*}[tb!]
\centering
\begin{tabular}{c|c|c|c|c|c}
    \toprule
    Category & Hyperparameter & Empty & Conveyor & Drop-Region & Random  \\
    \midrule
    \multirow{3}{*}{Environment} 
      & Robot Radius $R$     & 0.05 & 0.05 & 0.05 & 0.05 \\
      & Samples Size $M$   & 64  & 64 & 64  & 128  \\
      & Horizon $H$ & 64  & 64 & 64  & 64  \\
    \midrule
    \multirow{4}{*}{Diffusion} 
      & Temperature $\delta_d$     & 0.5 & 0.1 & 1e-12 & 1e-8 \\
      & Diffusion Steps $K$   & 100  & 200 & 250  & 100 \\
      & $\beta_0$  & 1e-5  & 1e-5 & 1e-5  & 1e-5  \\
      & $\beta_T$  & 1e-2  & 1e-2 & 1e-2  & 1e-2 \\
    \midrule
    \multirow{2}{*}{CBF} 
      & $\gamma$   & 0.8  & 0.8 & 1.5  & 1.2 \\
      & Clearance Buffer $\epsilon$ & 0.01  & 0.01 & 0.1  & 0.1  \\
    \midrule
    \multirow{7}{*}{Objective Function} 
      & $w_u$  & 1 & 1 & 1 & 1 \\
      & $w_d$   & 5  & 5 & 5  & 5 \\
      & $w_{\mathrm{time}}$  & 1  & 1 & 1  & 1  \\
      & $w_{\mathrm{vel}}$  & 1  & 1 & 1 & 1 \\
      & $w_{\mathrm{early}}$   & 0.5  & 0.5 & 0.5  & 0.5 \\
      & $w_{\mathrm{sdf}}$  & 5e3  & 5e3 & 5e3  & 5e3  \\
      & $c_f$  & 8  & 8 & 20 & 8 \\
    \bottomrule
    \end{tabular}
  \caption{We showcase the parameters that we use in MDOC-CBS in different environmental maps. We use same parameters for Larger maps expect changing Horizon $H$ to 128 and 196 for 4 $\times$ 4 map and 6 $\times$ 6 map.}
  \label{tab:parameters}
\end{table*}

\subsection{Performance of ECBS}
We integrate MDOC with ECBS and compare it against MDOC-CBS on Empty Circle and Conveyor Circle maps. In the Empty Circle map, due to the simple geometry and the fact that robots primarily face interaction conflicts rather than obstacle constraints, Monte Carlo score ascent can smoothly optimize each robot's trajectory while the CBF projection only applies minor safety corrections throughout the rollout. In this setting, even with heuristic relaxation, ECBS can reliably guide the conflict tree search without significantly perturbing the convergence direction of the lower-level MDOC. Moreover, because ECBS allows bounded suboptimality, it can bypass costly branches more aggressively, substantially reducing CT expansions and enabling MDOC-ECBS to achieve success rates comparable to MDOC-CBS while lowering planning time.

In the Conveyor Circle map, however, obstacles narrow the feasible region and increase the dependence of the lower-level MDOC rollout on high-level constraints and search direction. Here, score ascent pulls trajectories toward low-cost directions, while the CBF projection repeatedly pushes them away from obstacle boundaries, creating a sensitive coupling. With ECBS, heuristic relaxation can more easily choose suboptimal branches that drive the system into obstacle-incompatible regions, causing the optimization direction of MDOC to repeatedly conflict with CBF-imposed safety corrections and letting such biases accumulate over the rollout. As a consequence, the success rate of MDOC-ECBS decreases significantly on the Conveyor Circle map. Overall, ECBS improves efficiency in Empty environments by reducing CT expansions, whereas in obstacle-rich environments, its approximation can amplify the MDOC--CBF coupling and make strictly consistent CBS the more robust high-level choice.

Overall, in Empty environments, ECBS relaxation does not disrupt the MDOC optimization process and instead improves efficiency by reducing CT expansions. In contrast, in obstacle-rich environments, the approximation properties of ECBS and the dynamic coupling between Monte Carlo Score Ascent and CBF projection become more pronounced, leading to instability and making the strictly consistent CBS a more robust high-level choice.

\section{Experiment Hyperparameters}
We showcase the MRMP hyperparameters that we use for different environments on the main page in Table.~\ref{tab:parameters}.

\section{Additional Experiment Results}

\textbf{Result Tables.} We showcase the MRMP results in different environmental settings. Note that the metrics are calculated based on the number of successful trials.

\textbf{Visual Examples of Trajectories.} We also showcase some examples of trajectories of the MRMP results that we chose from successful trials for different environmental settings.

\input{figures/tex/app_results}
\input{figures/tex/app_vis}

%% file: figures/tex/baselines_table.tex
\begin{table*}[tb!]
\small
\newcolumntype{O}{>{          \arraybackslash}m{0.4 cm}}
\newcolumntype{A}{>{          \arraybackslash}m{3.3 cm}}
\newcolumntype{B}{>{\centering\arraybackslash}m{4.0 cm}}
\newcolumntype{C}{>{\centering\arraybackslash}m{2 cm}}
\newcolumntype{D}{>{\centering\arraybackslash}m{1.5 cm}}
\newcolumntype{E}{>{\centering\arraybackslash}m{1.5 cm}}
\newcolumntype{F}{>{\centering\arraybackslash}m{1.5 cm}}
\newcolumntype{G}{>{\centering\arraybackslash}m{1.5 cm}}
\newcolumntype{H}{>{\centering\arraybackslash}m{1.5 cm}}
\newcolumntype{I}{>{\centering\arraybackslash}m{3 cm}}
\setlength\tabcolsep{0pt}
\renewcommand{\arraystretch}{0.93}
\newcommand\B{\rule[-0.9ex]{0pt}{0pt}} 
\newcommand{\midsepremove}{\aboverulesep = 0mm \belowrulesep = 0mm}
\newcommand{\midsepdefault}{\aboverulesep = 0.605mm \belowrulesep = 0.984mm}
\label{tab:related}
\begin{center}
\begin{tabular}{O|ACDEFGHI}
\toprule
\multicolumn{2}{c}{\textbf{Method}} 
& {\textbf{Paradigm}}
& {Dynamics \smash{$\hphantom{^{X}}$}Aware}
& {Model- \smash{$\hphantom{^{X}}$}Based}
& {Non-Data \smash{$\hphantom{^{X}}$}Driven}
& {Diffusion \smash{$\hphantom{^{3}}$}Based}
& {Smoothness \smash{$\hphantom{^{X}}$}Guarantee}
& {Constraint Handling}\\
\midrule

\parbox[t]{2mm}{\multirow{6}{*}{\rotatebox[origin=c]{90}
{\textbf{Single Planar}\hspace{0.5em}}}}
& ~~CEM~\cite{botev2013cross} & Sampling Control & \cm & \cm & \cm & \xm  & \xm & Soft Constraints \\
& ~~MPPI~\cite{williams2018information} & Sampling Control & \cm & \cm & \cm & \xm  & \xm & Soft Constraints \\
& ~~RRT*~\cite{karaman2011anytime} & Geometric Search  & \xm & \xm & \cm & \xm & \xm & - \\
\cmidrule{2-9}
& ~~\textbf{MDOC (Ours)} & Diffusion + Control & \cm & \cm & \cm & \cm & \cm & Hard Constraints \\
\midrule

\parbox[t]{2mm}{\multirow{7}{*}{\rotatebox[origin=c]{90}
{\textbf{MRMP}\hspace{0em}}}}
& ~~A*-CBS/ECBS~\cite{sharon2015conflict, barer2014suboptimal}  & Search (MAPF) & \xm & \xm & \cm & \xm & \xm & Hard Constraints \\
& ~~KCBS~\cite{kottinger2022conflict} & Search + RRT* + MRMP  & \cm & \cm & \cm & \xm & \xm & Hard Constraints \\
& ~~MMD-CBS~\cite{shaoul2025multi} & Diffusion + MRMP & \xm & \xm & \xm & \cm & \xm & Guidance via Soft Constraints \\
\cmidrule{2-9}
& ~~\textbf{MDOC-CBS (Ours)} & Diffusion + Control + MRMP & \cm & \cm & \cm & \cm & \cm & Hard Constraints \\
\bottomrule

\end{tabular}
\caption{Comparison of single-robot and multi-robot motion baseline planners, highlighting our MDOC amd MDOC-CBS as the only approach that is simultaneously dynamics-aware, model-based, training-free, diffusion-based, and equipped with hard safety constraints through CBF-constrained projection within model-based diffusion rollouts.}
\label{tab:baselines}
\end{center}
\end{table*}

%% file: figures/tex/diffusion.tex
\begin{figure*}[t]
  \centering

  \newcommand{\LeftW}{0.16\textwidth}
  \newcommand{\RightW}{0.83\textwidth}
  \newcommand{\CellW}{0.32\linewidth}

  \begin{minipage}[t]{\LeftW}
    \vspace{0pt}%
    \centering
    \includegraphics[width=0.7\linewidth]{figures/exp2/conveyor/mdoc_single_trial_circle.png}\\
    \textbf{Conveyor Circle}\par\vspace{0.6em}%
    \includegraphics[width=0.7\linewidth]{figures/exp2/conveyor/mdoc_single_trial_weave.png}\\
    \textbf{Conveyor Weave}\par\vspace{0.6em}%
    \includegraphics[width=0.9\linewidth]{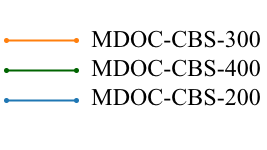} \\
  \end{minipage}%
  \hfill
  \begin{minipage}[t]{\RightW}
    \vspace{0pt}%

    \begin{minipage}[t]{\CellW}
      \vspace{0pt}\includegraphics[width=\linewidth]{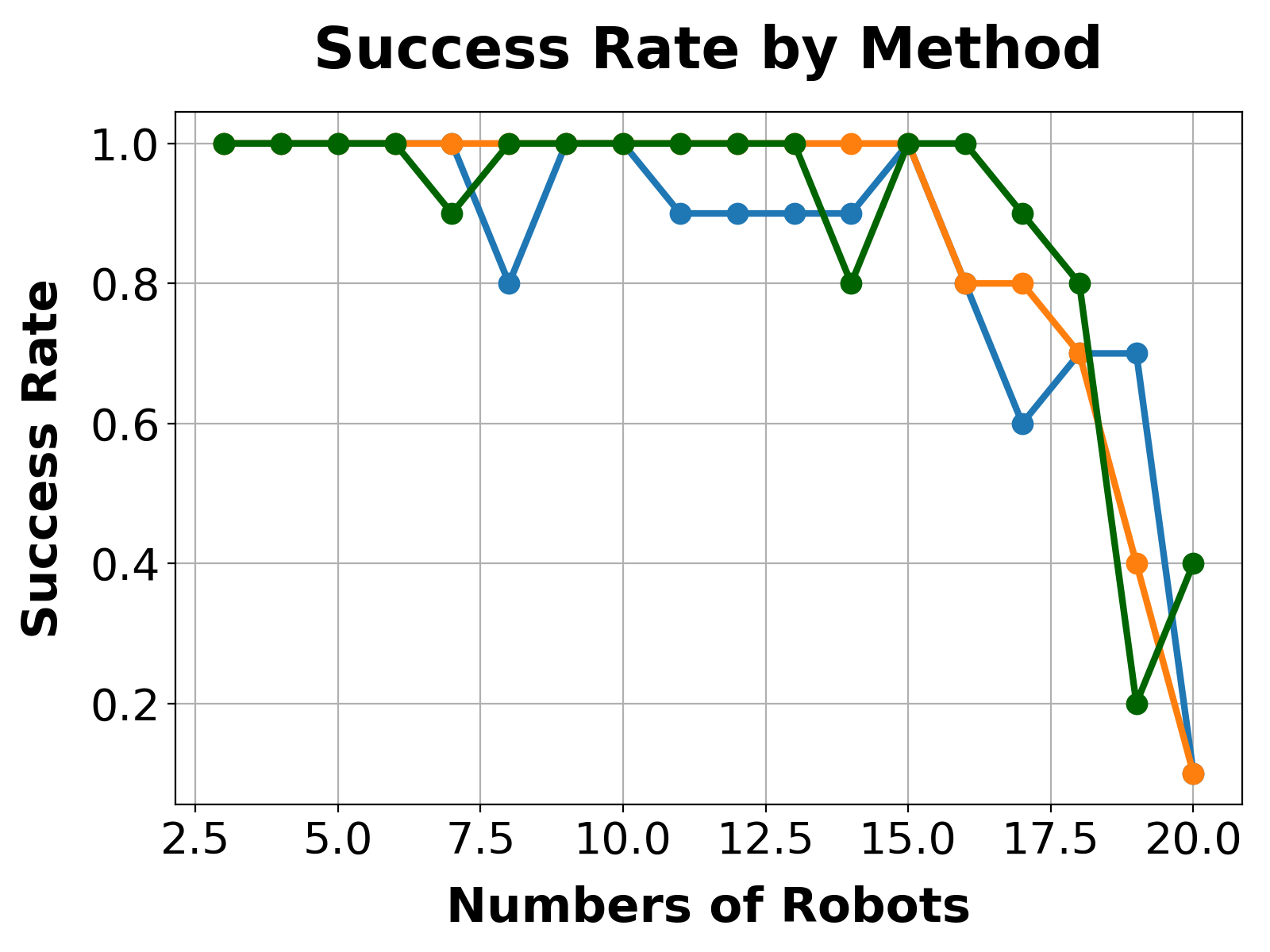}
    \end{minipage}\hfill
    \begin{minipage}[t]{\CellW}
      \vspace{0pt}\includegraphics[width=\linewidth]{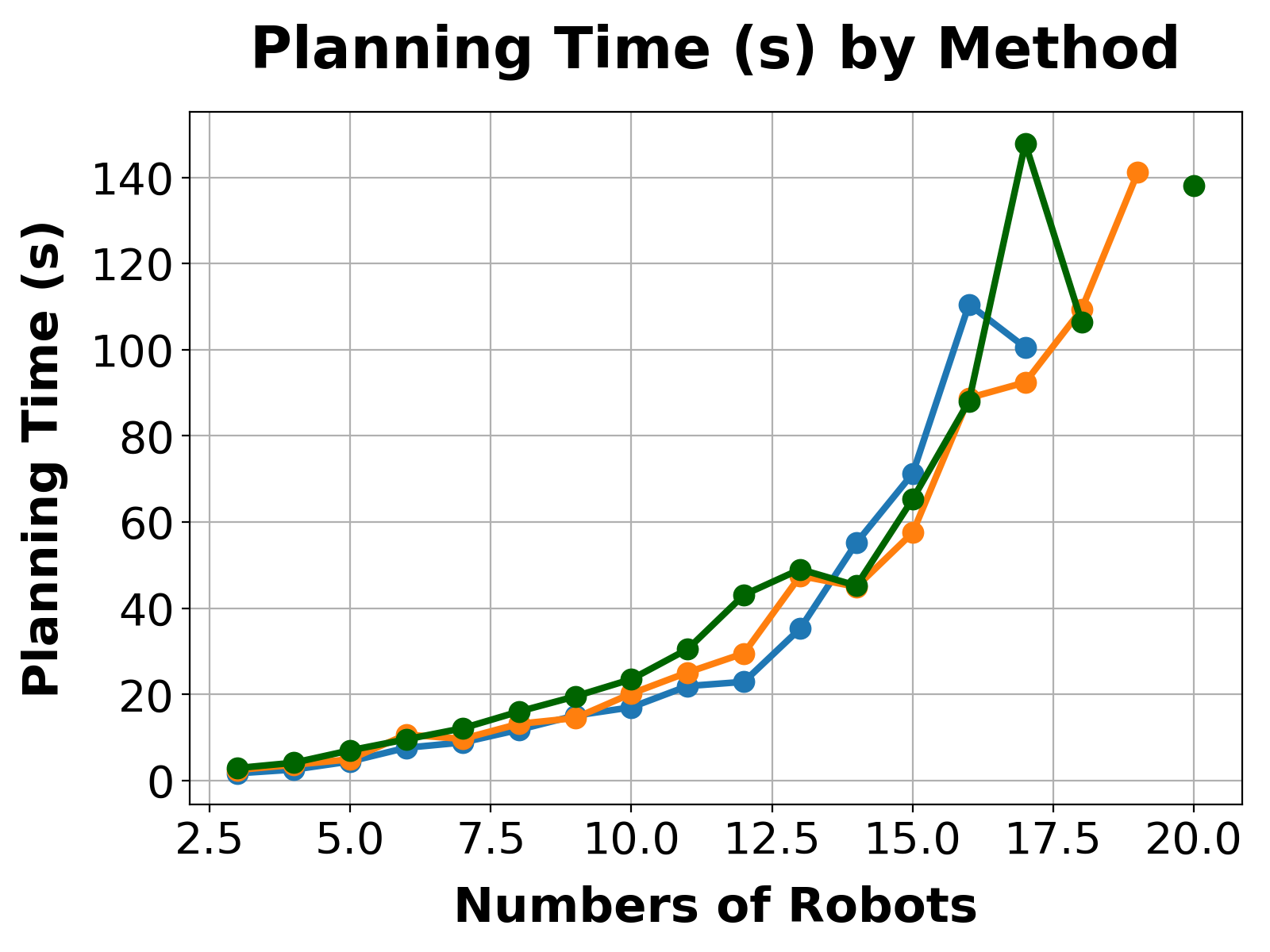}
    \end{minipage}\hfill
    \begin{minipage}[t]{\CellW}
      \vspace{0pt}\includegraphics[width=\linewidth]{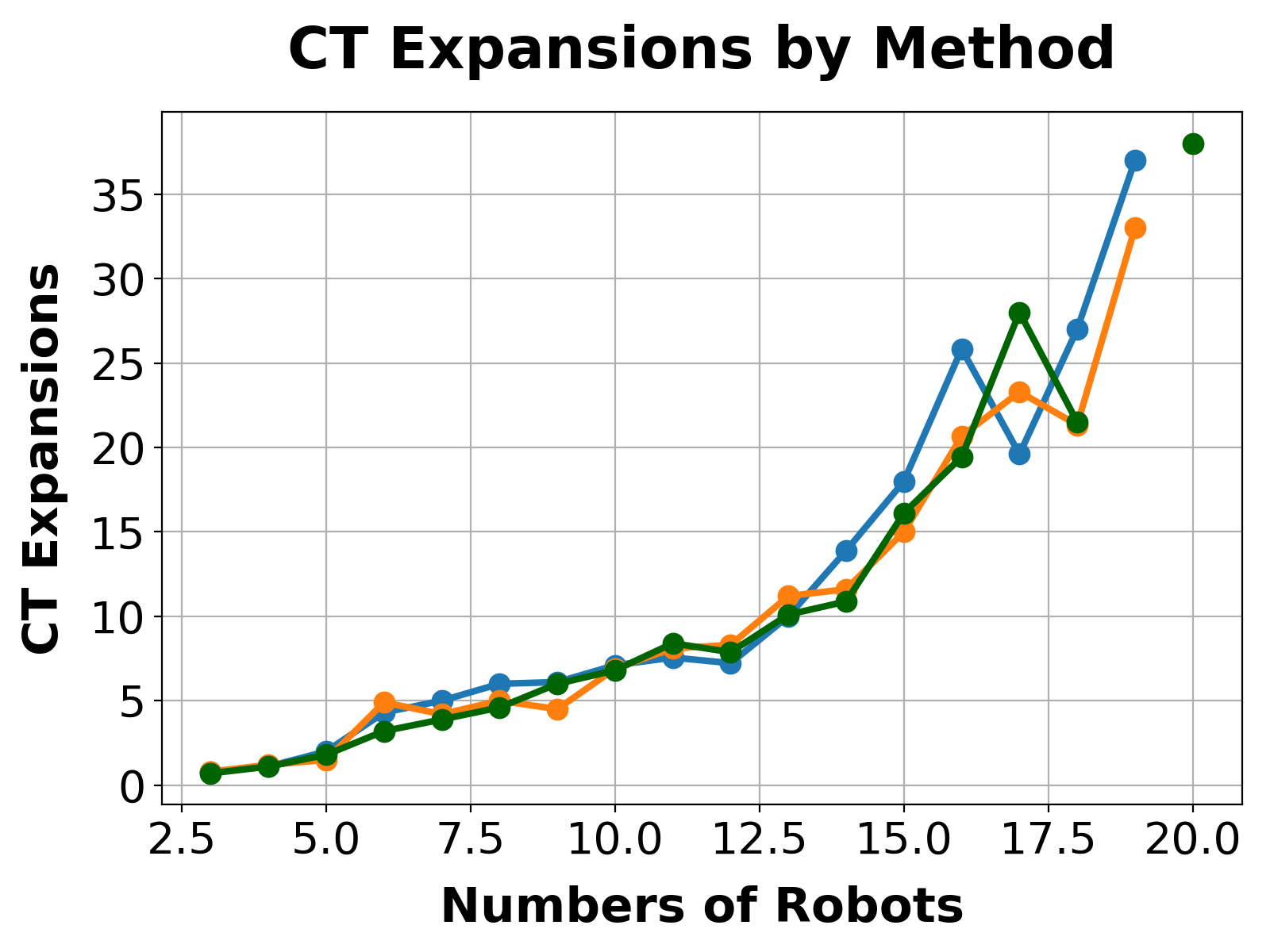}
    \end{minipage}

    \vspace{0.6em}

    \begin{minipage}[t]{\CellW}
      \vspace{0pt}\includegraphics[width=\linewidth]{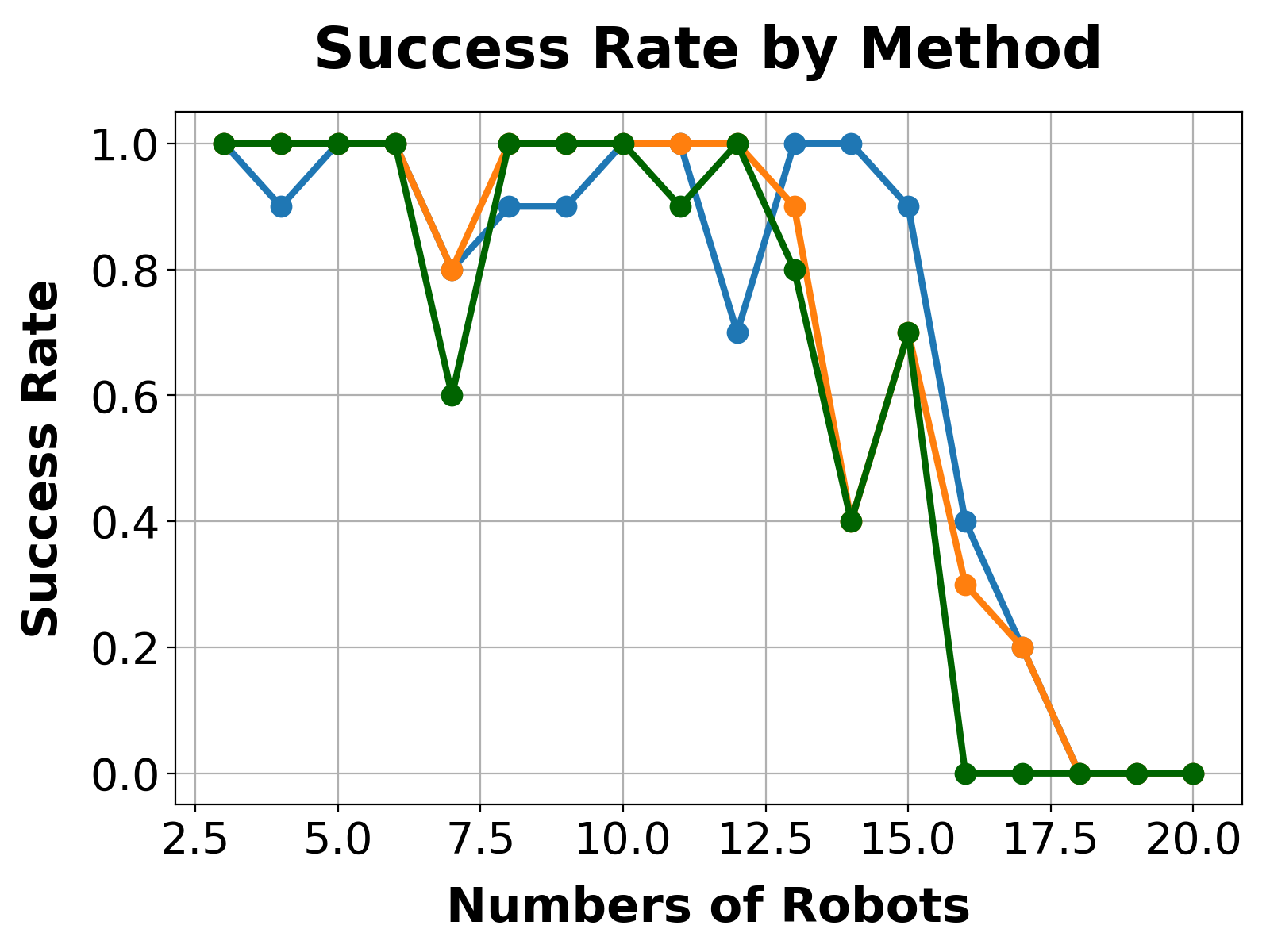}
    \end{minipage}\hfill
    \begin{minipage}[t]{\CellW}
      \vspace{0pt}\includegraphics[width=\linewidth]{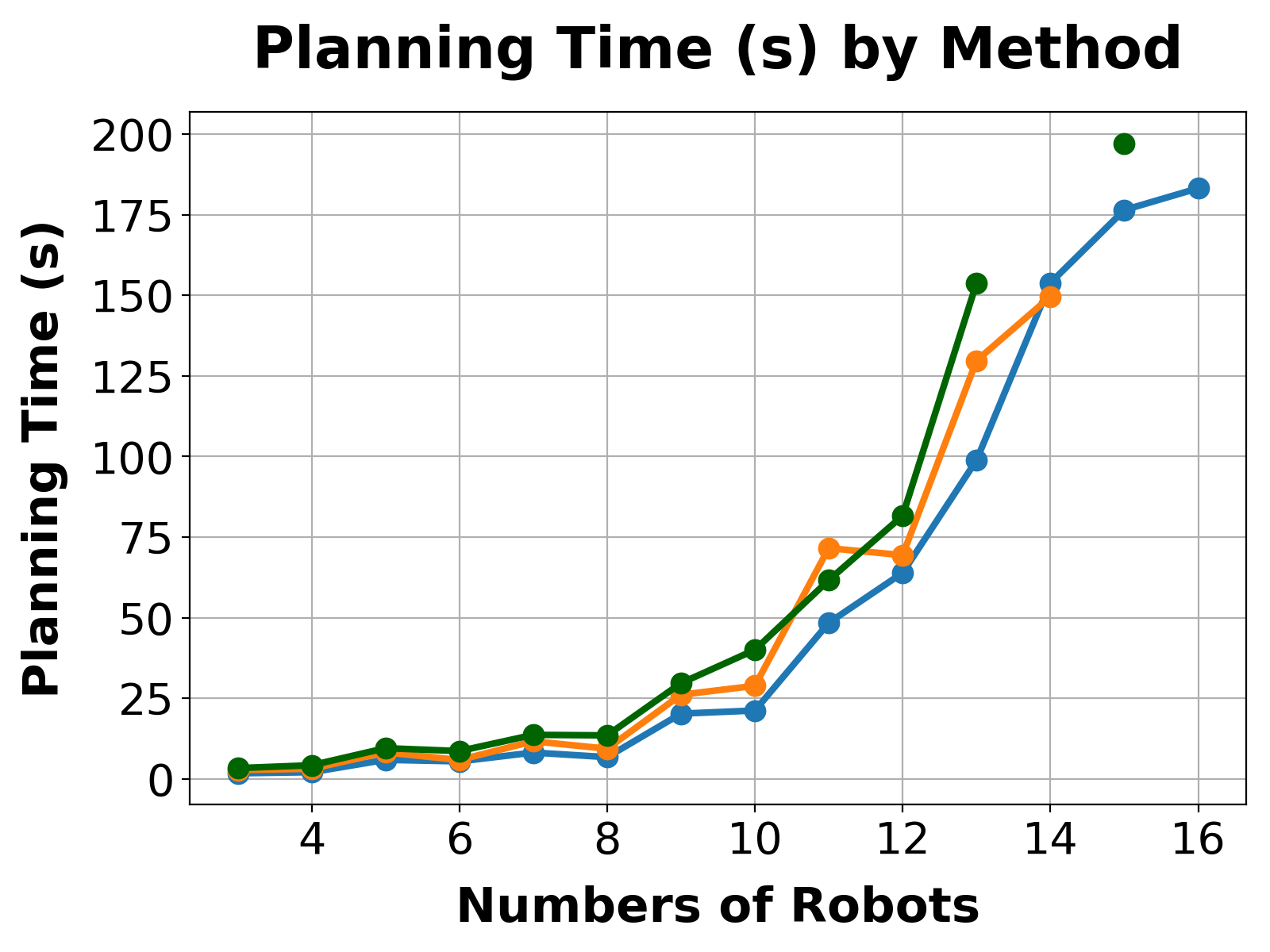}
    \end{minipage}\hfill
    \begin{minipage}[t]{\CellW}
      \vspace{0pt}\includegraphics[width=\linewidth]{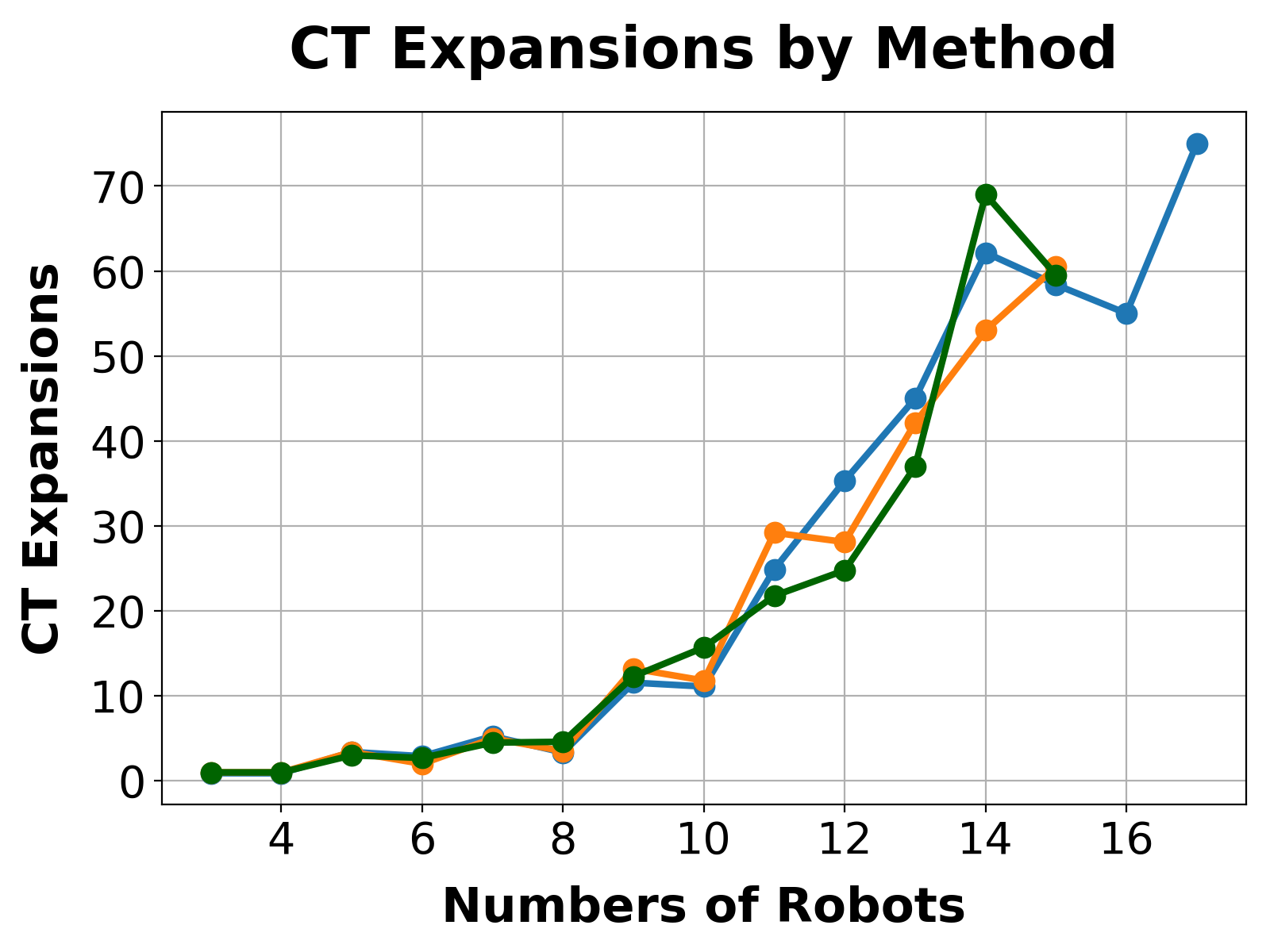}
    \end{minipage}
  \end{minipage}

  \caption{Performance of different diffusion steps 200 (MDOC-CBS-200), 300 (MDOC-CBS-300), 400 (MDOC-CBS-400) on Conveyor Circle map (Top Row), and Conveyor Weave (Below) for MRMP evaluated up to 20 robots with a 500-second time limit. Note that we use MDOC-CBS-200 as the result shown on the main page.} 
  \label{fig:additional}
\end{figure*}

%% file: figures/tex/double.tex
\begin{figure*}[t]
  \centering

  \newcommand{\LeftW}{0.20\textwidth}
  \newcommand{\RightW}{0.79\textwidth}
  \newcommand{\CellW}{0.33\linewidth}
  \newcommand{\CellP}{0.48\textwidth}
  \newcommand{\Legend}{1\textwidth}

  \begin{minipage}[t]{\LeftW}
    \vspace{0pt}%

    \begin{minipage}[t]{\CellP}
      \vspace{0pt}\includegraphics[width=\linewidth]{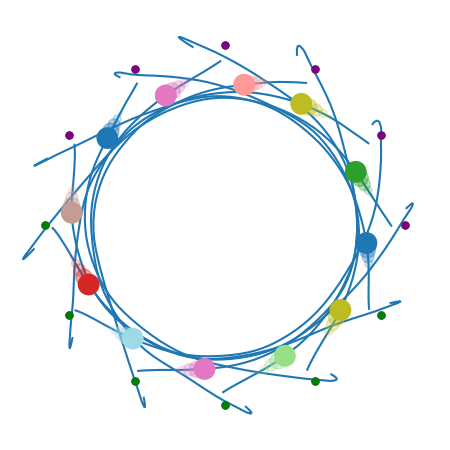}
    \end{minipage}\hfill
    \begin{minipage}[t]{\CellP}
      \vspace{0pt}\includegraphics[width=\linewidth]{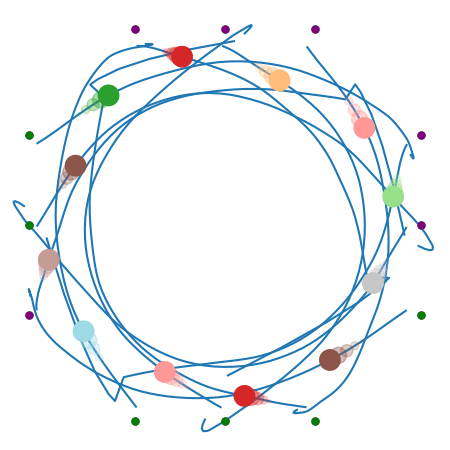}
    \end{minipage}\hfill\vspace{10pt}%

    \begin{minipage}[t]{\CellP}
    \centering
      \vspace{0pt}\includegraphics[width=\linewidth]{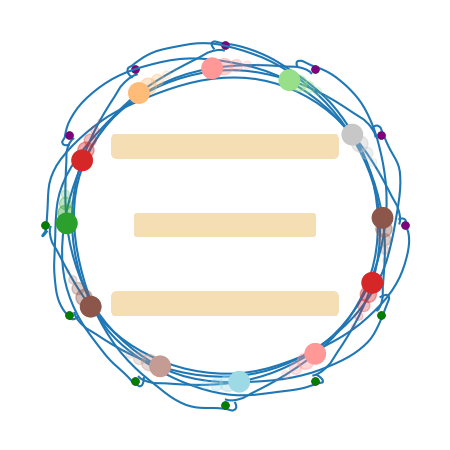}
      \textbf{Circle Setup}
    \end{minipage}\hfill
    \begin{minipage}[t]{\CellP}
    \centering
      \vspace{0pt}\includegraphics[width=\linewidth]{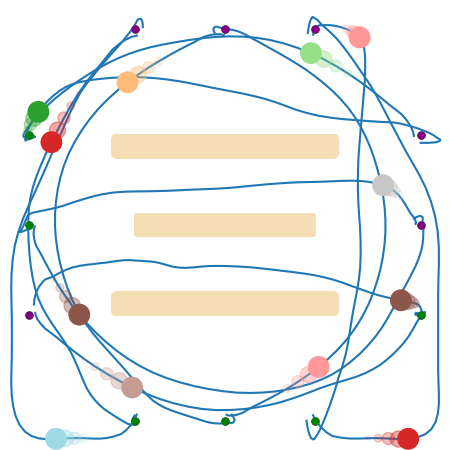}
      \textbf{Weave Setup}
    \end{minipage}\hfill

    \begin{minipage}[t]{\Legend}
      \vspace{10pt}\includegraphics[width=\linewidth]{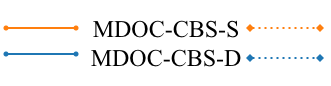} 
    \end{minipage}\hfill
    
  \end{minipage}%
  \hfill
  \begin{minipage}[t]{\RightW}
    \vspace{0pt}%

    \begin{minipage}[t]{\CellW}
      \vspace{0pt}\includegraphics[width=\linewidth]{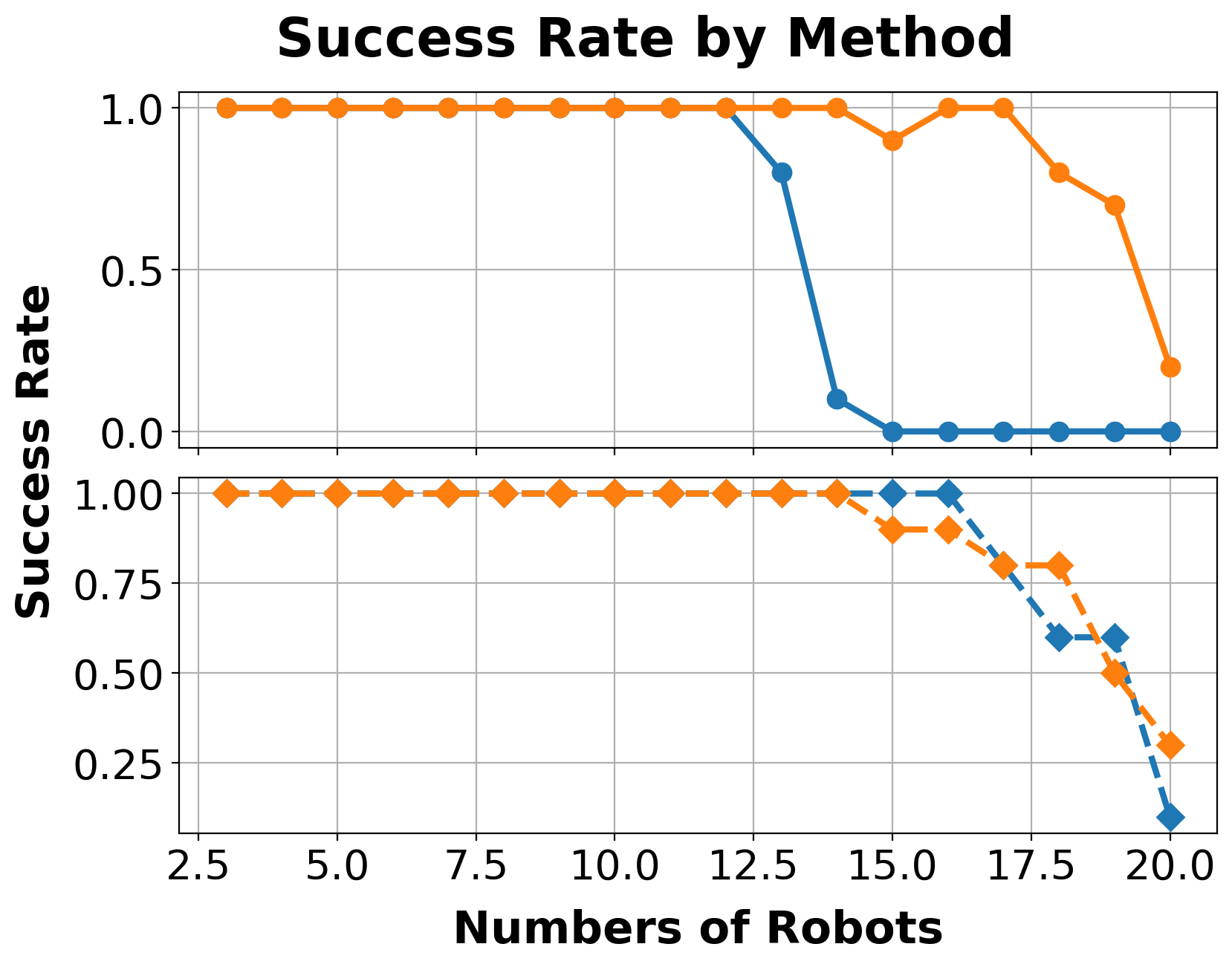}
    \end{minipage}\hfill
    \begin{minipage}[t]{\CellW}
      \vspace{0pt}\includegraphics[width=\linewidth]{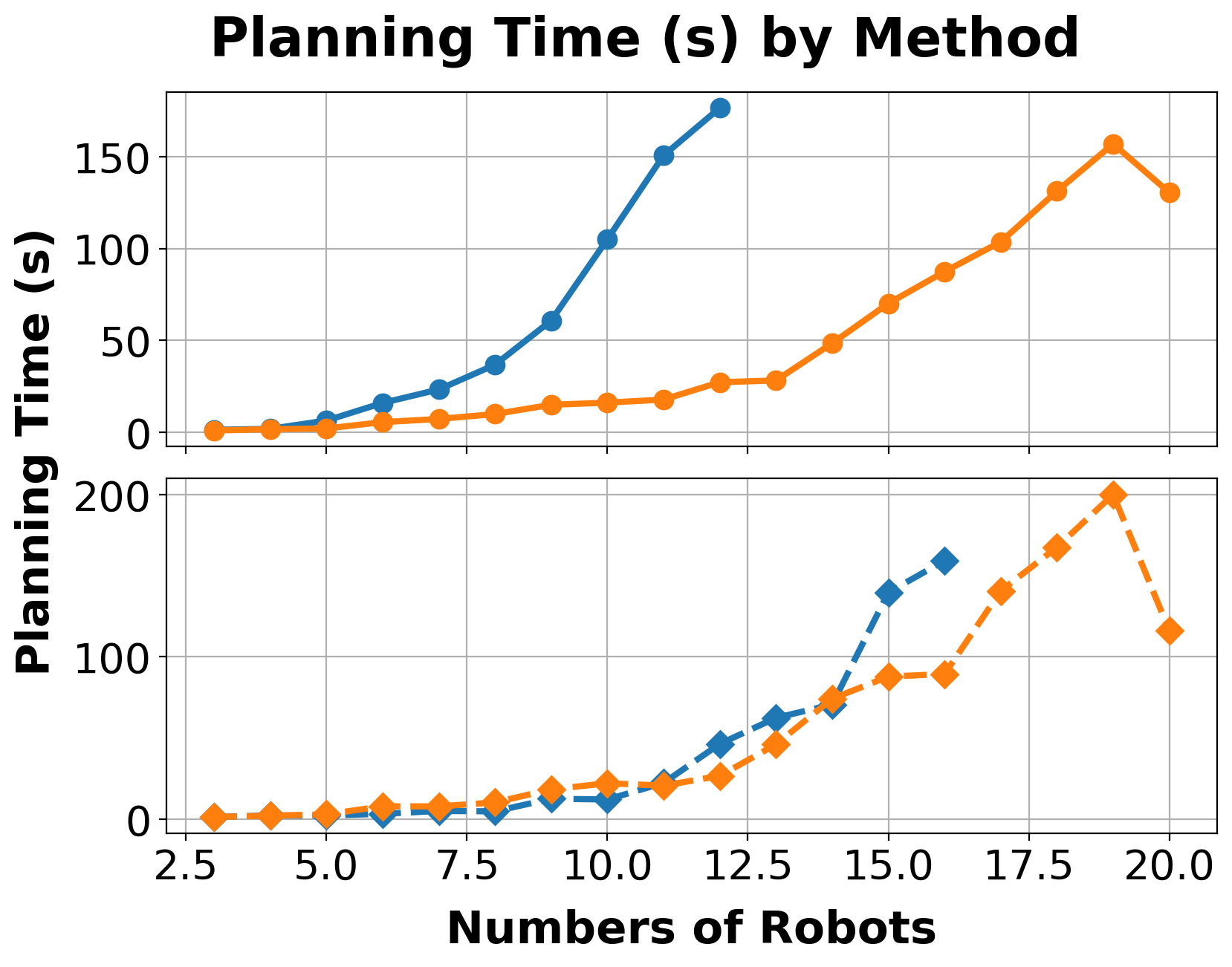}
    \end{minipage}\hfill
    \begin{minipage}[t]{\CellW}
      \vspace{0pt}\includegraphics[width=\linewidth]{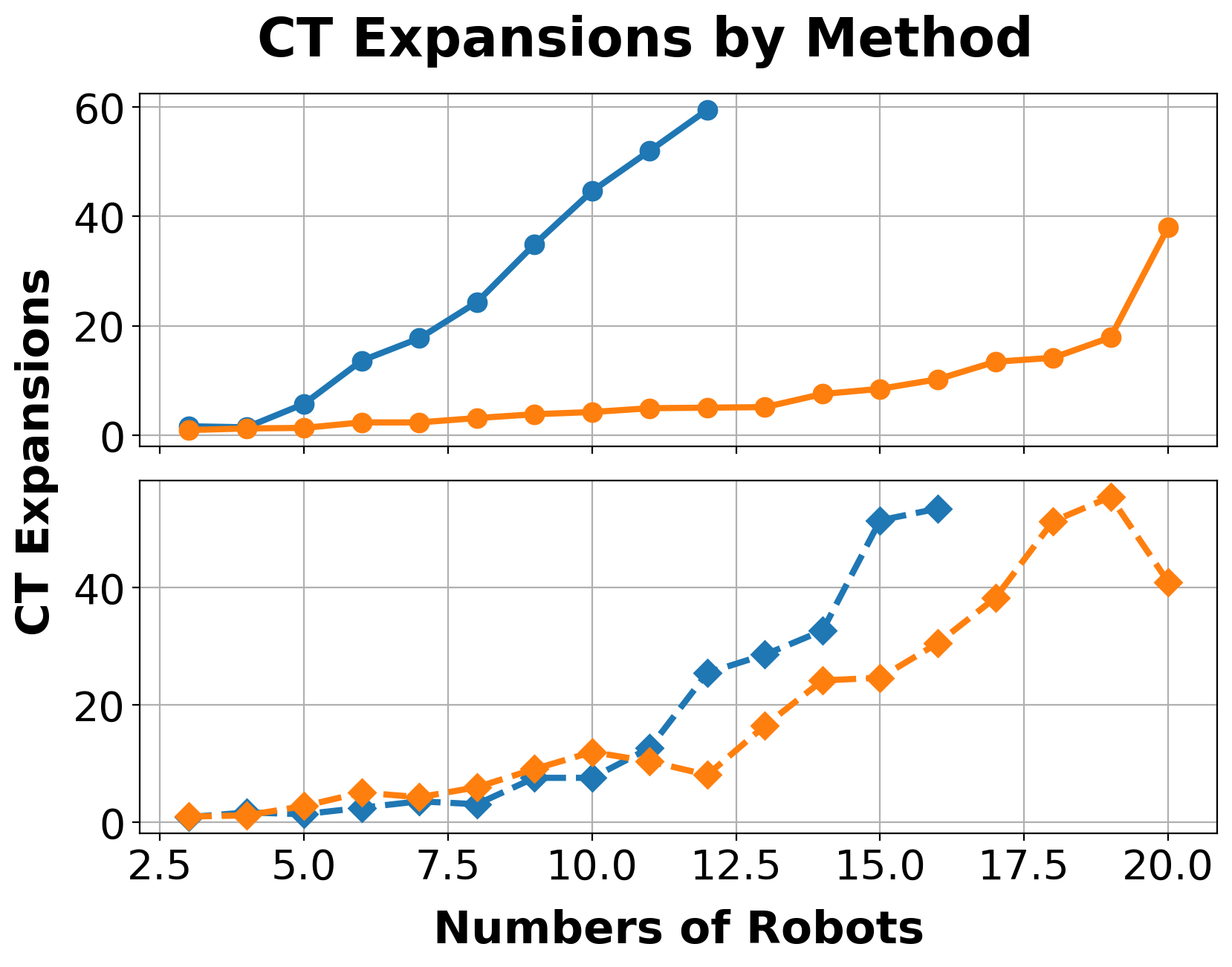}
    \end{minipage}

    \vspace{0.6em}

    \begin{minipage}[t]{\CellW}
      \vspace{0pt}\includegraphics[width=\linewidth]{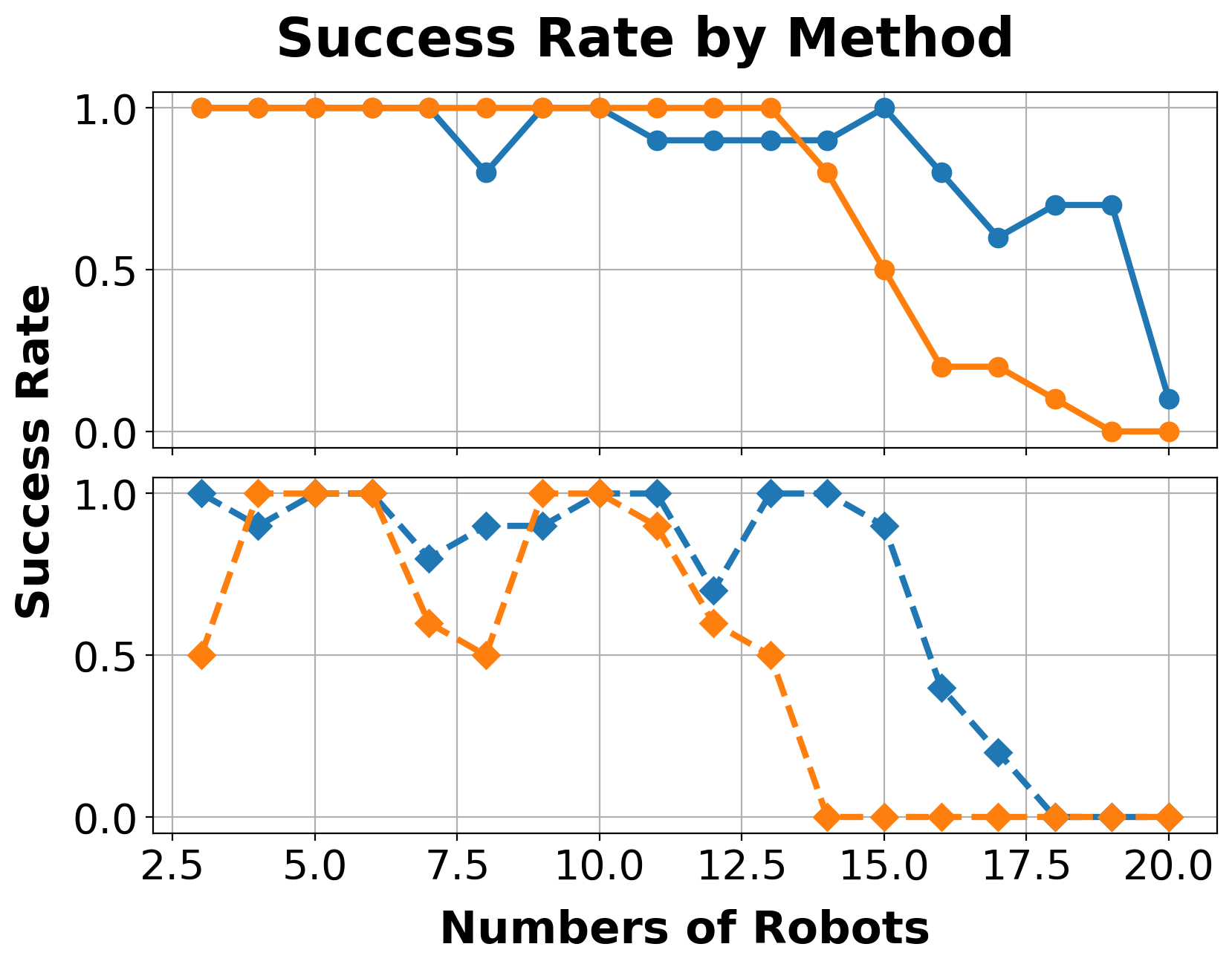}
    \end{minipage}\hfill
    \begin{minipage}[t]{\CellW}
      \vspace{0pt}\includegraphics[width=\linewidth]{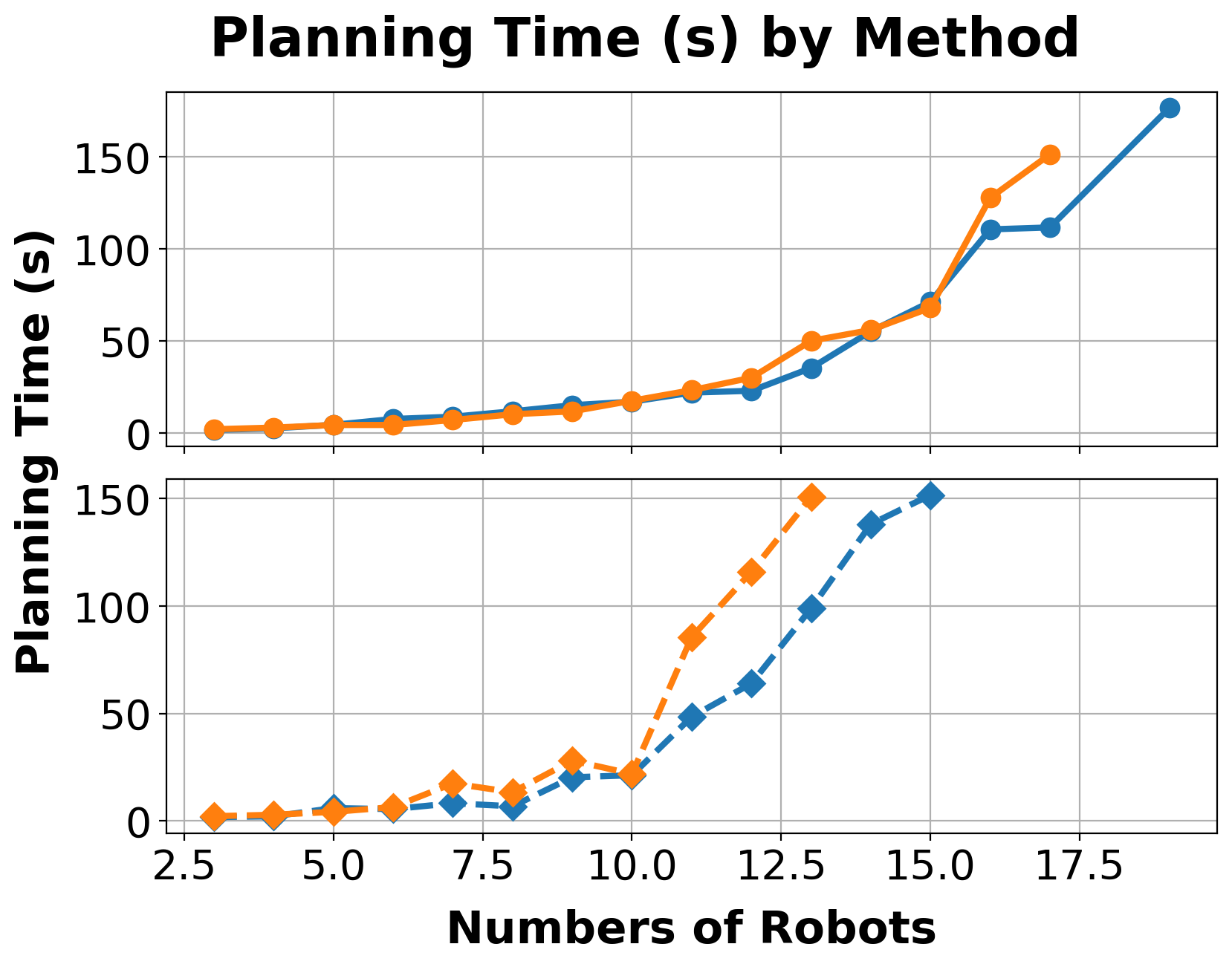}
    \end{minipage}\hfill
    \begin{minipage}[t]{\CellW}
      \vspace{0pt}\includegraphics[width=\linewidth]{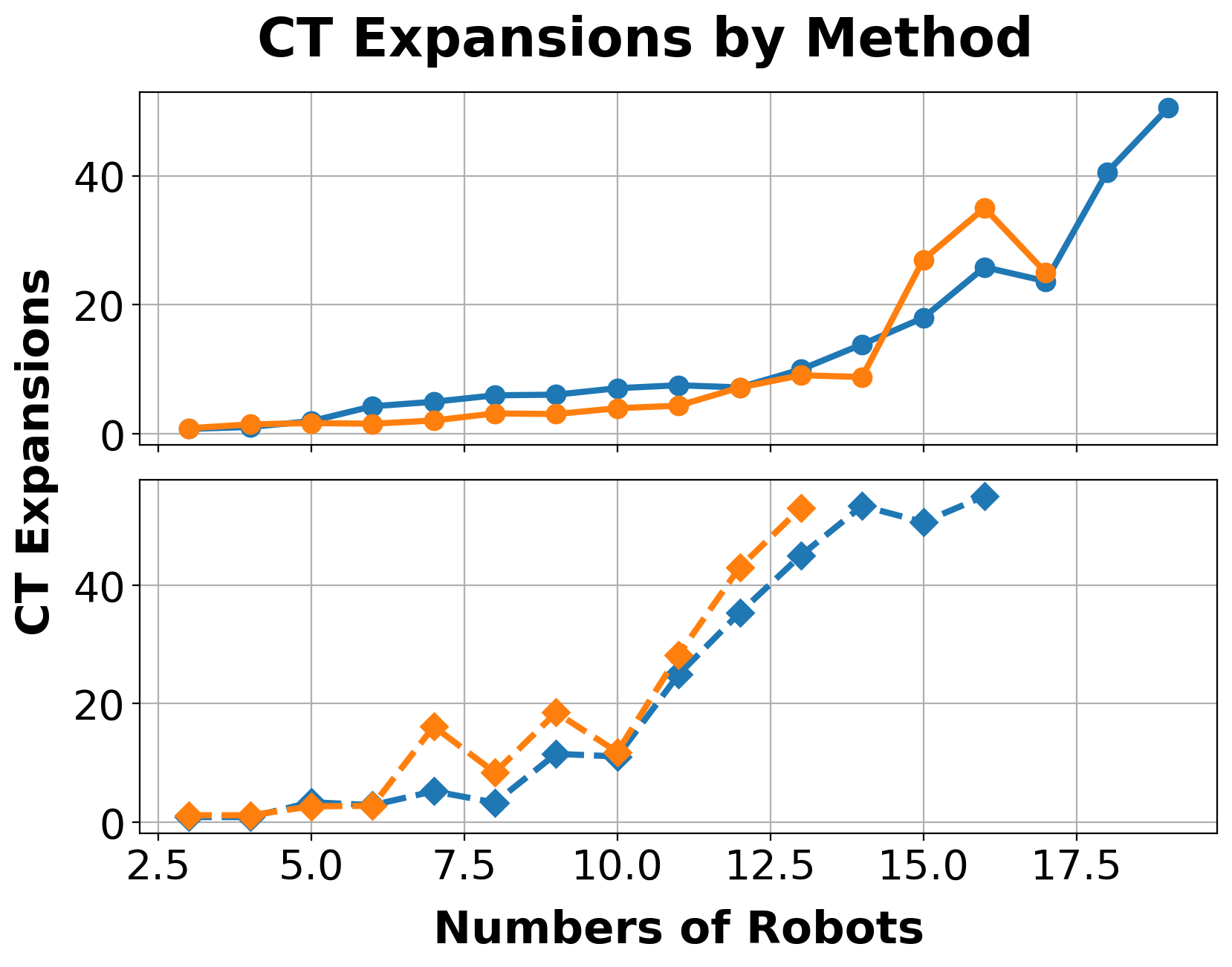}
    \end{minipage}
  \end{minipage}

  \caption{Performance of different dynamics, single-integrator (MDOC-CBS-S) and double-integrator (MDOC-CBS-D), on Empty map (Top Row) and Conveyor map (Below) up to 20 robots with a 500-second time limit. We showcase Circle Setup (circular markers with solid line) at the Top Row of each subfigure, while Weave Setup (diamond markers with dashed line) at the Bottom of each subfigure for each map. }
  \label{fig:double_empty_conveyor}
\end{figure*}

%% file: figures/tex/double_traj.tex
\begin{figure}[tb!]
  \centering
  \caption{Benchmark results on Empty map (Top Row) and Conveyor Map (Below) with 12 robots on Circle setup. Columns display different dynamics: single-integrator (Left) and double-integrator (Right). We show trajectories after post-smoothness.}
  \label{fig:double_path}
  \begin{subfigure}[b]{0.23\textwidth}
    \centering
    \includegraphics[width=0.8\linewidth]{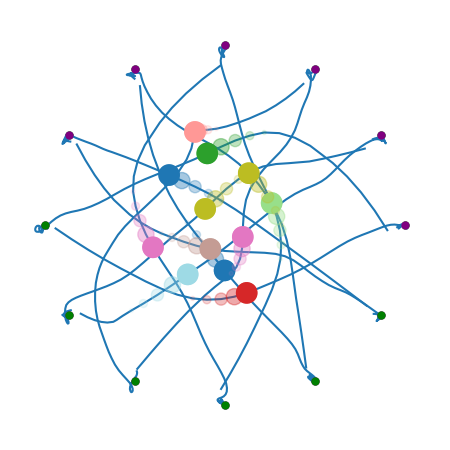}
  \end{subfigure}\hfill
  \begin{subfigure}[b]{0.23\textwidth}
    \centering
    \includegraphics[width=0.8\linewidth]{figures/app_double/mdoc_double_empty.png}
  \end{subfigure}\hfill
  \begin{subfigure}[b]{0.23\textwidth}
    \centering
    \includegraphics[width=0.8\linewidth]{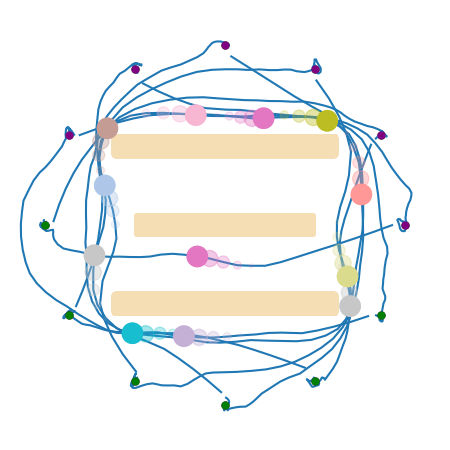}
  \end{subfigure}\hfill
  \begin{subfigure}[b]{0.23\textwidth}
    \centering
    \includegraphics[width=0.8\linewidth]{figures/app_double/mdoc_double_conveyor.png}
  \end{subfigure}\hfill
  \captionsetup{type=table}
  \begin{tabular}{c|c|c|c}
    \toprule
    Env & Method & {\textbf{P}$\downarrow$} & {\textbf{G }(1e-4)$\downarrow$} \\
    \midrule
    \multirow{2}{*}{Empty} 
      & Single   & \textbf{1.8 $\pm$ 0.03} & \textbf{6.7 $\pm$ 0.05}  \\
      & Double  & 2.5 $\pm$ 0.03 & 27.8 $\pm$ 13.80  \\
    \midrule
    \multirow{2}{*}{Conveyor} 
      & Single  & 2.8  $\pm$ 0.04 & 12.9 $\pm$ 1.40 \\
      & Double   & \textbf{2.5 $\pm$ 0.03} & \textbf{3.78 $\pm$ 1.71} \\
    \bottomrule
    \end{tabular}
  \caption{Comparison of single-integrator and double-integrator in terms of average path length \textbf{P} and geometric (Laplacian) smoothness \textbf{G} over success trails of all agent numbers, evaluated on the Empty and Conveyor maps in the Circle setup.}
\end{figure}

%% file: figures/tex/ecbs.tex
\begin{figure*}[t]
  \centering

  \newcommand{\LeftW}{0.16\textwidth}
  \newcommand{\RightW}{0.83\textwidth}
  \newcommand{\CellW}{0.31\linewidth}

  \begin{minipage}[t]{\LeftW}
    \vspace{0pt}%
    \centering
    \includegraphics[width=0.7\linewidth]{figures/exp2/empty/mdoc_single_trial_circle.png}
    \textbf{Empty Circle}\par\vspace{0.6em}%
    \includegraphics[width=0.7\linewidth]{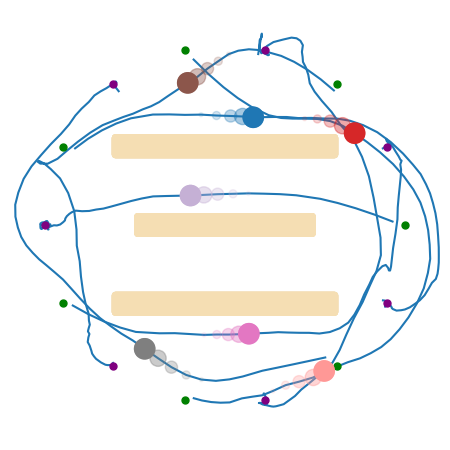}
    \textbf{Conveyor Circle}\par\vspace{0.6em}%
    \includegraphics[width=0.9\linewidth]{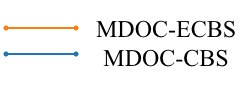}

  \end{minipage}%
  \hfill
  \begin{minipage}[t]{\RightW}
    \vspace{0pt}%

    \begin{minipage}[t]{\CellW}
      \vspace{0pt}\includegraphics[width=\linewidth]{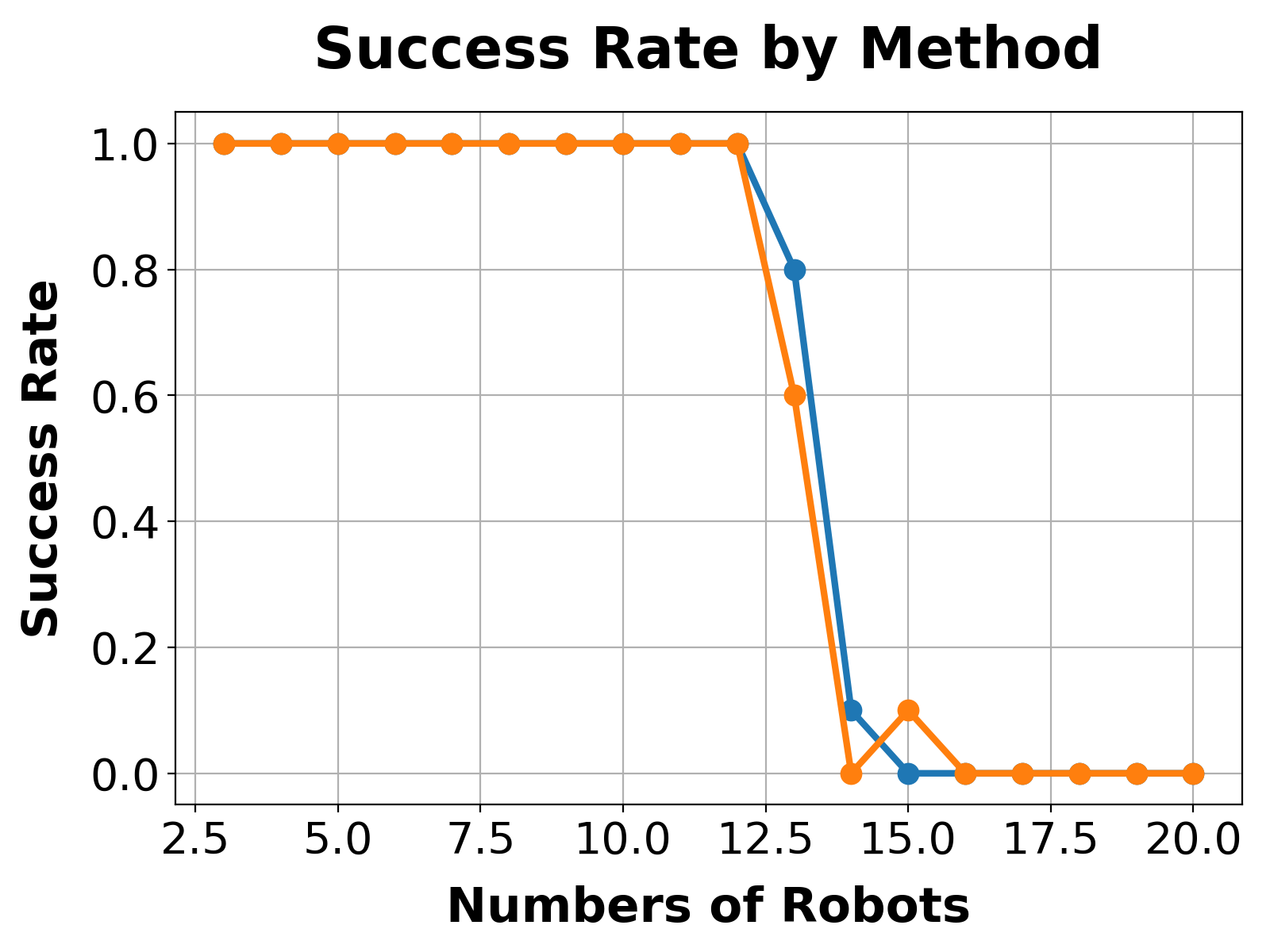}
    \end{minipage}\hfill
    \begin{minipage}[t]{\CellW}
      \vspace{0pt}\includegraphics[width=\linewidth]{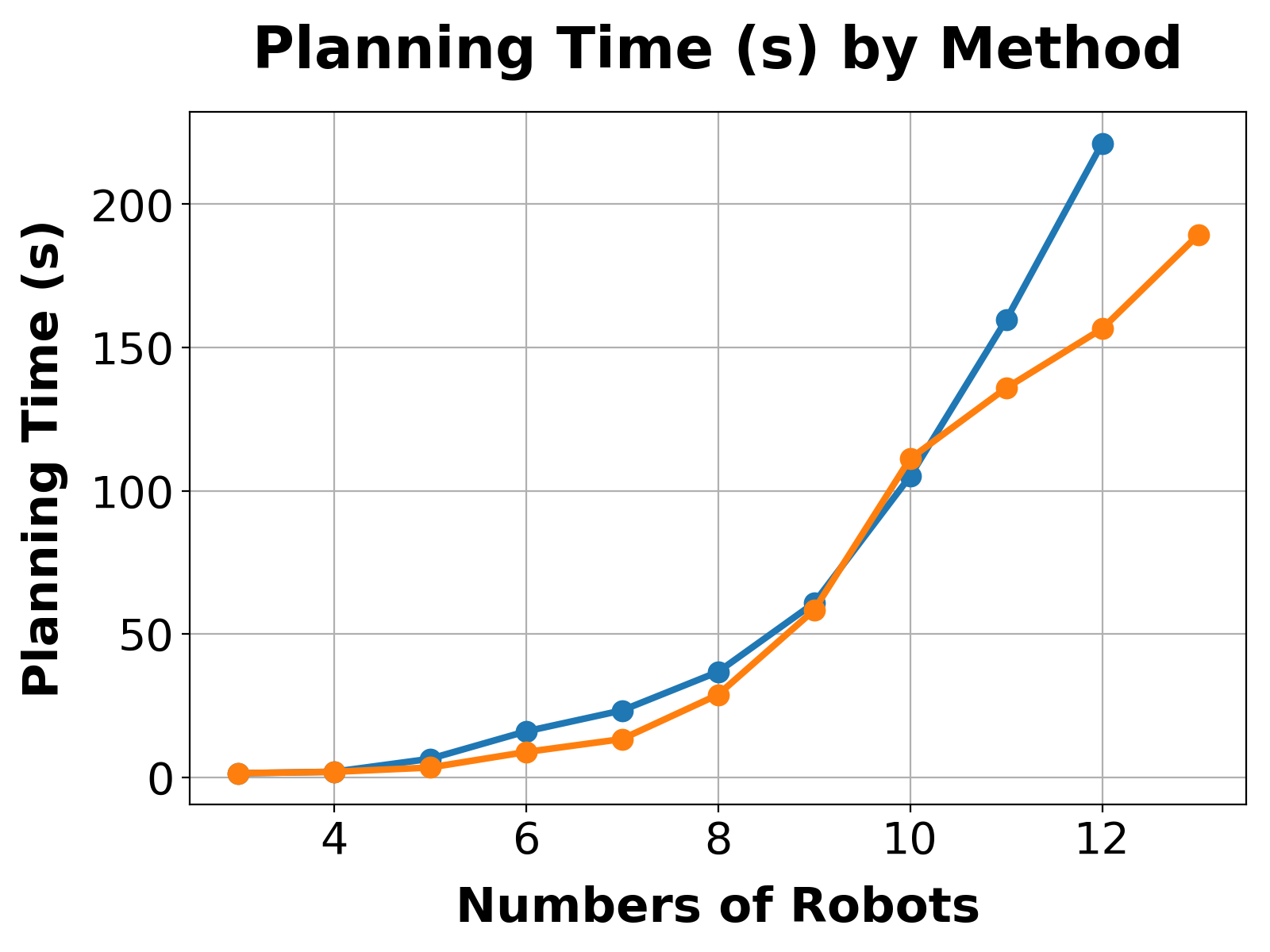}
    \end{minipage}\hfill
    \begin{minipage}[t]{\CellW}
      \vspace{0pt}\includegraphics[width=\linewidth]{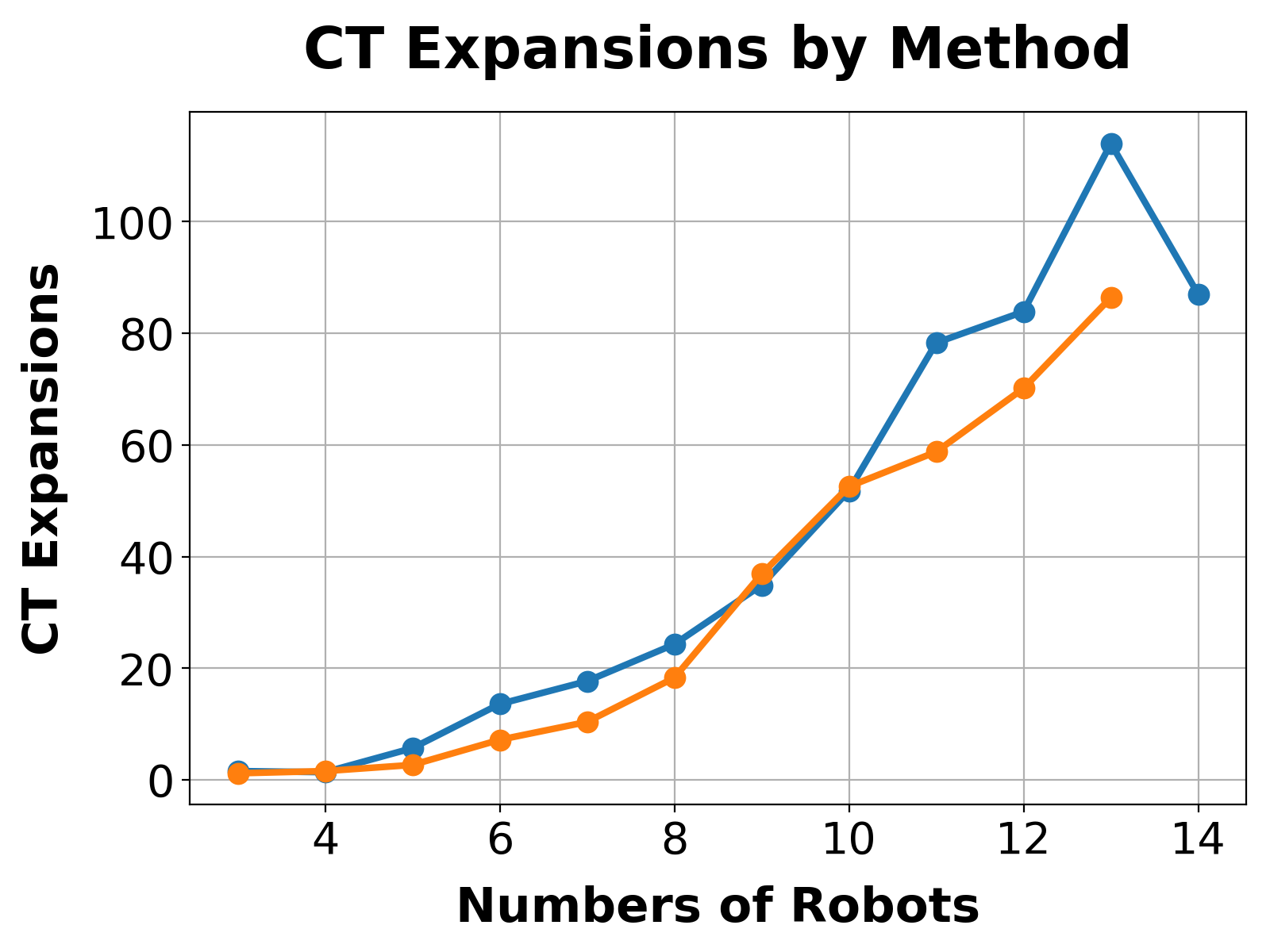}
    \end{minipage}

    \vspace{0.6em}

    \begin{minipage}[t]{\CellW}
      \vspace{0pt}\includegraphics[width=\linewidth]{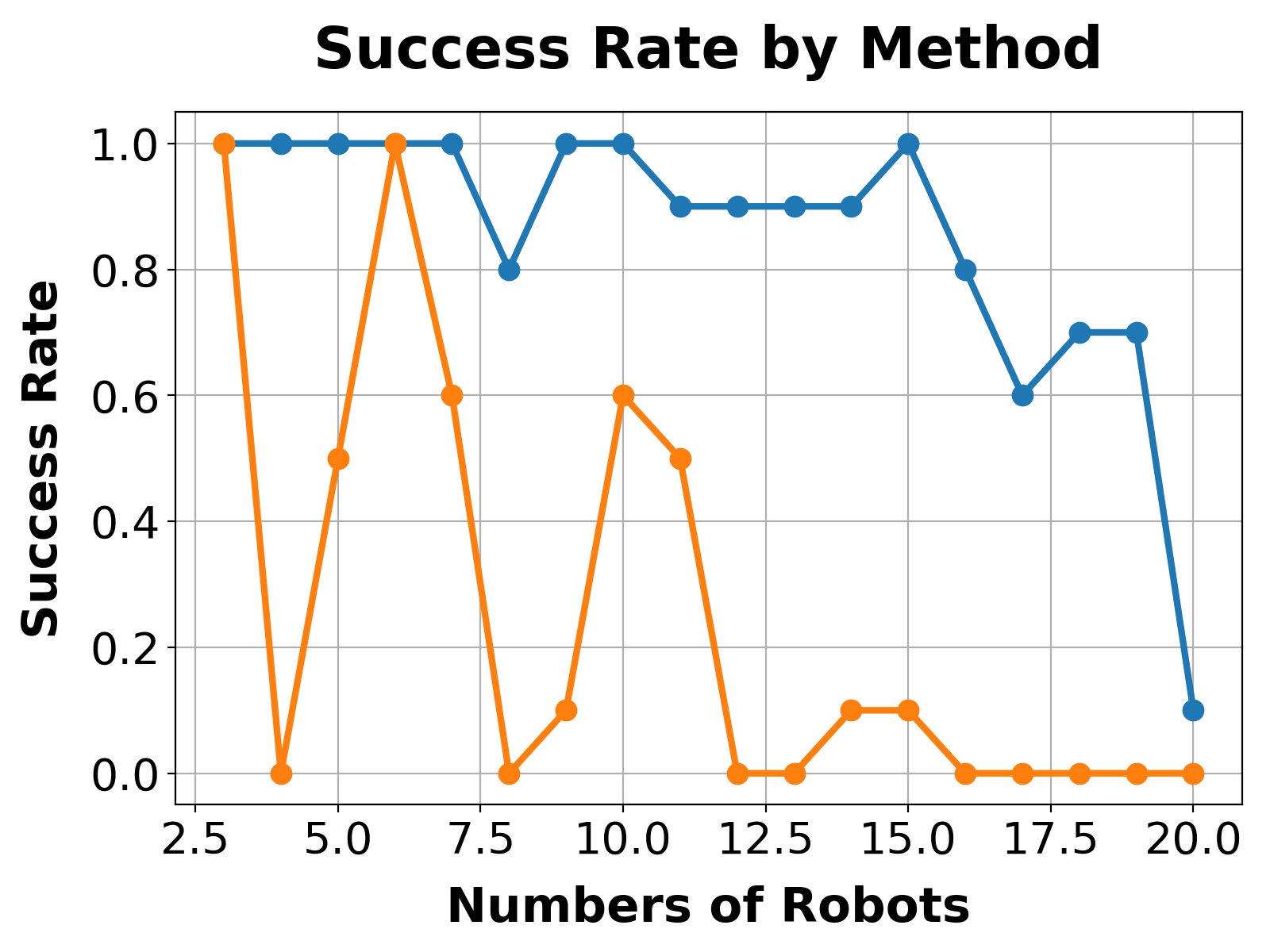}
    \end{minipage}\hfill
    \begin{minipage}[t]{\CellW}
      \vspace{0pt}\includegraphics[width=\linewidth]{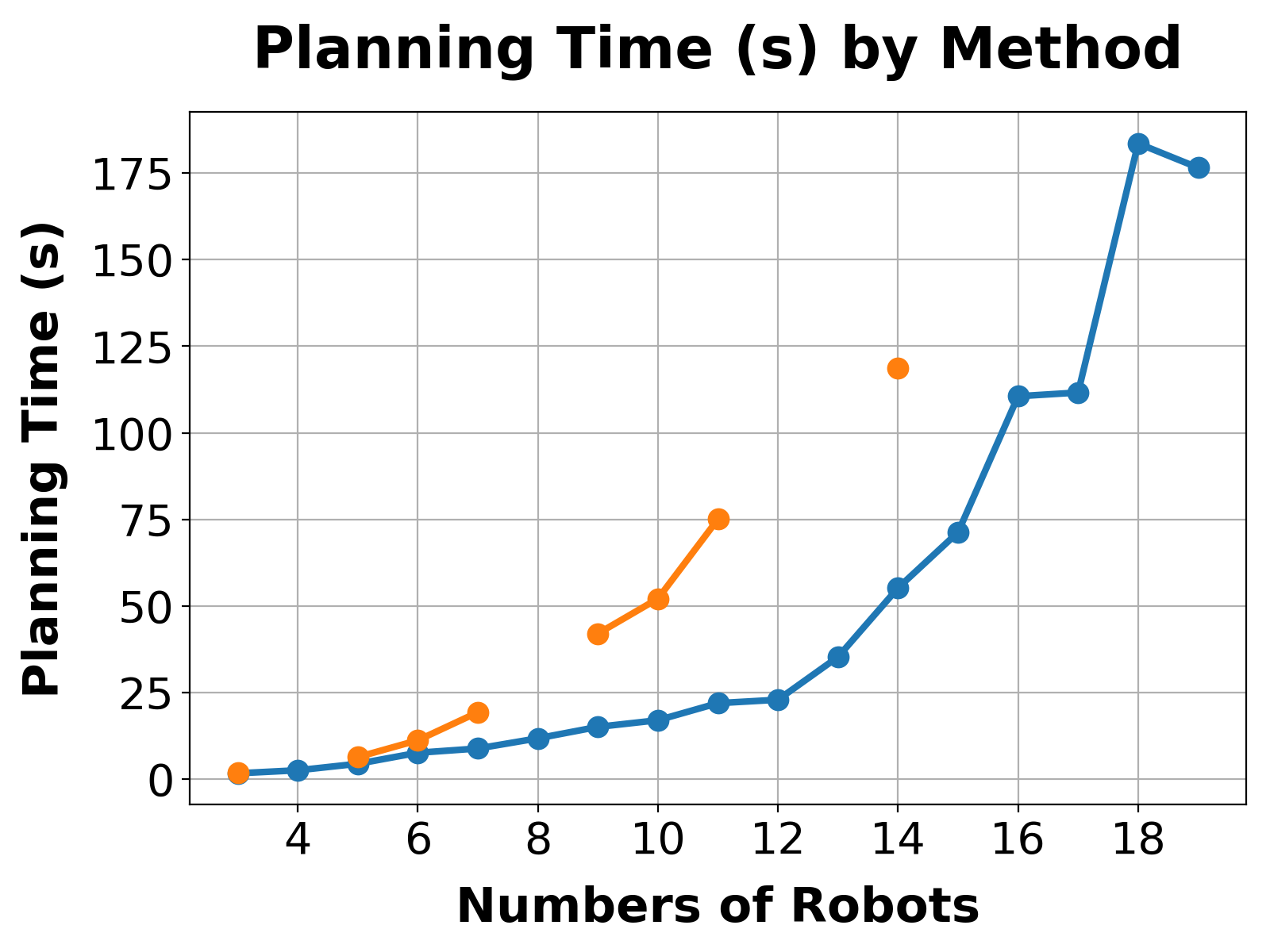}
    \end{minipage}\hfill
    \begin{minipage}[t]{\CellW}
      \vspace{0pt}\includegraphics[width=\linewidth]{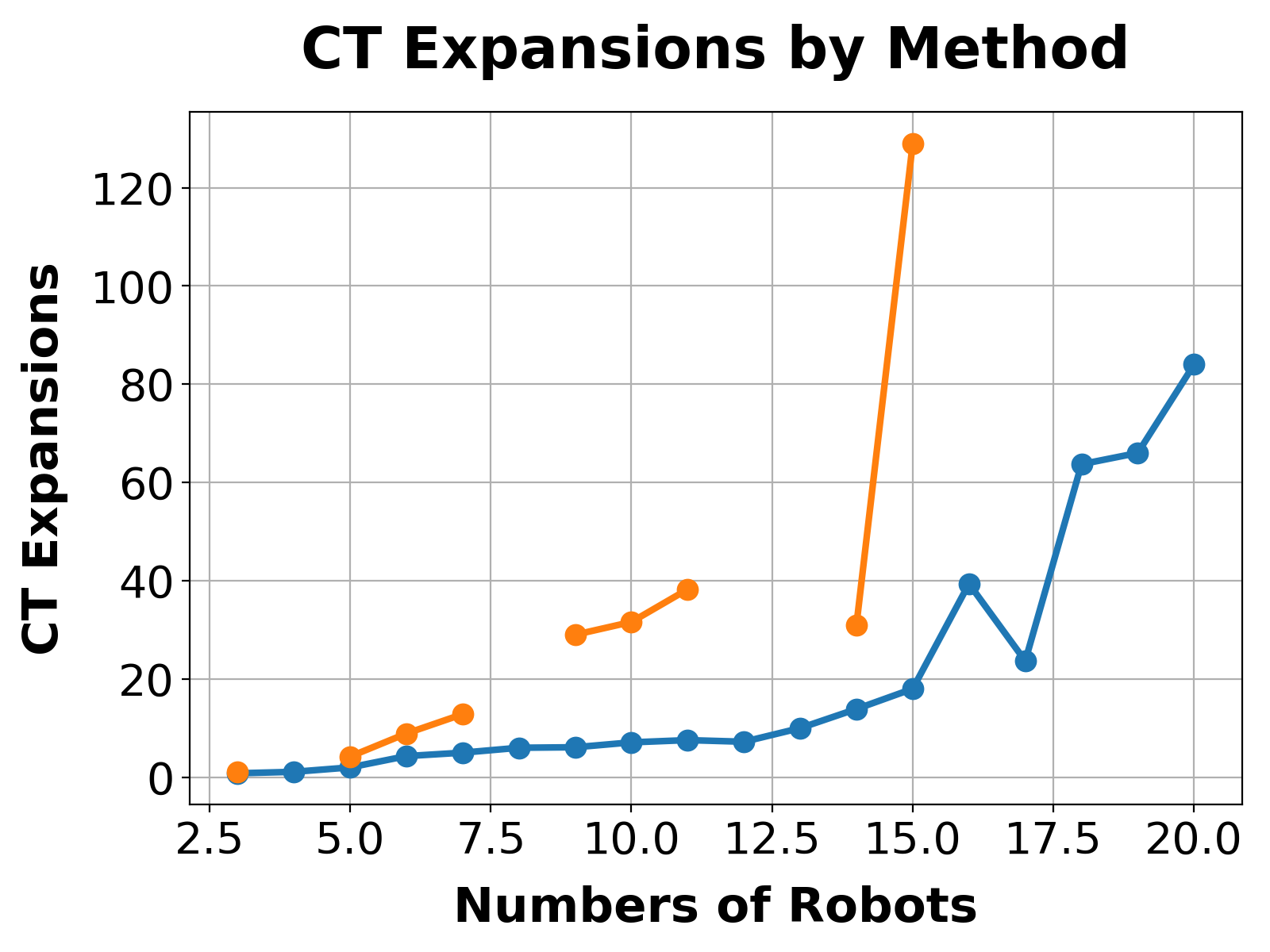}
    \end{minipage}
  \end{minipage}

  \caption{Performance of ECBS as a high-level planner on Empty Circle map (Top Row), and Conveyor Circle (Below) for MRMP evaluated up to 20 robots with a 500-second time limit.}
  \label{fig:aaai_grid_combo}
\end{figure*}

%% file: figures/tex/app_results.tex
\begin{table*}[tb!]
\centering
\small
\scriptsize  
\setlength{\tabcolsep}{5pt} 
\renewcommand{\arraystretch}{0.3}

\begin{tabular}{c|c|ccccc|ccccc}
    \toprule
    & & \multicolumn{5}{c|}{Empty Map (Circle)} 
      & \multicolumn{5}{c}{Empty Map (Weave)} \\
    \cmidrule(lr){3-7} \cmidrule(lr){8-12} 
    N & Method 
      & \textbf{S}($\uparrow$) & \textbf{T}($\downarrow$) & \textbf{E}($\downarrow$) & \textbf{P}($\downarrow$) & \textbf{G}($\downarrow$)  
      & \textbf{S}($\uparrow$) & \textbf{T}($\downarrow$) & \textbf{E}($\downarrow$) & \textbf{P}($\downarrow$) & \textbf{G}($\downarrow$)  \\
    \midrule
    \multirow{5}{*}{3}
      & A*-CBS & 100 & 0.29 $\pm$ 0.0 & 1.0  & 2.0 & 1.4e-3 & 100 & 0.6 $\pm$ 0.2 & 7.7 $\pm$ 2.9 & 1.9 $\pm$ 0.0 & 1.5e-3 $\pm$ 1e-4 \\
     & A*-ECBS & 100 & 0.35 $\pm$ 0.0 & 1.0 & 2.0 & 1.4e-3 & 100 & 0.1 $\pm$ 0.0 & 1.0 & 1.87 & 1.4e-3\\
     & KCBS & 90 & 0.1 $\pm$ 0.0 & 1.1 $\pm$ 0.6 & 2.0 $\pm$ 0.1 & 2.2e-3 $\pm$ 3e-4 & 90 & 0.1 $\pm$ 0.1 & 0.6 $\pm$ 0.7 & 2.1 $\pm$ 0.1 & 2.3e-3 $\pm$ 4e-4 \\
     & MMD-CBS & 100 & 20.8 $\pm$ 0.7 & 2.0 & 2.8 $\pm$ 0.0 & 4.8e-3 $\pm$ 3e-4 & 100 & 14.9 $\pm$ 0.3 & 1.0 & 2.8 $\pm$ 0.1 & 4.0e-3 $\pm$ 4e-4 \\
     & MDOC-CBS & 100 & 1.4 $\pm$ 0.6 & 1.6 $\pm$ 1.0 & 1.7 $\pm$ 0.0 & 6.3e-5 $\pm$ 2e-6 & 100 & 1.2 $\pm$ 0.3 & 0.9 $\pm$ 0.3 & 1.9 $\pm$ 0.0 & 6.2e-5 $\pm$ 3e-6 \\
    \midrule
     \multirow{5}{*}{4}
      & A*-CBS & 100 & 16.5 $\pm$ 14.7 & 211.7 $\pm$ 176.1 & 1.72 $\pm 0.0$ & 2.6e-3 $\pm$ 6e-4 & 100 & 21.0 $\pm$ 23.7 & 234.0$\pm$ 236.3 & 1.9 & 2.0e-3 $\pm$ 1e-4 \\
     & A*-ECBS & 100 & 1.5 $\pm$ 0.6 & 13.7 $\pm$ 5.0 & 1.7 $\pm$ 0.1 & 2.9e-3 $\pm$ 5e-4 & 100 & 1.8 $\pm$ 0.6 & 14.5 $\pm$ 4.5 & 2.0 $\pm$ 0.1 & 2.6e-3 $\pm$ 6e-4 \\
     & KCBS & 80 & 0.3 $\pm$ 0.1 & 3.4 $\pm$ 1.4 & 2.0 $\pm$ 0.1 & 2.1e-3 $\pm$ 3e-4 & 70 & 0.3 $\pm$ 0.1 & 3.3 $\pm$ 1.4 & 2.2 $\pm$ 0.1 & 2.1e-3 $\pm$ 3e-4 \\
     & MMD-CBS & 100 & 32.1 $\pm$ 1.0 & 3.0 & 2.9 $\pm$ 0.0 & 5.5e-3 $\pm$ 2e-4 & 100 & 32.5 $\pm$ 0.7 & 3.0 & 2.9 $\pm$ 0.1 & 5.6e-3 $\pm$ 6e-4 \\
     & MDOC-CBS & 100 & 2.0 $\pm$ 1.0 & 1.4 $\pm$ 0.8 & 1.7 $\pm$ 0.0 & 5.9e-5 $\pm$ 2e-6 & 100 & 2.2 $\pm$ 0.7 & 1.7 $\pm$ 0.7 & 1.9 $\pm$ 0.0 & 6.2e-5 $\pm$ 3e-6 \\
    \midrule
    \multirow{5}{*}{5}
      & A*-CBS & 100 & 0.5 $\pm$ 0.1 & 2.5 $\pm$ 0.5 & 2.0 & 1.81e-3 & 100 & 0.4 $\pm$ 0.2 & 5.9 $\pm$ 2.5 & 1.8 $\pm$ 0.0  & 1.1e-3 $\pm$ 7e-5 \\
     & A*-ECBS & 100 & 0.7 $\pm$ 0.1 & 2.2 $\pm$ 0.4 & 2.0 & 1.8e-3 & 100 & 0.2 $\pm$ 0.0 & 1.0 & 1.8 & 1.1e-3 \\
     & KCBS & 80 & 0.5 $\pm$ 0.3 & 6.0 $\pm$ 3.0 & 2.0 $\pm$ 0.1 & 2.1e-3 $\pm$ 3e-4 & 80 & 0.3 $\pm$ 0.1 & 2.5 $\pm$ 1.8 & 2.1 $\pm$ 0.1 & 2.2e-3 $\pm$ 3e-4 \\
     & MMD-CBS & 100 & 46.5 $\pm$ 6.0 & 4.5 $\pm$ 0.7 & 2.8 $\pm$ 0.1 & 5.5e-3 $\pm$ 5e-4 & 100 & 20.8 $\pm$ 0.8 & 1.0 & 2.7 $\pm$ 0.0 & 3.8e-3 $\pm$ 2e-4 \\
     & MDOC-CBS & 100 & 6.5 $\pm$ 2.0 & 5.7 $\pm$ 2.0 & 1.8 $\pm$ 0.0 & 6.2e-5 $\pm$ 3e-6 & 100 & 2.2 $\pm$ 0.6 & 1.4 $\pm$ 0.5 & 1.9 $\pm$ 0.0 & 6.1e-5 $\pm$ 3e-6 \\
    \midrule
    \multirow{5}{*}{6}
      & A*-CBS & 100 & 14.2 $\pm$ 12.2 & 120.5 $\pm$ 96.5 & 2.0 $\pm$ 0.0 & 2.5e-3 $\pm$ 3e-4 & 100 & 8.2 $\pm$ 3.6 & 116.4 $\pm$ 46.4 & 1.9 $\pm$ 0.0 & 2.1e-3 $\pm$ 2e-4 \\
     & A*-ECBS & 100 & 3.0 $\pm$ 0.8 & 11.6 $\pm$ 3.4 & 2.0 $\pm$ 0.0 & 2.4e-3 $\pm$ 3e-4 & 100 & 0.6 $\pm$ 0.2 & 4.2 $\pm$ 1.2 & 1.9 & 2.0e-3 $\pm$ 8e-5 \\
     & KCBS & 60 & 6.1 $\pm$ 13.6 & 6.8 $\pm$ 1.5 & 2.1 $\pm$ 0.1 & 2.1e-3 $\pm$ 2e-4 & 50 & 0.3 $\pm$ 0.1 & 3.2 $\pm$ 1.8 & 2.0 $\pm$ 0.0 & 2.2e-3 $\pm$ 4e-4 \\
     & MMD-CBS & 100 & 86.2 $\pm$ 16.3 & 9.1 $\pm$ 2.0 & 2.9 $\pm$ 0.1 & 6.1e-3 $\pm$ 7e-4 & 100 & 37.8 $\pm$ 1.3 & 3.0 & 2.8 $\pm$ 0.0 & 4.7e-3 $\pm$ 4e-4 \\
     & MDOC-CBS & 100 & 16.0 $\pm$ 2.4 & 13.6 $\pm$ 1.7 & 1.8 $\pm$ 0.0 & 6.7e-5 $\pm$ 4e-6 & 100 & 3.2 $\pm$ 0.5 & 2.5 $\pm$ 0.5 & 1.9 $\pm$ 0.0 & 6.1e-5 $\pm$ 2e-6 \\
    \midrule
    \multirow{5}{*}{7}
      & A*-CBS & 0 & - & - & - & - & 100 & 19.1 $\pm$ 18.7 & 259.7 $\pm$ 243.6 & 1.9 $\pm$ 0.0 & 1.9e-3 $\pm$ 2e-4 \\
     & A*-ECBS & 0 & - & - & - & - & 100 & 2.0 $\pm$ 0.5 & 11.2 $\pm$ 2.7 & 1.9 $\pm$ 0.0 & 1.9e-3 $\pm$ 2e-4 \\
     & KCBS & 40 & 9.2 $\pm$ 16.3 & 12.8 $\pm$ 2.9 & 2.1 $\pm$ 0.0 & 2.0e-3 $\pm$ 3e-4 & 80 & 0.5 $\pm$ 0.2 & 4.6 $\pm$ 2.1 & 2.1 $\pm$ 0.1 & 2.0e-3 $\pm$ 2e-4 \\
     & MMD-CBS & 100 & 159.7 $\pm$ 32.7 & 18.1 $\pm$ 4.4 & 3.0 $\pm$ 0.1 & 6.5e-3 $\pm$ 7e-4 & 100 & 49.2 $\pm$ 6.3 & 4.1 $\pm$ 0.7 & 2.8 $\pm$ 0.0 & 4.8e-3 $\pm$ 3e-4 \\
     & MDOC-CBS & 100 & 23.3 $\pm$ 5.6 & 17.7 $\pm$ 3.4 & 1.8 $\pm$ 0.0 & 7.6e-5 $\pm$ 2e-5 & 100 & 5.0 $\pm$ 1.3 & 3.6 $\pm$ 1.5 & 1.9 $\pm$ 0.0 & 6.3e-5 $\pm$ 2e-6 \\
    \midrule
    \multirow{5}{*}{8}
      & A*-CBS & 100 & 82.4 $\pm$ 52.4 & 965.0 $\pm$ 580.5 & 2.1 $\pm$ 0.0 & 2.4e-3 $\pm$ 4e-4 & 90 & 70.0 $\pm$ 45.2 & 898.9 $\pm$ 535.4 & 1.9 $\pm$ 0.0 & 2.1e-3 $\pm$ 1e-4 \\
     & A*-ECBS & 100 & 13.7 $\pm$ 5.9 & 70.4 $\pm$ 27.5 & 22.1 $\pm$ 0.0 & 2.6e-3 $\pm$ 5e-4 & 100 & 5.3 $\pm$ 2.5 & 32.5 $\pm$ 14.2 & 1.9 $\pm$ 0.0 & 2.3e-3 $\pm$ 2e-4 \\
     & KCBS & 60 & 6.8 $\pm$ 13.7 & 15.3 $\pm$ 3.2 & 2.2 $\pm$ 0.1 & 2.2e-3 $\pm$ 2e-4 & 60 & 0.5 $\pm$ 0.1 & 4.8 $\pm$ 1.5 & 2.1 $\pm$ 0.1 & 2.1e-3 $\pm$ 2e-4 \\
     & MMD-CBS & 100 & 270.9 $\pm$ 62.2 & 31.0 $\pm$ 8.4 & 3.0 $\pm$ 0.1 & 6.7e-3 $\pm$ 6e-4 & 100 & 57.5 $\pm$ 1.7 & 5.0 & 2.9 $\pm$ 0.0 & 4.9e-3 $\pm$ 2e-4 \\
     & MDOC-CBS & 100 & 36.8 $\pm$ 11.5 & 24.3 $\pm$ 10.1 & 1.8 $\pm$ 0.0 & 6.7e-5 $\pm$ 5e-6 & 100 & 4.7 $\pm$ 0.7 & 3.1 $\pm$ 0.6 & 1.9 $\pm$ 0.0 & 6.1e-5 $\pm$ 2e-6 \\
    \midrule
    \multirow{5}{*}{9}
      & A*-CBS & 0 & - & - & - & - & 100 & 31.8 $\pm$ 20.9 & 427.1 $\pm$ 274.1 & 1.9 $\pm$ 0.0 & 1.8e-3 $\pm$ 1e-4 \\
     & A*-ECBS & 0 & - & - & - & - & 100 & 6.3 $\pm$ 4.2 & 33.8 $\pm$ 21.4 & 1.9 $\pm$ 0.0 & 2.7e-3 $\pm$ 9e-5 \\
     & KCBS & 30 & 12.4 $\pm$ 19.7 & 17.0 $\pm$ 8.2 & 2.2 $\pm$ 0.1 & 2.0e-3 $\pm$ 4e-4 & 60 & 0.8 $\pm$ 0.6 & 8.7 $\pm$ 5.5 & 2.2 $ \pm $ 0.1 & 2.1e-3 $ \pm $ 4e-4 \\
     & MMD-CBS & 80 & 400.5 $\pm$ 33.8 & 42.5 $\pm$ 3.4 & 3.0 $\pm$ 0.1 & 7.3e-3 $\pm$ 8e-4 & 100 & 53.4 $\pm$ 4.4 & 4.1 $\pm$ 0.3 & 2.8 $ \pm $ 0.0 & 4.5e-3 $ \pm $ 3e-4 \\
     & MDOC-CBS & 100 & 60.8 $\pm$ 15.0 & 34.9 $\pm$ 10.1 & 1.8 $\pm$ 0.0 & 6.4e-5 $\pm$ 3e-6 & 100 & 12.5 $ \pm $ 2.8 & 7.6 $ \pm $ 1.8 & 1.9 $ \pm $ 0.0 & 6.2e-5 $ \pm $ 3e-6 \\
    \midrule
    \multirow{5}{*}{10}
      & A*-CBS & 0 & -- & -- & -- & -- & 100 & 159.1 $\pm$ 123.3 & 2066.6 $\pm$ 1464.4 & 1.9 $\pm$ 0.0 & 2.3e-3 $\pm$ 2e-4 \\
     & A*-ECBS & 0 & -- & -- & -- & -- & 100 & 2.04 $\pm$ 0.7 & 12.2 $\pm$ 4.8 & 1.9 $\pm$ 0.0 & 2.0e-3 $\pm$ 9e-5 \\
     & KCBS & 60 & 28.1 $\pm$ 52.9 & 26.5 $\pm$ 8.4 & 2.2 $\pm$ 0.0 & 2.1e-3 $\pm$ 2e-4 & 50 & 1.3 $ \pm $ 0.8 & 13.4 $ \pm $ 7.3 & 2.2 $ \pm $ 0.1 & 1.8e-3 $ \pm $ 2e-4 \\
     & MMD-CBS & 0 & -- & -- & -- & -- & 100 & 65.7 $ \pm $ 5.1 & 5.2 $ \pm $ 0.4 & 2.8 $ \pm $ 0.0 & 4.7e-3 $ \pm $ 2e-4 \\
     & MDOC-CBS & 100 & 105.0 $\pm$ 27.0 & 51.8 $\pm$ 17.0 & 1.8 $\pm$ 0.1 & 6.5e-5 $\pm$ 4e-6 & 100 & 12.0 $ \pm $ 2.3 & 7.6 $ \pm $ 2.0 & 1.9 $ \pm $ 0.0 & 6.2e-5 $ \pm $ 2e-6 \\
    \midrule
    \multirow{5}{*}{11}
      & A*-CBS & 0 & -- & -- & -- & -- & 50 & 353.6 $\pm$ 82.4 & 4751.8 $\pm$ 1138.2 & 1.9 $\pm$ 0.0 & 2.0e-3 $\pm$ 2e-4 \\
     & A*-ECBS & 100 & 224.75 $\pm$ 93.0 & 583.0 $\pm$ 241.2 & 2.1 $\pm$ 0.0 & 3.7e-3 $\pm$ 3e-4 & 100 & 8.57 $\pm$ 5.5 & 48.3 $\pm$ 32.3 & 1.9 $\pm$ 0.0 & 2.7e-3 $\pm$ 2e-4 \\
     & KCBS & 40 & 61.8 $\pm$ 51.5 & 34.5 $\pm$ 13.4 & 2.2 $\pm$ 0.1 & 1.9e-3 $\pm$ 2e-4 & 70 & 6.4 $ \pm $ 13.0 & 15.6 $ \pm $ 6.3 & 2.2 $ \pm $ 0.1 & 1.9e-3 $ \pm $ 3e-4 \\
     & MMD-CBS & 0 & -- & -- & -- & -- & 100 & 105.7 $ \pm $ 9.1 & 9.4 $ \pm $ 0.8 & 2.9 $ \pm $ 0.0 & 5.7e-3 $ \pm $ 3e-4 \\
     & MDOC-CBS & 100 & 192.4 $\pm$ 61.0 & 78.3 $\pm$ 31.8 & 1.9 $\pm$ 0.1 & 7.1e-5 $\pm$ 5e-6 & 100 & 22.2 $ \pm $ 5.6 & 12.7 $ \pm $ 2.4 & 1.9 $ \pm $ 0.0 & 6.3e-5 $ \pm $ 2e-6 \\
    \midrule
    \multirow{5}{*}{12}
      & A*-CBS & 0 & -- & -- & -- & -- & 0 & -- & -- & -- & -- \\
     & A*-ECBS & 0 & -- & -- & -- & -- & 100 & 100.1 $\pm$ 56.9 & 549.3 $\pm$ 299.3 & 1.9 $\pm$ 0.0 & 2.6e-3 $\pm$ 3e-4 \\
     & KCBS & 10 & 46.8 & 75.0 & 2.3 & 2.1e-3 & 20 & 1.7 $ \pm $ 0.2 & 17.0 $ \pm $ 1.4 & 2.4 $ \pm $ 0.1 & 1.9e-3 $ \pm $ 9e-5 \\
     & MMD-CBS & 0 & -- & -- & -- & -- & 100 & 145.4 $ \pm $ 9.8 & 13.2 $ \pm $ 0.9 & 3.0 $ \pm $ 0.0 & 6.2e-3 $ \pm $ 3e-4 \\
     & MDOC-CBS & 100 & 260.8 $\pm$ 79.9 & 97.8 $\pm$ 34.7 & 1.9 $\pm$ 0.0 & 7.1e-5 $\pm$ 3e-6 & 100 & 46.3 $ \pm $ 20.0 & 25.5 $ \pm $ 12.0 & 1.9 $ \pm $ 0.0 & 6.5e-5 $ \pm $ 4e-6 \\
    \midrule
    \multirow{5}{*}{13}
      & A*-CBS & 0 & -- & -- & -- & -- & 30 & 454.3 $\pm$ 18.3 & 5465.7 $\pm$ 425.5 & 1.9 $\pm$ 0.0 & 1.9e-3 $\pm$ 1e-4 \\
     & A*-ECBS & 0 & -- & -- & -- & -- & 100 & 43.9 $\pm$ 23.9 & 224.5 $\pm$ 116.6 & 1.9 $\pm$ 0.0 & 2.7e-3 $\pm$ 2e-4 \\
     & KCBS & 20 & 4.2 $\pm$ 2.4 & 39.5 $\pm$ 14.9 & 2.3 $\pm$ 0.0 & 2.0e-3 $\pm$ 2e-4 & 40 & 3.0 $ \pm $ 1.5 & 30.3 $ \pm $ 13.0 & 2.3 $ \pm $ 0.1 & 1.9e-3 $ \pm $ 4e-4 \\
     & MMD-CBS & 0 & -- & -- & -- & -- & 100 & 130.0 $ \pm $ 18.0 & 11.8 $ \pm $ 2.4 & 2.9 $ \pm $ 0.0 & 5.2e-3 $ \pm $ 4e-4 \\
     & MDOC-CBS & 80 & 440.1 $\pm$ 62.3 & 143.9 $\pm$ 22.8 & 2.0 $\pm$ 0.0 & 7.3e-5 $\pm$ 6e-6 & 100 & 62.3 $ \pm $ 20.0 & 28.6 $ \pm $ 11.0 & 1.9 $ \pm $ 0.0 & 6.3e-5 $ \pm $ 3e-6 \\
    \midrule
    \multirow{5}{*}{14}
      & A*-CBS & 0 & -- & -- & -- & -- & 0 & -- & -- & -- & -- \\
     & A*-ECBS & 0 & -- & -- & -- & -- & 100 & 46.7 $\pm$ 22.6 & 222.4 $\pm$ 109.8 & 2.0 $\pm$ 0.0 & 2.8e-3 $\pm$ 2e-4 \\
     & KCBS & 10 & 277.0 & 87.0 & 2.4 & 1.8e-3 & 40 & 12.0 $ \pm $ 19.0 & 34.0 $ \pm $ 15.0 & 2.3 $ \pm $ 0.1 & 1.9e-3 $ \pm $ 2e-4 \\
     & MMD-CBS & 0 & -- & -- & -- & -- & 100 & 170.0 $ \pm $ 26.0 & 16.3 $ \pm $ 3.3 & 3.0 $ \pm $ 0.0 & 5.8e-3 $ \pm $ 3e-4 \\
     & MDOC-CBS & 10 & 427.1 & 87.0 & 1.9 & 6.7e-5 & 100 & 70.6 $ \pm $ 21.0 & 32.7 $ \pm $ 11.0 & 1.9 $ \pm $ 0.0 & 6.5e-5 $ \pm $ 3e-6 \\
    \midrule
    \multirow{5}{*}{15}
      & A*-CBS & 0 & -- & -- & -- & -- & 0 & -- & -- & -- & -- \\
     & A*-ECBS & 0 & -- & -- & -- & -- & 100 & 37.0 $\pm$ 13.3 & 167.1 $\pm$ 59.3 & 1.9 $\pm$ 0.0 & 2.5e-3 $\pm$ 2e-4 \\
     & KCBS & 10 & 212.5 & 85.0 & 2.4 & 2.2e-3 & 20 & 3.7 $ \pm $ 1.5 & 33.0 $ \pm $ 8.5 & 2.3 $ \pm $ 0.0 & 1.9e-3 $ \pm $ 1e-4 \\
     & MMD-CBS & 0 & -- & -- & -- & -- & 100 & 208.6 $ \pm $ 29.0 & 19.7 $ \pm $ 3.8 & 3.0 $ \pm $ 0.0 & 6.3e-3 $ \pm $ 2e-4 \\
     & MDOC-CBS & 0 & -- & -- & -- & -- & 100 & 207.8 $ \pm $ 120.0 & 91.0 $ \pm $ 58.0 & 1.9 $ \pm $ 0.0 & 6.8e-5 $ \pm $ 3e-6 \\
    \midrule
    \multirow{5}{*}{16}
      & A*-CBS & 0 & -- & -- & -- & -- & 0 & -- & -- & -- & -- \\
     & A*-ECBS & 0 & -- & -- & -- & -- & 100 & 105.9 $\pm$ 68.9 & 470.4 $\pm$ 298.0 & 1.9 $\pm$ 0.0 & 2.5e-3 $\pm$ 3e-4 \\
     & KCBS & 10 & 183.5 & 100.0 & 2.5 & 2.0e-3 & 40 & 4.3 $ \pm $ 2.7 & 38.3 $ \pm $ 15.0 & 2.3 $ \pm $ 0.1 & 1.8e-3 $ \pm $ 2e-4 \\
     & MMD-CBS & 0 & -- & -- & -- & -- & 100 & 241.6 $ \pm $ 22.0 & 23.7 $ \pm $ 3.0 & 3.0 $ \pm $ 0.1 & 6.1e-3 $ \pm $ 5e-4 \\
     & MDOC-CBS & 0 & -- & -- & -- & -- & 100 & 182.7 $ \pm $ 59.0 & 78.6 $ \pm $ 30.0 & 1.9 $ \pm $ 0.0 & 6.5e-5 $ \pm $ 2e-6 \\
    \midrule
    \multirow{5}{*}{17}
      & A*-CBS & 0 & -- & -- & -- & -- & 0 & -- & -- & -- & -- \\
     & A*-ECBS & 0 & -- & -- & -- & -- & 100 & 176.9 $\pm$ 88.6 & 66.2 $\pm$ 342.9 & 1.9 $\pm$ 0.0 & 2.9e-3 $\pm$ 4e-4 \\
     & KCBS & 0 & -- & -- & -- & -- & 30 & 32.3 $ \pm $ 44.0 & 64.7 $ \pm $ 7.6 & 2.4 $ \pm $ 0.1 & 1.7e-3 $ \pm $ 2e-4 \\
     & MMD-CBS & 0 & -- & -- & -- & -- & 100 & 317.2 $ \pm $ 71.0 & 31.6 $ \pm $ 8.9 & 3.0 $ \pm $ 0.1 & 6.4e-3 $ \pm $ 6e-4 \\
     & MDOC-CBS & 0 & -- & -- & -- & -- & 80 & 328.3 $ \pm $ 44.0 & 110.3 $ \pm $ 17.0 & 1.9 $ \pm $ 0.0 & 6.7e-5 $ \pm $ 3e-6 \\
    \midrule
    \multirow{5}{*}{18}
      & A*-CBS & 0 & -- & -- & -- & -- & 0 & -- & -- & -- & -- \\
     & A*-ECBS & 0 & -- & -- & -- & -- & 90 & 221.5 $\pm$ 114.9 & 741.9 $\pm$ 370.6 & 2.0 $\pm$ 0.0 & 2.9e-3 $\pm$ 2e-4 \\
     & KCBS & 10 & 377.1 & 174.0 & 2.5 & 1.9e-3 & 10 & 7.0 & 71.0 & 2.3 & 1.8e-3 \\
     & MMD-CBS & 0 & -- & -- & -- & -- & 90 & 389.2 $ \pm $ 45.0 & 37.6 $ \pm $ 5.3 & 3.1 $ \pm $ 0.0 & 7.0e-3 $ \pm $ 6e-4 \\
     & MDOC-CBS & 0 & -- & -- & -- & -- & 60 & 337.1 $ \pm $ 80.0 & 108.7 $ \pm $ 23.0 & 2.0 $ \pm $ 0.0 & 7.1e-5 $ \pm $ 4e-6 \\
    \midrule
    \multirow{5}{*}{19}
      & A*-CBS & 0 & -- & -- & -- & -- & 0 & -- & -- & -- & -- \\
     & A*-ECBS & 0 & -- & -- & -- & -- & 50 & 260.1 $\pm$ 147.6 & 869.0 $\pm$ 476.2 & 2.0 $\pm$ 0.0 & 3e-3 $\pm$ 2e-4 \\
     & KCBS & 0 & -- & -- & -- & -- & 0 & -- & -- & -- & -- \\
     & MMD-CBS & 0 & -- & -- & -- & -- & 40 & 451.9 $ \pm $ 34.0 & 41.5 $ \pm $ 4.4 & 3.1 $ \pm $ 0.0 & 7.0e-3 $ \pm $ 4e-4 \\
     & MDOC-CBS & 0 & -- & -- & -- & -- & 60 & 399.0 $ \pm $ 28.0 & 151.5 $ \pm $ 45.0 & 2.0 $ \pm $ 0.0 & 6.9e-5 $ \pm $ 4e-6 \\
    \midrule
    \multirow{5}{*}{20}
      & A*-CBS & 0 & -- & -- & -- & -- & 0 & -- & -- & -- & -- \\
      & A*-ECBS & 0 & -- & -- & -- & -- & 10 & 323.32 & 1241.0 & 1.91 & 2.5e-3 \\
      & KCBS & 0 & -- & -- & -- & -- & 0 & -- & -- & -- & -- \\
      & MMD-CBS & 0 & -- & -- & -- & -- & 10 & 446.3 & 39.0 & 3.0 & 6.8e-3 \\
      & MDOC-CBS & 0 & -- & -- & -- & -- & 10 & 428.0 & 146.0 & 2.0 & 7.2e-5 \\
    \bottomrule

\end{tabular}
\caption{Results on Empty map tasks under a 500 running time-limit. Columns correspond to  \textbf{S}($\uparrow$): success rate(\%), \textbf{T}($\downarrow$): planning time, \textbf{E}($\downarrow$): CT expansions, \textbf{P}($\downarrow$): path length, and \textbf{G}($\downarrow$): geometric smoothness. All metrics are computed based on success trails.}
\end{table*}

\begin{table*}[tb!]
\centering
\small
\scriptsize  
\setlength{\tabcolsep}{4.5pt} 
\renewcommand{\arraystretch}{0.3}

\begin{tabular}{c|c|ccccc|ccccc}
    \toprule
    & & \multicolumn{5}{c|}{Conveyor Map (Circle)} 
      & \multicolumn{5}{c}{Conveyor Map (Weave)}\\
    \cmidrule(lr){3-7} \cmidrule(lr){8-12} 
    N & Method 
      & \textbf{S}($\uparrow$) & \textbf{T}($\downarrow$) & \textbf{E}($\downarrow$) & \textbf{P}($\downarrow$) & \textbf{G}($\downarrow$)  
      & \textbf{S}($\uparrow$) & \textbf{T}($\downarrow$) & \textbf{E}($\downarrow$) & \textbf{P}($\downarrow$) & \textbf{G}($\downarrow$) \\
    \midrule
    \multirow{5}{*}{3}
      & A*-CBS & 100 & 0.2 $\pm$ 0.0 & 1.0 & 2.4 & 2.5e-3 & 100 & 0.6 $\pm$ 0.2 & 5.5 $\pm$ 2.0 & 2.3 $\pm$ 0.0 & 3.0e-3 $\pm$ 1.4e-4\\
     & A*-ECBS & 100 & 0.2 $\pm$ 0.0 & 1.0 & 2.4 & 2.5e-3 & 100 & 0.2 $\pm$ 0.0 & 1.0 & 2.3 & 2.9e-3 \\
     & KCBS & 80 & 0.6 $\pm$ 0.2 & 0.6 $\pm$ 0.5 & 2.6 $\pm$ 0.1 & 2.1e-3 $\pm$ 5.0e-4 & 90 & 0.3 $\pm$ 0.1 & 0.3 $\pm$ 0.5 & 2.3 $\pm$ 0.2 & 2.2e-3 $\pm$ 3.5e-4 \\
     & MMD-CBS & 100 & 8.5 $\pm$ 0.3 & 0.0 & 3.6 $\pm$ 0.1 & 7.4e-3 $\pm$ 4.0e-4 & 100 & 14.3 $\pm$ 2.0 & 0.9 $\pm$ 0.3 & 4.3 $\pm$ 0.3 & 8.0e-3 $\pm$ 5.7e-4 \\
     & MDOC-CBS & 100 & 1.7 $\pm$ 0.4 & 0.8 $\pm$ 0.4 & 2.7 $\pm$ 0.0 & 1.1e-3 $\pm$ 1.2e-4 & 100 & 1.7 $\pm$ 0.3 & 0.9 $\pm$ 0.3 & 2.6 $\pm$ 0.1 & 1.2e-3 $\pm$ 1.2e-4 \\
    \midrule
     \multirow{5}{*}{4}
      & A*-CBS & 100 & 0.4 $\pm$ 0.1 & 2.8 $\pm$ 0.6 & 2.4 & 2.5e-3 $\pm$ 1.5e-4 & 100 & 0.7 $\pm$ 0.4 & 3.7 $\pm$ 2.1 & 2.6 $\pm$ 0.0 & 2.3e-3 $\pm$ 3.1e-5\\
     & A*-ECBS & 100 & 0.6 $\pm$ 0.1  & 3.1 $\pm$ 1.1 & 2.4 & 2.4e-3 $\pm$ 1.2e-4 & 100 & 0.8 $\pm$ 0.2 & 3.9 $\pm$ 1.1 & 2.6 & 2.3e-3 $\pm$ 1.7e-4 \\
     & KCBS & 80 & 0.8 $\pm$ 0.4 & 1.6 $\pm$ 1.0 & 2.5 $\pm$ 0.2 & 2.1e-3 $\pm$ 3.0e-4 & 90 & 0.8 $\pm$ 0.4 & 1.2 $\pm$ 0.8 & 2.7 $\pm$ 0.1 & 2.0e-3 $\pm$ 3.2e-4 \\
     & MMD-CBS & 100 & 18.5 $\pm$ 2.0 & 1.2 $\pm$ 0.4 & 3.9 $\pm$ 0.1 & 7.9e-3 $\pm$ 8.0e-4 & 100 & 19.6 $\pm$ 4.1 & 1.3 $\pm$ 0.7 & 4.0 $\pm$ 0.1 & 8.3e-3 $\pm$ 5.1e-4 \\
     & MDOC-CBS & 100 & 2.6 $\pm$ 0.6 & 1.1 $\pm$ 0.7 & 2.6 $\pm$ 0.0 & 1.2e-3 $\pm$ 1.0e-4 & 90 & 2.1 $\pm$ 0.7 & 0.9 $\pm$ 0.8 & 2.7 $\pm$ 0.0 & 1.1e-3 $\pm$ 7.7e-5 \\
    \midrule
    \multirow{5}{*}{5}
      & A*-CBS & 100 & 0.8 $\pm$ 0.2 & 6.6 $\pm$ 1.8 & 2.4 & 2.4e-3 $\pm$ 1.4e-4 & 100 & 1.0 $\pm$ 0.3 & 6.3 $\pm$ 1.9 & 2.2 $\pm$ 0.0 & 2.6e-3 $\pm$ 7.4e-5 \\
     & A*-ECBS & 100 & 0.54 $\pm$ 0.0 & 2.4 $\pm$ 0.5 & 2.4 & 2.4e-3 $\pm$ 7.7e-5 & 100 & 0.4 $\pm$ 0.0 & 1.0 & 2.2 & 2.5e-3 \\
     & KCBS & 80 & 1.1 $\pm$ 0.5 & 1.8 $\pm$ 1.0 & 2.6 $\pm$ 0.2 & 2.2e-3 $\pm$ 1.0e-4 & 70 & 0.8 $\pm$ 0.2 & 1.7 $\pm$ 0.8 & 2.4 $\pm$ 0.2 & 2.0e-3 $\pm$ 3.7e-4 \\
     & MMD-CBS & 100 & 22.0 $\pm$ 3.0 & 1.2 $\pm$ 0.4 & 3.6 $\pm$ 0.1 & 7.4e-3 $\pm$ 5.0e-4 & 100 & 36.0 $\pm$ 5.6 & 3.2 $\pm$ 0.8 & 4.2 $\pm$ 0.2 & 8.9e-3 $\pm$ 7.1e-4 \\
     & MDOC-CBS & 100 & 4.4 $\pm$ 2.0 & 2.0 $\pm$ 2.0 & 2.7 $\pm$ 0.0 & 1.2e-3 $\pm$ 1.0e-4 & 100 & 5.9 $\pm$ 1.4 & 3.4 $\pm$ 0.8 & 2.6 $\pm$ 0.0 & 1.2e-3 $\pm$ 1.1e-4 \\
    \midrule
    \multirow{5}{*}{6}
      & A*-CBS & 100 & 5.5 $\pm$ 3.8 & 45.5 $\pm$ 29.0 & 2.4 $\pm$ 0.0 & 2.8e-3 $\pm$ 2.3e-4 & 100 & 4.0 $\pm$ 1.5 & 36.8 $\pm$ 13.8 & 2.3 $\pm$ 0.0 & 3.9e-3 $\pm$ 1.1e-4 \\
     & A*-ECBS & 100 & 1.5 $\pm$ 0.6 & 5.1 $\pm$ 2.4 & 2.5 $\pm$ 0.0 & 2.8e-3 $\pm$ 3.3e-4 & 100 & 0.9 $\pm$ 0.2 & 4.4 $\pm$ 1.0 & 2.27 & 3.8e-3 \\
     & KCBS & 90 & 39.2 $\pm$ 53.0 & 5.1 $\pm$ 3.6 & 2.7 $\pm$ 0.1 & 2.2e-3 $\pm$ 1.7e-4 & 90 & 4.2 $\pm$ 10.0 & 1.6 $\pm$ 0.9 & 2.3 $\pm$ 0.1 & 2.0e-3 $\pm$ 3.5e-4 \\
     & MMD-CBS & 100 & 38.1 $\pm$ 6.9 & 3.0 $\pm$ 1.1 & 3.7 $\pm$ 0.1 & 8.0e-3 $\pm$ 2.6e-4 & 100 & 34.7 $\pm$ 4.3 & 2.5 $\pm$ 0.5 & 4.2 $\pm$ 0.1 & 8.6e-3 $\pm$ 6.3e-4 \\
     & MDOC-CBS & 100 & 7.6 $\pm$ 2.4 & 4.3 $\pm$ 1.6 & 2.7 $\pm$ 0.1 & 1.2e-3 $\pm$ 2.2e-4 & 100 & 5.4 $\pm$ 1.2 & 2.9 $\pm$ 0.7 & 2.6 $\pm$ 0.1 & 1.3e-3 $\pm$ 1.1e-4 \\
    \midrule
    \multirow{5}{*}{7}
      & A*-CBS & 100 & 16.1 $\pm$ 13.2 & 126.5 $\pm$ 93.7 & 2.3 $\pm$ 0.0 & 3.0e-3 $\pm$ 1.5e-4 & 100 & 66.9 $\pm$ 40.3 & 576.0 $\pm$ 324.1 & 2.3 $\pm$ 0.0 & 3.5e-3 $\pm$ 1.1e-4 \\
     & A*-ECBS & 100 & 5.4 $\pm$ 3.7 & 23.0 $\pm$ 15.9 & 2.4 $\pm$ 0.0 & 3.0e-3 $\pm$ 1.5e-4 & 100 & 1.5 $\pm$ 0.5 & 6.0 $\pm$ 2.0 & 2.3 $\pm$ 0.0 & 3.3e-3 $\pm$ 1.7e-4 \\
     & KCBS & 60 & 7.0 $\pm$ 13.0 & 4.3 $\pm$ 1.5 & 2.6 $\pm$ 0.1 & 2.1e-3 $\pm$ 3.0e-4 & 90 & 1.3 $\pm$ 0.3 & 2.6 $\pm$ 0.9 & 2.5 $\pm$ 0.2 & 1.9e-3 $\pm$ 2.2e-4 \\
     & MMD-CBS & 100 & 47.9 $\pm$ 7.5 & 4.1 $\pm$ 1.2 & 3.8 $\pm$ 0.1 & 8.4e-3 $\pm$ 5.4e-4 & 100 & 47.3 $\pm$ 9.8 & 3.9 $\pm$ 1.4 & 4.3 $\pm$ 0.2 & 8.5e-3 $\pm$ 6.0e-4 \\
     & MDOC-CBS & 100 & 8.8 $\pm$ 3.3 & 5.0 $\pm$ 3.0 & 2.7 $\pm$ 0.0 & 1.1e-3 $\pm$ 8.4e-5 & 80 & 8.2 $\pm$ 0.9 & 5.2 $\pm$ 1.4 & 2.7 $\pm$ 0.0 & 1.4e-3 $\pm$ 8.8e-5 \\
    \midrule
    \multirow{5}{*}{8}
      & A*-CBS & 100 & 121.5 $\pm$ 136.4 & 916.9 $\pm$ 928.8 & 2.4 $\pm$ 0.0 & 2.7e-3 $\pm$ 1.3e-4 & 100 & 28.8 $\pm$ 21.7 & 281.5 $\pm$ 199.4 & 2.3 $\pm$ 0.0 & 2.3e-3 $\pm$ 1.4e-4 \\
     & A*-ECBS & 100 & 46.9 $\pm$ 25.6 & 224.9 $\pm$ 115.1 & 2.4 $\pm$ 0.0 & 2.7e-3 $\pm$ 3.0e-4 & 100 & 6.5 $\pm$ 3.0 & 25.8 $\pm$ 8.6 & 2.3 $\pm$ 0.0 & 2.5e-3 $\pm$ 2.1e- \\
     & KCBS & 70 & 41.7 $\pm$ 110.0 & 7.4 $\pm$ 8.9 & 2.6 $\pm$ 0.1 & 2.1e-3 $\pm$ 1.3e-4 & 80 & 28.8 $\pm$ 78.0 & 4.6 $\pm$ 5.9 & 2.3 $\pm$ 0.1 & 1.8e-3 $\pm$ 1.7e-4 \\
     & MMD-CBS & 100 & 129.8 $\pm$ 79.0 & 14.6 $\pm$ 11.0 & 4.0 $\pm$ 0.2 & 9.2e-3 $\pm$ 8.5e-4 & 100 & 55.3 $\pm$ 20.0 & 4.5 $\pm$ 2.7 & 4.4 $\pm$ 0.1 & 8.9e-3 $\pm$ 3.8e-4 \\
     & MDOC-CBS & 80 & 11.8 $\pm$ 4.3 & 6.0 $\pm$ 3.3 & 2.7 $\pm$ 0.0 & 1.1e-3 $\pm$ 4.5e-5 & 90 & 6.8 $\pm$ 1.3 & 3.3 $\pm$ 1.2 & 2.7 $\pm$ 0.0 & 1.3e-3 $\pm$ 7.6e-5 \\
    \midrule
    \multirow{5}{*}{9}
      & A*-CBS & 90 & 199.0 $\pm$ 100.3 & 1956.0 $\pm$ 900.8 & 2.4 $\pm$ 0.0 & 3.0e-3 $\pm$ 1.4e-4 & 0 & -- & -- & -- & -- \\
     & A*-ECBS & 100 & 42.4 $\pm$ 20.7 & 163.4 $\pm$ 76.0 & 2.4 $\pm$ 0.0 & 3.0e-3 $\pm$ 2.1e-4 & 100 & 7.0 $\pm$ 4.1 & 24.1 $\pm$ 12.0 & 2.3 $\pm$ 0.0 & 3.2e-3 $\pm$ 1.1e-4 \\
     & KCBS & 40 & 42.6 $\pm$ 78.0 & 13.0 $\pm$ 7.6 & 2.7 $\pm$ 0.1 & 2.0e-3 $\pm$ 2.7e-4 & 40 & 2.8 $\pm$ 1.1 & 7.5 $\pm$ 3.1 & 2.4 $\pm$ 0.1 & 1.9e-3 $\pm$ 1.5e-4 \\
     & MMD-CBS & 100 & 131.8 $\pm$ 44.0 & 13.9 $\pm$ 6.2 & 3.9 $\pm$ 0.1 & 9.0e-3 $\pm$ 4.6e-4 & 100 & 111.6 $\pm$ 28.0 & 11.2 $\pm$ 3.8 & 4.2 $\pm$ 0.1 & 9.1e-3 $\pm$ 3.8e-4 \\
     & MDOC-CBS & 100 & 15.1 $\pm$ 3.9 & 6.1 $\pm$ 1.7 & 2.7 $\pm$ 0.0 & 1.2e-3 $\pm$ 9.2e-5 & 90 & 20.2 $\pm$ 2.2 & 11.6 $\pm$ 1.1 & 2.7 $\pm$ 0.1 & 1.4e-3 $\pm$ 1.0e-4 \\
    \midrule
    \multirow{5}{*}{10}
      & A*-CBS & 0 & -- & -- & -- & -- & 100 & 214.5 $\pm$ 163.7 & 2527 $\pm$ 1855.0 & 2.3 $\pm$ 0.0 & 3.0e-3 $\pm$ 2.1e-4 \\
     & A*-ECBS & 100 & 24.8 $\pm$ 11.6 & 61.1 $\pm$ 27.0 & 2.5 $\pm$ 0.0 & 3.3e-3 $\pm$ 3.9e-4 & 100 & 2.4 $\pm$ 0.8 & 8.5 $\pm$ 2.6 & 2.3 $\pm$ 0.0 & 2.7e-3 $\pm$ 5.2e-5 \\
     & KCBS & 40 & 117.0 $\pm$ 110.0 & 23.3 $\pm$ 12.0 & 2.7 $\pm$ 0.1 & 2.0e-3 $\pm$ 1.7e-4 & 70 & 7.1 $\pm$ 13.0 & 7.0 $\pm$ 3.7 & 2.5 $\pm$ 0.1 & 1.6e-3 $\pm$ 2.5e-4 \\
     & MMD-CBS & 90 & 240.4 $\pm$ 98.0 & 25.6 $\pm$ 13.0 & 4.1 $\pm$ 0.1 & 9.6e-3 $\pm$ 6.8e-4 & 100 & 216.2 $\pm$ 54.0 & 22.8 $\pm$ 7.2 & 4.2 $\pm$ 0.1 & 9.4e-3 $\pm$ 5.9e-4 \\
     & MDOC-CBS & 100 & 17.0 $\pm$ 2.8 & 7.1 $\pm$ 1.5 & 2.7 $\pm$ 0.0 & 1.1e-3 $\pm$ 6.5e-5 & 100 & 21.2 $\pm$ 3.4 & 11.1 $\pm$ 2.1 & 2.7 $\pm$ 0.1 & 1.3e-3 $\pm$ 1.4e-4 \\
    \midrule
    \multirow{5}{*}{11}
      & A*-CBS & 0 & -- & -- & -- & -- & 0 & -- & -- & -- & -- \\
     & A*-ECBS & 100 & 44.3 $\pm$ 28.6 & 121.9 $\pm$ 75.7 & 2.5 $\pm$ 0.0 & 3.2e-3 $\pm$ 3.7e-4 & 80 & 301.8 $\pm$ 86.0 & 721.8 $\pm$ 203.8 & 2.4 $\pm$ 0.0 & 3.7e-3 $\pm$ 1.8e-4 \\
     & KCBS & 70 & 98.6 $\pm$ 69.0 & 17.4 $\pm$ 6.2 & 2.8 $\pm$ 0.0 & 2.1e-3 $\pm$ 1.6e-4 & 50 & 35.5 $\pm$ 55.0 & 10.6 $\pm$ 2.9 & 2.5 $\pm$ 0.1 & 1.9e-3 $\pm$ 4.2e-4 \\
     & MMD-CBS & 80 & 314.3 $\pm$ 55.0 & 32.6 $\pm$ 6.5 & 4.1 $\pm$ 0.2 & 9.6e-3 $\pm$ 3.3e-4 & 100 & 236.0 $\pm$ 95.0 & 25.3 $\pm$ 13.0 & 4.4 $\pm$ 0.1 & 9.9e-3 $\pm$ 8.3e-4 \\
     & MDOC-CBS & 90 & 21.9 $\pm$ 5.0 & 7.6 $\pm$ 1.9 & 2.8 $\pm$ 0.0 & 1.2e-3 $\pm$ 5.4e-5 & 100 & 48.4 $\pm$ 15.0 & 24.9 $\pm$ 12.0 & 2.7 $\pm$ 0.1 & 1.4e-3 $\pm$ 1.5e-4 \\
    \midrule
    \multirow{5}{*}{12}
      & A*-CBS & 0 & -- & -- & -- & -- & 0 & -- & -- & -- & -- \\
     & A*-ECBS & 100 & 86.6 $\pm$ 38.1 & 259.2 $\pm$ 103.0 & 2.5 $\pm$ 0.0 & 3.1e-3 $\pm$ 1.5e-4 & 100 & 67.4 $\pm$ 32.7 & 245.4 $\pm$ 118.2 & 2.4 $\pm$ 0.0 & 3.3e-3 $\pm$ 1.4e-4 \\
     & KCBS & 40 & 197.3 $\pm$ 160.0 & 30.0 $\pm$ 11.0 & 2.9 $\pm$ 0.1 & 2.1e-3 $\pm$ 1.2e-4 & 80 & 75.3 $\pm$ 160.0 & 17.9 $\pm$ 16.0 & 2.6 $\pm$ 0.2 & 1.8e-3 $\pm$ 1.7e-4 \\
     & MMD-CBS & 0 & -- & -- & -- & -- & 60 & 359.8 $\pm$ 41.0 & 38.5 $\pm$ 5.2 & 4.3 $\pm$ 0.1 & 9.8e-3 $\pm$ 6.4e-4 \\
     & MDOC-CBS & 90 & 22.9 $\pm$ 6.4 & 7.2 $\pm$ 2.0 & 2.8 $\pm$ 0.0 & 1.2e-3 $\pm$ 1.6e-4 & 70 & 113.8 $\pm$ 130.0 & 66.0 $\pm$ 82.0 & 2.7 $\pm$ 0.0 & 1.3e-3 $\pm$ 1.1e-4 \\
    \midrule
    \multirow{5}{*}{13}
      & A*-CBS & 0 & -- & -- & -- & -- & 0 & -- & -- & -- & -- \\
     & A*-ECBS & 100 & 104.7 $\pm$ 54.5 & 271.2 $\pm$ 138.3 & 2.5 $\pm$ 0.0 & 3.0e-3 $\pm$ 1.4e-4 & 100 & 88.8 $\pm$ 43.6 & 215.6 $\pm$ 99.5 & 2.3 $\pm$ 0.0 & 3.3e-3 $\pm$ 1.2e-4 \\
     & KCBS & 40 & 191.7 $\pm$ 140.0 & 36.8 $\pm$ 15.0 & 2.9 $\pm$ 0.1 & 2.1e-3 $\pm$ 1.8e-4 & 20 & 36.9 $\pm$ 46.0 & 13.0 $\pm$ 5.7 & 2.6 $\pm$ 0.1 & 1.8e-3 $\pm$ 2.1e-4 \\
     & MMD-CBS & 0 & -- & -- & -- & -- & 20 & 369.2 $\pm$ 160.0 & 37.5 $\pm$ 19.0 & 4.3 $\pm$ 0.1 & 1.0e-2 $\pm$ 1.6e-3 \\
     & MDOC-CBS & 90 & 35.3 $\pm$ 9.6 & 10.0 $\pm$ 2.1 & 2.8 $\pm$ 0.1 & 1.2e-3 $\pm$ 1.5e-4 & 100 & 98.9 $\pm$ 28.0 & 45.0 $\pm$ 12.0 & 2.8 $\pm$ 0.0 & 1.4e-3 $\pm$ 1.1e-4 \\
    \midrule
    \multirow{5}{*}{14}
      & A*-CBS & 0 & -- & -- & -- & -- & 0 & -- & -- & -- & -- \\
     & A*-ECBS & 80 & 251.0 $\pm$ 96.6 & 632.7 $\pm$ 225.7  & 2.4 $\pm$ 0.0 & 3.5e-3 $\pm$ 1.4e-4 & 0 & -- & -- & -- & -- \\
     & KCBS & 0 & -- & -- & -- & -- & 20.0 & 136.1 $\pm$ 130.0 & 42.0 $\pm$ 8.5 & 2.7 $\pm$ 0.2 & 1.5e-3 $\pm$ 2.2e-4 \\
     & MMD-CBS & 0 & -- & -- & -- & -- & 10 & 495.4 & 50.0 & 4.4 & 1.0e-2 \\
     & MDOC-CBS & 90 & 55.2 $\pm$ 19.0 & 13.9 $\pm$ 5.2 & 2.8 $\pm$ 0.0 & 1.2e-3 $\pm$ 1.0e-4 & 100 & 153.8 $\pm$ 29.0 & 70.7 $\pm$ 17.0 & 2.8 $\pm$ 0.1 & 1.6e-3 $\pm$ 1.9e-4 \\
    \midrule
    \multirow{5}{*}{15}
      & A*-CBS & 0 & -- & -- & -- & -- & 0 & -- & -- & -- & -- \\
     & A*-ECBS & 40 & 303.53 $\pm$ 130.6 & 671.0 $\pm$ 245.7 & 2.5 $\pm$ 0.0 & 3.4e-3 $\pm$ 1.2e-4 & 0 & -- & -- & -- & -- \\
     & KCBS & 0 & -- & -- & -- & -- & 10 & 78.7 & 37.0 & 2.6 & 1.9e-3 \\
     & MMD-CBS & 0 & -- & -- & -- & -- & 0 & -- & -- & -- & -- \\
     & MDOC-CBS & 100 & 71.3 $\pm$ 29.0 & 18.0 $\pm$ 6.9 & 2.8 $\pm$ 0.1 & 1.4e-3 $\pm$ 2.9e-4 & 90 & 182.2 $\pm$ 38.0 & 60.9 $\pm$ 14.0 & 2.8 $\pm$ 0.0 & 1.6e-3 $\pm$ 1.9e-4 \\
    \midrule
    \multirow{5}{*}{16}
      & A*-CBS & 0 & -- & -- & -- & -- & 0 & -- & -- & -- & -- \\
     & A*-ECBS & 100 & 20.6 & 63.0 & 2.4 & 3.3e-3 & 0 & -- & -- & -- & -- \\
     & KCBS & 0 & -- & -- & -- & -- & 10 & 7.6 & 22.0 & 2.5 & 1.9e-3 \\
     & MMD-CBS & 0 & -- & -- & -- & -- & 0 & -- & -- & -- & -- \\
     & MDOC-CBS & 80 & 147.2 $\pm$ 75.0 & 39.4 $\pm$ 26.0 & 2.8 $\pm$ 0.0 & 1.4e-3 $\pm$ 2.0e-4 & 40 & 315.7 $\pm$ 100.0 & 97.8 $\pm$ 37.0 & 2.8 $\pm$ 0.0 & 1.4e-3 $\pm$ 7.1e-5 \\
    \midrule
    \multirow{5}{*}{17}
      & A*-CBS & 0 & -- & -- & -- & -- & 0 & -- & -- & -- & -- \\
     & A*-ECBS & 0 & -- & -- & -- & -- & 0 & -- & -- & -- & -- \\
     & KCBS & 10 & 228.0 & 50.0 & 2.9 & 1.8e-3 & 10 & 9.5 & 26.0 & 2.6 & 1.5e-3 \\
     & MMD-CBS & 0 & -- & -- & -- & -- & 0 & -- & -- & -- & -- \\
     & MDOC-CBS & 60 & 111.6 $\pm$ 46.0 & 23.7 $\pm$ 15.0 & 2.8 $\pm$ 0.0 & 1.4e-3 $\pm$ 1.8e-4 & 20 & 387.7 $\pm$ 130.0 & 132.5 $\pm$ 81.0 & 2.8 $\pm$ 0.0 & 1.7e-3 $\pm$ 1.4e-4 \\
    \midrule
    \multirow{5}{*}{18}
      & A*-CBS & 0 & -- & -- & -- & -- & 0 & -- & -- & -- & -- \\
     & A*-ECBS & 0 & -- & -- & -- & -- & 0 & -- & -- & -- & -- \\
     & KCBS & 0 & -- & -- & -- & -- & 0 & -- & -- & -- & -- \\
     & MMD-CBS & 0 & -- & -- & -- & -- & 0 & -- & -- & -- & -- \\
     & MDOC-CBS & 70 & 286.8 $\pm$ 100.0 & 63.7 $\pm$ 30.0 & 2.9 $\pm$ 0.1 & 1.6e-3 $\pm$ 2.3e-4 & 0 & -- & -- & -- & -- \\
    \midrule
    \multirow{5}{*}{19}
      & A*-CBS & 0 & -- & -- & -- & -- & 0 & -- & -- & -- & -- \\
     & A*-ECBS & 0 & -- & -- & -- & -- & 0 & -- & -- & -- & -- \\
     & KCBS & 0 & -- & -- & -- & -- & 0 & -- & -- & -- & -- \\
     & MMD-CBS & 0 & -- & -- & -- & -- & 0 & -- & -- & -- & -- \\
     & MDOC-CBS & 70 & 297.3 $\pm$ 94.0 & 66.0 $\pm$ 24.0 & 2.9 $\pm$ 0.1 & 1.6e-3 $\pm$ 1.9e-4 & 0 & -- & -- & -- & -- \\
    \midrule
    \multirow{5}{*}{20}
      & A*-CBS & 0 & -- & -- & -- & -- & 0 & -- & -- & -- & -- \\
      & A*-ECBS & 0 & -- & -- & -- & -- & 0 & -- & -- & -- & -- \\
      & KCBS & 0 & -- & -- & -- & -- & 0 & -- & -- & -- & -- \\
      & MMD-CBS & 0 & -- & -- & -- & -- & 0 & -- & -- & -- & -- \\
      & MDOC-CBS & 10 & 369.6 & 84.0 & 2.9 & 1.9e-3 & 0 & -- & -- & -- & -- \\
    \bottomrule

\end{tabular}
\caption{Results on Conveyor map tasks under a 500 running time-limit. Columns correspond to  \textbf{S}($\uparrow$): success rate(\%), \textbf{T}($\downarrow$): planning time, \textbf{E}($\downarrow$): CT expansions, \textbf{P}($\downarrow$): path length, and \textbf{G}($\downarrow$): geometric smoothness. All metrics are computed based on success trails.}
\end{table*}

\begin{table*}[tb!]
\centering
\small
\scriptsize  
\setlength{\tabcolsep}{5pt} 
\renewcommand{\arraystretch}{0.8}

\begin{tabular}{c|c|ccccc|ccccc}
    \toprule
    & & \multicolumn{5}{c|}{Drop-Region Map (Weave)} 
      & \multicolumn{5}{c}{Random Map} \\
    \cmidrule(lr){3-7} \cmidrule(lr){8-12}
    N & Method 
      & \textbf{S}($\uparrow$) & \textbf{T}($\downarrow$) & \textbf{E}($\downarrow$) & \textbf{P}($\downarrow$) & \textbf{G}($\downarrow$)  
      & \textbf{S}($\uparrow$) & \textbf{T}($\downarrow$) & \textbf{E}($\downarrow$) & \textbf{P}($\downarrow$) & \textbf{G}($\downarrow$) \\
    \midrule
    \multirow{5}{*}{3}
      & A*-CBS & 100 & 0.1 $\pm$ 0.0  & 1.0 & 2.2 & 1.88e-3 & 100 & 0.1 $ \pm $ 0.1 & 0.4 $ \pm $ 0.7 & 1.3 $ \pm $ 0.4 & 4.1e-3 $ \pm $ 1.8e-3 \\
     & A*-ECBS & 100 & 0.2 $\pm$ 0.0 & 1.0 & 2.2 & 1.88e-3 & 100 & 0.1 $ \pm $ 0.1 & 0.2 $ \pm $ 0.4 & 1.3 $ \pm $ 0.4 & 4.1e-3 $ \pm $ 1.8e-3 \\
     & KCBS & 90 & 0.6 $ \pm $ 0.2 & 1.4 $ \pm $ 0.5 & 2.5 $ \pm $ 0.2 & 2.1e-3 $ \pm $ 2.6e-4 & 90 & 0.3 $ \pm $ 0.2 & 0.3 $ \pm $ 0.5 & 1.5 $ \pm $ 0.5 & 2.0e-3 $ \pm $ 6.6e-4 \\
     & MMD-CBS & 100 & 15.6 $ \pm $ 3.0 & 1.3 $ \pm $ 0.5 & 3.4 $ \pm $ 0.1 & 7.2e-3 $ \pm $ 5.7e-4 & 100 & 10.1 $ \pm $ 3.2 & 0.4 $ \pm $ 0.5 & 2.8 $ \pm $ 0.3 & 7.1e-3 $ \pm $ 1.3e-3 \\
     & MDOC-CBS & 100 & 2.8 $ \pm $ 0.9 & 1.5 $ \pm $ 0.5 & 3.3 $ \pm $ 0.1 & 3.9e-3 $ \pm $ 8.0e-4 & 100 & 1.1 $ \pm $ 0.8 & 0.3 $ \pm $ 0.7 & 2.6 $ \pm $ 0.2 & 2.8e-3 $ \pm $ 4.4e-4 \\
    \midrule
     \multirow{5}{*}{4}
      & A*-CBS & 100 & 13.6 $\pm$ 11.3 & 159.0 $\pm$ 119.7 & 1.9 $\pm$ 0.0 & 2.2e-3 $\pm$ 4.1e-4 & 100 & 0.1 $ \pm $ 0.1 & 0.5 $ \pm $ 0.7 & 1.4 $ \pm $ 0.3 & 2.8e-3 $ \pm $ 1.2e-3 \\
     & A*-ECBS & 100 & 2.1 $\pm$ 0.7 & 17.8 $\pm$ 5.6 & 2.0 $\pm$ 0.0 & 2.7e-3 $\pm$ 4.2e-4 & 100 & 0.1 $ \pm $ 0.1 & 0.3 $ \pm $ 0.5 & 1.4 $ \pm $ 0.3 & 2.8e-3 $ \pm $ 1.2e-3 \\
     & KCBS & 90 & 1.0 $ \pm $ 0.3 & 3.7 $ \pm $ 0.7 & 2.4 $ \pm $ 0.2 & 1.9e-3 $ \pm $ 1.8e-4 & 100 & 7.1 $ \pm $ 14.0 & 0.9 $ \pm $ 1.0 & 1.5 $ \pm $ 0.4 & 1.8e-3 $ \pm $ 2.9e-4 \\
     & MMD-CBS & 100 & 26.6 $ \pm $ 3.1 & 2.6 $ \pm $ 0.5 & 3.3 $ \pm $ 0.2 & 7.1e-3 $ \pm $ 4.1e-4 & 100 & 15.6 $ \pm $ 3.5 & 0.9 $ \pm $ 0.6 & 2.9 $ \pm $ 0.3 & 7.8e-3 $ \pm $ 1.3e-3 \\
     & MDOC-CBS & 100 & 6.4 $ \pm $ 2.3 & 3.5 $ \pm $ 2.3 & 3.5 $ \pm $ 0.2 & 6.7e-3 $ \pm $ 3.4e-3 & 100 & 2.7 $ \pm $ 2.2 & 1.1 $ \pm $ 1.2 & 2.8 $ \pm $ 0.3 & 3.2e-3 $ \pm $ 4.5e-4 \\
    \midrule
    \multirow{5}{*}{5}
      & A*-CBS & 100 & 0.5 $\pm$ 0.1 & 6.1 $\pm$ 1.7 & 2.1 & 2.6e-3 $\pm$ 9.5e-5 & 100 & 0.7  $ \pm $ 0.9 & 6.6 $ \pm $ 7.4 & 1.4 $ \pm $ 0.3 & 3.3e-3 $ \pm $ 1.0e-3 \\
     & A*-ECBS & 100 & 0.2 $\pm$ 0.0 & 1.0 & 2.1 & 2.5e-3 & 100 & 2.0  $ \pm $ 3.7 & 7.3 $ \pm $ 10.0 & 1.5 $ \pm $ 0.4 & 3.6e-3 $ \pm $ 1.5e-3 \\
     & KCBS & 70 & 1.5 $ \pm $ 0.5 & 4.1 $ \pm $ 1.2 & 2.6 $ \pm $ 0.2 & 2.0e-3 $ \pm $ 2.3e-4 & 80 & 9.2  $ \pm $ 24.0 & 1.9 $ \pm $ 1.5 & 1.6 $ \pm $ 0.3 & 1.6e-3 $ \pm $ 2.5e-4 \\
     & MMD-CBS & 100 & 27.8 $ \pm $ 4.8 & 2.4 $ \pm $ 0.7 & 3.4 $ \pm $ 0.1 & 7.4e-3 $ \pm $ 4.4e-4 & 100 & 16.2 $ \pm $ 5.4 & 0.6 $ \pm $ 0.8 & 2.8 $ \pm $ 0.2 & 7.3e-3 $ \pm $ 6.1e-4 \\
     & MDOC-CBS & 100 & 2.7 $ \pm $ 1.1 & 0.5 $ \pm $ 1.0 & 3.0 $ \pm $ 0.1 & 2.9e-3 $ \pm $ 3.1e-4 & 100 & 4.6  $ \pm $ 3.5 & 1.9 $ \pm $ 1.7 & 2.7 $ \pm $ 0.1 & 3.2e-3 $ \pm $ 4.1e-4 \\
    \midrule
    \multirow{5}{*}{6}
      & A*-CBS & 100 & 3.7 $\pm$ 3.6 & 43.2 $\pm$ 41.0 & 2.2 $\pm$ 0.0 & 2.5e-3 $\pm$ 2.7e-4 & 100 & 2.5 $ \pm $ 6.6 & 23.6 $ \pm $ 64.0 & 1.4 $ \pm $ 0.3 & 2.9e-3 $ \pm $ 8.4e-4 \\
     & A*-ECBS & 100 & 1.2 $\pm$ 0.5 & 8.4 $\pm$ 4.3 & 2.3 & 2.4e-3 $\pm$ 9e-5 & 100 & 1.4 $ \pm $ 3.0 & 7.3 $ \pm $ 16.0 & 1.4 $ \pm $ 0.3 & 2.9e-3 $ \pm $ 8.2e-4 \\
     & KCBS & 40 & 2.5 $ \pm $ 1.1 & 6.3 $ \pm $ 3.2 & 2.7 $ \pm $ 0.1 & 1.8e-3 $ \pm $ 2.0e-4 & 80 & 13.9 $ \pm $ 18.0 & 1.5 $ \pm $ 0.9 & 1.6 $ \pm $ 0.2 & 1.6e-3 $ \pm $ 3.5e-4 \\
     & MMD-CBS & 100 & 63.4 $ \pm $ 15.3 & 7.1 $ \pm $ 2.3 & 3.7 $ \pm $ 0.2 & 7.9e-3 $ \pm $ 6.8e-4 & 100 & 28.0 $ \pm $ 14.0 & 1.9 $ \pm $ 1.9 & 2.9 $ \pm $ 0.2 & 7.6e-3 $ \pm $ 8.8e-4 \\
     & MDOC-CBS & 100 & 9.8 $ \pm $ 8.8 & 6.1 $ \pm $ 7.5 & 3.2 $ \pm $ 0.1 & 4.2e-3 $ \pm $ 1.6e-3 & 90 & 6.3 $ \pm $ 7.5 & 1.9 $ \pm $ 2.9 & 2.7 $ \pm $ 0.2 & 3.1e-3 $ \pm $ 4.8e-4 \\
    \midrule
    \multirow{5}{*}{7}
      & A*-CBS & 70 & 281.2 $\pm$ 120.3 & 3036.9 $\pm$ 1214.9 & 2.2 $\pm$ 0.0 & 4.7e-3 $\pm$ 2.1e-4 & 100 & 12.2 $ \pm $ 36.0 & 130.7 $ \pm $ 390.0 & 1.4 $ \pm $ 0.2 & 3.3e-3 $ \pm $ 6.7e-4 \\
     & A*-ECBS & 100 & 9.4 $\pm$ 3.1 & 51.1 $\pm$ 16.4 & 2.3 $\pm$ 0.0 & 5.0e-3 $\pm$ 3.6e-4 & 100 & 0.8  $ \pm $ 1.2 & 4.7 $ \pm $ 9.3 & 1.4 $ \pm $ 0.2 & 3.1e-3 $ \pm $ 6.4e-4 \\
     & KCBS & 90 & 18.0 $ \pm $ 31.6 & 7.8 $ \pm $ 3.2 & 2.5 $ \pm $ 0.1 & 2.0e-3 $ \pm $ 1.8e-4 & 40 & 0.6  $ \pm $ 0.3 & 1.3 $ \pm $ 1.0 & 1.3 $ \pm $ 0.2 & 1.5e-3 $ \pm $ 2.7e-4 \\
     & MMD-CBS & 100 & 76.8 $ \pm $ 7.7 & 8.6 $ \pm $ 1.1 & 3.7 $ \pm $ 0.2 & 8.4e-3 $ \pm $ 5.1e-4 & 100 & 33.9 $ \pm $ 6.8 & 2.4 $ \pm $ 1.0 & 2.9 $ \pm $ 0.1 & 8.0e-3 $ \pm $ 7.4e-4 \\
     & MDOC-CBS & 100 & 22.3 $ \pm $ 12.6 & 12.9 $ \pm $ 8.4 & 3.2 $ \pm $ 0.1 & 5.2e-3 $ \pm $ 1.7e-3 & 90 & 10.3 $ \pm $ 14.0 & 3.1 $ \pm $ 3.4 & 2.7 $ \pm $ 0.1 & 3.3e-3 $ \pm $ 6.1e-4 \\
    \midrule
     \multirow{5}{*}{8}
      & A*-CBS & 90 & 138.1 $\pm$ 134.13 & 1625.8 $\pm$ 1519.4 & 2.0 $\pm$ 0.0 & 3.5e-3 & 100 & 4.5 $ \pm $ 6.8 & 34.1 $ \pm $ 57.0 & 1.4 $ \pm $ 0.3 & 4.0e-3 $ \pm $ 1.0e-3 \\
     & A*-ECBS & 100 & 6.5 $\pm$ 3.6 & 41.6 $\pm$ 23.0 & 2.0 $\pm$ 0.0 & 3.9e-3 $\pm$ 3.7e-4 & 100 & 2.1 $ \pm $ 3.5 & 10.4 $ \pm $ 18.0 & 1.4 $ \pm $ 0.3 & 3.8e-3 $ \pm $ 9.9e-4 \\
     & KCBS & 50 & 54.4 $ \pm $ 97.0 & 17.8 $ \pm $ 11.0 & 2.6 $ \pm $ 0.2 & 1.9e-3 $ \pm $ 3.0e-4 & 60 & 78.3 $ \pm $ 170.0 & 10.2 $ \pm $ 13.0 & 1.8 $ \pm $ 0.2 & 1.7e-3 $ \pm $ 4.7e-4 \\
     & MMD-CBS & 100 & 117.8 $ \pm $ 19.2 & 13.6 $ \pm $ 3.0 & 3.9 $ \pm $ 0.1 & 8.9e-3 $ \pm $ 6.7e-4 & 100 & 36.2 $ \pm $ 10.0 & 2.4 $ \pm $ 1.5 & 2.8 $ \pm $ 0.1 & 7.5e-3 $ \pm $ 6.2e-4 \\
     & MDOC-CBS & 100 & 54.6 $ \pm $ 119.9 & 31.4 $ \pm $ 76.9 & 3.4 $ \pm $ 0.1 & 7.3e-3 $ \pm $ 2.3e-3 & 90 & 9.8 $ \pm $ 3.2 & 2.9 $ \pm $ 1.2 & 2.7 $ \pm $ 0.1 & 3.3e-3 $ \pm $ 3.6e-4 \\
    \midrule
    \multirow{5}{*}{9}
      & A*-CBS & 60 & 191.0 $\pm$ 123.4 & 2013.7 $\pm$ 1228.8 & 2.1 $\pm$ 0.0 & 2.9e-3 $\pm$ 4.3e-5 & 100 & 1.1 $ \pm $ 1.7 & 12.9 $ \pm $ 25.0 & 1.4 $ \pm $ 0.3 & 3.3e-3 $ \pm $ 9.1e-4 \\
     & A*-ECBS & 100 & 11.3 $\pm$ 5.2 & 57.2 $\pm$ 27.8 & 2.2 $\pm$ 0.0 & 3.4e-3 $\pm$ 1.5e-4 & 100 & 1.3 $ \pm $ 1.7 & 8.1 $ \pm $ 12.0 & 1.4 $ \pm $ 0.3 & 3.4e-3 $ \pm $ 9.5e-4 \\
     & KCBS & 40 & 15.3 $ \pm $ 13.6 & 18.8 $ \pm $ 5.6 & 2.8 $ \pm $ 0.1 & 1.8e-3 $ \pm $ 9.3e-5 & 10 & 35.8 & 3.0 & 1.0 & 1.0e-3 \\
     & MMD-CBS & 100 & 184.4 $ \pm $ 39.0 & 22.5 $ \pm $ 5.3 & 3.8 $ \pm $ 0.1 & 8.8e-3 $ \pm $ 4.5e-4 & 100 & 47.4 $ \pm $ 12.0 & 3.5 $ \pm $ 1.7 & 2.9 $ \pm $ 0.1 & 8.2e-3 $ \pm $ 6.3e-4 \\
     & MDOC-CBS & 90 & 48.7 $ \pm $ 22.7 & 18.2 $ \pm $ 8.8 & 3.3 $ \pm $ 0.2 & 5.2e-3 $ \pm $ 2.3e-3 & 80 & 14.6 $ \pm $ 6.6 & 4.0 $ \pm $ 1.4 & 2.7 $ \pm $ 0.1 & 3.3e-3 $ \pm $ 2.4e-4 \\
    \midrule
    \multirow{5}{*}{10}
      & A*-CBS & 0 & -- & -- & -- & -- & 90 & 0.9 $ \pm $ 0.5 & 10.1 $ \pm $ 5.4 & 1.4 $ \pm $ 0.3 & 3.0e-3 $ \pm $ 5.6e-4 \\
     & A*-ECBS & 100 & 19.6 $\pm$ 9.0 & 101.2 $\pm$ 44.7 & 2.2 $\pm$ 0.0 & 4.2e-3 $\pm$ 2.4e-4 & 100 & 4.0 $ \pm $ 9.8 & 19.3 $ \pm $ 41.0 & 1.4 $ \pm $ 0.3 & 2.9e-3 $ \pm $ 5.3e-4 \\
     & KCBS & 20 & 19.3 $ \pm $ 16.3 & 16.5 $ \pm $ 6.4 & 2.5 $ \pm $ 0.2 & 1.8e-3 $ \pm $ 6.8e-5 & 40 & 36.9 $ \pm $ 70.0 & 6.0 $ \pm $ 3.2 & 1.6 $ \pm $ 0.3 & 1.6e-3 $ \pm $ 4.1e-4 \\
     & MMD-CBS & 100 & 258.0 $ \pm $ 68.2 & 31.7 $ \pm $ 10.3 & 3.8 $ \pm $ 0.1 & 8.9e-3 $ \pm $ 3.6e-4 & 100 & 58.6 $ \pm $ 16.0 & 4.6 $ \pm $ 2.1 & 3.0 $ \pm $ 0.2 & 8.3e-3 $ \pm $ 9.2e-4 \\
     & MDOC-CBS & 100 & 64.7 $ \pm $ 32.9 & 39.8 $ \pm $ 23.5 & 3.5 $ \pm $ 0.1 & 5.4e-3 $ \pm $ 2.1e-3 & 90 & 65.6 $ \pm $ 78.0 & 16.8 $ \pm $ 18.0 & 2.8 $ \pm $ 0.2 & 3.4e-3 $ \pm $ 6.0e-4 \\
    \midrule
    \multirow{5}{*}{11}
      & A*-CBS & 0 & -- & -- & -- & -- & 100 & 1.9 $ \pm $ 2.1 & 19.9 $ \pm $ 16.0 & 1.5 $ \pm $ 0.2 & 3.4e-3 $ \pm $ 6.4e-4 \\
     & A*-ECBS & 100 & 64.7 $\pm$ 71.2 & 255.6 $\pm$ 275.1 & 2.2 $\pm$ 0.0 & 3.7e-3 $\pm$ 1.6e-4 & 100 & 40.1 $ \pm $ 110.0 & 139.6 $ \pm $ 400.0 & 1.6 $ \pm $ 0.2 & 3.6e-3 $ \pm $ 6.0e-4 \\
     & KCBS & 10 & 8.2 & 20.0 & 2.7 & 2.0e-3 & 0 & -- & -- & -- & -- \\
     & MMD-CBS & 10 & 192.6 & 22.0 & 3.8 & 8.4e-3 & 100 & 67.0 $ \pm $ 18.0 & 5.3 $ \pm $ 2.3 & 3.0 $ \pm $ 0.2 & 8.3e-3 $ \pm $ 1.0e-3 \\
     & MDOC-CBS & 80 & 130.9 $ \pm $ 86.2 & 73.3 $ \pm $ 49.0 & 3.5 $ \pm $ 0.1 & 6.9e-3 $ \pm $ 1.7e-3 & 90 & 75.5 $ \pm $ 110.0 & 21.3 $ \pm $ 31.0 & 2.8 $ \pm $ 0.1 & 3.5e-3 $ \pm $ 3.6e-4 \\
    \midrule
     \multirow{5}{*}{12}
      & A*-CBS & 0 & -- & -- & -- & -- & 100 & 11.4 $ \pm $ 30.0 & 171.4 $ \pm $ 460.0 & 1.3 $ \pm $ 0.2 & 3.2e-3 $ \pm $ 5.6e-4 \\
     & A*-ECBS & 100 & 268.6 $\pm$ 132.6 & 1227.5 $\pm$ 606.7 & 2.3 $\pm$ 0.0 & 3.4e-3 $\pm$ 3.4e-4 & 100 & 1.4 $ \pm $ 1.0 & 7.4 $ \pm $ 4.6 & 1.4 $ \pm $ 0.2 & 3.2e-3 $ \pm $ 2.6e-4 \\
     & KCBS & 20 & 52.9 $ \pm $ 16.2 & 37.5 $ \pm $ 0.7 & 2.8 $ \pm $ 0.1 & 1.7e-3 $ \pm $ 1.7e-4 & 10 & 143.0 & 19.0 & 2.0 & 1.4e-3 \\
     & MMD-CBS & 20 & 371.1 $ \pm $ 14.4 & 41.0 $ \pm $ 2.8 & 3.9 $ \pm $ 0.2 & 9.2e-3 $ \pm $ 9.6e-5 & 100 & 77.8 $ \pm $ 24.0 & 6.4 $ \pm $ 3.0 & 2.9 $ \pm $ 0.1 & 8.1e-3 $ \pm $ 5.5e-4 \\
     & MDOC-CBS & 80 & 206.6 $ \pm $ 90.9 & 95.3 $ \pm $ 45.6 & 3.6 $ \pm $ 0.1 & 7.4e-3 $ \pm $ 2.3e-3 & 100 & 75.7 $ \pm $ 51.0 & 14.9 $ \pm $ 8.7 & 2.9 $ \pm $ 0.2 & 3.9e-3 $ \pm $ 4.8e-4 \\
    \midrule
    \multirow{5}{*}{13}
      & A*-CBS & 0 & -- & -- & -- & -- & 90  & 4.6 $ \pm $ 7.3 & 30.4 $ \pm $ 40.5 & 1.3 $ \pm $ 0.2 & 3.1e-3 $ \pm $ 4.9e-4  \\
      & A*-ECBS & 100 & 214.6 $\pm$ 142.1 & 745.4 $\pm$ 476.7  & 2.3 $\pm$ 0.0 & 3.8e-3 $\pm$ 2.2e-4 & 100 & 2.3 $ \pm $ 1.9 & 11.1 $ \pm $ 7.5 & 1.4 $ \pm $ 0.2 & 3.0e-3 $ \pm $ 5.5e-4  \\
      & KCBS & 50  & 107.1 $ \pm $ 106.0 & 42.8 $ \pm $ 16.7 & 2.8 $ \pm $ 0.2 & 1.9e-3 $ \pm $ 3.0e-4 & 10  & 210.2 & 13.0 & 1.7 & 1.7e-3   \\
      & MMD-CBS & 0 & -- & -- & -- & -- & 100 & 128.1 $ \pm $ 62.7 & 12.8 $ \pm $ 8.6 & 3.0 $ \pm $ 0.2 & 8.8e-3 $ \pm $ 7.4e-4  \\
      & MDOC-CBS & 60  & 304.3 $ \pm $ 114.1 & 145.2 $ \pm $ 63.0 & 3.7 $ \pm $ 0.1 & 6.6e-3 $ \pm $ 1.5e-3 & 90  & 76.0 $ \pm $ 48.1 & 16.0 $ \pm $ 1.1e1 & 2.9 $ \pm $ 0.1 & 3.6e-3 $ \pm $ 3.7e-4 \\
    \midrule
    \multirow{5}{*}{14}
      & A*-CBS & 0 & -- & -- & -- & -- & 90 & 4.6 $ \pm $ 7.3 & 30.4 $ \pm $ 40.0 & 1.3 $ \pm $ 0.2 & 3.1e-3 $ \pm $ 4.9e-4 \\
     & A*-ECBS & 0 & -- & -- & -- & -- & 100 & 2.3 $ \pm $ 1.9 & 11.1 $ \pm $ 7.5 & 1.4 $ \pm $ 0.2 & 3.0e-3 $ \pm $ 5.5e-4 \\
     & KCBS & 20 & 485.2 $ \pm $ 15.9 & 89.5 $ \pm $ 20.5 & 3.0 $ \pm $ 0.1 & 1.7e-3 $ \pm $ 1.8e-5 & 10 & 210.2 & 13.0 & 1.7 & 1.7e-3 \\
     & MMD-CBS & 0 & -- & -- & -- & -- & 100 & 128.1 $ \pm $ 63.0 & 12.8 $ \pm $ 8.6 & 3.0 $ \pm $ 0.2 & 8.8e-3 $ \pm $ 7.4e-4 \\
     & MDOC-CBS & 70 & 345.3 $ \pm $ 95.8 & 142.0 $ \pm $ 51.8 & 3.7 $ \pm $ 0.1 & 7.9e-3 $ \pm $ 1.8e-3 & 90 & 76.0 $ \pm $ 48.0 & 16.0 $ \pm $ 11.0 & 2.9 $ \pm $ 0.1 & 3.6e-3 $ \pm $ 3.7e-4 \\
    \midrule
    \multirow{5}{*}{15}
      & A*-CBS & 0 & -- & -- & -- & -- & 80 & 2.8   $ \pm $ 3.5 & 18.9 $ \pm $ 13.0 & 1.3 $ \pm $ 0.2 & 3.0e-3 $ \pm $ 1.0e-3 \\
     & A*-ECBS & 40 & 308.7 $\pm$ 56.2 & 882.2 $\pm$ 155.3 & 2.3 $\pm$ 0.0 & 3.9e-3 $\pm$ 2.1e-4 & 100 & 6.5   $ \pm $ 8.2 & 21.3 $ \pm $ 24.0 & 1.4 $ \pm $ 0.2 & 3.3e-3 $ \pm $ 1.1e-3 \\
     & KCBS & 10 & 18.3 & 40.0 & 2.9 & 2.0e-3 & 0 & -- & -- & -- & -- \\
     & MMD-CBS & 0 & -- & -- & -- & -- & 100 & 152.3 $ \pm $ 50.0 & 14.2 $ \pm $ 6.4 & 3.0 $ \pm $ 0.1 & 8.9e-3 $ \pm $ 8.1e-4 \\
     & MDOC-CBS & 20 & 374.1 $ \pm $ 137.7 & 214.5 $ \pm $ 78.5 & 3.8 $ \pm $ 0.0 & 7.5e-3 $ \pm $ 2.3e-3 & 90 & 174.6 $ \pm $ 94.0 & 28.2 $ \pm $ 14.0 & 2.8 $ \pm $ 0.1 & 3.4e-3 $ \pm $ 1.6e-4 \\
    \bottomrule

\end{tabular}
\caption{Results on Drop-Region map (Weave) and Random map under a 500 running time-limit. Columns correspond to  \textbf{S}($\uparrow$): success rate(\%), \textbf{T}($\downarrow$): planning time, \textbf{E}($\downarrow$): CT expansions, \textbf{P}($\downarrow$): path length, and \textbf{G}($\downarrow$): geometric smoothness. All metrics are computed based on success trails.}
\end{table*}

\begin{table*}[tb!]
\centering
\small
\scriptsize  
\setlength{\tabcolsep}{5pt} 
\renewcommand{\arraystretch}{0.9}

\begin{tabular}{c|c|ccccc|ccccc}
    \toprule
    & & \multicolumn{5}{c|}{Empty Large Map $4 \times 4$} 
      & \multicolumn{5}{c}{Empty Extra Large Map $6 \times 6$} \\
    \cmidrule(lr){3-7} \cmidrule(lr){8-12}
    N & Method 
      & \textbf{S}($\uparrow$) & \textbf{T}($\downarrow$) & \textbf{E}($\downarrow$) & \textbf{P}($\downarrow$) & \textbf{G}($\downarrow$)  
      & \textbf{S}($\uparrow$) & \textbf{T}($\downarrow$) & \textbf{E}($\downarrow$) & \textbf{P}($\downarrow$) & \textbf{G}($\downarrow$) \\
    \midrule
    \multirow{5}{*}{3}
      & A*-CBS   & 100 & 0.3 $ \pm $ 0.2 & 0.1 $ \pm $ 0.3 & 2.6 $ \pm $ 0.7 & 8.3e-4 $ \pm $ 2.9e-4 & 100 & 0.6 $ \pm $ 0.3  & 0.0 & 3.9 $ \pm $ 1.2  & 5.4e-4 $ \pm $ 2.1e-4 \\
      & A*-ECBS  & 100 & 0.3 $ \pm $ 0.2 & 0.1 $ \pm $ 0.3 & 2.6 $ \pm $ 0.7 & 8.3e-4 $ \pm $ 2.9e-4 & 100 & 0.7 $ \pm $ 0.4  & 0.0 & 3.9 $ \pm $ 1.2  & 5.4e-4 $ \pm $ 2.1e-4 \\
      & KCBS     & 90  & 0.1 $ \pm $ 0.0 & 0.0             & 2.5 $ \pm $ 0.8 & 1.7e-3 $ \pm $ 2.0e-4 & 100 & 5.1 $ \pm $ 13.4 & 0.1 $ \pm $ 0.4 & 2.8 $ \pm $ 0.8  & 1.3e-3 $ \pm $ 1.9e-4 \\
      & MMD-CBS  & 100 & 8.3 $ \pm $ 1.9 & 0.1 $ \pm $ 0.3 & 9.3 $ \pm $ 0.3 & 2.0e-2 $ \pm $ 1.5e-3 & 100 & 9.5 $ \pm $ 2.8  & 0.4 $ \pm $ 0.5 & 22.2 $ \pm $ 0.8 & 5.0e-2 $ \pm $ 4.1e-3 \\
      & MDOC-CBS & 100 & 1.1 $ \pm $ 0.4 & 0.3 $ \pm $ 0.5 & 2.8 $ \pm $ 0.6 & 2.9e-4 $ \pm $ 3.3e-5 & 100 & 0.9 $ \pm $ 0.1  & 0.0 & 3.6 $ \pm $ 1.0  & 2.1e-3 $ \pm $ 3.0e-3 \\
    \midrule
     \multirow{5}{*}{6}
      & A*-CBS   & 100 & 0.7 $ \pm $ 0.3 & 0.5 $ \pm $ 1.1 & 2.5 $ \pm $ 0.5 & 6.9e-4 $ \pm $ 2.3e-4 & 100 & 1.8 $ \pm $ 0.7  & 0.7 $ \pm $ 0.7 & 3.9 $ \pm $ 0.7 & 5.4e-4 $ \pm $ 1.5e-4  \\
      & A*-ECBS  & 100 & 0.8 $ \pm $ 0.3 & 0.3 $ \pm $ 0.7 & 2.5 $ \pm $ 0.5 & 6.9e-4 $ \pm $ 2.3e-4 & 100 & 2.0 $ \pm $ 0.7  & 0.5 $ \pm $ 0.5 & 3.9 $ \pm $ 0.7 & 5.5e-4 $ \pm $ 1.7e-4 \\
      & KCBS     & 90  & 4.3 $ \pm $ 12.0 & 0.4 $ \pm $ 0.7 & 2.3 $ \pm $ 0.4 & 1.4e-3 $ \pm $ 1.9e-4 & 100 & 7.3 $ \pm $ 15.0 & 0.5 $ \pm $ 0.7 & 3.5 $ \pm $ 0.9 & 1.4e-3 $ \pm $ 2.7e-4 \\
      & MMD-CBS  & 100 & 19.3 $ \pm $ 3.6 & 0.8 $ \pm $ 0.6 & 9.5 $ \pm $ 0.2 & 2.0e-2 $ \pm $ 1.3e-3 & 100 & 16.7 $ \pm $ 4.5 & 0.4 $ \pm $ 0.7 & 21.7 $ \pm $ 0.8 & 5.0e-2 $ \pm $ 3.9e-3 \\
      & MDOC-CBS & 100 & 1.9 $ \pm $ 0.6 & 0.4 $ \pm $ 0.5 & 2.8 $ \pm $ 0.3 & 3.0e-4 $ \pm $ 3.1e-5 & 100 & 2.0 $ \pm $ 0.5  & 0.3 $ \pm $ 0.5 & 3.6 $ \pm $ 0.4 & 1.6e-3 $ \pm $ 1.6e-3 \\
    \midrule
    \multirow{5}{*}{9}
      & A*-CBS   & 90  & 1.7  $ \pm $ 1.3 & 2.2 $ \pm $ 1.9 & 2.6 $ \pm $ 0.3 & 7.8e-4 $ \pm $ 1.7e-4 & 80 & 2.6  $ \pm $ 1.9  & 1.0 $ \pm $ 1.7  & 3.6  $ \pm $ 0.4 & 4.5e-4 $ \pm $ 5.6e-5  \\
      & A*-ECBS  & 100 & 3.2  $ \pm $ 2.9 & 2.4 $ \pm $ 2.5 & 2.6 $ \pm $ 0.3 & 8.2e-4 $ \pm $ 2.0e-4 & 100 & 48.6 $ \pm $ 141.1 & 4.3 $ \pm $ 10.2 & 3.7  $ \pm $ 0.4 & 6.2e-4 $ \pm $ 5.0e-4  \\
      & KCBS     & 80  & 4.7  $ \pm $ 12.0 & 1.3 $ \pm $ 1.3 & 2.4 $ \pm $ 0.5 & 1.4e-3 $ \pm $ 2.4e-4 & 90 & 8.3  $ \pm $ 15.9 & 0.8 $ \pm $ 1.6  & 3.6  $ \pm $ 0.6 & 1.2e-3 $ \pm $ 3.1e-4  \\
      & MMD-CBS  & 100 & 33.4 $ \pm $ 9.2 & 1.7 $ \pm $ 1.3 & 9.7 $ \pm $ 0.3 & 2.0e-2 $ \pm $ 1.5e-3 & 100 & 30.0 $ \pm $ 7.5  & 1.4 $ \pm $ 1.1  & 22.3 $ \pm $ 0.5 & 5.0e-2 $ \pm $ 2.9e-3  \\
      & MDOC-CBS & 100 & 5.4  $ \pm $ 3.9 & 1.9 $ \pm $ 1.9 & 2.9 $ \pm $ 0.3 & 3.2e-4 $ \pm $ 4.2e-5 & 100 & 4.2  $ \pm $ 1.5  & 0.8 $ \pm $ 0.8  & 3.8  $ \pm $ 0.4 & 2.1e-3 $ \pm $ 1.7e-3  \\
    \midrule
    \multirow{5}{*}{12}
      & A*-CBS   & 70  & 5.7  $ \pm $ 7.7 & 11.7 $ \pm $ 22.0 & 2.6 $ \pm $ 0.5 & 7.3e-4 $ \pm $ 1.9e-4 & 100 & 4.5  $ \pm $ 2.6  & 3.5 $ \pm $ 3.6 & 3.8 $ \pm $ 0.5 & 4.9e-4 $ \pm $ 7.6e-5  \\
      & A*-ECBS  & 100 & 13.7 $ \pm $ 32.0 & 15.2 $ \pm $ 41.0 & 2.5 $ \pm $ 0.4 & 7.8e-4 $ \pm $ 2.1e-4 & 100 & 5.4  $ \pm $ 2.6  & 1.1 $ \pm $ 1.4 & 3.8 $ \pm $ 0.5 & 4.9e-4 $ \pm $ 8.6e-5  \\
      & KCBS     & 50  & 21.1 $ \pm $ 31.0 & 3.0 $ \pm $ 1.2 & 2.3 $ \pm $ 0.4 & 1.4e-3 $ \pm $ 1.6e-4 & 40  & 35.5 $ \pm $ 28.7 & 2.0 $ \pm $ 0.8 & 3.2 $ \pm $ 0.1 & 1.0e-3 $ \pm $ 4.9e-5  \\
      & MMD-CBS  & 100 & 50.5 $ \pm $ 8.0 & 3.3 $ \pm $ 1.2 & 9.8 $ \pm $ 0.3 & 2.0e-2 $ \pm $ 1.8e-3 & 100 & 44.5 $ \pm $ 11.5 & 2.2 $ \pm $ 1.6 & 22.4 $ \pm $ 0.6 & 5.0e-2 $ \pm $ 3.3e-3  \\
      & MDOC-CBS & 100 & 8.9  $ \pm $ 4.7 & 3.1  $ \pm $ 2.1 & 2.9 $ \pm $ 0.2 & 3.8e-4 $ \pm $ 1.5e-4 & 100 & 7.5  $ \pm $ 5.6  & 1.6 $ \pm $ 1.8 & 3.7 $ \pm $ 0.4 & 1.9e-3 $ \pm $ 1.9e-3 \\
    \midrule
    \multirow{5}{*}{15}
      & A*-CBS   & 100 & 14.9 $ \pm $ 25.0 & 23.0 $ \pm $ 26.0 & 2.8 $ \pm $ 0.3 & 8.0e-4 $ \pm $ 2.4e-4 & 100 & 44.0 $ \pm $ 90.9 & 29.2 $ \pm $ 56.2 & 4.1 $ \pm $ 0.6 & 6.3e-4 $ \pm $ 5.0e-4  \\
      & A*-ECBS  & 100 & 20.1 $ \pm $ 39.0 & 11.4 $ \pm $ 17.0 & 2.8 $ \pm $ 0.3 & 8.0e-4 $ \pm $ 2.1e-4 & 100 & 18.0 $ \pm $ 15.9 & 5.5  $ \pm $ 5.3  & 4.1 $ \pm $ 0.5 & 5.0e-4 $ \pm $ 1.1e-4  \\
      & KCBS     & 20  & 0.5  $ \pm $ 0.5  & 3.5 $ \pm $ 5.0  & 1.9 $ \pm $ 0.4 & 1.4e-3 $ \pm $ 1.9e-4 & 60 & 71.9 $ \pm $ 64.1 & 6.0 $ \pm $ 3.6 & 3.6  $ \pm $ 0.5 & 1.2e-3 $ \pm $ 2.3e-4  \\
      & MMD-CBS  & 100 & 87.1 $ \pm $ 24.0 & 6.6 $ \pm $ 2.8  & 10.1 $ \pm $ 0.3 & 3.0e-2 $ \pm $ 2.0e-3 & 100 & 76.3 $ \pm $ 18.4 & 5.4 $ \pm $ 2.1 & 22.8 $ \pm $ 0.6 & 6.0e-2 $ \pm $ 3.0e-3  \\
      & MDOC-CBS & 100 & 22.4 $ \pm $ 15.0 & 7.1 $ \pm $ 4.2  & 2.9 $ \pm $ 0.2 & 4.3e-4 $ \pm $ 3.0e-4 & 100 & 14.8 $ \pm $ 5.8  & 4.0 $ \pm $ 1.9 & 4.0  $ \pm $ 0.3 & 2.6e-3 $ \pm $ 1.1e-3  \\
    \midrule
     \multirow{5}{*}{20}
      & A*-CBS   & 40  & 5.4  $ \pm $ 3.9  & 11.0 $ \pm $ 12.0 & 2.6 $ \pm $ 0.4 & 7.6e-4 $ \pm $ 7.9e-5 & 60  & 42.5 $ \pm $ 61.4 & 57.5 $ \pm $ 103.2 & 3.7 $ \pm $ 0.4 & 5.3e-4 $ \pm $ 1.3e-4  \\
      & A*-ECBS  & 100 & 31.1 $ \pm $ 57.0 & 22.3 $ \pm $ 36.0 & 2.7 $ \pm $ 0.3 & 7.5e-4 $ \pm $ 1.4e-4 & 80  & 51.4 $ \pm $ 75.1 & 11.9 $ \pm $ 11.4 & 3.9 $ \pm $ 0.5 & 5.4e-4 $ \pm $ 1.2e-4  \\
      & KCBS     & 0   & -- & -- & -- & -- & 10  & 1.0  & 2.0  & 3.7 & 1.3e-3   \\
      & MMD-CBS  & 100 & 161.8 $ \pm $ 48.0 & 13.1 $ \pm $ 5.2 & 10.2 $ \pm $ 0.2 & 3.0e-2 $ \pm $ 9.4e-4 & 100 & 121.0 $ \pm $ 29.7 & 8.9 $ \pm $ 3.1 & 23.4 $ \pm $ 0.6 & 6.0e-2 $ \pm $ 3.2e-3  \\
      & MDOC-CBS & 100 & 39.1 $ \pm $ 16.0 & 12.2 $ \pm $ 4.8 & 2.8 $ \pm $ 0.2 & 4.0e-4 $ \pm $ 1.1e-4 & 100 & 38.9 $ \pm $ 27.6 & 8.2 $ \pm $ 6.1 & 4.0 $ \pm $ 0.3 & 3.8e-3 $ \pm $ 1.9e-3 \\
    \midrule
    \multirow{5}{*}{25}
      & A*-CBS   & 60  & 42.4 $ \pm $ 45.0 & 111.3 $ \pm $ 140.0 & 2.4 $ \pm $ 0.2 & 8.5e-4 $ \pm $ 1.4e-4 & 60 & 39.0 $ \pm $ 42.0  & 63.0 $ \pm $ 68.4  & 3.8 $ \pm $ 0.5  & 5.6e-4 $ \pm $ 4.7e-5  \\
      & A*-ECBS  & 100 & 48.7 $ \pm $ 65.0 & 24.4  $ \pm $ 17.0  & 2.5 $ \pm $ 0.3 & 8.9e-4 $ \pm $ 1.4e-4 & 80 & 84.2 $ \pm $ 102.8 & 39.5 $ \pm $ 49.9 & 3.9 $ \pm $ 0.5  & 5.6e-4 $ \pm $ 4.9e-5  \\
      & KCBS     & 0   & -- & -- & -- & -- & 40 & 105.0 $ \pm $ 55.6 & 10.2 $ \pm $ 4.0  & 3.8 $ \pm $ 0.5  & 1.3e-3 $ \pm $ 1.1e-4   \\
      & MMD-CBS  & 100 & 221.8 $ \pm $ 34.0 & 17.5 $ \pm $ 3.6 & 10.3 $ \pm $ 0.2 & 3.0e-2 $ \pm $ 1.1e-3 & 100 & 194.2 $ \pm $ 58.6 & 14.8 $ \pm $ 5.7  & 23.6 $ \pm $ 0.5 & 6.0e-2 $ \pm $ 2.9e-3  \\
      & MDOC-CBS & 100 & 92.7  $ \pm $ 46.0 & 22.4 $ \pm $ 13.0 & 2.8 $ \pm $ 0.1 & 3.7e-4 $ \pm $ 7.6e-5 & 100 & 54.5 $ \pm $ 22.4  & 10.3 $ \pm $ 4.7  & 3.9 $ \pm $ 0.3  & 2.9e-3 $ \pm $ 1.8e-3 \\
    \midrule
    \multirow{5}{*}{30}
      & A*-CBS   & 10  & 84.0              & 54.0              & 2.5             & 1.3e-3  & 40  & 27.6 $ \pm $ 24.7 & 30.8 $ \pm $ 28.9 & 3.9 $ \pm $ 0.1 & 5.5e-4 $ \pm $ 1.5e-4 \\
      & A*-ECBS  & 70  & 99.4 $ \pm $ 54.2 & 56.7 $ \pm $ 53.9 & 2.6 $ \pm $ 0.2 & 1.1e-3 $ \pm $ 1.9e-4 & 70  & 57.9 $ \pm $ 23.7 & 20.1 $ \pm $ 12.4 & 3.9 $ \pm $ 0.2 & 4.9e-4 $ \pm $ 6.9e-5 \\
      & KCBS     & 0   & -- & -- & -- & -- & 0   & --               & --               & --              & --  \\
      & MMD-CBS  & 80  & 329.1 $ \pm $ 55.1 & 25.4 $ \pm $ 6.2 & 10.5 $ \pm $ 0.2 & 3.0e-2 $ \pm $ 1.0e-3 & 100 & 314.1 $ \pm $ 61.6 & 22.1 $ \pm $ 5.7 & 24.0 $ \pm $ 0.6 & 6.0e-2 $ \pm $ 3.2e-3 \\
      & MDOC-CBS & 100 & 180.3 $ \pm $ 88.3 & 38.6 $ \pm $ 18.0 & 2.9 $ \pm $ 0.2 & 7.6e-4 $ \pm $ 3.8e-4 & 100 & 130.2 $ \pm $ 69.2 & 19.0 $ \pm $ 9.0 & 4.0 $ \pm $ 0.2  & 3.2e-3 $ \pm $ 1.6e-3 \\
    \midrule
    \multirow{5}{*}{35}
      & A*-CBS   & 40 & 39.8 $ \pm $ 27.0 & 90.3 $ \pm $ 59.0 & 2.3 $ \pm $ 0.2 & 9.7e-4 $ \pm $ 2.1e-4 & 40  & 80.7 $ \pm $ 62.0   & 56.8 $ \pm $ 31.3 & 4.0 $ \pm $ 0.2  & 6.7e-4 $ \pm $ 1.5e-4  \\
      & A*-ECBS  & 100 & 90.1 $ \pm $ 61.0 & 62.2 $ \pm $ 43.0 & 2.6 $ \pm $ 0.3 & 9.7e-4 $ \pm $ 1.9e-4 & 80  & 162.2 $ \pm $ 128.4 & 28.8 $ \pm $ 8.7  & 4.1 $ \pm $ 0.2  & 6.2e-4 $ \pm $ 1.4e-4  \\
      & KCBS     & 0 & -- & -- & -- & -- & 0 & -- & -- & -- & --  \\
      & MMD-CBS  & 40 & 426.0 $ \pm $ 33.0 & 30.0 $ \pm $ 1.8 & 10.6 $ \pm $ 0.1 & 3.0e-2 $ \pm $ 3.2e-4 & 70  & 342.8 $ \pm $ 86.1  & 24.1 $ \pm $ 6.7  & 23.8 $ \pm $ 0.5 & 6.0e-2 $ \pm $ 2.4e-3  \\
      & MDOC-CBS & 80 & 304.1 $ \pm $ 100.0 & 51.6 $ \pm $ 15.0 & 2.9 $ \pm $ 0.1 & 1.2e-3 $ \pm $ 8.0e-4 & 90  & 212.2 $ \pm $ 76.4  & 24.6 $ \pm $ 7.7  & 3.8 $ \pm $ 0.2  & 2.7e-3 $ \pm $ 1.1e-3  \\
    \midrule
    \multirow{5}{*}{40}
      & A*-CBS   & 20  & 86.7 $ \pm $ 43.0  & 111.0 $ \pm $ 37.0 & 2.7 $ \pm $ 0.1 & 1.4e-3 $ \pm $ 1.2e-5 & 10 & 207.3              & 172.0             & 4.2            & 5.5e-4  \\
      & A*-ECBS  & 60  & 173.8 $ \pm $ 110.0 & 60.3 $ \pm $ 30.0 & 2.7 $ \pm $ 0.2 & 1.1e-3 $ \pm $ 1.9e-4 & 40 & 222.9 $ \pm $ 122.3 & 45.3 $ \pm $ 29.4 & 4.2 $ \pm $ 0.1 & 5.4e-4 $ \pm $ 4.4e-5  \\
      & KCBS     & 0   & -- & -- & -- & -- & 0 & -- & -- & -- & --  \\
      & MMD-CBS  & 10  & 424.6 & 29.0 & 10.5 & 3.0e-2 & 0 & -- & -- & -- & --  \\
      & MDOC-CBS & 60  & 402.4 $ \pm $ 51.0 & 55.7 $ \pm $ 5.1 & 2.8 $ \pm $ 0.2 & 7.9e-4 $ \pm $ 8.2e-4 & 100 & 274.8 $ \pm $ 111.2 & 35.0 $ \pm $ 13.0 & 3.8 $ \pm $ 0.2 & 2.8e-3 $ \pm $ 1.3e-3 \\
    \bottomrule

\end{tabular}
\caption{Results on Empty Large map $4 \times 4$ and Empty ExtraLarge map $6 \times 6$ under a 500 running time-limit. All robots have random start-goal pairs. Columns correspond to  \textbf{S}($\uparrow$): success rate(\%), \textbf{T}($\downarrow$): planning time, \textbf{E}($\downarrow$): CT expansions, \textbf{P}($\downarrow$): path length, and \textbf{G}($\downarrow$): geometric smoothness. All metrics are computed based on success trails.}
\end{table*}

%% file: figures/tex/app_vis.tex
\begin{figure*}[ht]
    \centering
    \setlength{\tabcolsep}{1em} 
    \renewcommand{\arraystretch}{0.5} 
    
    \begin{tabular}{p{0.02\textwidth}ccccc}
        & \textbf{A*-CBS} & \textbf{A*ECBS} & \textbf{KCBS} & \textbf{MMD-CBS} & \textbf{MDOC-CBS} \\
        \rotatebox{90}{\textbf{3 Robots}} &
        \subcaptionbox*{}[0.15\textwidth]{%
            \includegraphics[width=\linewidth]{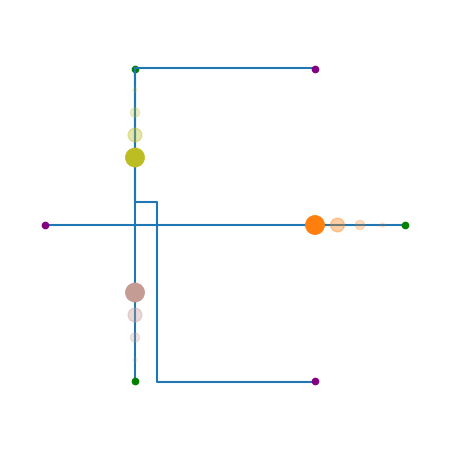}
        } &
        \subcaptionbox*{}[0.15\textwidth]{%
            \centering
            \includegraphics[width=\linewidth]{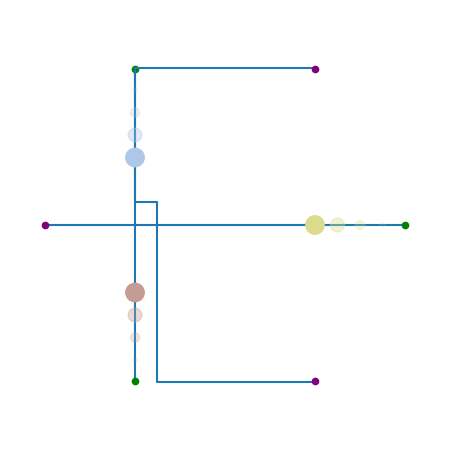}
        } &
        \subcaptionbox*{}[0.15\textwidth]{%
            \includegraphics[width=\linewidth]{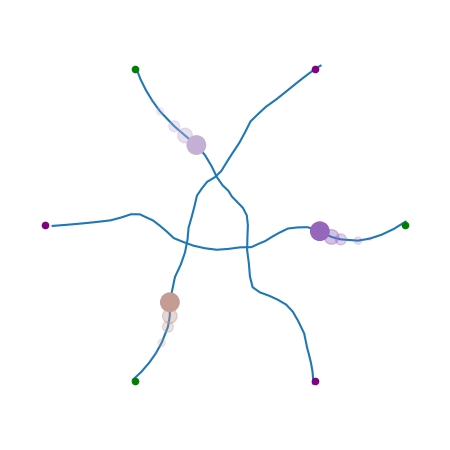}
        } &
        \subcaptionbox*{}[0.15\textwidth]{%
            \includegraphics[width=\linewidth]{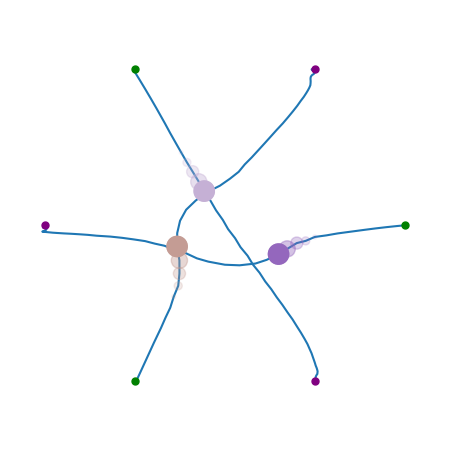}
        } &
        \subcaptionbox*{}[0.15\textwidth]{%
            \includegraphics[width=\linewidth]{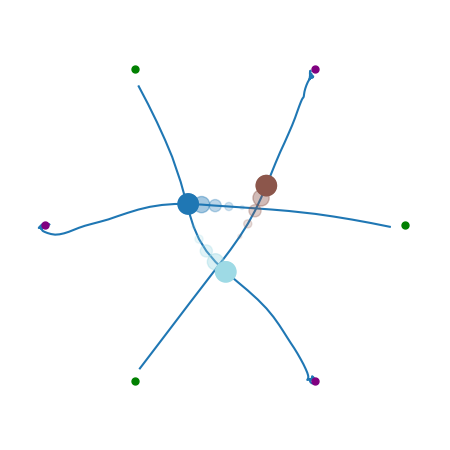}
        } \\

        \rotatebox{90}{\textbf{6 Robots}} &
        \subcaptionbox*{}[0.15\textwidth]{%
            \includegraphics[width=\linewidth]{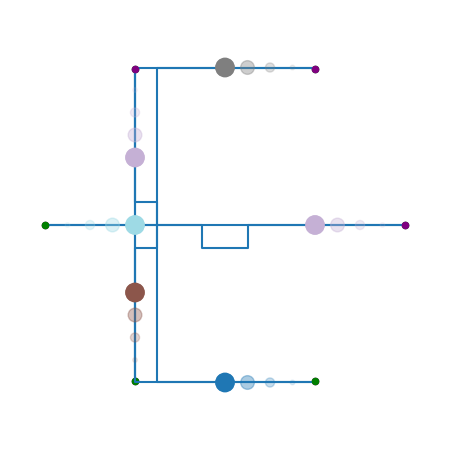}
        } &
        \subcaptionbox*{}[0.15\textwidth]{%
            \centering
            \includegraphics[width=\linewidth]{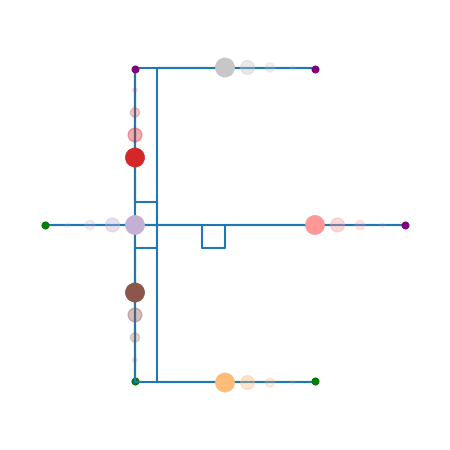}
        } &
        \subcaptionbox*{}[0.15\textwidth]{%
            \includegraphics[width=\linewidth]{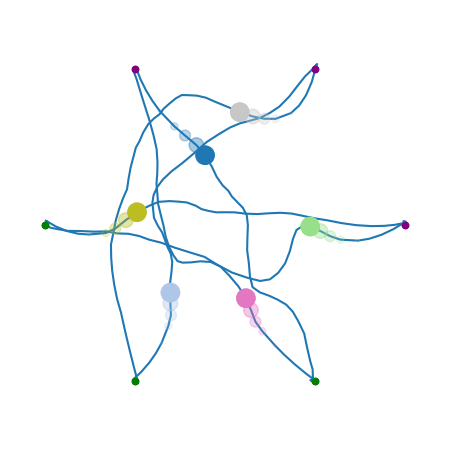}
        } &
        \subcaptionbox*{}[0.15\textwidth]{%
            \includegraphics[width=\linewidth]{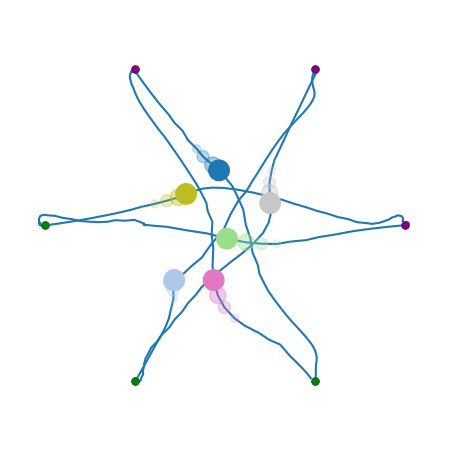}
        } &
        \subcaptionbox*{}[0.15\textwidth]{%
            \includegraphics[width=\linewidth]{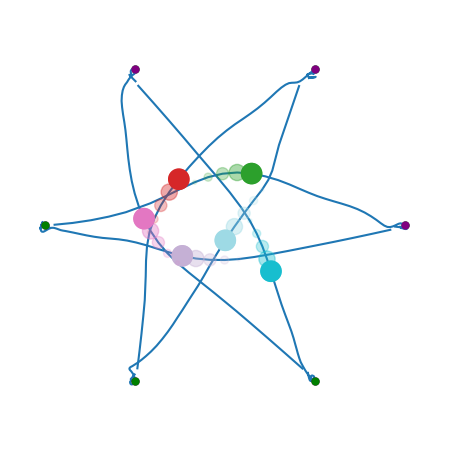}
        } \\

        \rotatebox{90}{\textbf{9 Robots}} &
        \subcaptionbox*{}[0.15\textwidth]{%
            \includegraphics[width=\linewidth]{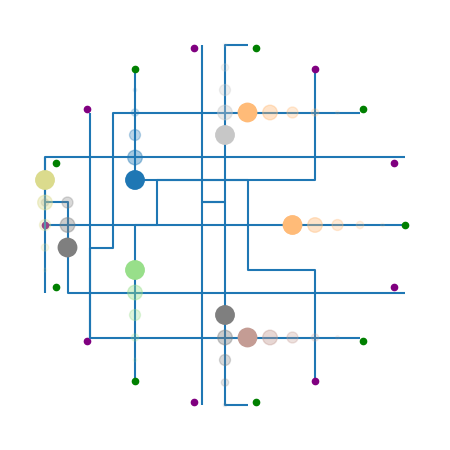}
        } &
        \subcaptionbox*{}[0.15\textwidth]{%
            \centering
            \includegraphics[width=\linewidth]{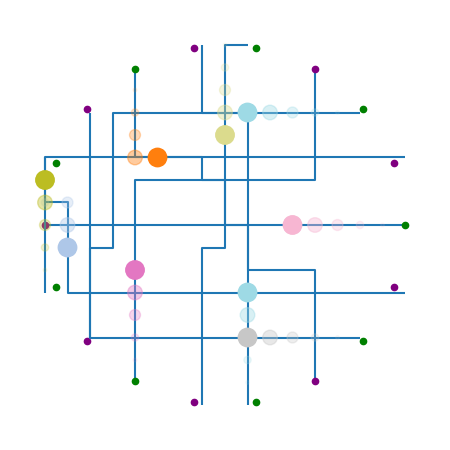}
        } &
        \subcaptionbox*{}[0.15\textwidth]{%
            \includegraphics[width=\linewidth]{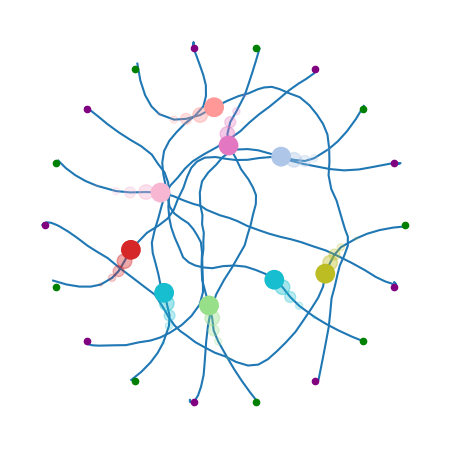}
        } &
        \subcaptionbox*{}[0.15\textwidth]{%
            \includegraphics[width=\linewidth]{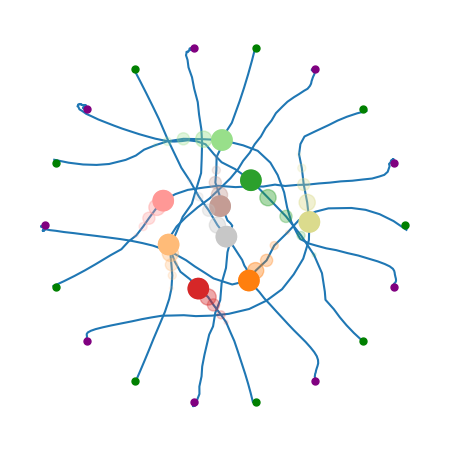}
        } &
        \subcaptionbox*{}[0.15\textwidth]{%
            \includegraphics[width=\linewidth]{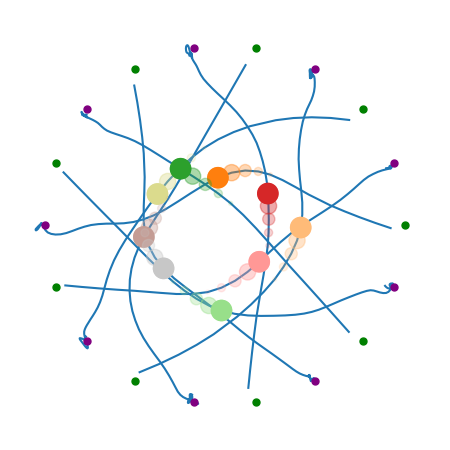}
        } \\

        \rotatebox{90}{\textbf{12 Robots}} &
        \subcaptionbox*{}[0.15\textwidth]{%
            N/A
        } &
        \subcaptionbox*{}[0.15\textwidth]{%
            \centering
            \includegraphics[width=\linewidth]{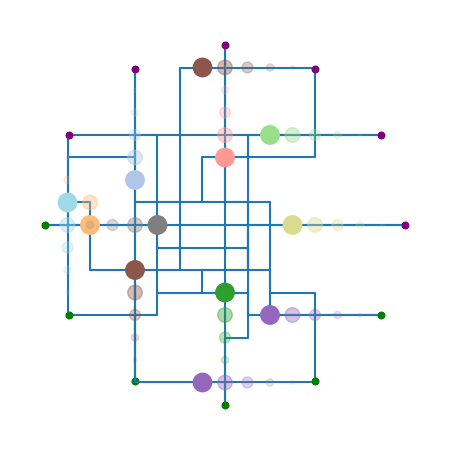}
        } &
        \subcaptionbox*{}[0.15\textwidth]{%
            \includegraphics[width=\linewidth]{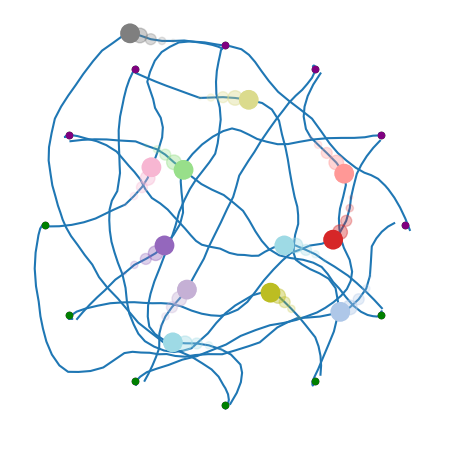}
        } &
        \subcaptionbox*{}[0.15\textwidth]{%
            N/A
        } &
        \subcaptionbox*{}[0.15\textwidth]{%
            \includegraphics[width=\linewidth]{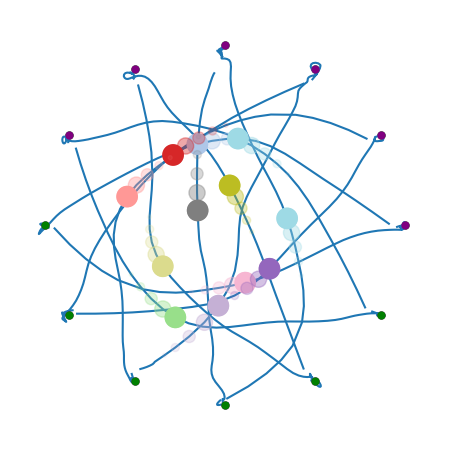}
        } \\

        \rotatebox{90}{\textbf{15 Robots}} &
        \subcaptionbox*{}[0.15\textwidth]{%
            N/A
        } &
        \subcaptionbox*{}[0.15\textwidth]{%
            \centering
            \includegraphics[width=\linewidth]{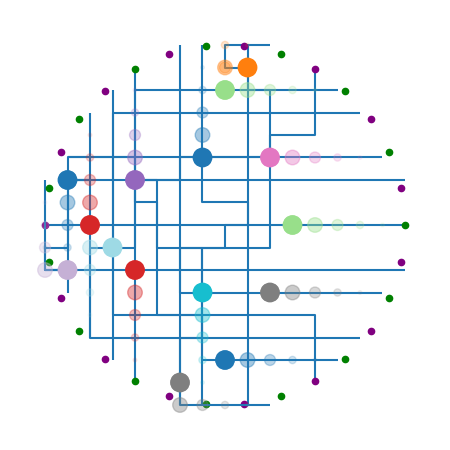}
        } &
        \subcaptionbox*{}[0.15\textwidth]{%
            \includegraphics[width=\linewidth]{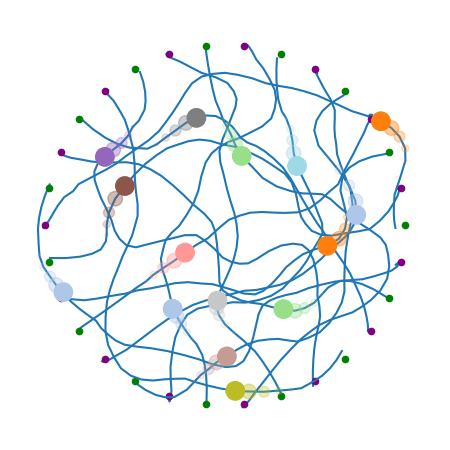}
        } &
        \subcaptionbox*{}[0.15\textwidth]{%
            N/A
        } &
        \subcaptionbox*{}[0.15\textwidth]{%
            N/A
        } \\

        \rotatebox{90}{\textbf{18 Robots}} &
        \subcaptionbox*{}[0.15\textwidth]{%
            N/A
        } &
        \subcaptionbox*{}[0.15\textwidth]{%
            N/A
        } &
        \subcaptionbox*{}[0.15\textwidth]{%
            \includegraphics[width=\linewidth]{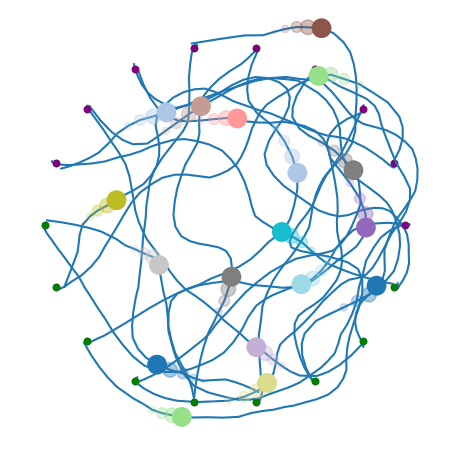}
        } &
        \subcaptionbox*{}[0.15\textwidth]{%
            N/A
        } &
        \subcaptionbox*{}[0.15\textwidth]{%
            N/A
        } \\

        \rotatebox{90}{\textbf{20 Robots}} &
        \subcaptionbox*{}[0.15\textwidth]{%
            N/A
        } &
        \subcaptionbox*{}[0.15\textwidth]{%
            N/A   
        } &
        \subcaptionbox*{}[0.15\textwidth]{%
            N/A
        } &
        \subcaptionbox*{}[0.15\textwidth]{%
            N/A
        } &
        \subcaptionbox*{}[0.15\textwidth]{%
            N/A
        } \\
        
    \end{tabular}
    \caption{Some visual examples of trajectories generated by A*-CBS, A*-ECBS, KCBS, MMD-CBS, and MDOC-CBS in the Empty map with the Circle setup. We choose robot number \{3, 6, 9, 12, 15, 18, 20\}. N/A denotes the failure among all trials. Trajectories of KCBS, MMD-CBS, and MDOC-CBS are displayed after post-smoothing for better visualization.}
    \label{fig:empty_circle_trajs}
\end{figure*}

\begin{figure*}[ht]
    \centering
    \setlength{\tabcolsep}{1em} 
    \renewcommand{\arraystretch}{0.5} 
    
    \begin{tabular}{p{0.02\textwidth}ccccc}
        & \textbf{A*-CBS} & \textbf{A*ECBS} & \textbf{KCBS} & \textbf{MMD-CBS} & \textbf{MDOC-CBS} \\
        \rotatebox{90}{\textbf{3 Robots}} &
        \subcaptionbox*{}[0.15\textwidth]{%
            \includegraphics[width=\linewidth]{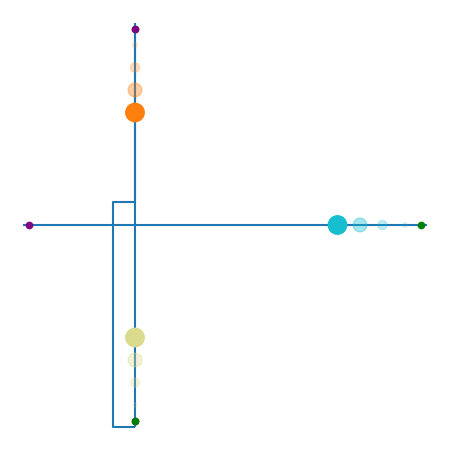}
        } &
        \subcaptionbox*{}[0.15\textwidth]{%
            \centering
            \includegraphics[width=\linewidth]{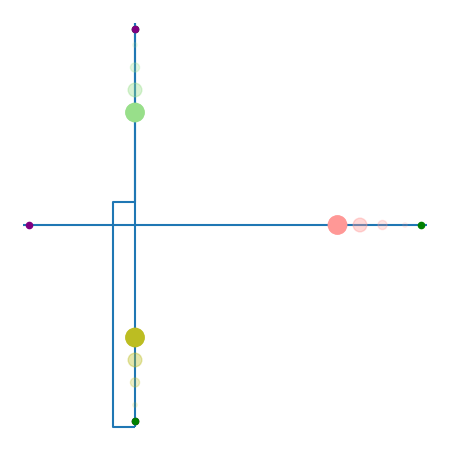}
        } &
        \subcaptionbox*{}[0.15\textwidth]{%
            \includegraphics[width=\linewidth]{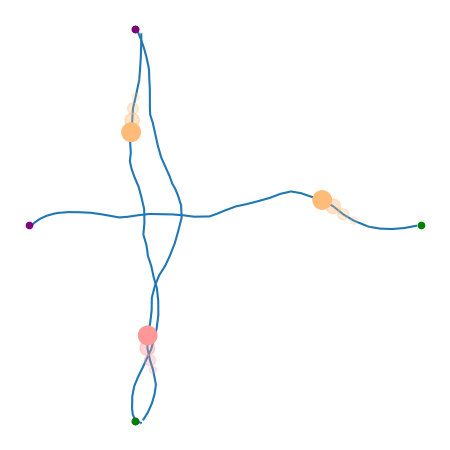}
        } &
        \subcaptionbox*{}[0.15\textwidth]{%
            \includegraphics[width=\linewidth]{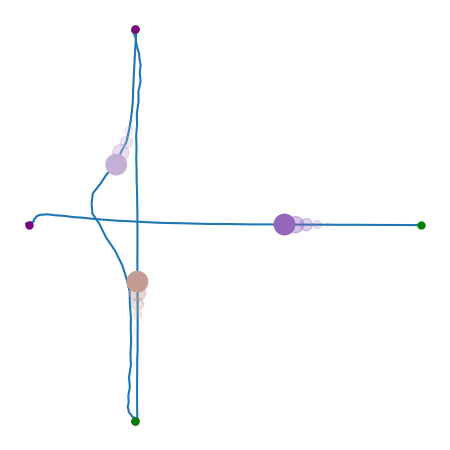}
        } &
        \subcaptionbox*{}[0.15\textwidth]{%
            \includegraphics[width=\linewidth]{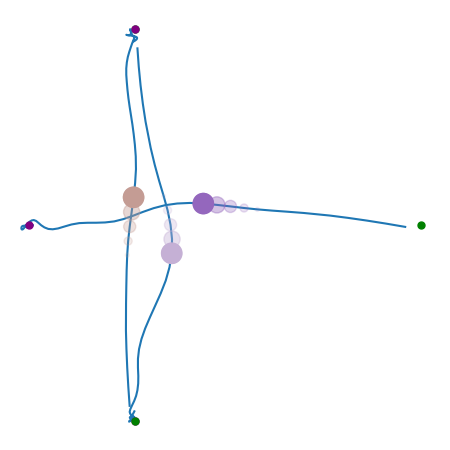}
        } \\

        \rotatebox{90}{\textbf{6 Robots}} &
        \subcaptionbox*{}[0.15\textwidth]{%
            \includegraphics[width=\linewidth]{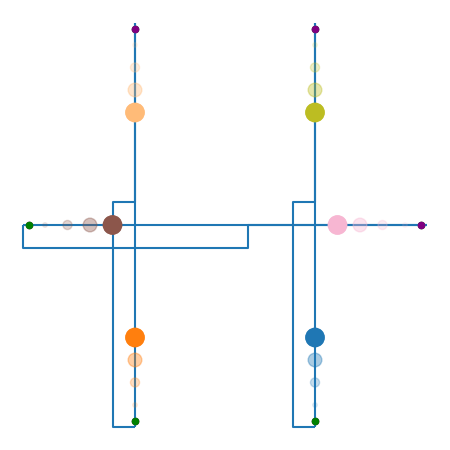}
        } &
        \subcaptionbox*{}[0.15\textwidth]{%
            \centering
            \includegraphics[width=\linewidth]{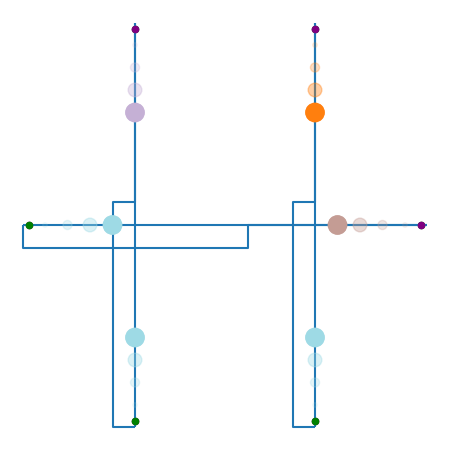}
        } &
        \subcaptionbox*{}[0.15\textwidth]{%
            \includegraphics[width=\linewidth]{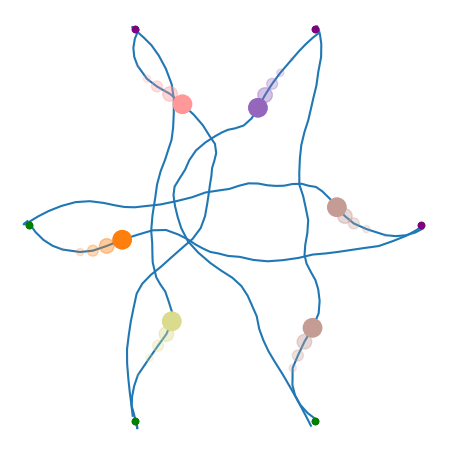}
        } &
        \subcaptionbox*{}[0.15\textwidth]{%
            \includegraphics[width=\linewidth]{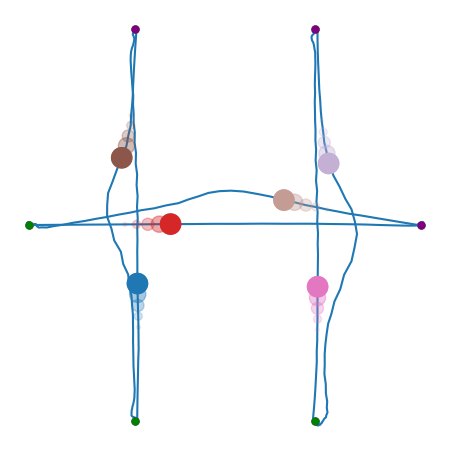}
        } &
        \subcaptionbox*{}[0.15\textwidth]{%
            \includegraphics[width=\linewidth]{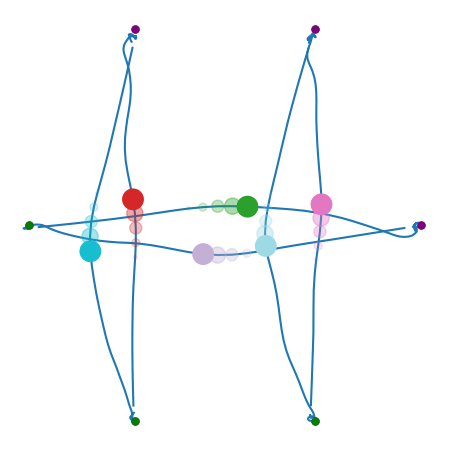}
        } \\

        \rotatebox{90}{\textbf{9 Robots}} &
        \subcaptionbox*{}[0.15\textwidth]{%
            \includegraphics[width=\linewidth]{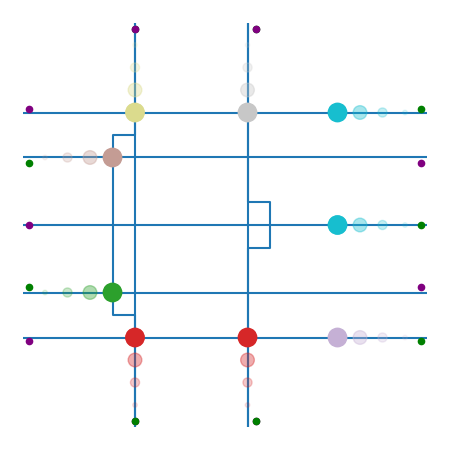}
        } &
        \subcaptionbox*{}[0.15\textwidth]{%
            \centering
            \includegraphics[width=\linewidth]{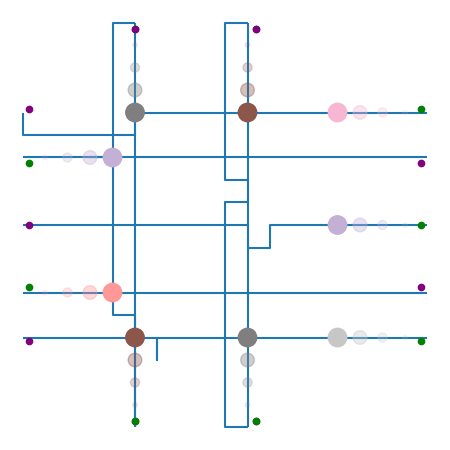}
        } &
        \subcaptionbox*{}[0.15\textwidth]{%
            \includegraphics[width=\linewidth]{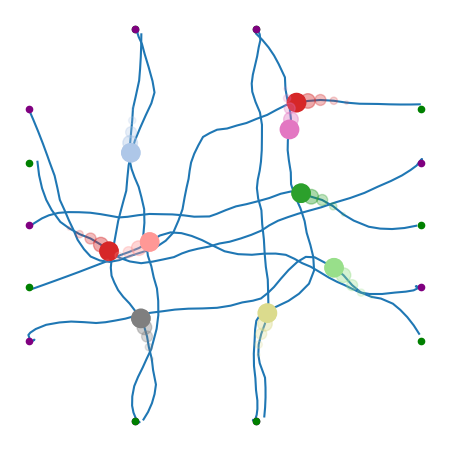}
        } &
        \subcaptionbox*{}[0.15\textwidth]{%
            \includegraphics[width=\linewidth]{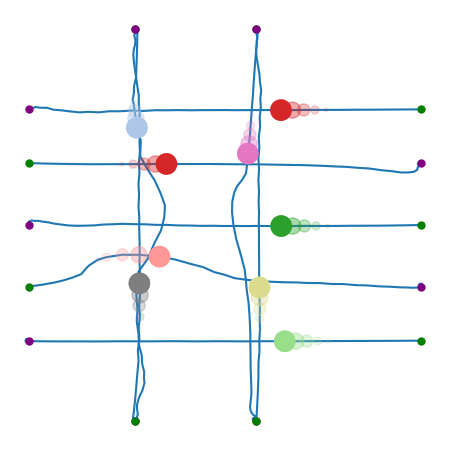}
        } &
        \subcaptionbox*{}[0.15\textwidth]{%
            \includegraphics[width=\linewidth]{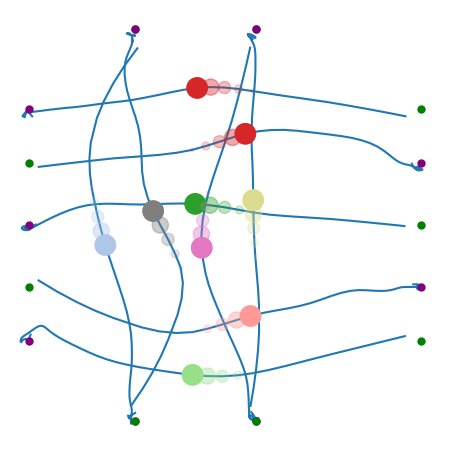}
        } \\

        \rotatebox{90}{\textbf{12 Robots}} &
        \subcaptionbox*{}[0.15\textwidth]{%
            \includegraphics[width=\linewidth]{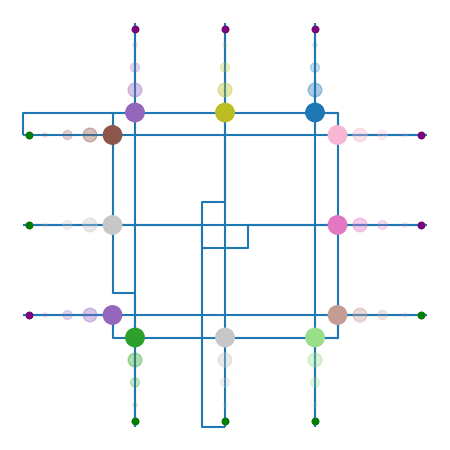}
        } &
        \subcaptionbox*{}[0.15\textwidth]{%
            \centering
            \includegraphics[width=\linewidth]{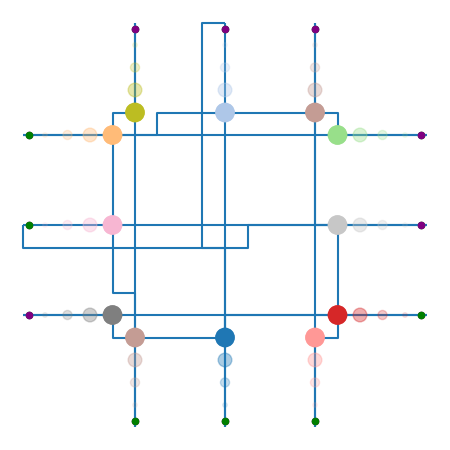}
        } &
        \subcaptionbox*{}[0.15\textwidth]{%
            \includegraphics[width=\linewidth]{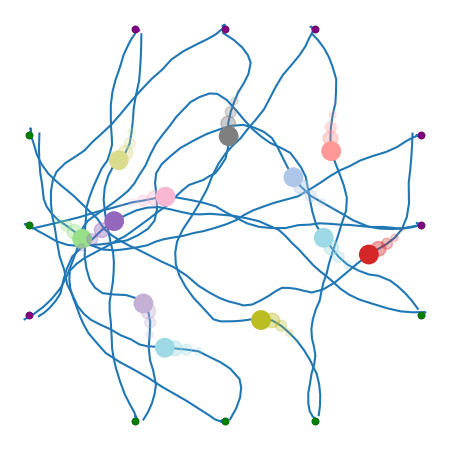}
        } &
        \subcaptionbox*{}[0.15\textwidth]{%
            \includegraphics[width=\linewidth]{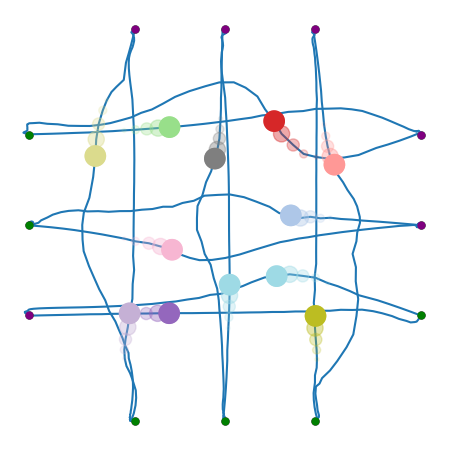}
        } &
        \subcaptionbox*{}[0.15\textwidth]{%
            \includegraphics[width=\linewidth]{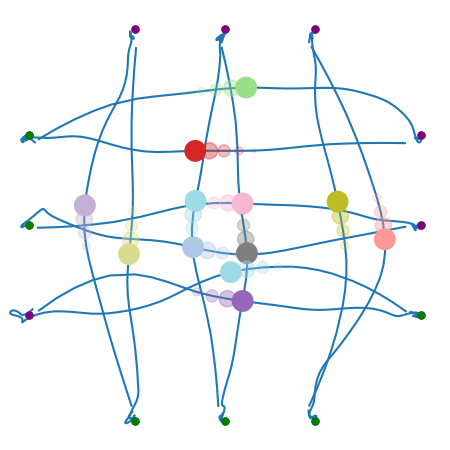}
        } \\

        \rotatebox{90}{\textbf{15 Robots}} &
        \subcaptionbox*{}[0.15\textwidth]{%
            \includegraphics[width=\linewidth]{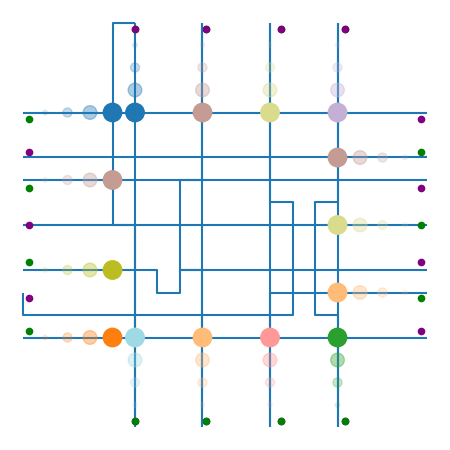}
        } &
        \subcaptionbox*{}[0.15\textwidth]{%
            \centering
            \includegraphics[width=\linewidth]{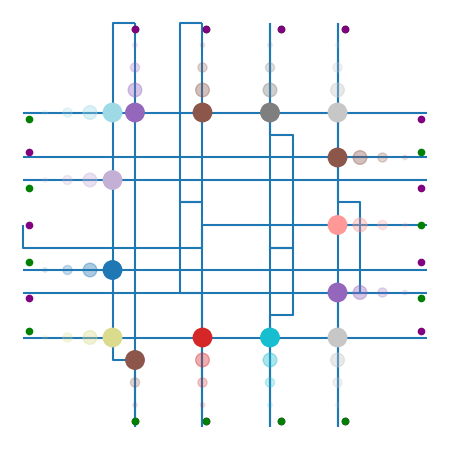}
        } &
        \subcaptionbox*{}[0.15\textwidth]{%
            \includegraphics[width=\linewidth]{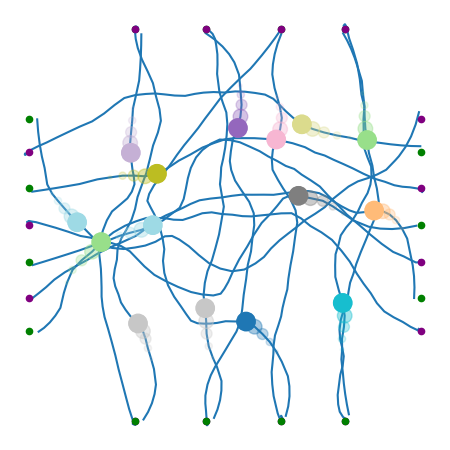}
        } &
        \subcaptionbox*{}[0.15\textwidth]{%
            \includegraphics[width=\linewidth]{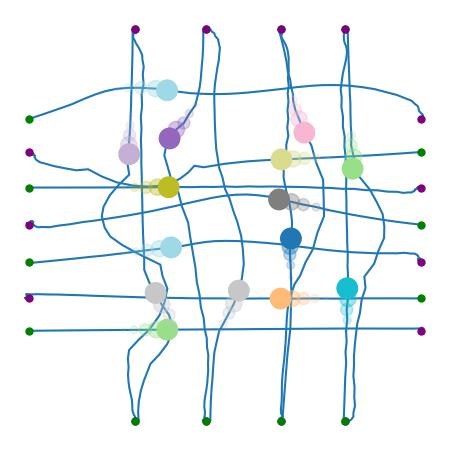}
        } &
        \subcaptionbox*{}[0.15\textwidth]{%
            \includegraphics[width=\linewidth]{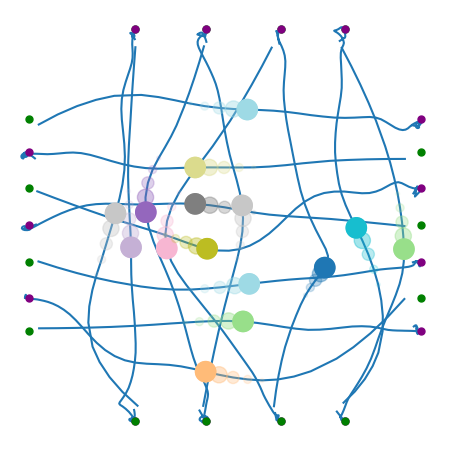}
        } \\

        \rotatebox{90}{\textbf{18 Robots}} &
        \subcaptionbox*{}[0.15\textwidth]{%
            \includegraphics[width=\linewidth]{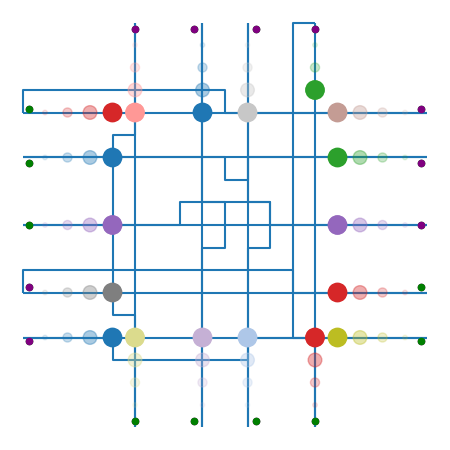}
        } &
        \subcaptionbox*{}[0.15\textwidth]{%
            \centering
            \includegraphics[width=\linewidth]{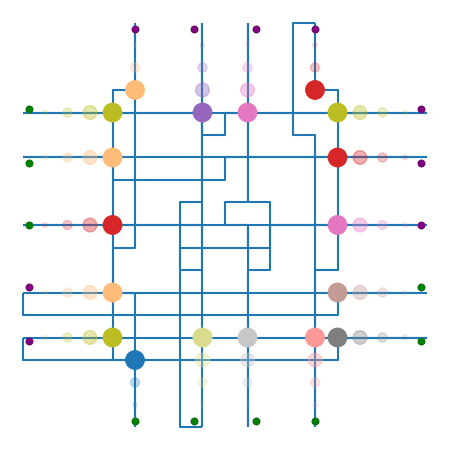}
        } &
        \subcaptionbox*{}[0.15\textwidth]{%
            \includegraphics[width=\linewidth]{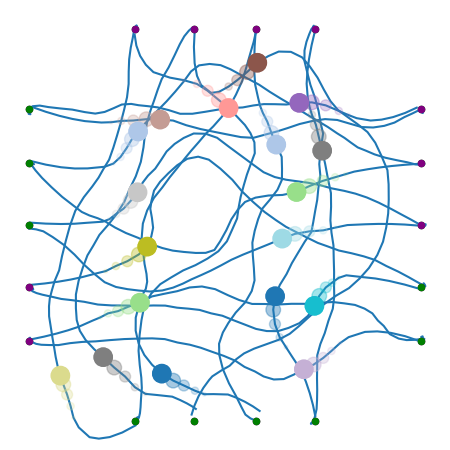}
        } &
        \subcaptionbox*{}[0.15\textwidth]{%
            \includegraphics[width=\linewidth]{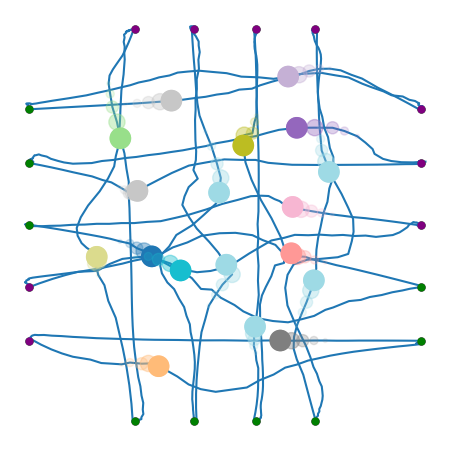}
        } &
        \subcaptionbox*{}[0.15\textwidth]{%
            \includegraphics[width=\linewidth]{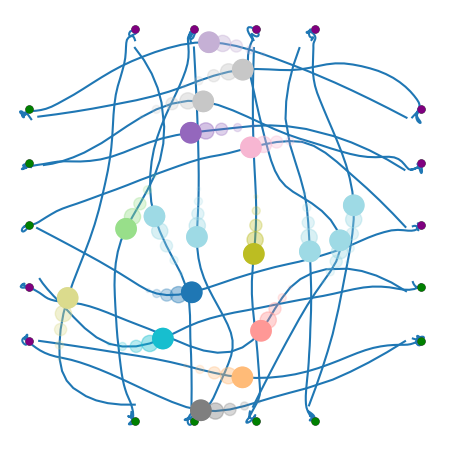}
        } \\

        \rotatebox{90}{\textbf{20 Robots}} &
        \subcaptionbox*{}[0.15\textwidth]{%
            \includegraphics[width=\linewidth]{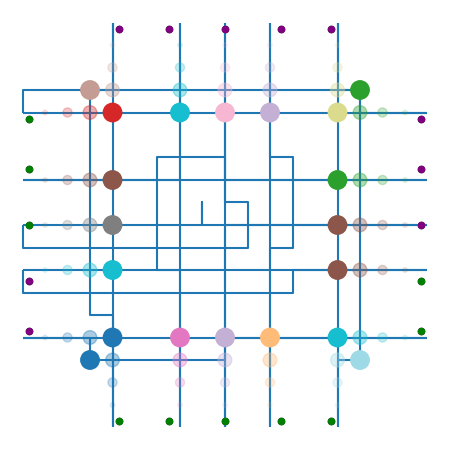}
        } &
        \subcaptionbox*{}[0.15\textwidth]{%
            \includegraphics[width=\linewidth]{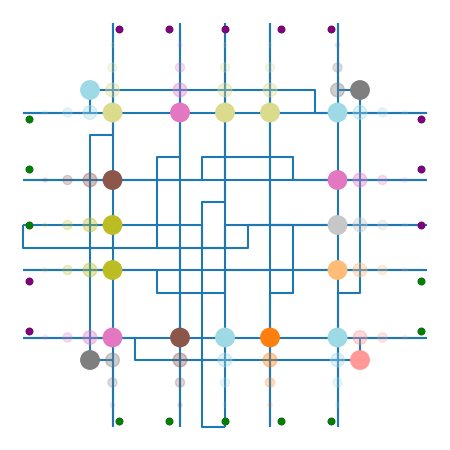}  
        } &
        \subcaptionbox*{}[0.15\textwidth]{%
            N/A
        } &
        \subcaptionbox*{}[0.15\textwidth]{%
            \includegraphics[width=\linewidth]{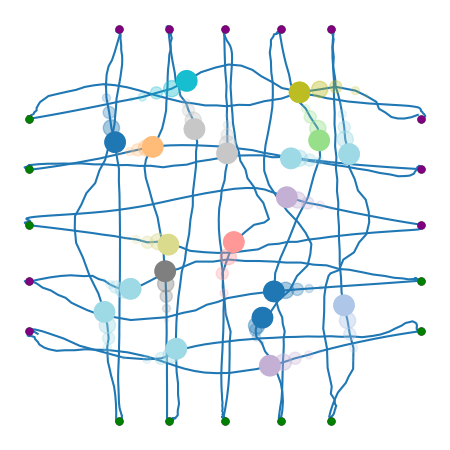}
        } &
        \subcaptionbox*{}[0.15\textwidth]{%
            \includegraphics[width=\linewidth]{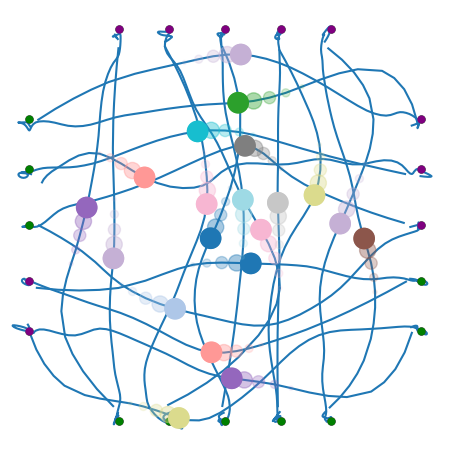}
        } \\
        
    \end{tabular}
    \caption{Some visual examples of trajectories generated by A*-CBS, A*-ECBS, KCBS, MMD-CBS, and MDOC-CBS in the Empty map with the Weave setup. We choose robot number \{3, 6, 9, 12, 15, 18, 20\}. N/A denotes the failure among all trials. Trajectories of KCBS, MMD-CBS, and MDOC-CBS are displayed after post-smoothing for better visualization.}
    \label{fig:emptyweave_trajs}
\end{figure*}

\begin{figure*}[ht]
    \centering
    \setlength{\tabcolsep}{1em} 
    \renewcommand{\arraystretch}{0.5} 
    
    \begin{tabular}{p{0.02\textwidth}ccccc}
        & \textbf{A*-CBS} & \textbf{A*ECBS} & \textbf{KCBS} & \textbf{MMD-CBS} & \textbf{MDOC-CBS} \\
        \rotatebox{90}{\textbf{3 Robots}} &
        \subcaptionbox*{}[0.15\textwidth]{%
            \includegraphics[width=\linewidth]{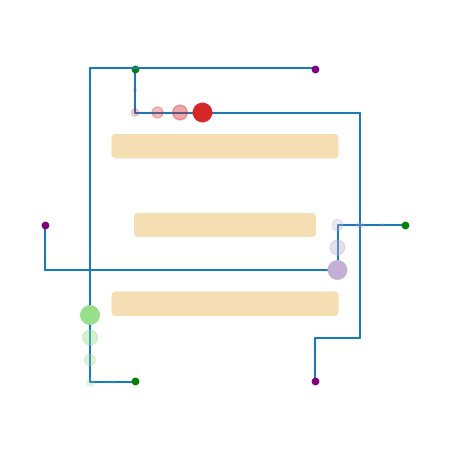}
        } &
        \subcaptionbox*{}[0.15\textwidth]{%
            \centering
            \includegraphics[width=\linewidth]{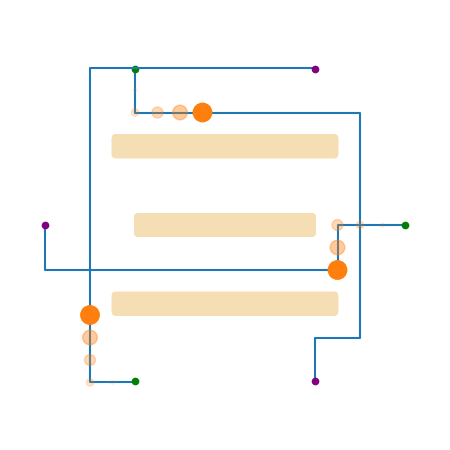}
        } &
        \subcaptionbox*{}[0.15\textwidth]{%
            \includegraphics[width=\linewidth]{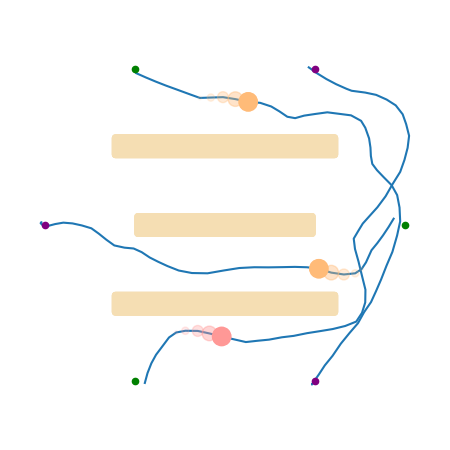}
        } &
        \subcaptionbox*{}[0.15\textwidth]{%
            \includegraphics[width=\linewidth]{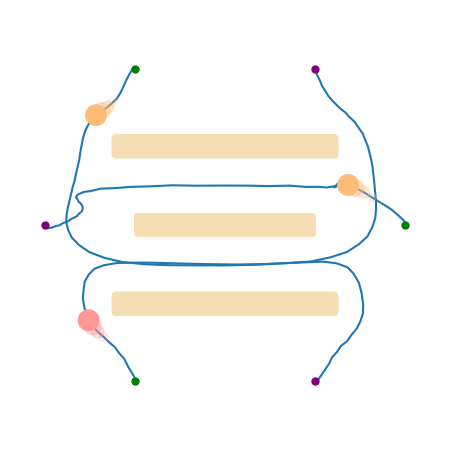}
        } &
        \subcaptionbox*{}[0.15\textwidth]{%
            \includegraphics[width=\linewidth]{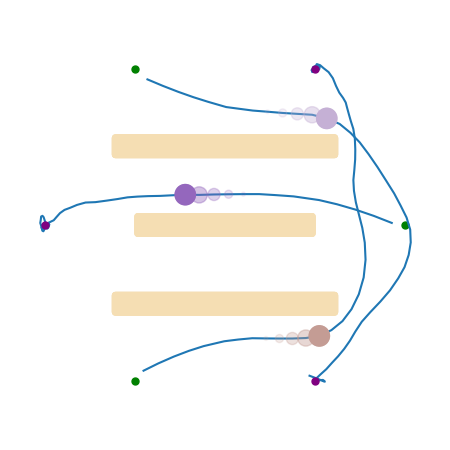}
        } \\

        \rotatebox{90}{\textbf{6 Robots}} &
        \subcaptionbox*{}[0.15\textwidth]{%
            \includegraphics[width=\linewidth]{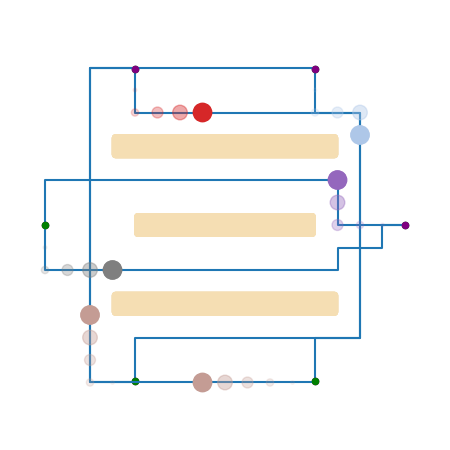}
        } &
        \subcaptionbox*{}[0.15\textwidth]{%
            \centering
            \includegraphics[width=\linewidth]{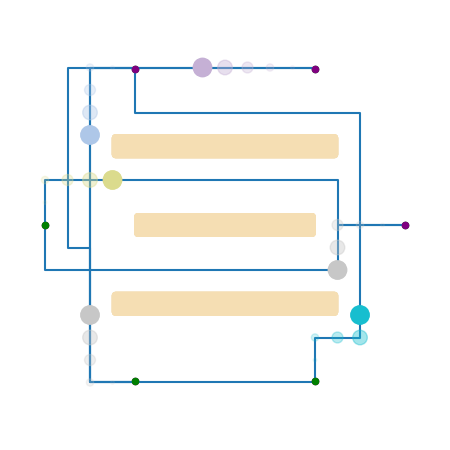}
        } &
        \subcaptionbox*{}[0.15\textwidth]{%
            \includegraphics[width=\linewidth]{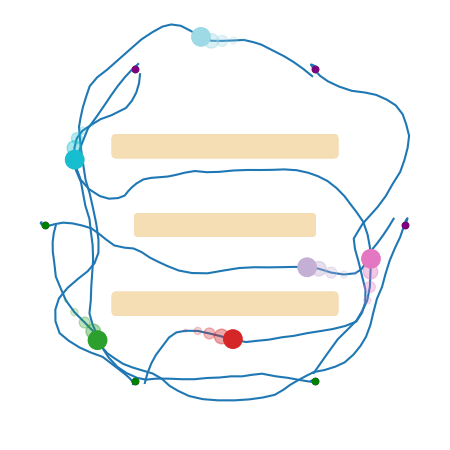}
        } &
        \subcaptionbox*{}[0.15\textwidth]{%
            \includegraphics[width=\linewidth]{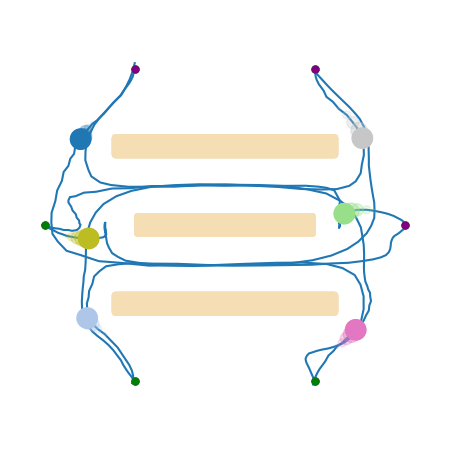}
        } &
        \subcaptionbox*{}[0.15\textwidth]{%
            \includegraphics[width=\linewidth]{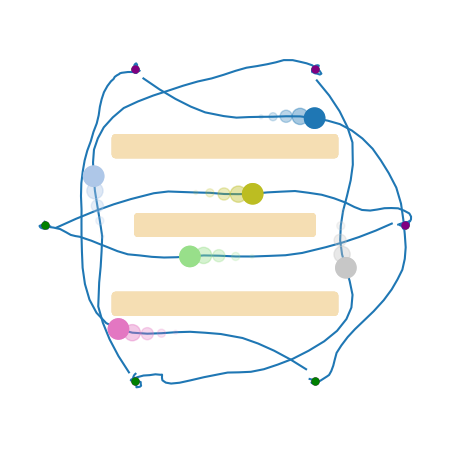}
        } \\

        \rotatebox{90}{\textbf{9 Robots}} &
        \subcaptionbox*{}[0.15\textwidth]{%
            \includegraphics[width=\linewidth]{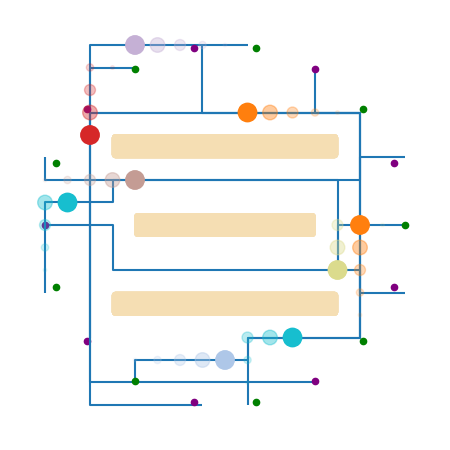}
        } &
        \subcaptionbox*{}[0.15\textwidth]{%
            \centering
            \includegraphics[width=\linewidth]{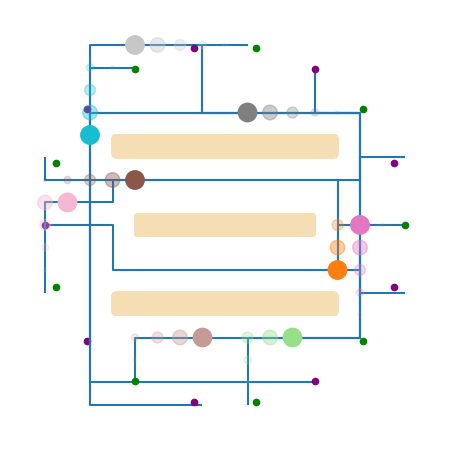}
        } &
        \subcaptionbox*{}[0.15\textwidth]{%
            \includegraphics[width=\linewidth]{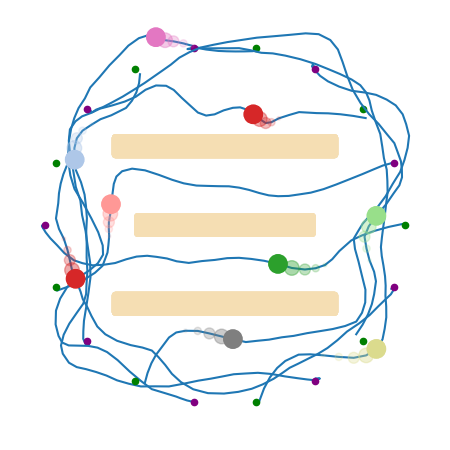}
        } &
        \subcaptionbox*{}[0.15\textwidth]{%
            \includegraphics[width=\linewidth]{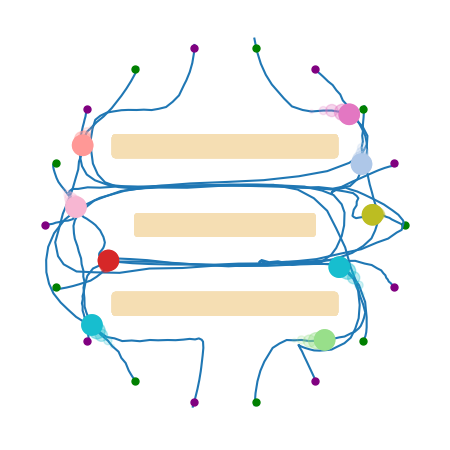}
        } &
        \subcaptionbox*{}[0.15\textwidth]{%
            \includegraphics[width=\linewidth]{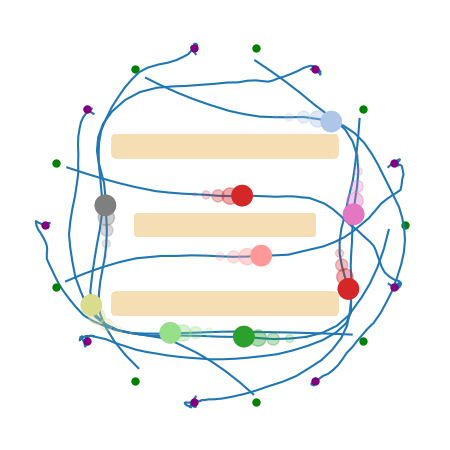}
        } \\

        \rotatebox{90}{\textbf{12 Robots}} &
        \subcaptionbox*{}[0.15\textwidth]{%
            \includegraphics[width=\linewidth]{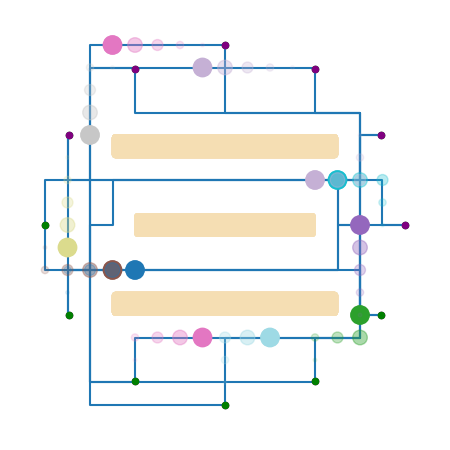}
        } &
        \subcaptionbox*{}[0.15\textwidth]{%
            \centering
            \includegraphics[width=\linewidth]{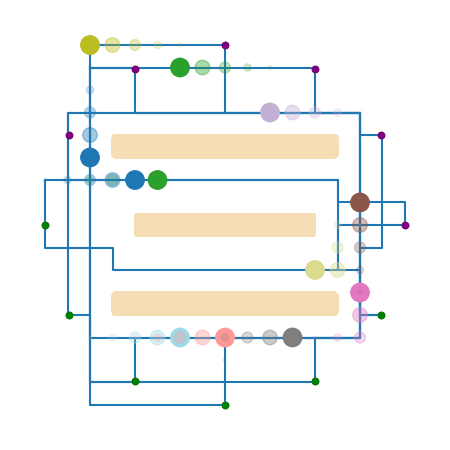}
        } &
        \subcaptionbox*{}[0.15\textwidth]{%
            \includegraphics[width=\linewidth]{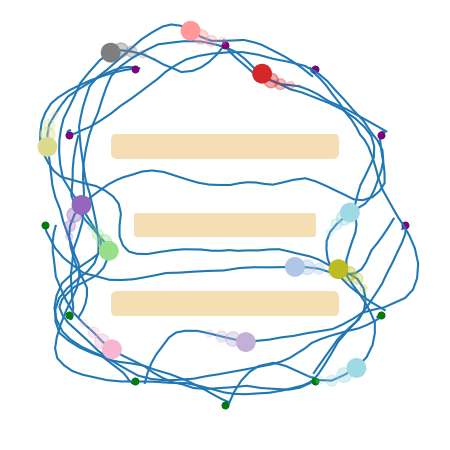}
        } &
        \subcaptionbox*{}[0.15\textwidth]{%
            N/A
        } &
        \subcaptionbox*{}[0.15\textwidth]{%
            \includegraphics[width=\linewidth]{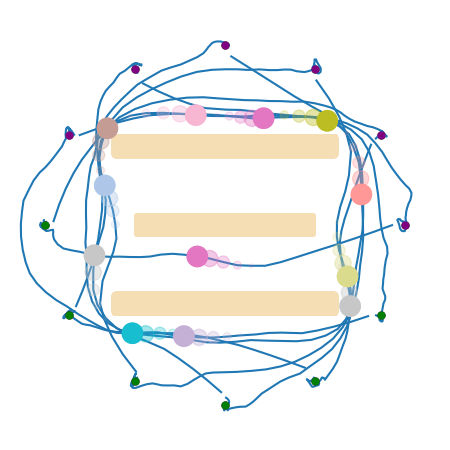}
        } \\

        \rotatebox{90}{\textbf{15 Robots}} &
        \subcaptionbox*{}[0.15\textwidth]{%
            N/A
        } &
        \subcaptionbox*{}[0.15\textwidth]{%
            \centering
            \includegraphics[width=\linewidth]{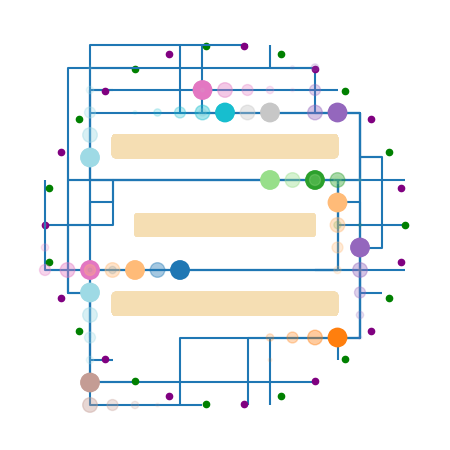}
        } &
        \subcaptionbox*{}[0.15\textwidth]{%
            N/A
        } &
        \subcaptionbox*{}[0.15\textwidth]{%
            N/A
        } &
        \subcaptionbox*{}[0.15\textwidth]{%
            \includegraphics[width=\linewidth]{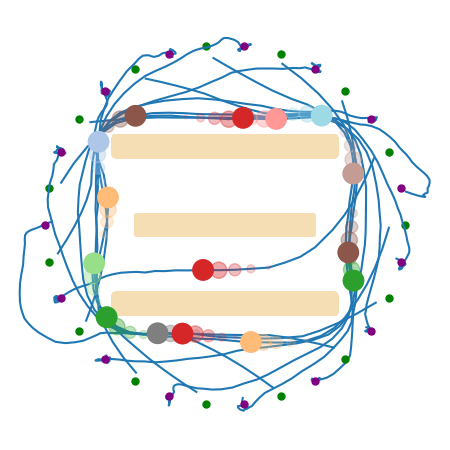}
        } \\

        \rotatebox{90}{\textbf{18 Robots}} &
        \subcaptionbox*{}[0.15\textwidth]{%
            N/A
        } &
        \subcaptionbox*{}[0.15\textwidth]{%
            \centering
            \includegraphics[width=\linewidth]{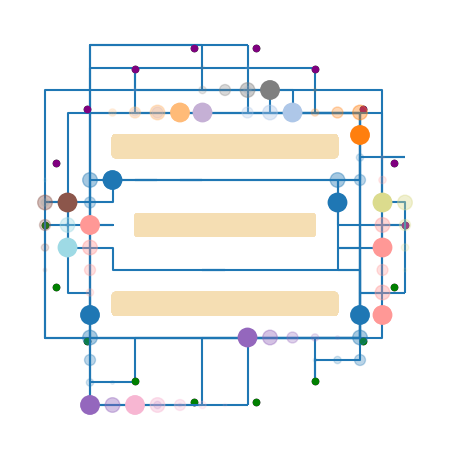}
        } &
        \subcaptionbox*{}[0.15\textwidth]{%
            N/A
        } &
        \subcaptionbox*{}[0.15\textwidth]{%
            N/A
        } &
        \subcaptionbox*{}[0.15\textwidth]{%
            \includegraphics[width=\linewidth]{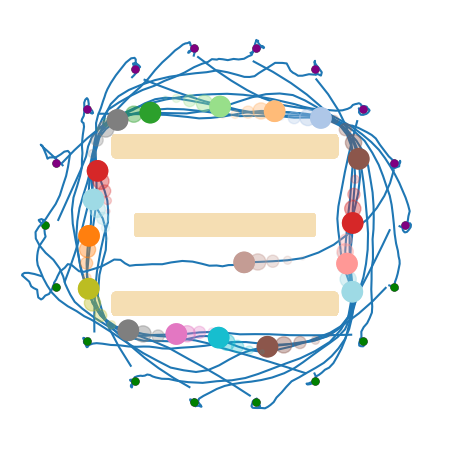}
        } \\

        \rotatebox{90}{\textbf{20 Robots}} &
        \subcaptionbox*{}[0.15\textwidth]{%
            N/A
        } &
        \subcaptionbox*{}[0.15\textwidth]{%
            N/A
        } &
        \subcaptionbox*{}[0.15\textwidth]{%
            N/A
        } &
        \subcaptionbox*{}[0.15\textwidth]{%
            N/A
        } &
        \subcaptionbox*{}[0.15\textwidth]{%
            \includegraphics[width=\linewidth]{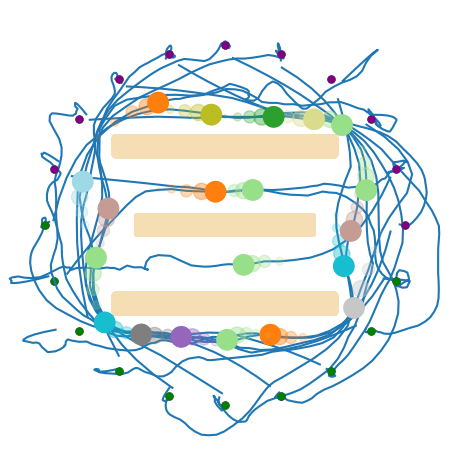}
        } \\
        
    \end{tabular}
    \caption{Some visual examples of trajectories generated by A*-CBS, A*-ECBS, KCBS, MMD-CBS, and MDOC-CBS in the Conveyor map with the Circle setup. We choose robot number \{3, 6, 9, 12, 15, 18, 20\}. N/A denotes the failure among all trials. Trajectories of KCBS, MMD-CBS, and MDOC-CBS are displayed after post-smoothing for better visualization.}
    \label{fig:conveyorcircle_trajs}
\end{figure*}

\begin{figure*}[ht]
    \centering
    \setlength{\tabcolsep}{1em} 
    \renewcommand{\arraystretch}{0.5} 
    
    \begin{tabular}{p{0.02\textwidth}ccccc}
        & \textbf{A*-CBS} & \textbf{A*ECBS} & \textbf{KCBS} & \textbf{MMD-CBS} & \textbf{MDOC-CBS} \\
        \rotatebox{90}{\textbf{3 Robots}} &
        \subcaptionbox*{}[0.15\textwidth]{%
            \includegraphics[width=\linewidth]{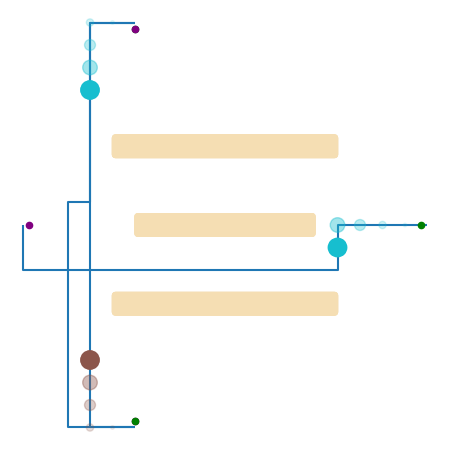}
        } &
        \subcaptionbox*{}[0.15\textwidth]{%
            \centering
            \includegraphics[width=\linewidth]{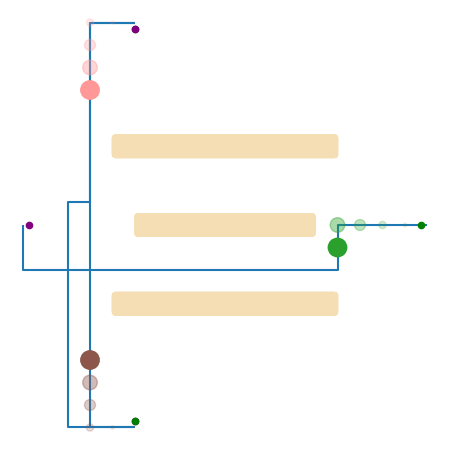}
        } &
        \subcaptionbox*{}[0.15\textwidth]{%
            \includegraphics[width=\linewidth]{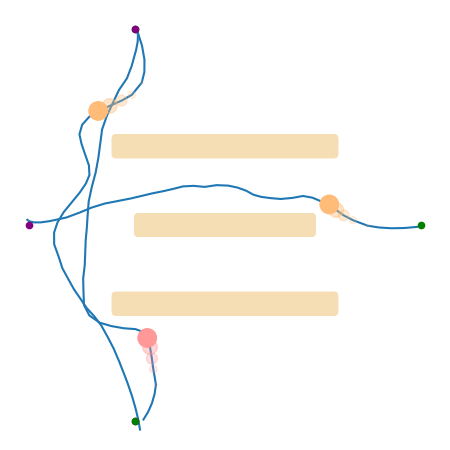}
        } &
        \subcaptionbox*{}[0.15\textwidth]{%
            \includegraphics[width=\linewidth]{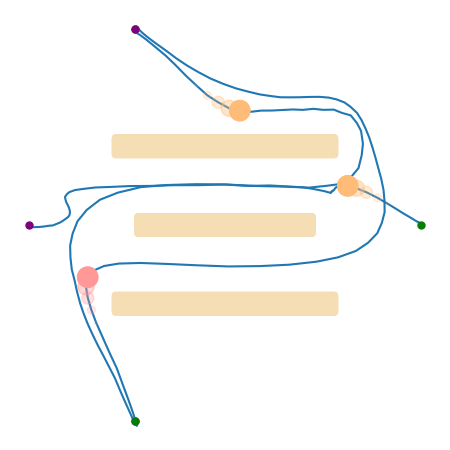}
        } &
        \subcaptionbox*{}[0.15\textwidth]{%
            \includegraphics[width=\linewidth]{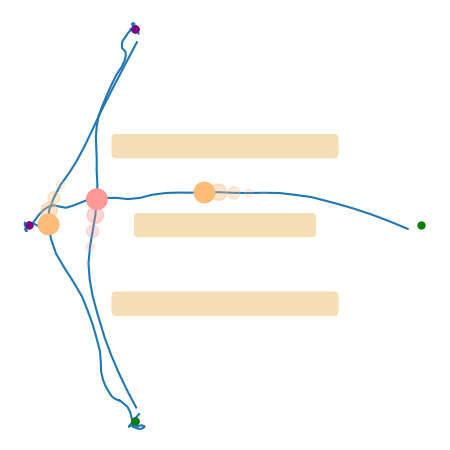}
        } \\

        \rotatebox{90}{\textbf{6 Robots}} &
        \subcaptionbox*{}[0.15\textwidth]{%
            \includegraphics[width=\linewidth]{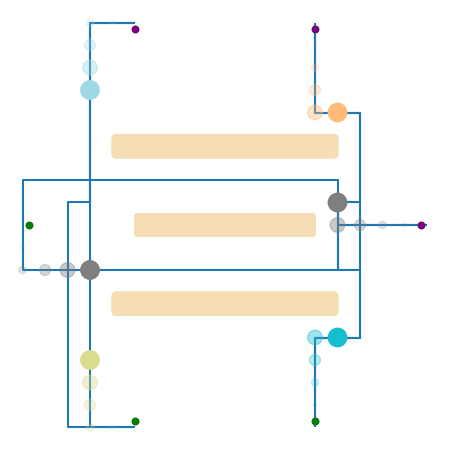}
        } &
        \subcaptionbox*{}[0.15\textwidth]{%
            \centering
            \includegraphics[width=\linewidth]{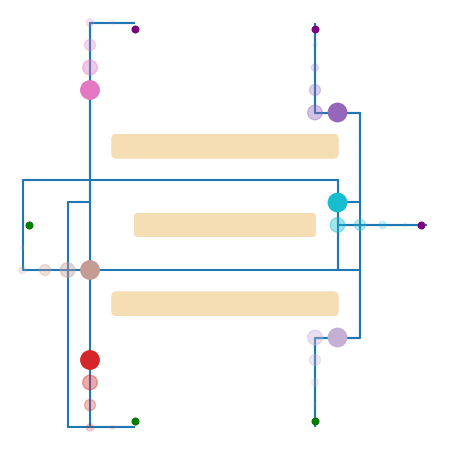}
        } &
        \subcaptionbox*{}[0.15\textwidth]{%
            \includegraphics[width=\linewidth]{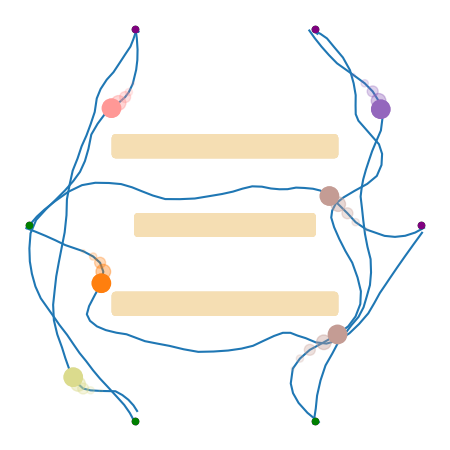}
        } &
        \subcaptionbox*{}[0.15\textwidth]{%
            \includegraphics[width=\linewidth]{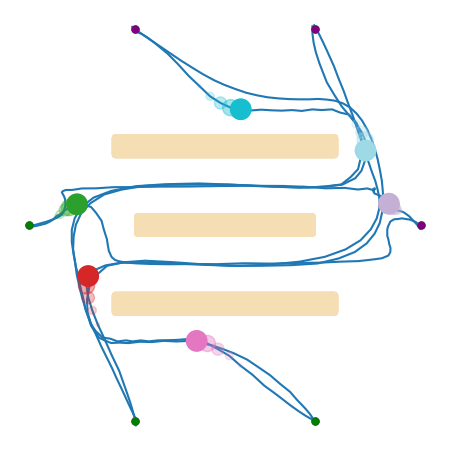}
        } &
        \subcaptionbox*{}[0.15\textwidth]{%
            \includegraphics[width=\linewidth]{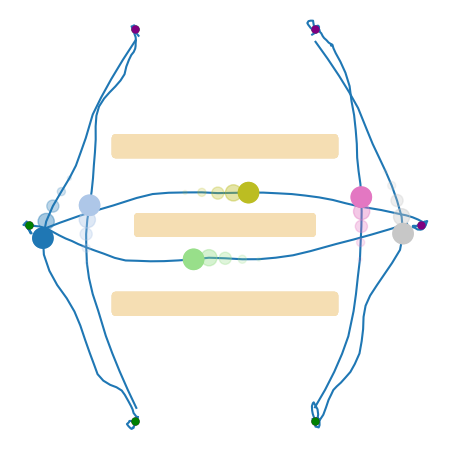}
        } \\

        \rotatebox{90}{\textbf{9 Robots}} &
        \subcaptionbox*{}[0.15\textwidth]{%
            \includegraphics[width=\linewidth]{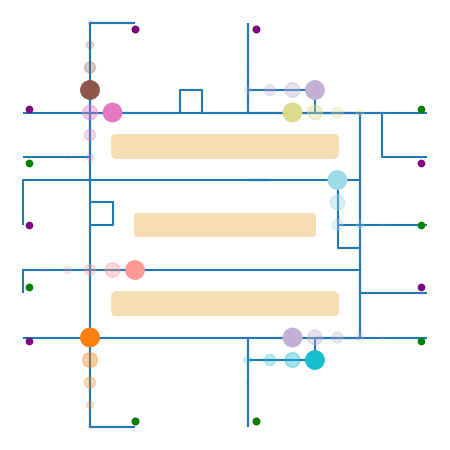}
        } &
        \subcaptionbox*{}[0.15\textwidth]{%
            \centering
            \includegraphics[width=\linewidth]{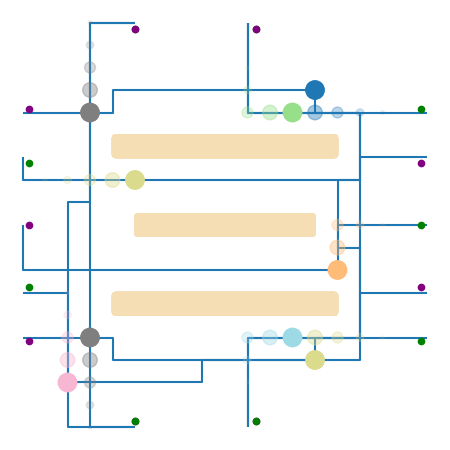}
        } &
        \subcaptionbox*{}[0.15\textwidth]{%
            \includegraphics[width=\linewidth]{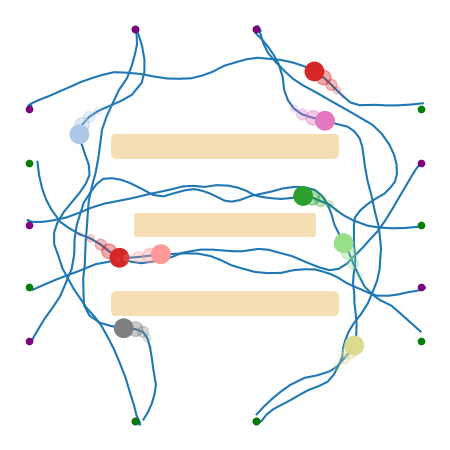}
        } &
        \subcaptionbox*{}[0.15\textwidth]{%
            \includegraphics[width=\linewidth]{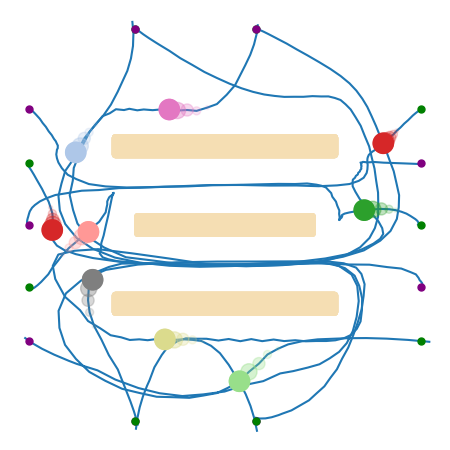}
        } &
        \subcaptionbox*{}[0.15\textwidth]{%
            \includegraphics[width=\linewidth]{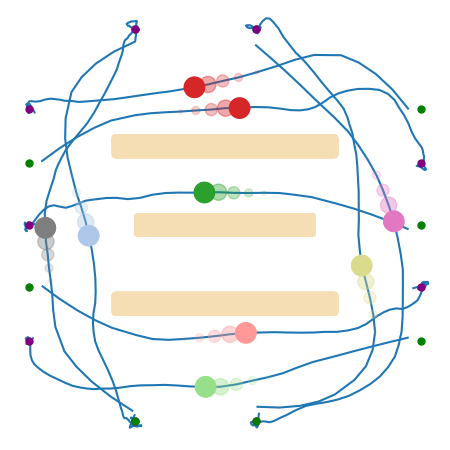}
        } \\

        \rotatebox{90}{\textbf{12 Robots}} &
        \subcaptionbox*{}[0.15\textwidth]{%
            \includegraphics[width=\linewidth]{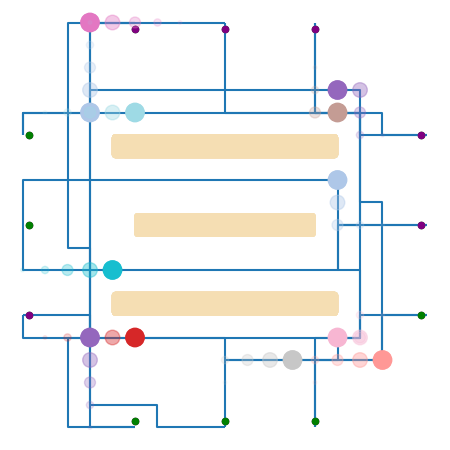}
        } &
        \subcaptionbox*{}[0.15\textwidth]{%
            \centering
            \includegraphics[width=\linewidth]{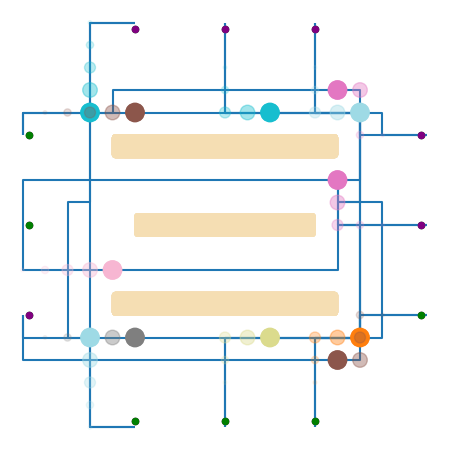}
        } &
        \subcaptionbox*{}[0.15\textwidth]{%
            \includegraphics[width=\linewidth]{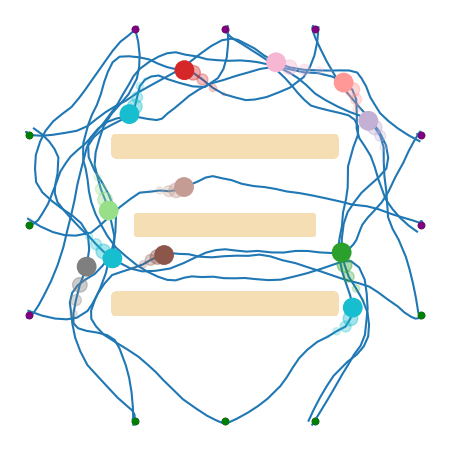}
        } &
        \subcaptionbox*{}[0.15\textwidth]{%
            \includegraphics[width=\linewidth]{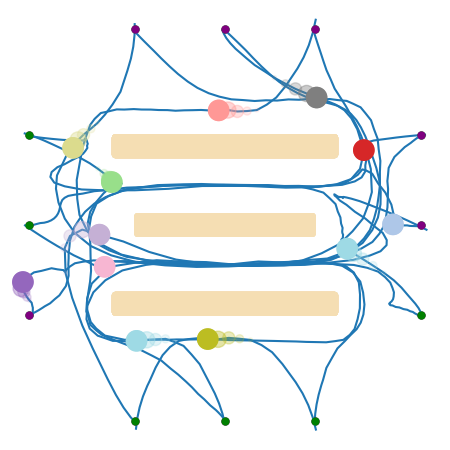}
        } &
        \subcaptionbox*{}[0.15\textwidth]{%
            \includegraphics[width=\linewidth]{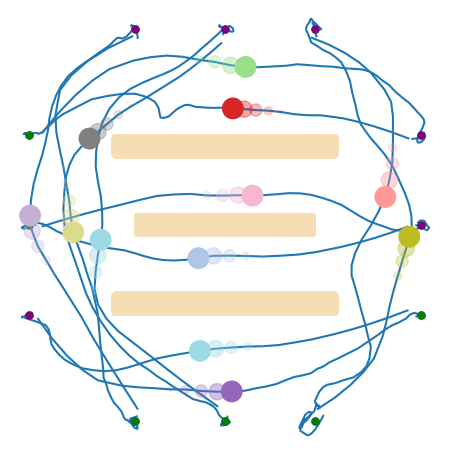}
        } \\

        \rotatebox{90}{\textbf{15 Robots}} &
        \subcaptionbox*{}[0.15\textwidth]{%
            N/A
        } &
        \subcaptionbox*{}[0.15\textwidth]{%
            \centering
            \includegraphics[width=\linewidth]{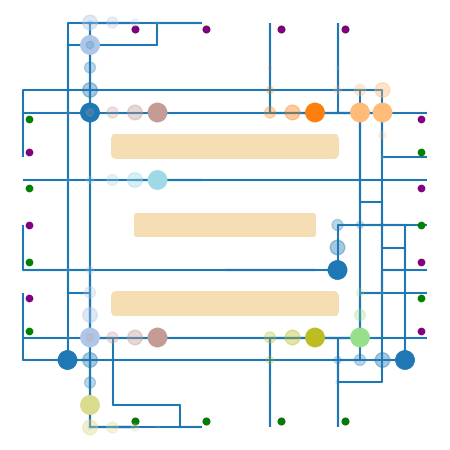}
        } &
        \subcaptionbox*{}[0.15\textwidth]{%
            N/A
        } &
        \subcaptionbox*{}[0.15\textwidth]{%
            N/A
        } &
        \subcaptionbox*{}[0.15\textwidth]{%
            \includegraphics[width=\linewidth]{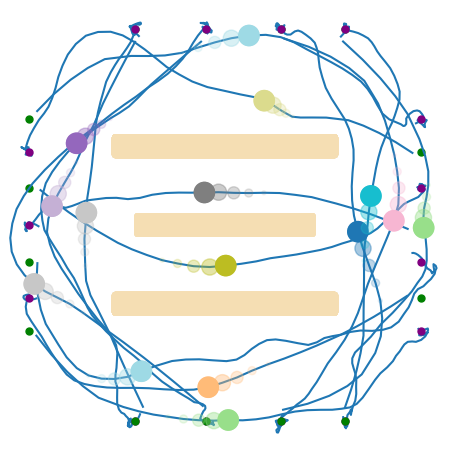}
        } \\

        \rotatebox{90}{\textbf{18 Robots}} &
        \subcaptionbox*{}[0.15\textwidth]{%
            N/A
        } &
        \subcaptionbox*{}[0.15\textwidth]{%
            \centering
            \includegraphics[width=\linewidth]{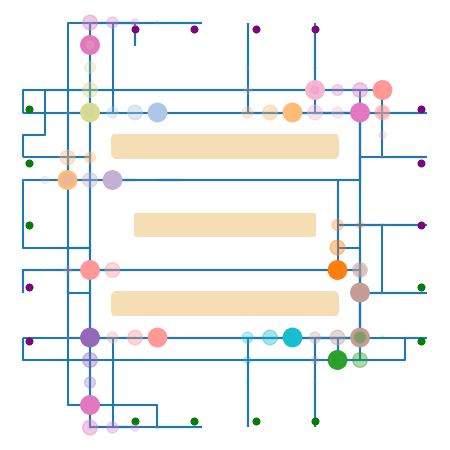}
        } &
        \subcaptionbox*{}[0.15\textwidth]{%
            N/A
        } &
        \subcaptionbox*{}[0.15\textwidth]{%
            N/A
        } &
        \subcaptionbox*{}[0.15\textwidth]{%
            N/A
        } \\

        \rotatebox{90}{\textbf{20 Robots}} &
        \subcaptionbox*{}[0.15\textwidth]{%
            N/A
        } &
        \subcaptionbox*{}[0.15\textwidth]{%
            N/A
        } &
        \subcaptionbox*{}[0.15\textwidth]{%
            N/A
        } &
        \subcaptionbox*{}[0.15\textwidth]{%
            N/A
        } &
        \subcaptionbox*{}[0.15\textwidth]{%
            N/A
        } \\
        
    \end{tabular}
    \caption{Some visual examples of trajectories generated by A*-CBS, A*-ECBS, KCBS, MMD-CBS, and MDOC-CBS in the Conveyor map with the Weave setup. We choose robot number \{3, 6, 9, 12, 15, 18, 20\}. N/A denotes the failure among all trials. Trajectories of KCBS, MMD-CBS, and MDOC-CBS are displayed after post-smoothing for better visualization.}
    \label{fig:conveyorweave_trajs}
\end{figure*}

\begin{figure*}[ht]
    \centering
    \setlength{\tabcolsep}{1em} 
    \renewcommand{\arraystretch}{0.5} 
    
    \begin{tabular}{p{0.02\textwidth}ccccc}
        & \textbf{A*-CBS} & \textbf{A*ECBS} & \textbf{KCBS} & \textbf{MMD-CBS} & \textbf{MDOC-CBS} \\
        \rotatebox{90}{\textbf{3 Robots}} &
        \subcaptionbox*{}[0.15\textwidth]{%
            \includegraphics[width=\linewidth]{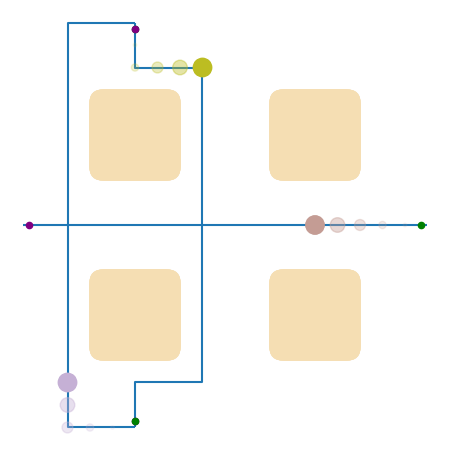}
        } &
        \subcaptionbox*{}[0.15\textwidth]{%
            \centering
            \includegraphics[width=\linewidth]{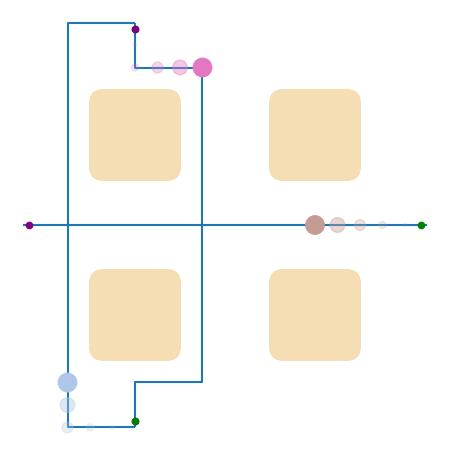}
        } &
        \subcaptionbox*{}[0.15\textwidth]{%
            \includegraphics[width=\linewidth]{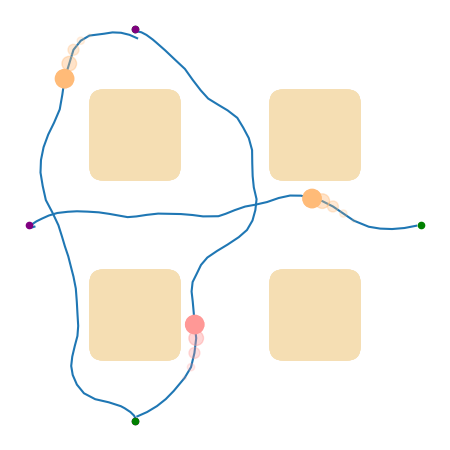}
        } &
        \subcaptionbox*{}[0.15\textwidth]{%
            \includegraphics[width=\linewidth]{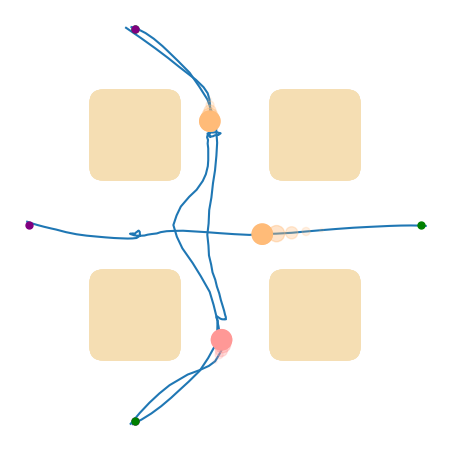}
        } &
        \subcaptionbox*{}[0.15\textwidth]{%
            \includegraphics[width=\linewidth]{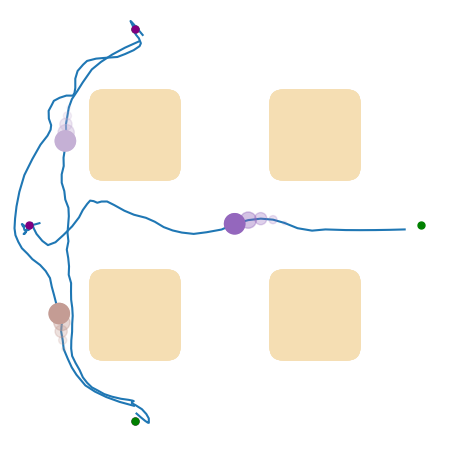}
        } \\

        \rotatebox{90}{\textbf{6 Robots}} &
        \subcaptionbox*{}[0.15\textwidth]{%
            \includegraphics[width=\linewidth]{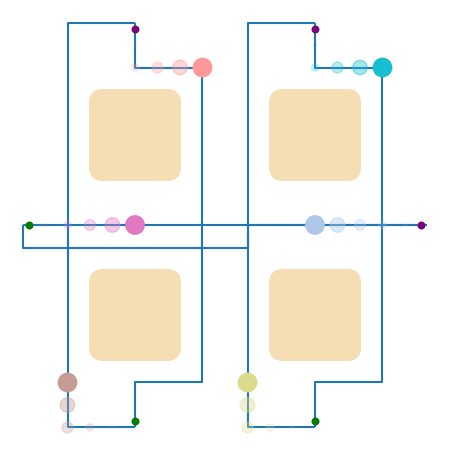}
        } &
        \subcaptionbox*{}[0.15\textwidth]{%
            \centering
            \includegraphics[width=\linewidth]{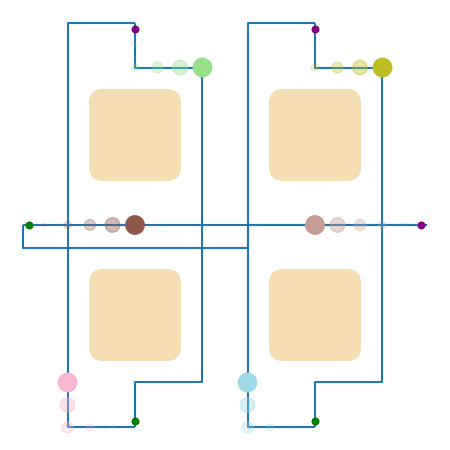}
        } &
        \subcaptionbox*{}[0.15\textwidth]{%
            \includegraphics[width=\linewidth]{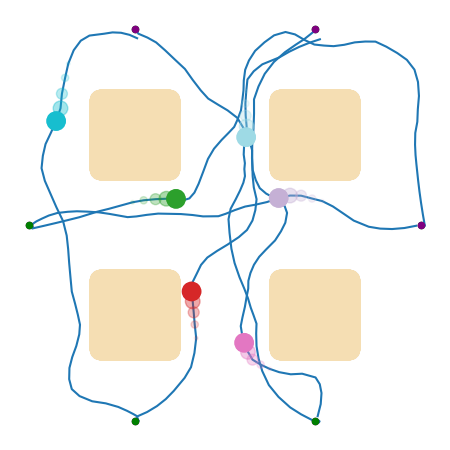}
        } &
        \subcaptionbox*{}[0.15\textwidth]{%
            \includegraphics[width=\linewidth]{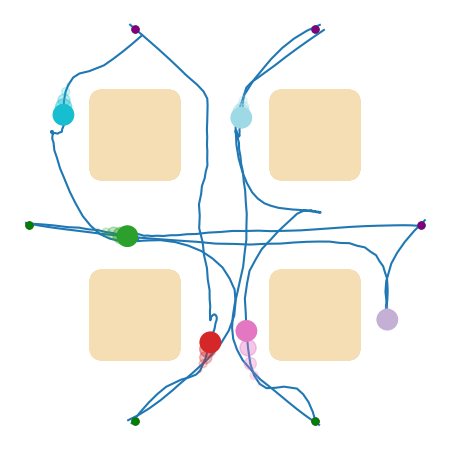}
        } &
        \subcaptionbox*{}[0.15\textwidth]{%
            \includegraphics[width=\linewidth]{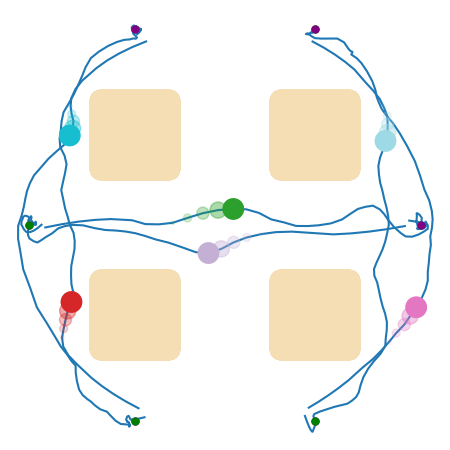}
        } \\

        \rotatebox{90}{\textbf{9 Robots}} &
        \subcaptionbox*{}[0.15\textwidth]{%
            \includegraphics[width=\linewidth]{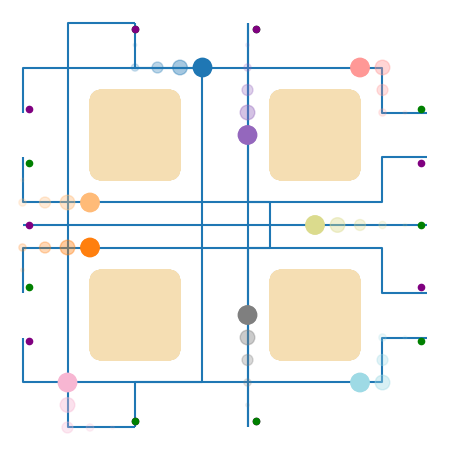}
        } &
        \subcaptionbox*{}[0.15\textwidth]{%
            \centering
            \includegraphics[width=\linewidth]{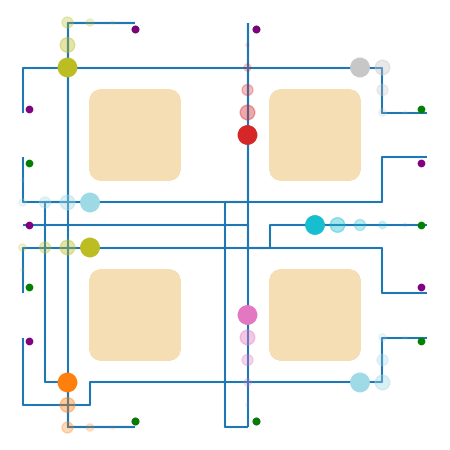}
        } &
        \subcaptionbox*{}[0.15\textwidth]{%
            \includegraphics[width=\linewidth]{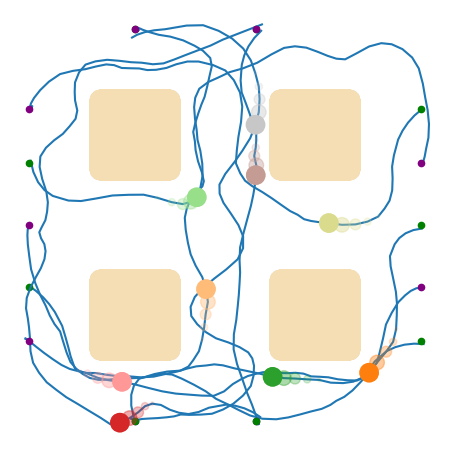}
        } &
        \subcaptionbox*{}[0.15\textwidth]{%
            \includegraphics[width=\linewidth]{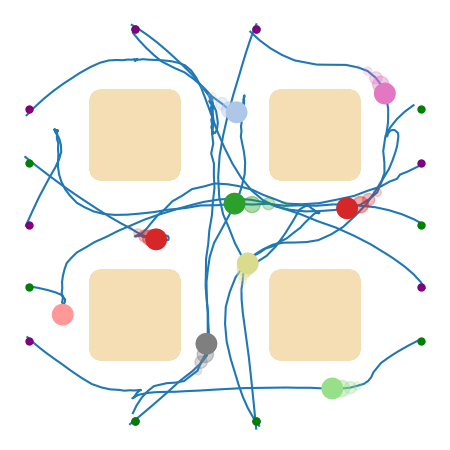}
        } &
        \subcaptionbox*{}[0.15\textwidth]{%
            \includegraphics[width=\linewidth]{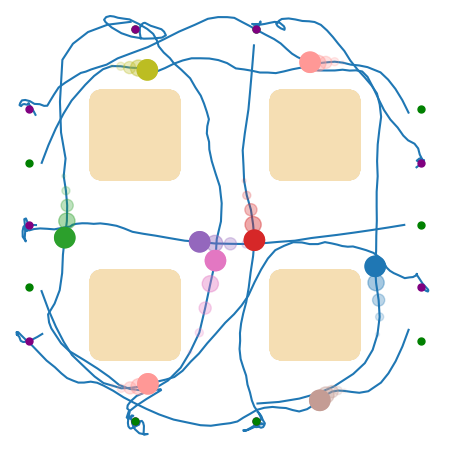}
        } \\

        \rotatebox{90}{\textbf{12 Robots}} &
        \subcaptionbox*{}[0.15\textwidth]{%
            \includegraphics[width=\linewidth]{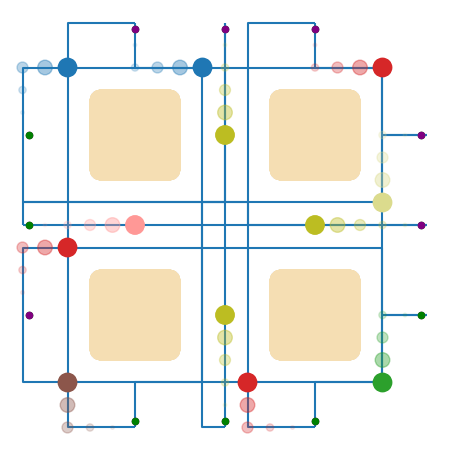}
        } &
        \subcaptionbox*{}[0.15\textwidth]{%
            \centering
            \includegraphics[width=\linewidth]{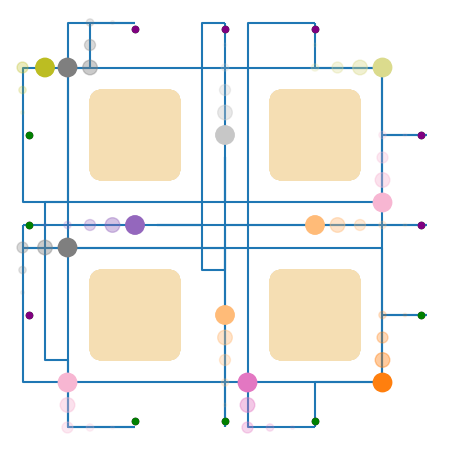}
        } &
        \subcaptionbox*{}[0.15\textwidth]{%
            \includegraphics[width=\linewidth]{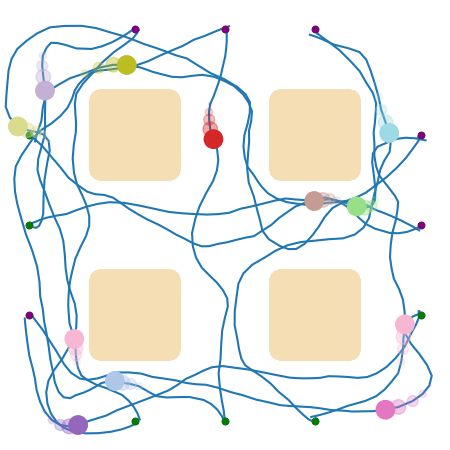}
        } &
        \subcaptionbox*{}[0.15\textwidth]{%
            \includegraphics[width=\linewidth]{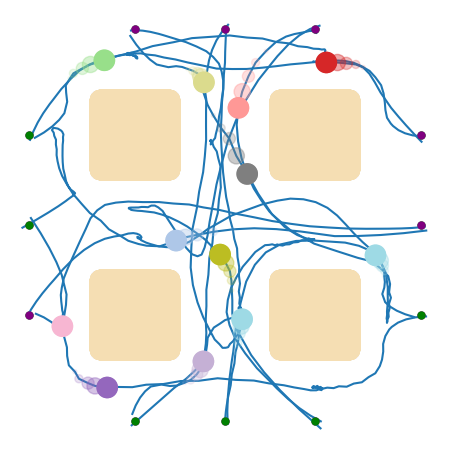}
        } &
        \subcaptionbox*{}[0.15\textwidth]{%
            \includegraphics[width=\linewidth]{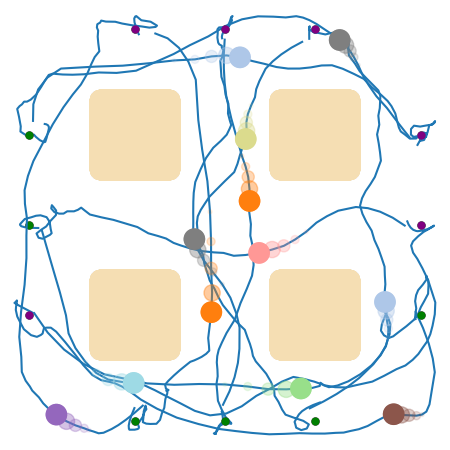}
        } \\

        \rotatebox{90}{\textbf{15 Robots}} &
        \subcaptionbox*{}[0.15\textwidth]{%
            N/A
        } &
        \subcaptionbox*{}[0.15\textwidth]{%
            \centering
            \includegraphics[width=\linewidth]{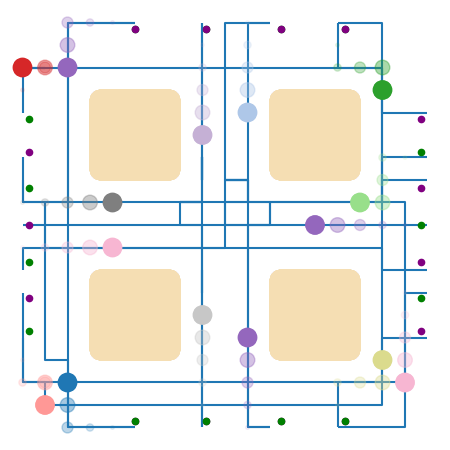}
        } &
        \subcaptionbox*{}[0.15\textwidth]{%
            \includegraphics[width=\linewidth]{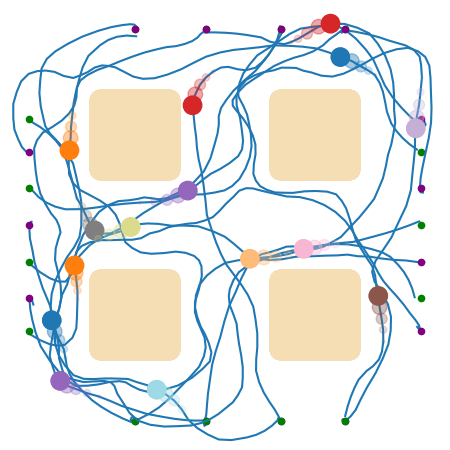}
        } &
        \subcaptionbox*{}[0.15\textwidth]{%
            N/A
        } &
        \subcaptionbox*{}[0.15\textwidth]{%
            \includegraphics[width=\linewidth]{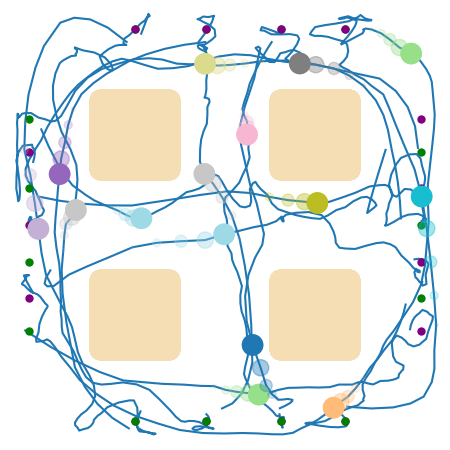}
        } \\
        
    \end{tabular}
    \caption{Some visual examples of trajectories generated by A*-CBS, A*-ECBS, KCBS, MMD-CBS, and MDOC-CBS in the Drop-Region map with the Weave setup. We choose robot number \{3, 6, 9, 12, 15\}. N/A denotes the failure among all trials. Trajectories of KCBS, MMD-CBS, and MDOC-CBS are displayed after post-smoothing for better visualization.}
    \label{fig:dropregion_trajs}
\end{figure*}

\begin{figure*}[ht]
    \centering
    \setlength{\tabcolsep}{1em} 
    \renewcommand{\arraystretch}{0.5} 
    
    \begin{tabular}{p{0.02\textwidth}ccccc}
        & \textbf{A*-CBS} & \textbf{A*ECBS} & \textbf{KCBS} & \textbf{MMD-CBS} & \textbf{MDOC-CBS} \\
        \rotatebox{90}{\textbf{3 Robots}} &
        \subcaptionbox*{}[0.15\textwidth]{%
            \includegraphics[width=\linewidth]{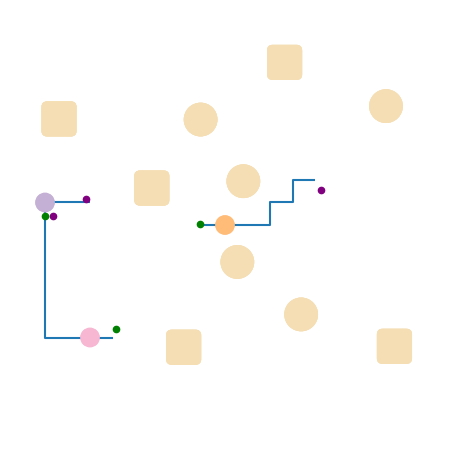}
        } &
        \subcaptionbox*{}[0.15\textwidth]{%
            \centering
            \includegraphics[width=\linewidth]{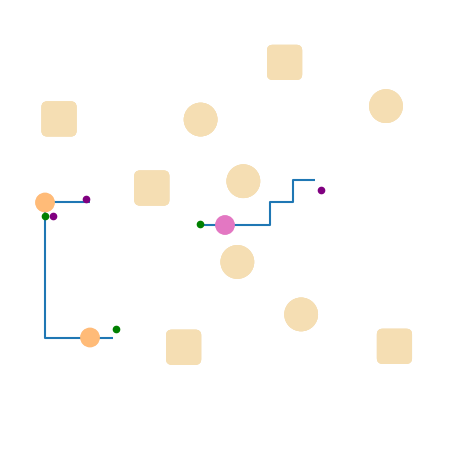}
        } &
        \subcaptionbox*{}[0.15\textwidth]{%
            \includegraphics[width=\linewidth]{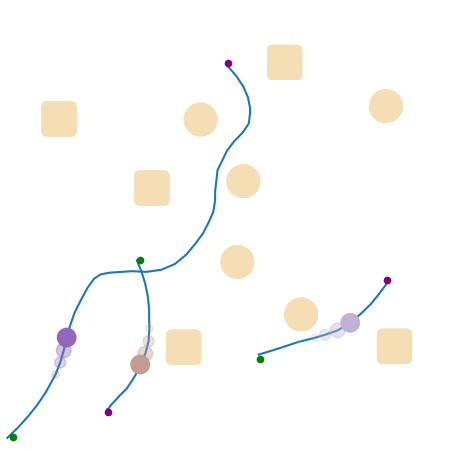}
        } &
        \subcaptionbox*{}[0.15\textwidth]{%
            \includegraphics[width=\linewidth]{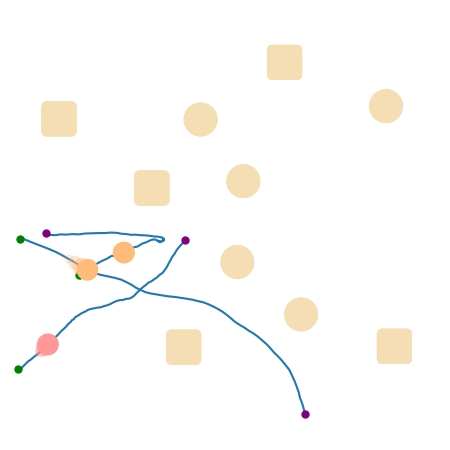}
        } &
        \subcaptionbox*{}[0.15\textwidth]{%
            \includegraphics[width=\linewidth]{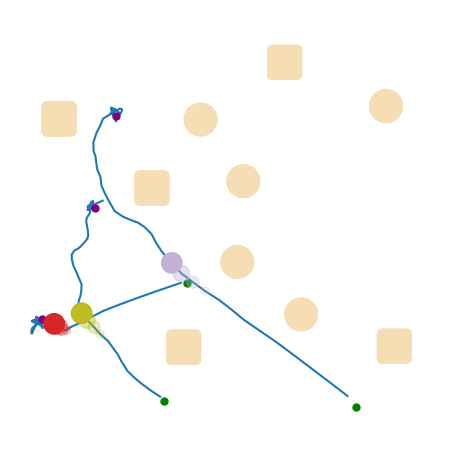}
        } \\

        \rotatebox{90}{\textbf{6 Robots}} &
        \subcaptionbox*{}[0.15\textwidth]{%
            \includegraphics[width=\linewidth]{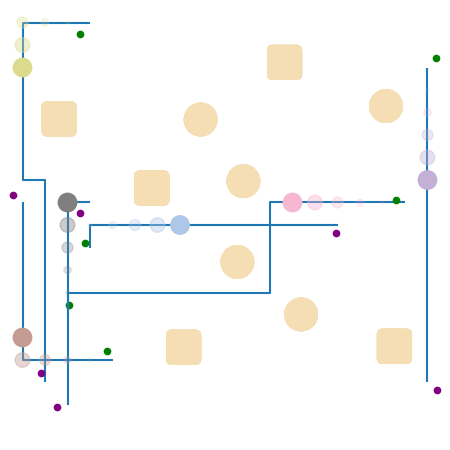}
        } &
        \subcaptionbox*{}[0.15\textwidth]{%
            \centering
            \includegraphics[width=\linewidth]{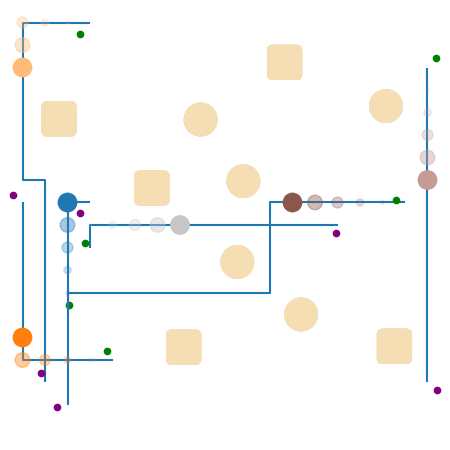}
        } &
        \subcaptionbox*{}[0.15\textwidth]{%
            \includegraphics[width=\linewidth]{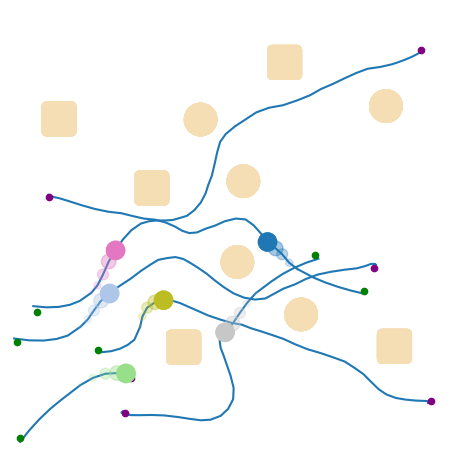}
        } &
        \subcaptionbox*{}[0.15\textwidth]{%
            \includegraphics[width=\linewidth]{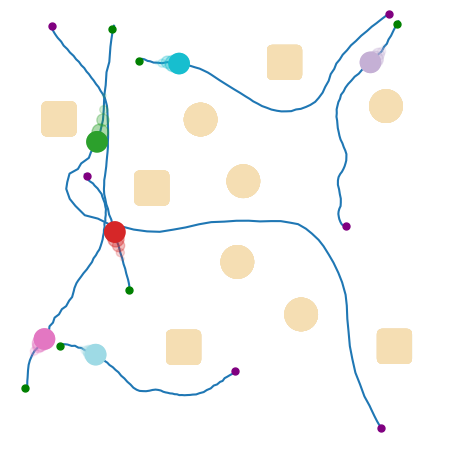}
        } &
        \subcaptionbox*{}[0.15\textwidth]{%
            \includegraphics[width=\linewidth]{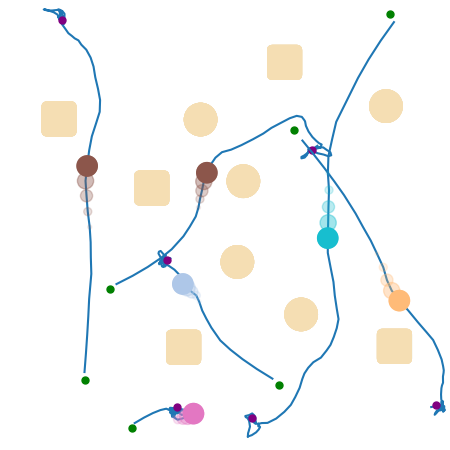}
        } \\

        \rotatebox{90}{\textbf{9 Robots}} &
        \subcaptionbox*{}[0.15\textwidth]{%
            \includegraphics[width=\linewidth]{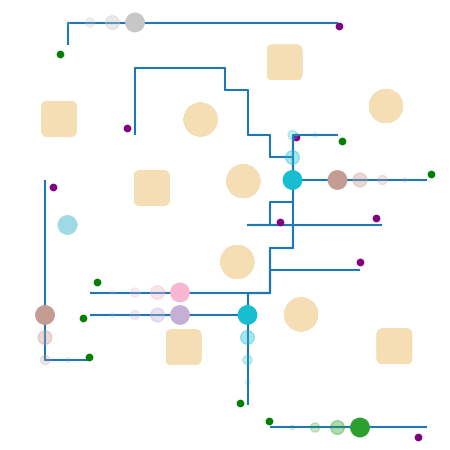}
        } &
        \subcaptionbox*{}[0.15\textwidth]{%
            \centering
            \includegraphics[width=\linewidth]{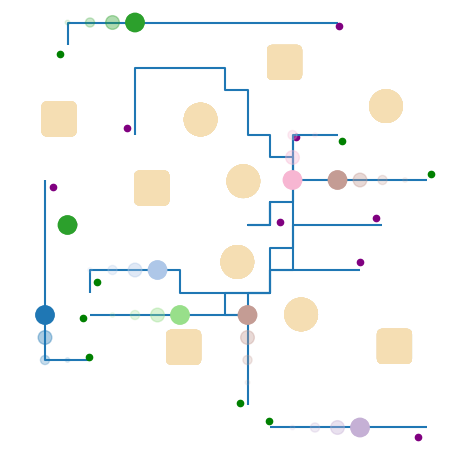}
        } &
        \subcaptionbox*{}[0.15\textwidth]{%
            \includegraphics[width=\linewidth]{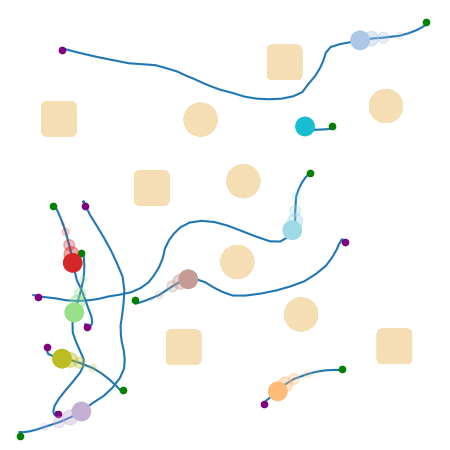}
        } &
        \subcaptionbox*{}[0.15\textwidth]{%
            \includegraphics[width=\linewidth]{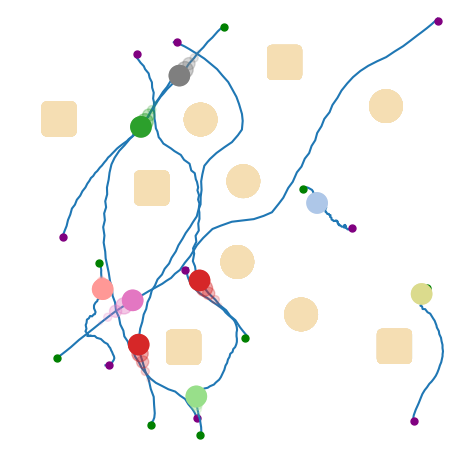}
        } &
        \subcaptionbox*{}[0.15\textwidth]{%
            \includegraphics[width=\linewidth]{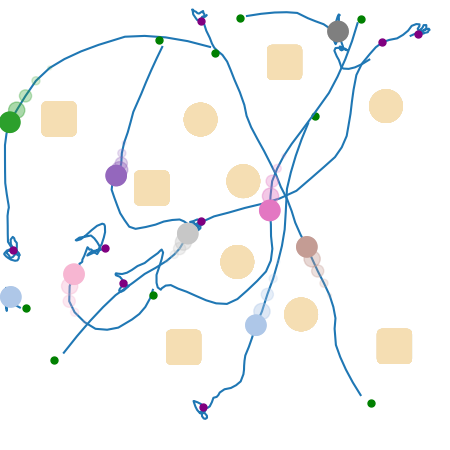}
        } \\

        \rotatebox{90}{\textbf{12 Robots}} &
        \subcaptionbox*{}[0.15\textwidth]{%
            \includegraphics[width=\linewidth]{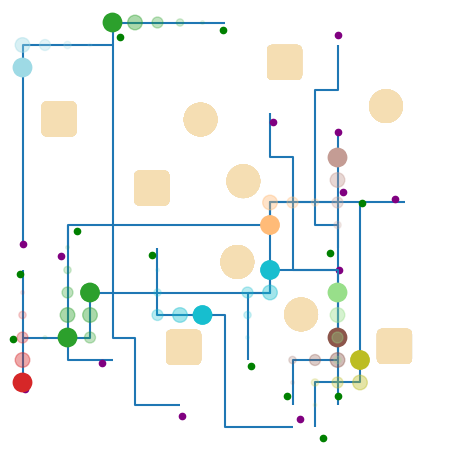}
        } &
        \subcaptionbox*{}[0.15\textwidth]{%
            \centering
            \includegraphics[width=\linewidth]{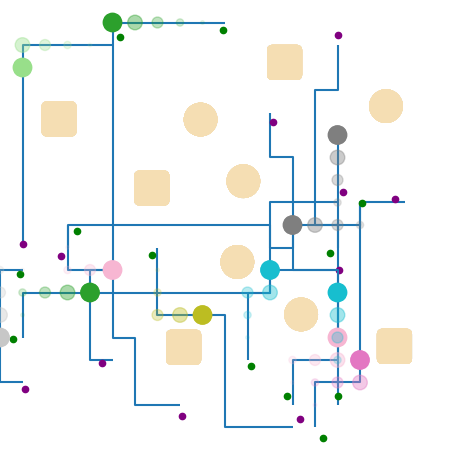}
        } &
        \subcaptionbox*{}[0.15\textwidth]{%
            \includegraphics[width=\linewidth]{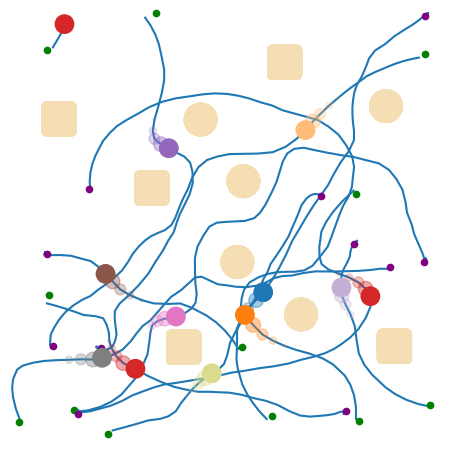}
        } &
        \subcaptionbox*{}[0.15\textwidth]{%
            \includegraphics[width=\linewidth]{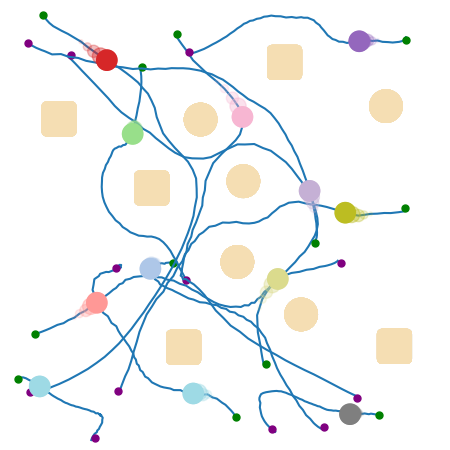}
        } &
        \subcaptionbox*{}[0.15\textwidth]{%
            \includegraphics[width=\linewidth]{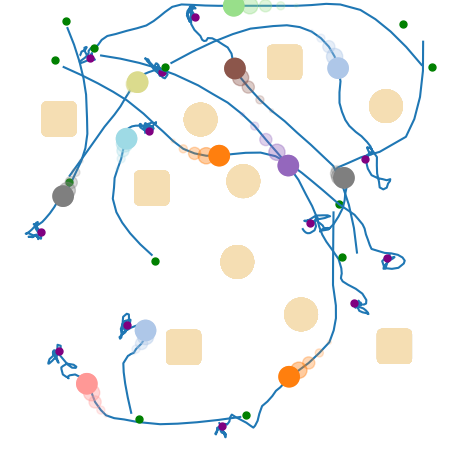}
        } \\

        \rotatebox{90}{\textbf{15 Robots}} &
        \subcaptionbox*{}[0.15\textwidth]{%
            \includegraphics[width=\linewidth]{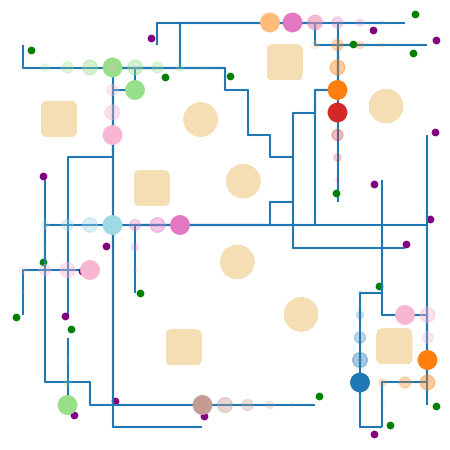}
        } &
        \subcaptionbox*{}[0.15\textwidth]{%
            \centering
            \includegraphics[width=\linewidth]{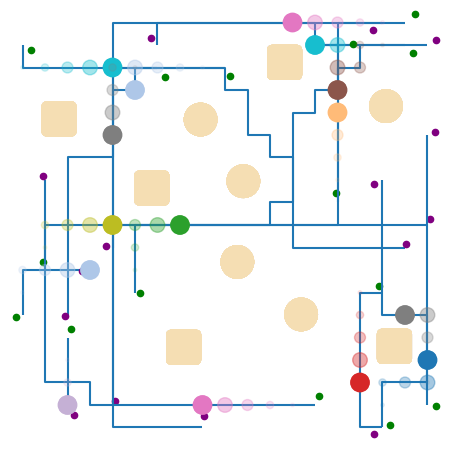}
        } &
        \subcaptionbox*{}[0.15\textwidth]{%
            N/A
        } &
        \subcaptionbox*{}[0.15\textwidth]{%
            \includegraphics[width=\linewidth]{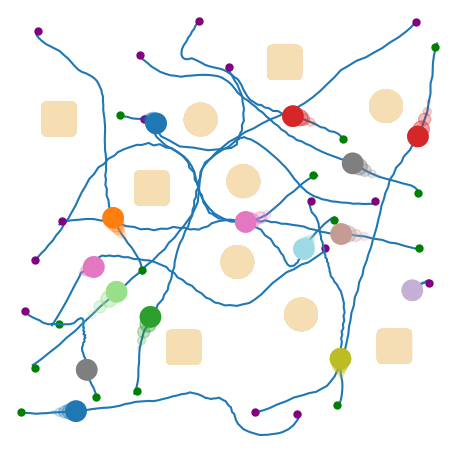}
        } &
        \subcaptionbox*{}[0.15\textwidth]{%
            \includegraphics[width=\linewidth]{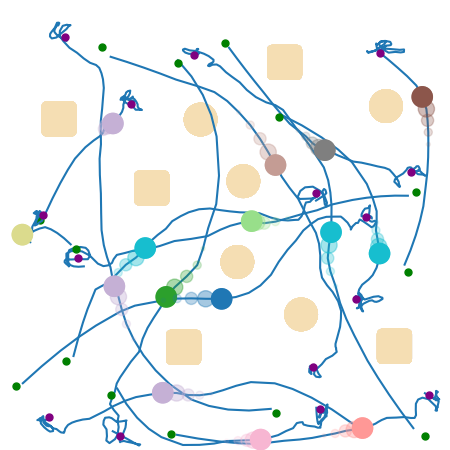}
        } \\
        
    \end{tabular}
    \caption{Some visual examples of trajectories generated by A*-CBS, A*-ECBS, KCBS, MMD-CBS, and MDOC-CBS in the Random map with the random start-goal pairs. We choose robot number \{3, 6, 9, 12, 15\}. N/A denotes the failure among all trials. Trajectories of KCBS, MMD-CBS, and MDOC-CBS are displayed after post-smoothing for better visualization.}
    \label{fig:random_trajs}
\end{figure*}

\begin{figure*}[ht]
    \centering
    \setlength{\tabcolsep}{1em} 
    \renewcommand{\arraystretch}{0.5} 
    
    \begin{tabular}{p{0.02\textwidth}ccccc}
        & \textbf{A*-CBS} & \textbf{A*ECBS} & \textbf{KCBS} & \textbf{MMD-CBS} & \textbf{MDOC-CBS} \\

        \rotatebox{90}{\textbf{12 Robots}} &
        \subcaptionbox*{}[0.15\textwidth]{%
            \includegraphics[width=\linewidth]{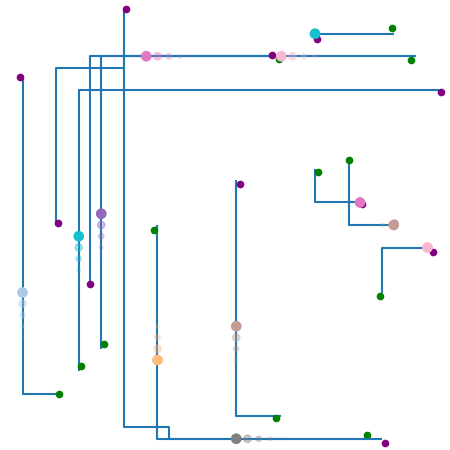}
        } &
        \subcaptionbox*{}[0.15\textwidth]{%
            \centering
            \includegraphics[width=\linewidth]{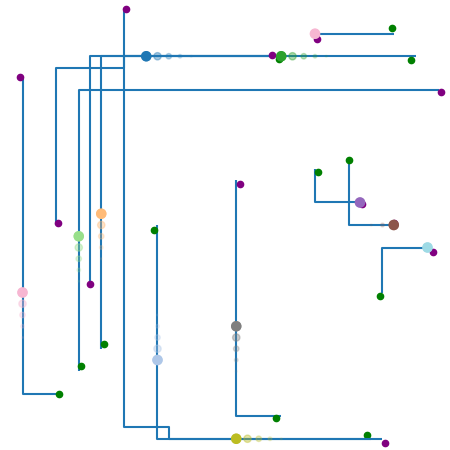}
        } &
        \subcaptionbox*{}[0.15\textwidth]{%
            \includegraphics[width=\linewidth]{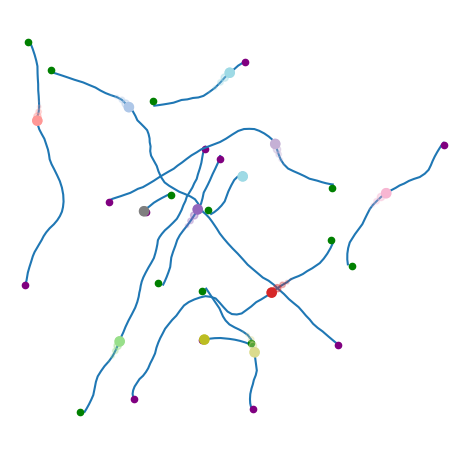}
        } &
        \subcaptionbox*{}[0.15\textwidth]{%
            \includegraphics[width=\linewidth]{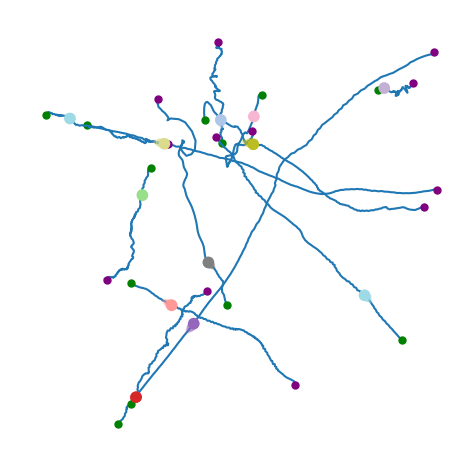}
        } &
        \subcaptionbox*{}[0.15\textwidth]{%
            \includegraphics[width=\linewidth]{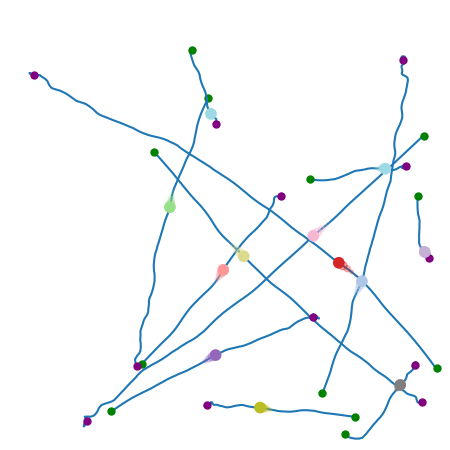}
        } \\

        \rotatebox{90}{\textbf{15 Robots}} &
        \subcaptionbox*{}[0.15\textwidth]{%
            \includegraphics[width=\linewidth]{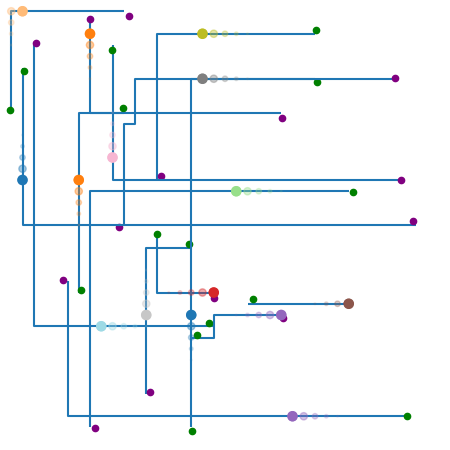}
        } &
        \subcaptionbox*{}[0.15\textwidth]{%
            \centering
            \includegraphics[width=\linewidth]{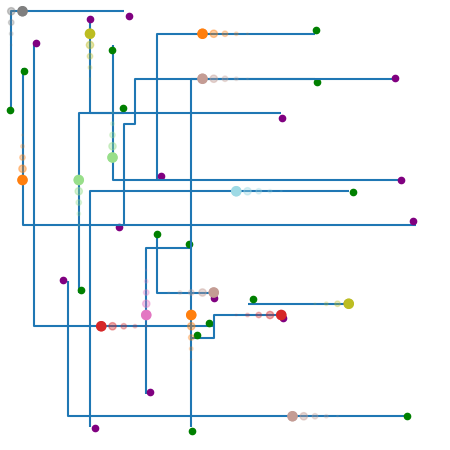}
        } &
        \subcaptionbox*{}[0.15\textwidth]{%
            \includegraphics[width=\linewidth]{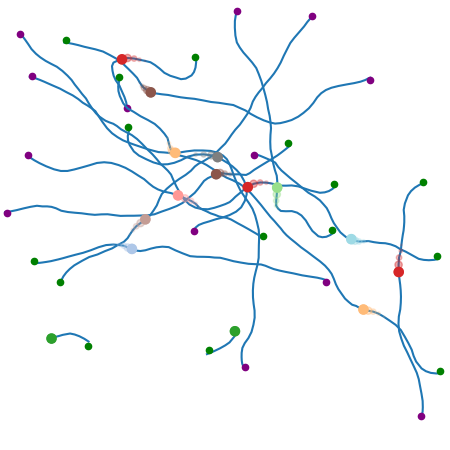}
        } &
        \subcaptionbox*{}[0.15\textwidth]{%
            \includegraphics[width=\linewidth]{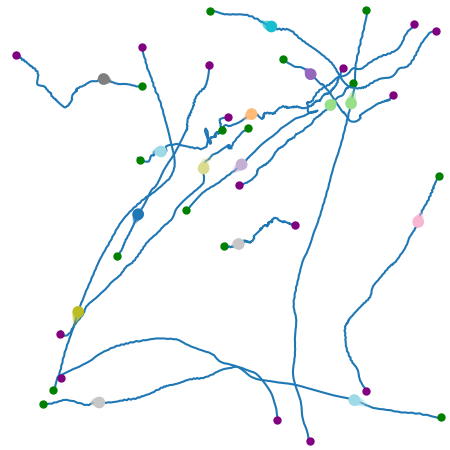}
        } &
        \subcaptionbox*{}[0.15\textwidth]{%
            \includegraphics[width=\linewidth]{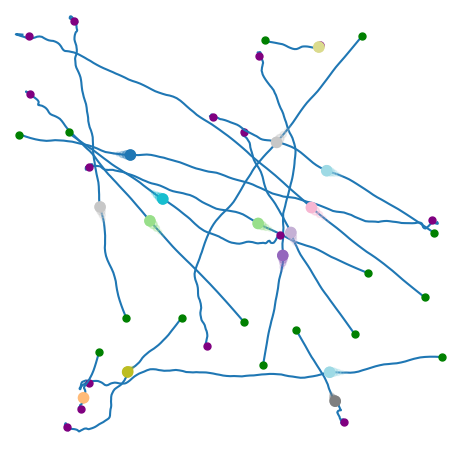}
        } \\

        \rotatebox{90}{\textbf{20 Robots}} &
        \subcaptionbox*{}[0.15\textwidth]{%
            \includegraphics[width=\linewidth]{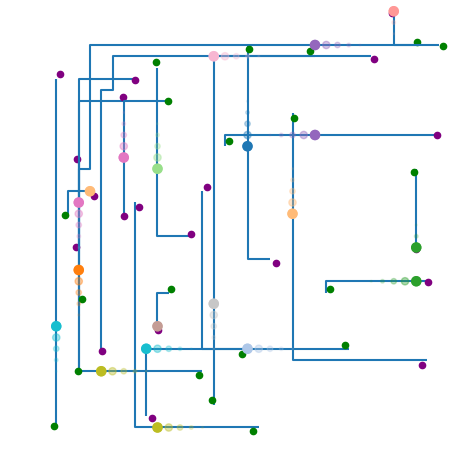}
        } &
        \subcaptionbox*{}[0.15\textwidth]{%
            \includegraphics[width=\linewidth]{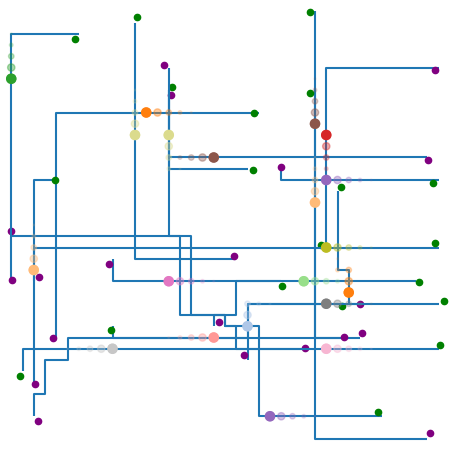}  
        } &
        \subcaptionbox*{}[0.15\textwidth]{%
            N/A
        } &
        \subcaptionbox*{}[0.15\textwidth]{%
            \includegraphics[width=\linewidth]{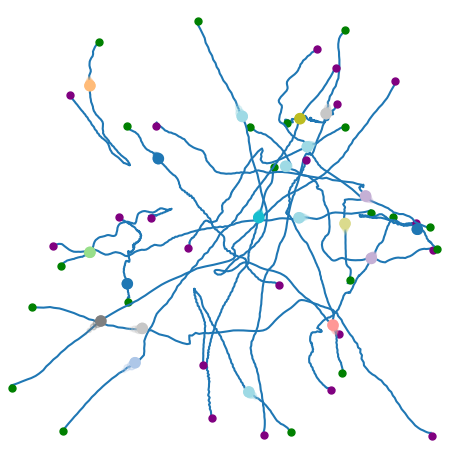}
        } &
        \subcaptionbox*{}[0.15\textwidth]{%
            \includegraphics[width=\linewidth]{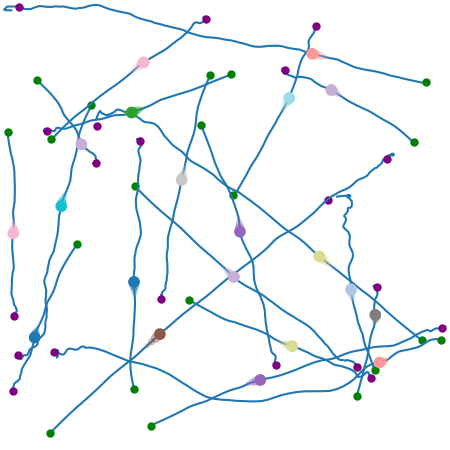}
        } \\

        \rotatebox{90}{\textbf{25 Robots}} &
        \subcaptionbox*{}[0.15\textwidth]{%
            \includegraphics[width=\linewidth]{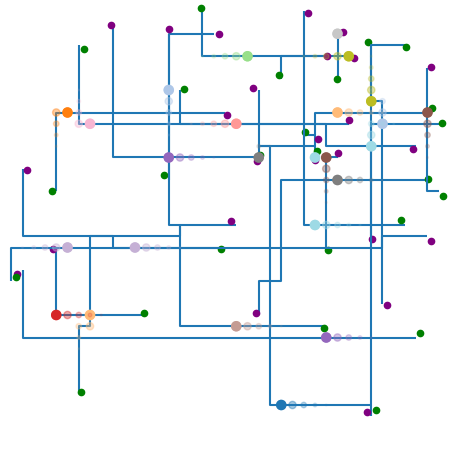}
        } &
        \subcaptionbox*{}[0.15\textwidth]{%
            \centering
            \includegraphics[width=\linewidth]{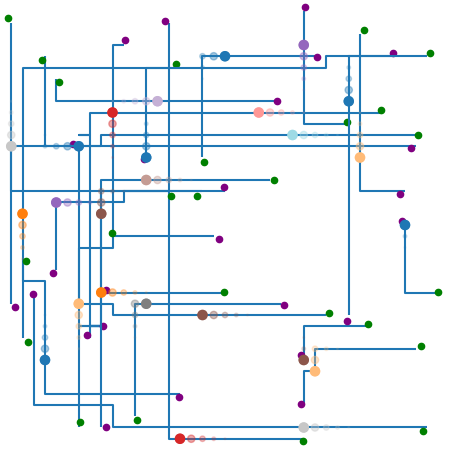}
        } &
        \subcaptionbox*{}[0.15\textwidth]{%
            N/A
        } &
        \subcaptionbox*{}[0.15\textwidth]{%
            \includegraphics[width=\linewidth]{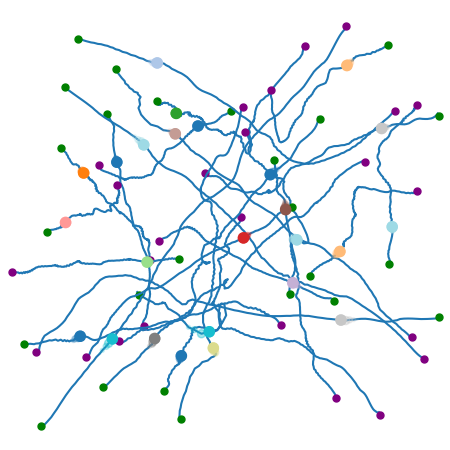}
        } &
        \subcaptionbox*{}[0.15\textwidth]{%
            \includegraphics[width=\linewidth]{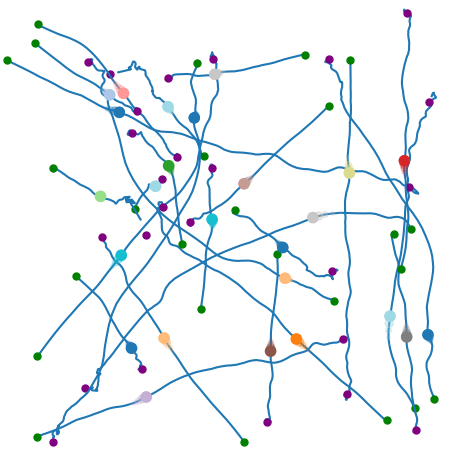}
        } \\

        \rotatebox{90}{\textbf{30 Robots}} &
        \subcaptionbox*{}[0.15\textwidth]{%
            \includegraphics[width=\linewidth]{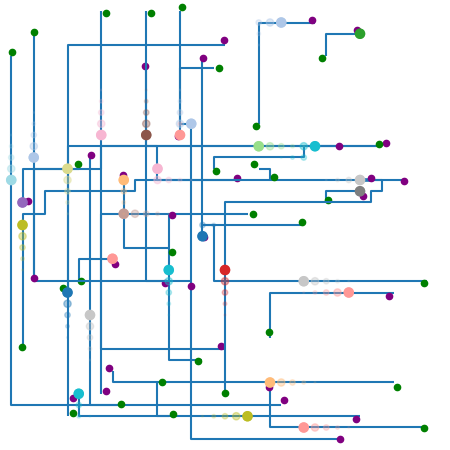}
        } &
        \subcaptionbox*{}[0.15\textwidth]{%
            \includegraphics[width=\linewidth]{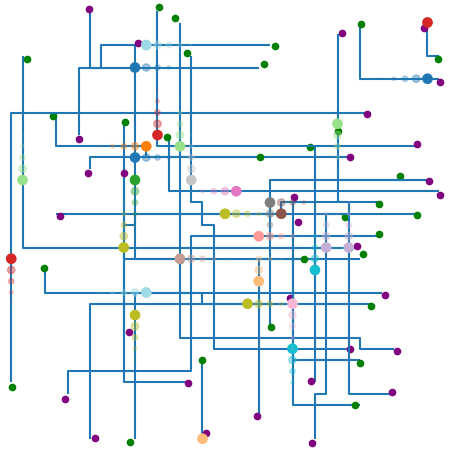}  
        } &
        \subcaptionbox*{}[0.15\textwidth]{%
            N/A
        } &
        \subcaptionbox*{}[0.15\textwidth]{%
            \includegraphics[width=\linewidth]{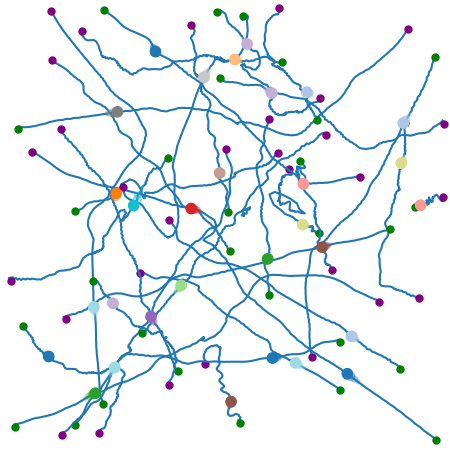}
        } &
        \subcaptionbox*{}[0.15\textwidth]{%
            \includegraphics[width=\linewidth]{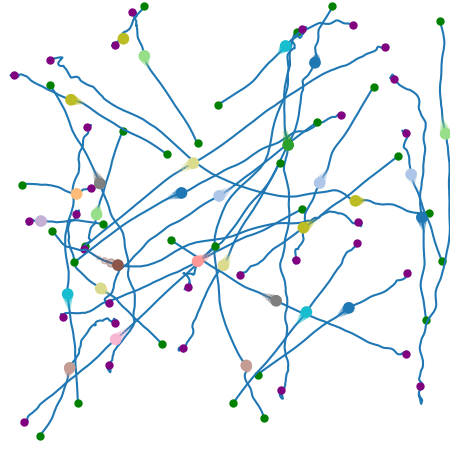}
        } \\

        \rotatebox{90}{\textbf{35 Robots}} &
        \subcaptionbox*{}[0.15\textwidth]{%
            \includegraphics[width=\linewidth]{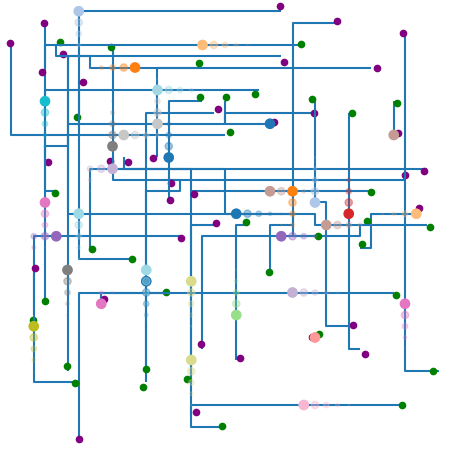}
        } &
        \subcaptionbox*{}[0.15\textwidth]{%
            \centering
            \includegraphics[width=\linewidth]{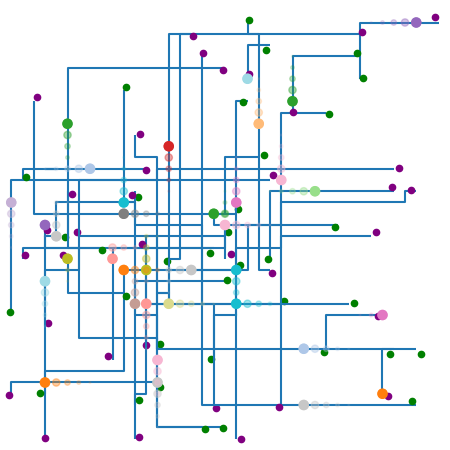}
        } &
        \subcaptionbox*{}[0.15\textwidth]{%
            N/A
        } &
        \subcaptionbox*{}[0.15\textwidth]{%
            \includegraphics[width=\linewidth]{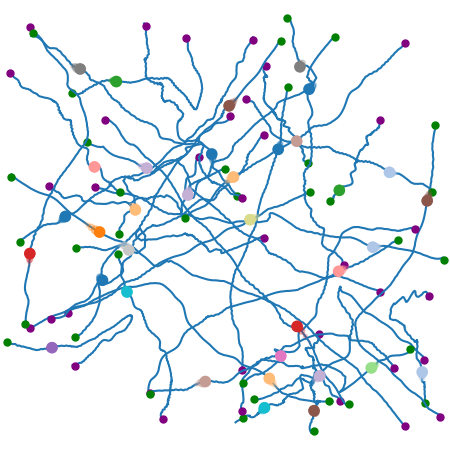}
        } &
        \subcaptionbox*{}[0.15\textwidth]{%
            \includegraphics[width=\linewidth]{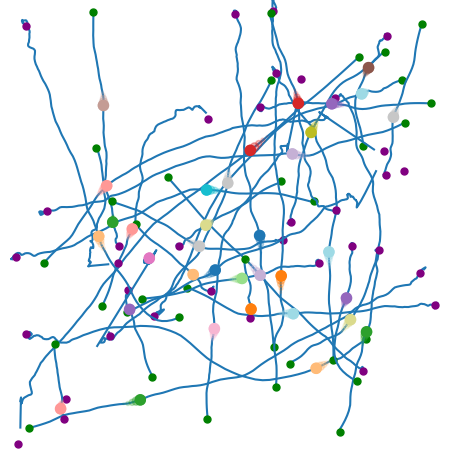}
        } \\

        \rotatebox{90}{\textbf{40 Robots}} &
        \subcaptionbox*{}[0.15\textwidth]{%
            \includegraphics[width=\linewidth]{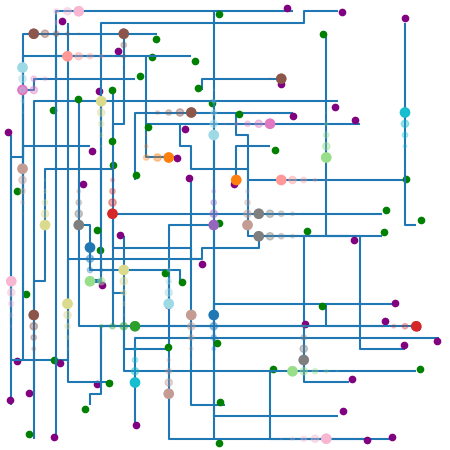}
        } &
        \subcaptionbox*{}[0.15\textwidth]{%
            \includegraphics[width=\linewidth]{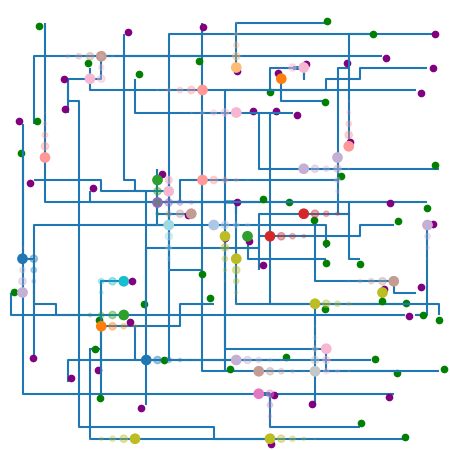}  
        } &
        \subcaptionbox*{}[0.15\textwidth]{%
            N/A
        } &
        \subcaptionbox*{}[0.15\textwidth]{%
            \includegraphics[width=\linewidth]{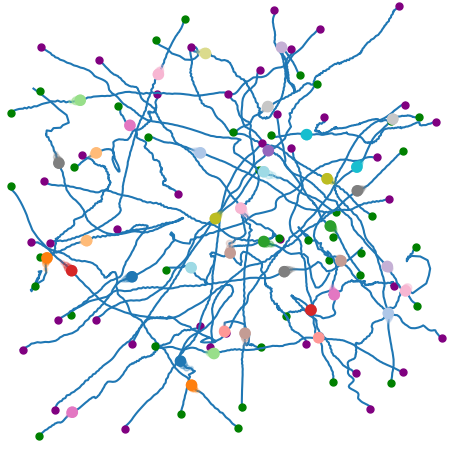}
        } &
        \subcaptionbox*{}[0.15\textwidth]{%
            \includegraphics[width=\linewidth]{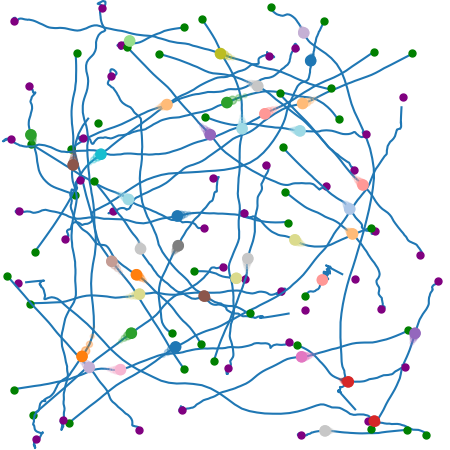}
        } \\
        
    \end{tabular}
    \caption{Some visual examples of trajectories generated by A*-CBS, A*-ECBS, KCBS, MMD-CBS, and MDOC-CBS in the Empty Large $4 \times 4$ map with the random start-goal pairs. We choose robot number \{12, 15, 20, 25, 30, 35, 40\}. N/A denotes the failure among all trials. Trajectories of KCBS, MMD-CBS, and MDOC-CBS are displayed after post-smoothing for better visualization.}
    \label{fig:emptylarge_trajs}
\end{figure*}

\begin{figure*}[ht]
    \centering
    \setlength{\tabcolsep}{1em} 
    \renewcommand{\arraystretch}{0.5} 
    
    \begin{tabular}{p{0.02\textwidth}ccccc}
        & \textbf{A*-CBS} & \textbf{A*ECBS} & \textbf{KCBS} & \textbf{MMD-CBS} & \textbf{MDOC-CBS} \\

        \rotatebox{90}{\textbf{12 Robots}} &
        \subcaptionbox*{}[0.15\textwidth]{%
            \includegraphics[width=\linewidth]{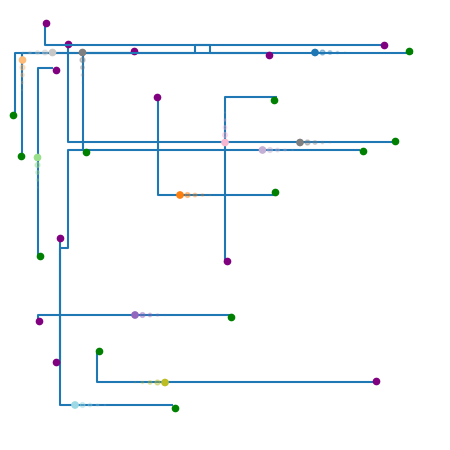}
        } &
        \subcaptionbox*{}[0.15\textwidth]{%
            \centering
            \includegraphics[width=\linewidth]{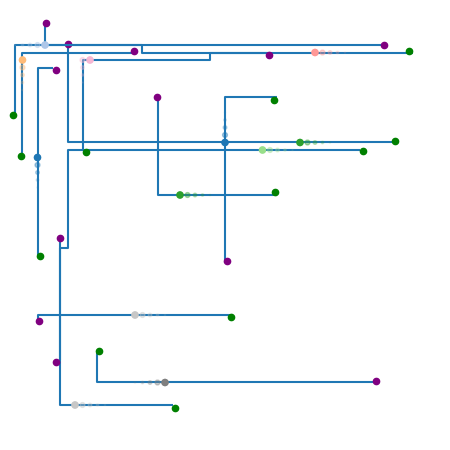}
        } &
        \subcaptionbox*{}[0.15\textwidth]{%
            \includegraphics[width=\linewidth]{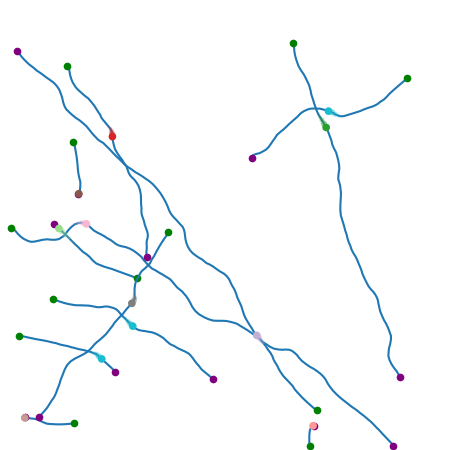}
        } &
        \subcaptionbox*{}[0.15\textwidth]{%
            \includegraphics[width=\linewidth]{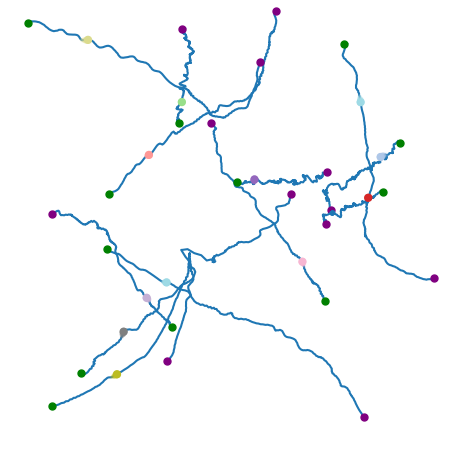}
        } &
        \subcaptionbox*{}[0.15\textwidth]{%
            \includegraphics[width=\linewidth]{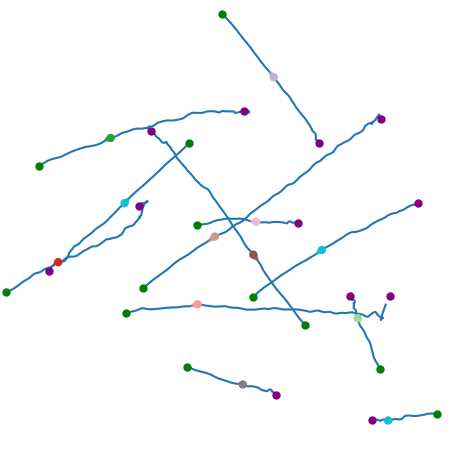}
        } \\

        \rotatebox{90}{\textbf{15 Robots}} &
        \subcaptionbox*{}[0.15\textwidth]{%
            \includegraphics[width=\linewidth]{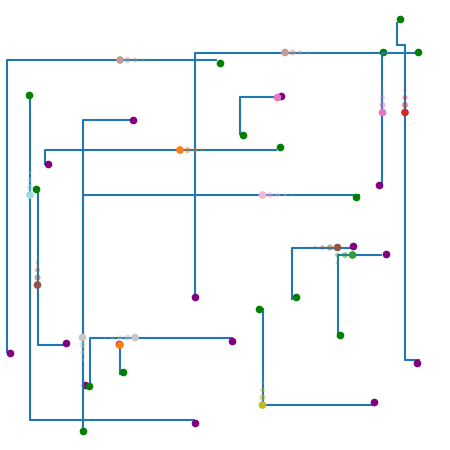}
        } &
        \subcaptionbox*{}[0.15\textwidth]{%
            \centering
            \includegraphics[width=\linewidth]{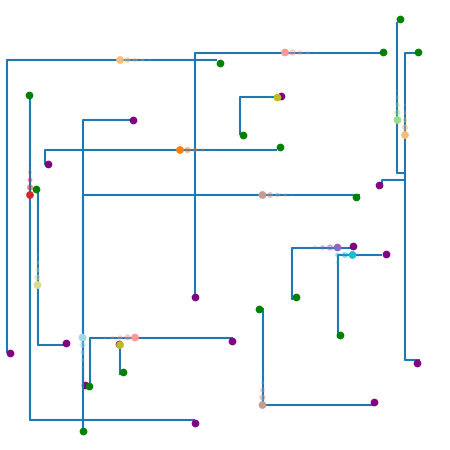}
        } &
        \subcaptionbox*{}[0.15\textwidth]{%
            \includegraphics[width=\linewidth]{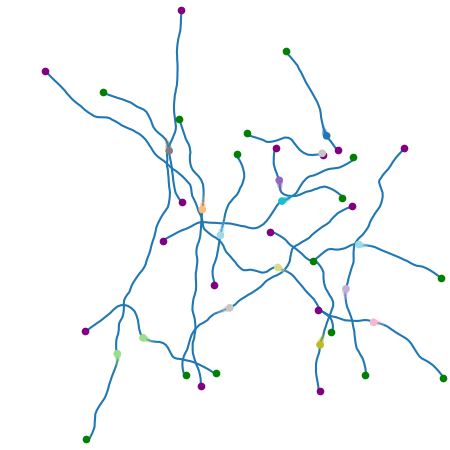}
        } &
        \subcaptionbox*{}[0.15\textwidth]{%
            \includegraphics[width=\linewidth]{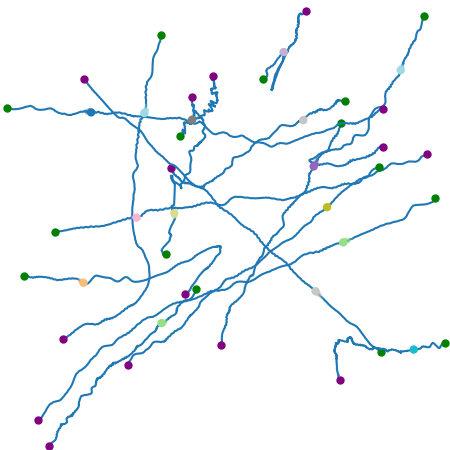}
        } &
        \subcaptionbox*{}[0.15\textwidth]{%
            \includegraphics[width=\linewidth]{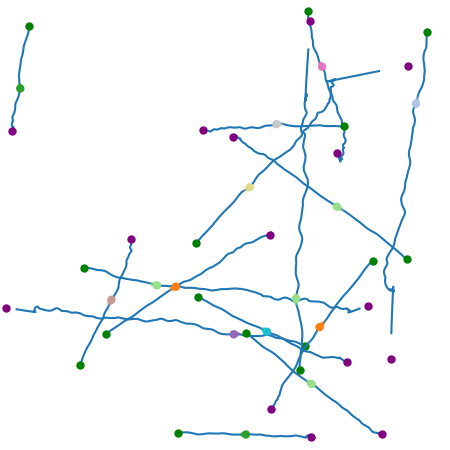}
        } \\

        \rotatebox{90}{\textbf{20 Robots}} &
        \subcaptionbox*{}[0.15\textwidth]{%
            \includegraphics[width=\linewidth]{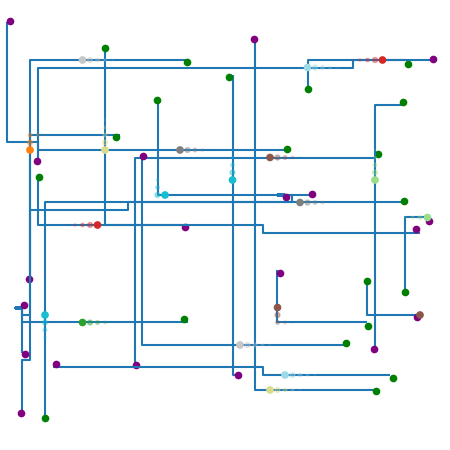}
        } &
        \subcaptionbox*{}[0.15\textwidth]{%
            \includegraphics[width=\linewidth]{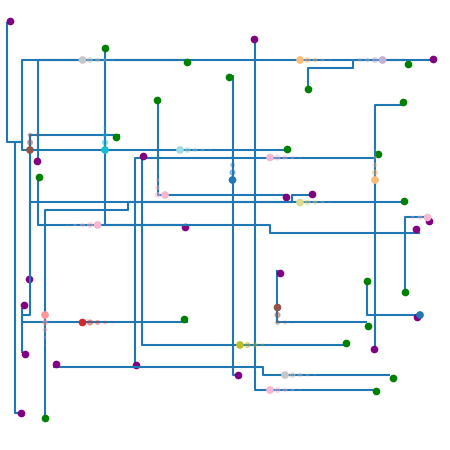}  
        } &
        \subcaptionbox*{}[0.15\textwidth]{%
            \includegraphics[width=\linewidth]{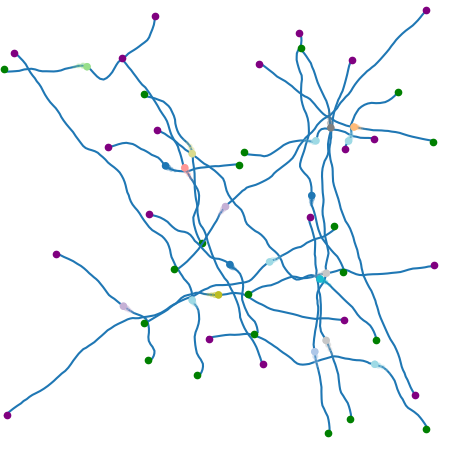}
        } &
        \subcaptionbox*{}[0.15\textwidth]{%
            \includegraphics[width=\linewidth]{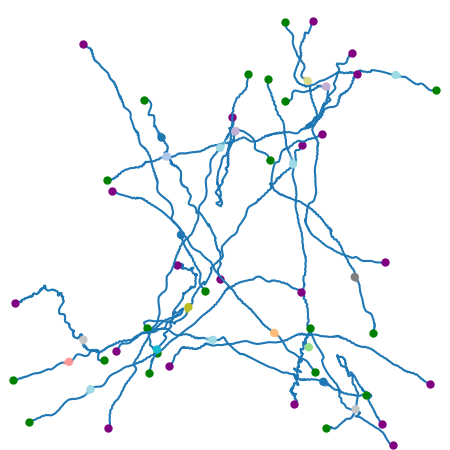}
        } &
        \subcaptionbox*{}[0.15\textwidth]{%
            \includegraphics[width=\linewidth]{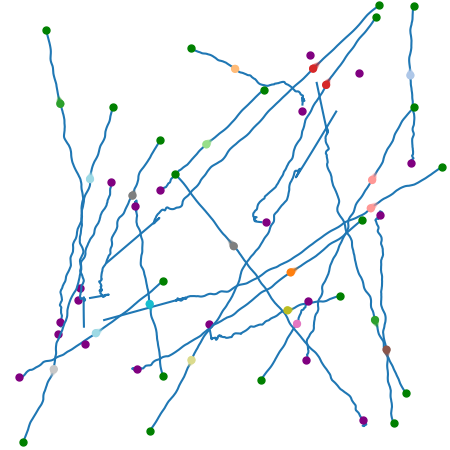}
        } \\

        \rotatebox{90}{\textbf{25 Robots}} &
        \subcaptionbox*{}[0.15\textwidth]{%
            \includegraphics[width=\linewidth]{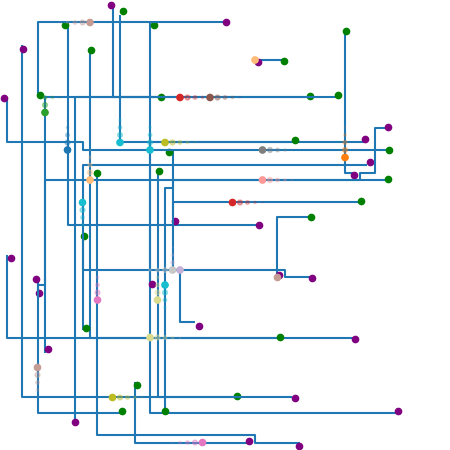}
        } &
        \subcaptionbox*{}[0.15\textwidth]{%
            \centering
            \includegraphics[width=\linewidth]{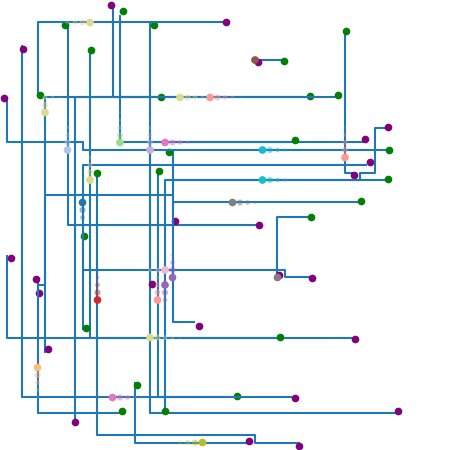}
        } &
        \subcaptionbox*{}[0.15\textwidth]{%
            \includegraphics[width=\linewidth]{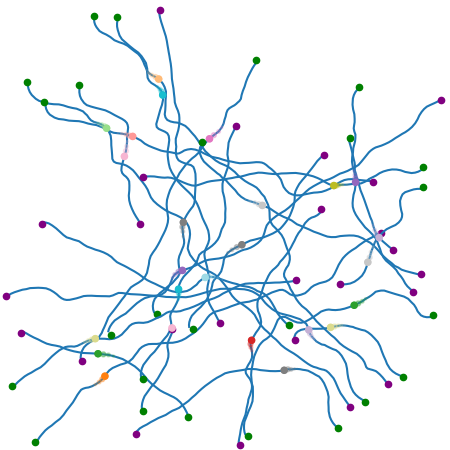}
        } &
        \subcaptionbox*{}[0.15\textwidth]{%
            \includegraphics[width=\linewidth]{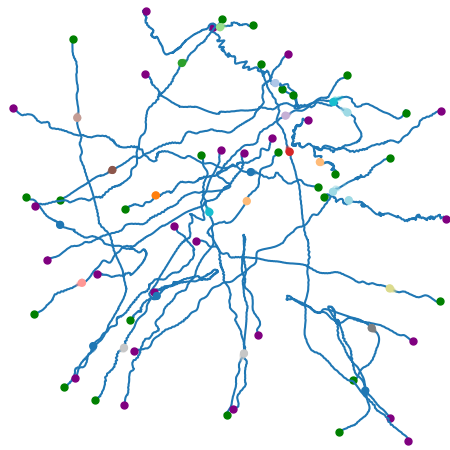}
        } &
        \subcaptionbox*{}[0.15\textwidth]{%
            \includegraphics[width=\linewidth]{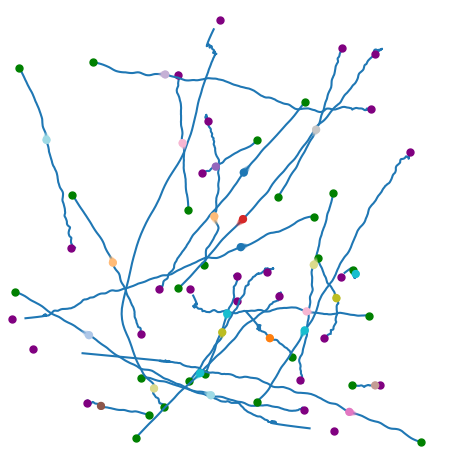}
        } \\

        \rotatebox{90}{\textbf{30 Robots}} &
        \subcaptionbox*{}[0.15\textwidth]{%
            \includegraphics[width=\linewidth]{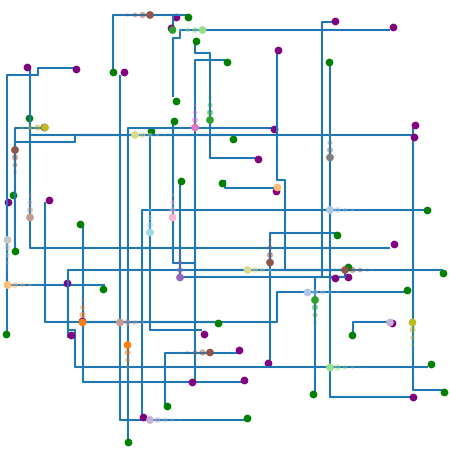}
        } &
        \subcaptionbox*{}[0.15\textwidth]{%
            \includegraphics[width=\linewidth]{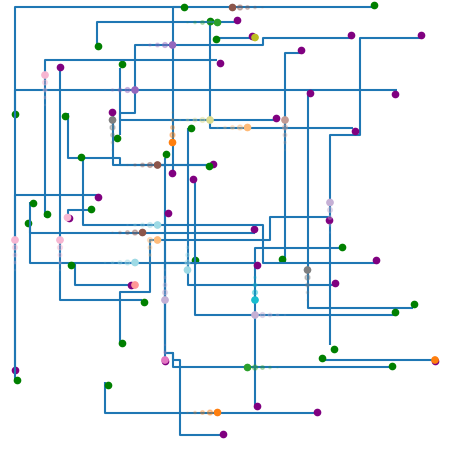}  
        } &
        \subcaptionbox*{}[0.15\textwidth]{%
            N/A
        } &
        \subcaptionbox*{}[0.15\textwidth]{%
            \includegraphics[width=\linewidth]{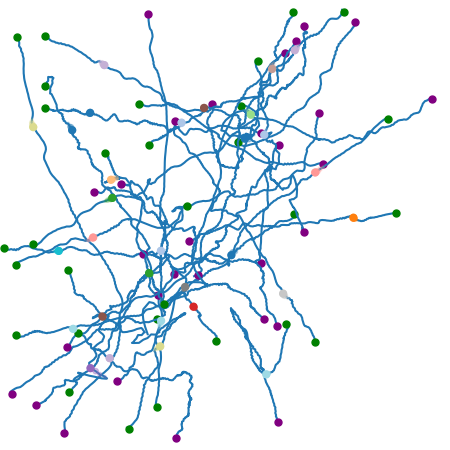}
        } &
        \subcaptionbox*{}[0.15\textwidth]{%
            \includegraphics[width=\linewidth]{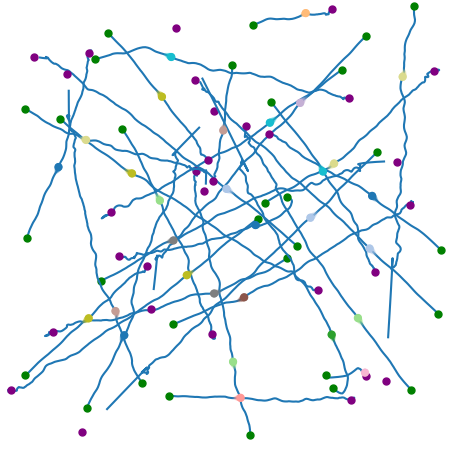}
        } \\

        \rotatebox{90}{\textbf{35 Robots}} &
        \subcaptionbox*{}[0.15\textwidth]{%
            \includegraphics[width=\linewidth]{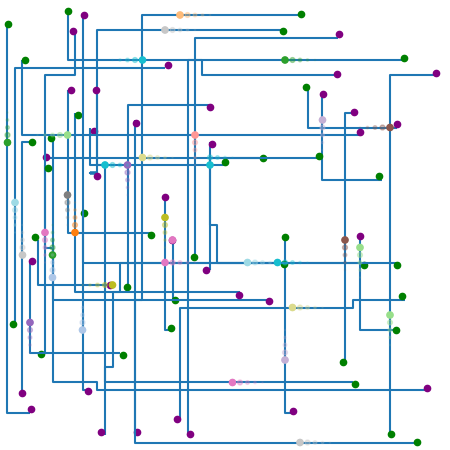}
        } &
        \subcaptionbox*{}[0.15\textwidth]{%
            \centering
            \includegraphics[width=\linewidth]{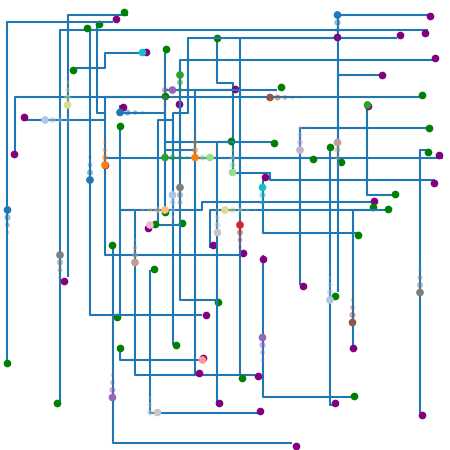}
        } &
        \subcaptionbox*{}[0.15\textwidth]{%
            N/A
        } &
        \subcaptionbox*{}[0.15\textwidth]{%
            \includegraphics[width=\linewidth]{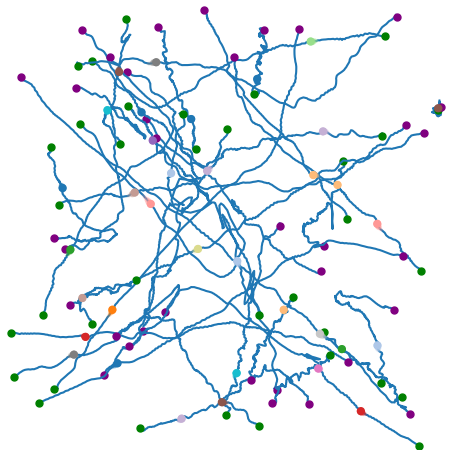}
        } &
        \subcaptionbox*{}[0.15\textwidth]{%
            \includegraphics[width=\linewidth]{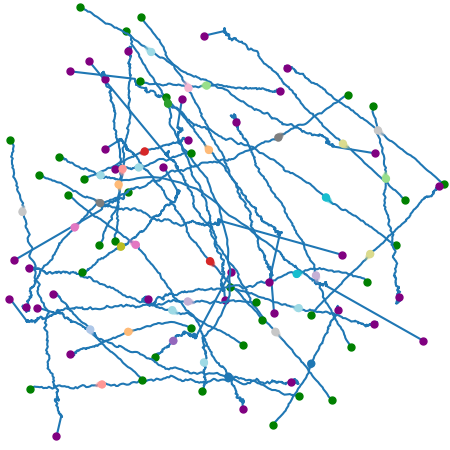}
        } \\

        \rotatebox{90}{\textbf{40 Robots}} &
        \subcaptionbox*{}[0.15\textwidth]{%
            \includegraphics[width=\linewidth]{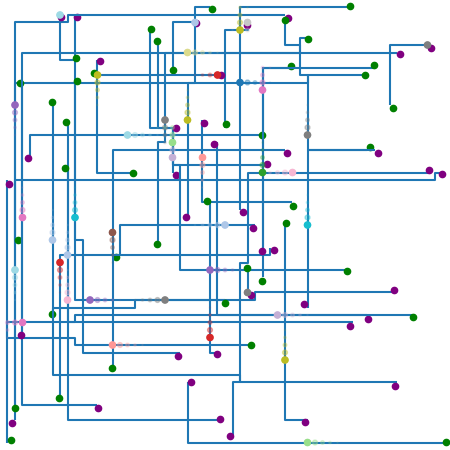}
        } &
        \subcaptionbox*{}[0.15\textwidth]{%
            \includegraphics[width=\linewidth]{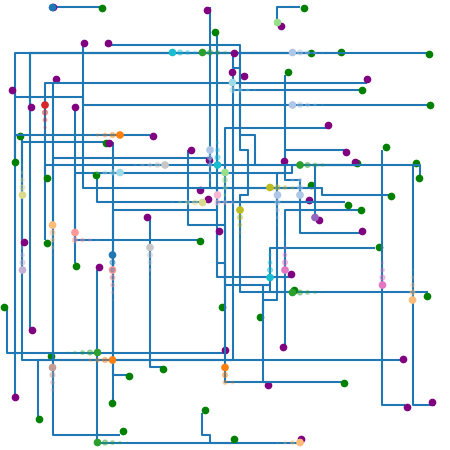}  
        } &
        \subcaptionbox*{}[0.15\textwidth]{%
            N/A
        } &
        \subcaptionbox*{}[0.15\textwidth]{%
            N/A
        } &
        \subcaptionbox*{}[0.15\textwidth]{%
            \includegraphics[width=\linewidth]{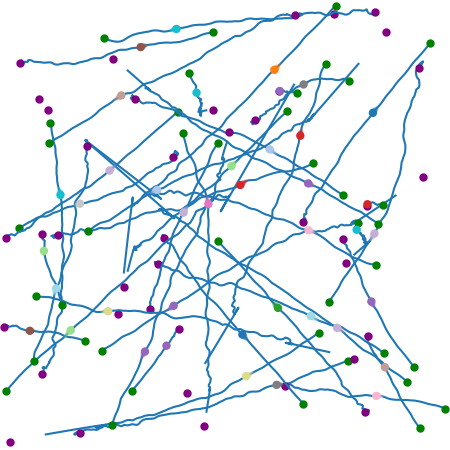}
        } \\
        
    \end{tabular}
    \caption{Some visual examples of trajectories generated by A*-CBS, A*-ECBS, KCBS, MMD-CBS, and MDOC-CBS in the Empty Extra Large $6 \times 6$ map with the random start-goal pairs. We choose robot number \{12, 15, 20, 25, 30, 35, 40\}. N/A denotes the failure among all trials. Trajectories of KCBS, MMD-CBS, and MDOC-CBS are displayed after post-smoothing for better visualization.}
    \label{fig:emptyextralarge_trajs}
\end{figure*}